\pdfoutput=1

\documentclass[12pt]{thesis-umich}[thesis]

\usepackage{url}
\usepackage{graphicx}
\usepackage{subcaption}
\usepackage{amsmath}
\usepackage[inline]{enumitem}
\usepackage{booktabs}
\usepackage{multicol}
\usepackage{color}
\usepackage{booktabs}
\usepackage{multirow}
\usepackage{xspace}
\usepackage{xcolor}
\usepackage{natbib}
\usepackage{amsmath,bm,amsfonts,dsfont,bbm}
\usepackage{hyperref}
\usepackage{mdframed}
\usepackage{csquotes}
\usepackage{epigraph}
\usepackage{geometry}
\usepackage{pdfpages}
\usepackage{graphicx}
\usepackage{fullpage}
\usepackage{amsfonts}
\usepackage{amsmath}
\usepackage{algpseudocode}
\usepackage{algorithm}
\usepackage{subcaption}
\usepackage{amssymb}
\usepackage{appendix}
\usepackage{bbm}
\usepackage{multirow}
\usepackage{pifont}%
\usepackage{latexsym}

\usepackage{tikz}
\usetikzlibrary{positioning}
\usetikzlibrary{calc}
\usetikzlibrary{arrows}
\usetikzlibrary{decorations.markings}
\usetikzlibrary{shapes.misc}
\usetikzlibrary{matrix,shapes,arrows,fit,tikzmark}
\usetikzlibrary{arrows.meta}
\usetikzlibrary{patterns}
\usetikzlibrary{patterns.meta}
\usetikzlibrary{decorations.pathreplacing,calc,positioning}
\usetikzlibrary{arrows,shapes}
\usetikzlibrary{arrows,chains,positioning,scopes,quotes}
\usetikzlibrary{calc,trees,positioning,arrows,chains,shapes.geometric,%
	decorations.pathreplacing,decorations.pathmorphing,shapes,%
	matrix,shapes.symbols}
\usetikzlibrary{shapes.callouts}
\usepackage[most]{tcolorbox}
\usepackage{array}
\usepackage{colortbl}
\usepackage{dcolumn}
\newcolumntype{d}[1]{D{.}{.}{#1}}  %

\usepackage{geometry}
\usepackage[activate={true,nocompatibility}, kerning=true,spacing=true, stretch=10,shrink=30]{microtype}
\microtypecontext{spacing=nonfrench}

\tikzset{
	>=stealth',
	punktchain/.style={
		rectangle, 
		rounded corners, 
		draw=black, %
		text width=5cm, 
		minimum height=2cm, 
		text centered, 
		inner sep=0,outer sep=0,
	},
	line/.style={draw, thick, <-},
	element/.style={
		tape,
		top color=white,
		bottom color=blue!50!black!60!,
		minimum width=8em,
		draw=blue!40!black!90, very thick,
		text width=10em, 
		minimum height=3.5em, 
		text centered, 
		on chain},
	every join/.style={->, thick,shorten >=1pt},
	decoration={brace},
	tuborg/.style={decorate},
	tubnode/.style={midway, right=2pt},
}

\newcommand{\hlent}[2]{\colorbox{gray!20}{#1\textsubscript{#2}}}
\definecolor{dkgreen}{RGB}{0,130,0}
\definecolor{aqua}{rgb}{0.0, 0.4, 1.0}
\def\vec#1{\ensuremath{\boldsymbol{{#1}}}}

\newcommand{\actignore}{\textsc{ignore}\xspace}
\newcommand{\actcoref}{\textsc{coref}\xspace}
\newcommand{\actoverwrite}{\textsc{overwrite}\xspace}
\newcommand{\mysim}{\mathit{sim}}
\newcommand{\mlp}{\mathrm{MLP}}
\newcommand{\cs}{\mathit{cs}}

\newcommand{\unbounded}{U-MEM\xspace}
\newcommand{\learned}{LB-MEM\xspace}
\newcommand{\lru}{RB-MEM\xspace}

\newcommand{\bertbase}{BERT\textsubscript{BASE}\xspace}
\newcommand{\bertlarge}{BERT\textsubscript{LARGE}\xspace}

\newcommand*\pct{\scalebox{.8}{\%}}
\newcommand{\pos}[1]{\texttt{#1}}

\newcommand{\legalmove}{LgM\xspace}
\newcommand{\exactmove}{ExM\xspace}
\newcommand{\piecetype}{AP\xspace}

\newenvironment{itemizesquish}{\begin{list}{\setcounter{enumi}{0}\labelitemi}{\setlength{\itemsep}{-0.25em}\setlength{\labelwidth}{0.5em}\setlength{\leftmargin}{\labelwidth}\addtolength{\leftmargin}{\labelsep}}}{\end{list}}

\newcommand{\cmark}{\ding{51}}%
\newcommand{\xmark}{\ding{55}}
\newcommand{\state}[0]{\texttt{[STATE]}}

\department{Computer Science}

\title{\large{Efficient and Interpretable Neural Models for Entity Tracking}}

\author{Shubham Toshniwal}

\department{Computer Science}

\email{shtoshni@gmail.com}

\frontpagestyle{7} %

	\chapter*{Acknowledgments}
	\noindent
	\input{frontmatter/thanks}
	\vspace*{\fill} \newpage
[7]{
The famous proverb "It takes a village to raise a child" is quite apt for my PhD. 
It has been a long winding journey which would've been impossible without some of the smartest, nicest, and hard-working people I've met along the way.

To begin with, I want to thank Karen Livescu, Kevin Gimpel, and Sam Wiseman. Karen has been my advisor from day one and has been a source of constant support and wisdom throughout these years. I've learned so much from her and am still in awe of her infinite reservoir of calm. 
Karen's meticulousness and research ethics are values that I hope to carry forward in my research career. 
Talking of Kevin, while he officially became my advisor much later, he has guided me from the very start. His love for NLP can rub off on anyone. I'm amazed by his ability to remember even the most minor
details while juggling countless projects. Finally, Sam has been an unofficial third advisor for me. He has been critical to many of this thesis's ideas, which I'm sure he would humbly minimize. I started working with him somewhat late in my Ph.D. which is why I'm envious of his students at
Duke :) Along with Karen, Kevin, and Sam, I want to thank Kenton and Yejin for making time in their busy schedules to be on the thesis committee.

I want to thank Allyson Ettinger, Mohit Bansal, Liang Lu, and Mrinmaya Sachan with whom I have collaborated during their stints as Research Assistant Professor (RAP) at TTIC. 
Thanks to Tara Sainath and Ron Weiss for being amazing intern hosts at Google in summer of 2017.
Tara and Ron continued to support me beyond my internship and even when I moved away from research in speech. 
Thanks to Daniel Gillick and Alessandro Presta for hosting me at Google in 2018. 

Thanks to Hao Tang, Trang Tran, Kalpesh Krishna, Patrick Xia, Freda Shi, Bowen Shi, and Lingyu Gao for being amazing collaborators. 
Hao is one of the most principled person I've come across and
was lucky to have him as my senior/mentor and for also teaching me tennis. 
With Trang I was involved in one of the most annoying at first but ultimately one of the most satisfying projects during my PhD. 
I've never met Patrick but he is my go to person to talk about coreference. 

Thanks to Hao Tang, Shane Settle, Herman Kamper, Ankita Pasad, Davis Yoshida, Freda Shi, Bowen Shi, David Yunis who made Karen's and SLATTIC group meetings fun and enriching to attend.  
Thanks to my cohort of Ruotian Luo, Rachit Nimavat, Shane Settle, Nick Kolkin, Blake Woodworth, and Falcon Dai with whom I had a lot of fun taking courses and discussing research directions. 
Shane's probably the funniest guy I've met in person and I had an absolute blast attending my initial conferences with him.
Rachit was my first flatmate when I moved to Chicago. 
I'm still searching for someone as carefree and happy as Rachit.
Ruotian was my cubicle neighbor and gosh he could code up things in sleep which I couldn't while awake.
I want to thank Behnam, Shubhendu, Somaya, Haris, and Mrinal for being amazing seniors. Shubhendu, my ex-flatmate, is probably the most interesting person I've met till now. 
I hope his claim of smoking cigars for a long life turns out to be true. 
Thanks to Sudarshan, Andrea, Pushkar, Igor, Ankita, Chip, Kevin, Shashank, Akilesh, Renyu, Omar, Naren, Kshitij for making TTIC a fun place. 
I've bonded with Sudarshan, Pushkar, Ankita, and Shashank over so many topics besides research. 
I want to thank Ankita, Shane, and Sudarshan for proof-reading the thesis. 
I'm grateful to Sudarshan and Igor for bringing in the latest gossip at lightning speeds in the comfort of our cubicles. 
Andrea was my gym/fitness buddy which meant we went to gym a handful of times over the course of our PhDs. 
I'm thankful to Erika and Andrea for inviting me to their elaborate dinners. 

Outside of TTIC, I want to thank my undergraduate friends Baba (Palak), Prof (Prashant), DKM (Dipendra), Gitesh, Gaur (Siddharth), Aritro, Chirag who have been a constant presence even after so many years. 
I want to thank Archita, Palak, Kshiteej, Prashant, Gitesh, Rachit, Udbhav, Sujaya, Jahn, Eeshit for some memorable trips along the way.  

Returning to TTIC, I want to thank Greg and Madhur for teaching two of my favorite courses. TAing for Greg meant seeing more of his humor at close quarters. 
Thanks to Sunanda and Madhur for some fun times outside of work.  
Admininstrators at TTIC have always been top notch and a special shout out to Adam, Chrissy, Amy, and Mary. 
Finally, I want to remember our late ex-president Sadaoki Furui who recently passed away. 
Playing tennis with him was one of the highlights of my PhD. 
We remained in contact till a couple of months back and he was still on top of the latest developments in machine learning.  
I'm still in awe of his youthful energy, humility, and openness. 

I want to thank the city of Chicago, which has such a rich cultural heritage and iconic architecture. 
The view of the Chicago skyline from the lakeshore trail remains one of the most beautiful sights I've ever seen. 
Taking walks along Lake Michigan, Jackson Park, and UChicago campus remain the highlight of my Hyde Park stay. 
Now that I've moved on introducing myself via Chicago has endowed me with an American identity that fills me with pride.

I want to thank my teachers in DPS Agra and Allen Institute Kota who paved my path to IIT Kanpur. 
In particular, I want to thank Vinod Arora Sir (Principal, DPS Agra), Haraprasad Sahu Sir (Maths, DPS Agra), Jeevan Jyoti Sir (Maths, Allen Kota), and Parijat Sir (Admin, Allen Kota). 
I ended up at both these institutes by sheer luck and they remain seminal in my life trajectory. 

I'm forever indebted to my parents who're still not sure what I do but have unconditionally supported me regardless. 
Mummy and Papa I promise to keep my work a mystery :)
My late grandfather wanted me to become a doctor. He didn't say which one, so I'm counting this one :)
Thanks to my elder brother Arvind and my sister-in-law Neelam for giving us Aadvik and Karthik to pamper.  
While I'm earning the first doctorate in my extended family, the emphasis of education in my upbringing has a big role in that. 
Thanks to my extended family who have always found time for me when I'm back home.

}

\committee{ %
	Professor Kevin Gimpel, Co-Chair \\
	Professor Karen Livescu, Co-Chair \\
	Professor Sam Wiseman \\
	Kenton Lee \\
	Professor Yejin Choi \\
} 

\cochair{Co-chair One \& Co-chair Two}

\showlistoftables

\abstract{
	\begin{center}
	{\Large\bf Efficient and Interpretable Neural Models for Entity Tracking} \\[0.1cm]
\end{center}

\vspace{1cm}

What would it take for a natural language model to \emph{understand} a  novel, such as \emph{The Lord of the Rings}? 
Among other things, such a model must be able to: 
(a) identify and record new characters (entities) and their attributes as they are introduced in the text, and (b) identify subsequent references to the characters previously introduced and update their attributes. This problem of \emph{entity tracking} is essential for language understanding, and thus, useful for a wide array of downstream applications in NLP such as question-answering, summarization. 

In this thesis, we focus on two key problems in relation to facilitating the use of entity tracking models: (i) scaling entity tracking models to long documents, such as a novel, and (ii) integrating entity tracking into language models.  
Applying language technologies to long documents has garnered interest recently, but computational constraints are a significant bottleneck in scaling up current methods. 
In this thesis, we argue that computationally efficient entity tracking models can be developed by representing entities with rich, fixed-dimensional vector representations derived from pretrained language models, and by  exploiting the ephemeral nature of entities. 
We also argue for the integration of entity tracking into language models as it will allow for: (i) wider application given the current ubiquitous use of pretrained language models in NLP applications, and (ii) easier adoption since it is much easier to swap in a new pretrained language model than to  integrate a separate standalone entity tracking model.

The thesis is divided into two parts. In the first half, we focus on a specific class of entity tracking problem referred to as \emph{coreference resolution}. The goal here is to identify text spans referring to the same entity. 
We propose memory models where the external memory module is trained to \emph{explicitly} track the entities mentioned in the text. 
We first discuss a sparsely supervised memory model for the pronoun resolution task. This model outperforms prior work on both the end task and  interpretability measures. 
We then adapt this memory model for the full coreference resolution task. 
The proposed memory models can effectively scale to long documents, and in particular, the proposed bounded memory model offers a linear runtime complexity in document length while remaining competitive with the state-of-the-art models. 
Next, we test the presented models for their generalization capability, specifically their zero-shot performance on other coreference benchmarks. 
We find that domain shift is a challenge in coreference resolution, though annotation differences across datasets partly exaggerate this challenge. 
We also find that joint training on multiple datasets moderately alleviates the domain shift challenge.
Finally, the presented models have achieved state-of-the-art performance on multiple coreference benchmarks.

In the latter half, we focus on integrating entity tracking capability into neural language models. 
As a first step, we propose the task of language modeling for the game of chess to evaluate the entity tracking capabilities of transformer LMs. 
Our experiments on chess suggest that augmenting LM training instances with board state information (represented as text tokens) aids the state tracking and language modeling performance. 
Training LMs with state-augmented instances also allows probing for entity state at inference time simply via prompting. 
Next, we extend these findings from chess to natural language. 
We first experiment in a closed domain where we show that state-augmented training improves the state tracking performance and the text generation quality. 
Finally, we adapt the state-augmented training for baking coreference knowledge into natural language models and show improvements on a popular cloze task.

}

\begin{document}

\chapter{Introduction}

\label{sec:intro}

\begin{figure}[h]

	\centering{
			\fbox{
	\begin{minipage}{0.92\textwidth}
\texttt{[\textcolor{aqua}{Bilbo}]\textsubscript{1} celebrates [\textcolor{aqua}{his}]\textsubscript{1} eleventy-first birthday and leaves [\textcolor{dkgreen}{the Shire}]\textsubscript{2} suddenly, passing [\textcolor{orange}{the Ring}]\textsubscript{3} to [\textcolor{purple}{Frodo Baggins}]\textsubscript{4}, [[\textcolor{aqua}{his}]\textsubscript{1} \textcolor{purple}{cousin and heir}]\textsubscript{4}. %
Neither hobbit is aware of [\textcolor{orange}{the Ring's}]\textsubscript{3} origin, but [\textcolor{gray}{the wizard Gandalf}]\textsubscript{5} suspects [\textcolor{orange}{it}]\textsubscript{3} is a Ring of Power. 
Seventeen years later, [\textcolor{gray}{Gandalf}]\textsubscript{5} tells [\textcolor{purple}{Frodo}]\textsubscript{4} that [\textcolor{gray}{he}]\textsubscript{5} has confirmed that [\textcolor{orange}{the Ring}]\textsubscript{3} is the one lost by [\textbf{the Dark Lord Sauron}]\textsubscript{6} long ago and counsels [\textcolor{purple}{him}]\textsubscript{4} to take [\textcolor{orange}{it}]\textsubscript{3} away from [\textcolor{dkgreen}{the Shire}]\textsubscript{2}. 
}
	\end{minipage}
}
\caption{Text excerpt from plot summary of \emph{The Fellowship of the Ring}.}
	\label{fig:coref_example}
}
\end{figure}

Understanding text narratives requires: 
(a) identifying and recording new characters (entities) and their attributes as they are introduced in the text, and (b) identifying later references to the characters previously introduced and updating their attributes.  
To appreciate the challenges of this \emph{entity tracking} task, consider the above text. 
The text introduces some of the key characters of \emph{The Lord of the Rings} novel, such as \emph{Bilbo Baggins}, \emph{Frodo Baggins}, \emph{Gandalf}, \emph{the Ring}, and \emph{Sauron}. In the text, the character \textit{Frodo Baggins} is referenced in multiple surface forms, namely \emph{Frodo Baggins}, \textit{his cousin and heir}, \textit{Frodo}, and \textit{him}.   
At the same time, personal pronouns \emph{his}, \emph{he}, and \emph{him} have been used in different contexts to refer to \emph{Bilbo}, \emph{Gandalf}, and \emph{Frodo}, respectively. 
We also see the ownership of \textit{the Ring} change from \textit{Bilbo} to \textit{Frodo}, and that it originally belonged to \textit{Sauron}. 
Thus, even in this short snippet, we see that understanding the text requires dealing with the many-to-many relationship between the surface forms and the entities and keeping track of the dynamically evolving state of entities introduced in the text.  
Finally, given the importance of entity tracking in text comprehension, the entity tracking task is helpful for many downstream applications in NLP, such as question answering, summarization.
This thesis focuses on building efficient and interpretable neural models for entity tracking.

\section{Entity Tracking Task}
\label{sec:entity_tracking_def}
In this section, we first define the key preliminary concepts for the entity tracking task. Next, we give an overview of the instantiations of the entity tracking task and approaches in the literature. We will be using Figure~\ref{fig:coref_example} as our running example.

\subsection{Task Definitions}

\paragraph{An Entity} is any individual, location, organization, or object mentioned in the text. For the example text, the entity set is \{\emph{Bilbo} (people), \emph{the Shire} (location), \emph{the Ring} (object), \emph{Frodo Baggins} (people), \emph{Gandalf} (people), \emph{Sauron} (people)\}.

\paragraph{A Mention} is any text span that refers to an entity. In our example, the entity \emph{Frodo Baggins} is referenced via the following mentions:  \emph{Frodo Baggins}, \textit{his cousin and heir}, \textit{Frodo}, and \textit{him}. 

\paragraph{An Antecedent} of a mention is any mention that: (a) refers to the same entity, and (b) occurs earlier in the document. A mention can have multiple antecedents, as in the case of the mention \emph{Frodo}  which has  \emph{Frodo Baggins} and \emph{his cousin and heir} as its antecedents.

\paragraph{Entity Attribute} is any entity property, such as location, age. 
In our example, \emph{the Ring} has the attribute \texttt{owner} whose value evolves through the story. The \texttt{location} attribute of \emph{the Ring} remains \emph{the Shire} in the narrative, but it's suggested that it would change soon.

\paragraph{Entity State} is the collection of all the attribute-value pairs for an entity, essentially the complete  representation of an entity. Any change in any entity attribute implies a change in the entity state.

\subsection{Task Instantiations}
The entity tracking tasks in the literature can broadly be categorized into two categories, namely coreference resolution and entity state tracking.\footnote{Prior work generally refers to entity state tracking when using the term entity tracking, though there have been exceptions where the term has been used to denote coreference~\cite{hoang-etal-2018-entity, clark-etal-2018-neural}.} 
The coreference resolution task focuses on identifying the entity to which a given text span refers, while the entity state tracking task is concerned with the state/attributes of entities mentioned in the text.\footnote{Coreference resolution can be an intermediate task for the entity state tracking task.} 
We define the two tasks in more detail below.

\paragraph{Coreference Resolution} is a class of entity tracking problem where the task is to identify text spans that refer to the same entity. The task derives its name from the linguistic phenomenon of \emph{coreference} where two or more mentions refer to the same entity. The coreference resolution task takes the document as input, and the goal is to output mention clusters such that only mentions which refer to the same entity are co-clustered. The task assumes that every mention refers to one and only one entity, implying that clustering the mentions is equivalent to partitioning the mention set. The ground truth clusters for our running example are \{\emph{Bilbo}, \emph{his}, \emph{his}\},  \{\emph{the Shire}, \emph{the Shire}\}, \{\emph{the Ring}, \emph{the Ring's}, \emph{it}, \emph{the Ring}, \emph{it}\},  \{\emph{Frodo Baggins}, \emph{his cousin and heir}, \emph{Frodo}, \emph{him}\}, \{\emph{the wizard Gandalf}, \emph{Gandalf}, \emph{he}\}, and \{\emph{the Dark Lord Sauron}\}. 

Formally, let us assume that input document $\mathcal{D}$ has $T$ tokens and thus, $T(T + 1) / 2$ spans, which also upper bounds the maximum number of entities/clusters.\footnote{In practice we upper bound the length of candidate mentions with a constant, which reduces the number of potential spans from $\mathcal{O}(T^2)$ to $\mathcal{O}(T)$.}
We define $\mathcal{C}(\mathcal{D}) = \{\epsilon, 1, \cdots, T(T + 1) / 2\}$ as the set of all cluster IDs for document $\mathcal{D}$, where the dummy cluster $\epsilon$ serves as a target for invalid spans, i.e.\ spans that don't refer to any entity. The goal of coreference resolution is to output a clustering assignment $ c(x) \in \mathcal{C}(\mathcal{D})$ for all spans $x$ in $\mathcal{D}$  such that:
\begin{enumerate}
	\item $c(x) \ne  \epsilon$ is true only for all valid mentions i.e. spans that correspond to entities, and  
	\item  $c(x_1) = c(x_2) \ne \epsilon$ is true only for spans $x_1$ and $x_2$ which refer to the same discourse entity.
\end{enumerate}

Most of the earliest approaches for coreference used linguistically and cognitively motivated heuristics for the task~\citep{Hobbs1978ResolvingPR, brennan87centering, Carter1987InterpretingAI}.  
The current paradigm is to train supervised models on coreference resolution datasets~\citep{weischedel2013ontonotes, webster2018gap, bamman2019annotated}.
The statistical learning paradigm in coreference resolution has had a rich history, with recent past seeing the development of several neural models~\citep{wiseman-etal-2015-learning,clark-manning-2016-deep,lee-etal-2017-end,lee-etal-2018-higher, joshi-etal-2019-bert, joshi-etal-2020-spanbert, wu2019coreference}. 

Current learning approaches for coreference resolution typically break down the task into two steps, namely \textit{mention detection} and \textit{mention clustering}. 
The mention detection step outputs a pruned set of candidate spans which serve as input to the mention clustering step, which finally outputs the entity clusters. 
While earlier state-of-the-art models used a pipeline approach i.e.\ separately learning the two steps~\citep{durrett2013easy,wiseman-etal-2015-learning},  \citet{lee-etal-2017-end} proposed the first end-to-end neural model, which has become the dominant paradigm. 
In \citet{lee-etal-2017-end} and much of the follow up work since then~\citep[\emph{inter alia}]{lee-etal-2018-higher, joshi-etal-2019-bert, xu-choi-2020-revealing, kantor-globerson-2019-coreference}, mention clustering is done via \emph{mention-ranking}, where for each candidate mention the model ranks the top antecedent. Clustering naturally follows from these ranking predictions by chaining together mentions based on their top antecedent picks. 
One limitation of the mention-ranking approach is that it requires keeping all the past mentions in memory, which leads to linear growth in memory and quadratic increase in runtime with the length of the document. 
In practice, models cap the number of candidate antecedents, but even then, the models have large memory requirements \cite{xia-etal-2020-incremental}, making them inefficient, if not infeasible, for long documents.

\paragraph{Entity State Tracking} refers to the broad umbrella of tasks concerning maintaining a record of the entities and their attributes, which can evolve as the narrative progresses.\footnote{In linguistics, this task of representing entities and their properties is formally treated under the subject \emph{dynamic semantics}~\cite{heim1983file}.} 
In our example, the character \emph{Frodo} is: (a) \texttt{located} in \emph{the Shire},  (b) \texttt{owns} \emph{the Ring}, (c) a \texttt{person of type} hobbit. Note that attributes such as location, ownership, etc.\ can be dynamic, while \emph{Frodo} being a hobbit is a static attribute. 
What exactly constitutes an entity state can be domain-specific, making entity state tracking less standardized compared to coreference resolution. 
For example, the ProPara task~\cite{dalvi-etal-2018-tracking} is concerned with tracking an entity's existence and location in text describing scientific processes, while the bAbI tasks are regarding entity locations and their relations with other entities~\cite{weston2015aicomplete}. 
The ``closed vocabulary'' nature of entity attributes i.e.\ restricting them to a predefined set can be limiting; recent work by \citet{tandon-etal-2020-dataset} explores an ``open vocabulary'' approach to tracking entity state changes where the task is to generate the modified attributes and corresponding values.  

Formally, let us assume that the input text $\mathcal{S}$ consists of sentences $\{s_1, s_2, ..,  s_T\}$. Let $(e_i)_{i=1}^k$ represent the entities introduced in the discourse till sentence $s_t$. 
The entity tracking task is to learn a function $f(.)$ which, given an entity $e_i$ and an appropriate attribute $a$ for entity $e_i$, outputs the value $v$ of the attribute $a$ for entity $e_i$ at time step $t$:  
$$f(e_i, a, \{s_1, .., s_t\}) = v $$

A popular approach for this task is entity-centric memory models where the entity state for each entity is maintained and updated in an external memory ~\cite{henaff2016tracking, bosselut-18} (see Section~\ref{sec:memory_model} for a review of memory models). \citet{gupta-durrett-2019-effective} explore the use of pretrained transformer models for entity tracking.

\section{Motivations for This Thesis}

Prior work has shown that using the predictions of an entity tracking model leads to improved performance in a variety of downstream tasks such as cloze-style reading comprehension~\citep{dhingra-etal-2018-neural, hoang-etal-2018-entity, cheng2020entity}, question answering~\citep{dasigi-etal-2019-quoref}, dialog systems~\citep{gao-etal-2019-interconnected}, and  summarization~\citep{sharma-etal-2019-entity, narayan-etal-2021-planning}.
However, \textbf{for most current state-of-the-art  models for downstream tasks, integration of explicit entity tracking is more an exception than a norm.}
Typical state-of-the-art  models are built on top of massive pretrained language models (LMs) \citep[\textit{inter alia}]{ devlin-etal-2019-bert, radford2019language, brown2020language}. 
In this paradigm, the tacit assumption is that either pretrained LMs are already doing entity tracking, or that entity tracking can be learnt via the end task supervision.

\emph{Are language models/downstream task models doing implicit entity tracking?}
Probing results suggest that the current pretrained LMs are limited in their entity tracking capabilities~\citep{tenney2019probing, liu2019linguistic, sorodoc-etal-2020-probing}. 
\citet{schuster-linzen-2022-sentence} find that even models at the scale of GPT-3~\cite{brown2020language} lack basic entity tracking capabilities.  
Similarly, \citet{tulio2020checklist} show that state-of-the-art machine comprehension models fail on a variety of simple out-of-domain  evaluations, including 100\% failure rate on coreference-based evaluations. 
In neural summarization models, entity hallucination is a pertinent issue, where the generated summary has entities not mentioned in the source document~\cite{kryscinski-etal-2019-neural}. 
These results suggest that \emph{current state-of-the-art models lack entity tracking capabilities.} 
Moreover, the role of entity tracking is even more important to understand longer narratives~\cite{wu2021recursive}, such as book-length texts, which has seen a recent surge of interest in NLP~\cite{shaham-etal-2022-scrolls}. 

In this thesis, we take a two-pronged approach to facilitating the use of entity tracking models. The first line of work focuses on developing standalone entity tracking models which can scale to long text. The second line of work focuses on integrating entity tracking into pretrained LMs to allow for a wider application of entity tracking, given the ubiquity of pretrained LMs in current NLP pipeline. We discuss these two lines of work in more detail next.

\subsection{Efficient Entity Tracking  Models for Long Context Understanding} 
There has been a recent interest in extending NLP models~\cite{beltagy2020longformer, roy-etal-2021-efficient} to longer contexts, such as book-length texts~\cite{kocisky2018narrative, chen-etal-2022-summscreen, pang-etal-2022-quality, shaham-etal-2022-scrolls, xu-etal-2022-beyond}. 
However, analysis of state-of-the-art LMs trained for long context suggests that the models rarely use the full context, which suggests the inability of the LMs to capture discourse-level phenomena~\cite{sun-etal-2021-long}.  
\citet{shuster-etal-2022-state} show that dialog agents struggle to maintain an identity through a chat session, and the problem becomes worse with longer history. 
In question-answering (QA), methods which work for short context QA tasks~\cite{lewis-etal-2020-retrieval, karpukhin-etal-2020-dense}, struggle with long context tasks such as Book QA which requires modeling recurring characters and plots~\cite{mou-etal-2021-narrative}.      
These results highlight the importance of modeling discourse phenomena such as entity tracking for long context tasks.  

However, long context modeling brings computational challenges as well. 
In the context of coreference resolution, the current state-of-the-art model by~\citet{wu2019coreference}, which frames coreference resolution as a QA task, scales quadratically in runtime with the length of the document. 
On the other hand, the popular mention-ranking paradigm, which requires keeping past mentions in memory, has high memory requirement~\citep{xia-etal-2020-incremental}. 
Thus, prior work in coreference resolution suffers from scalability issues, in terms of both memory and running time. 

We show that computationally efficient coreference resolution models can be developed by maintaining just  compressed entity representations rather than individual mention representations. 
Specifically, we show that entities can be represented with fixed-dimensional vector representations derived from pretrained language models.\footnote{Concurrent work  has shown the efficacy of similar fixed-dimensional representations for representing documents~\cite{karpukhin-etal-2020-dense}.}  
Further efficiency can be achieved by exploiting the transient nature of entities. Concretely, an entity tracking model can be trained to ``forget'' entities that are less relevant for future discourse and only keep around a small, bounded number of entities in its memory.\footnote{A similar idea of forgetting memories for long context LMs was proposed by \citet{Sukhbaatar2021NotAM} concurrently with our work.}

\subsection{Integrating Entity Tracking into Pretrained Language Models}

Given the ubiquity of the current paradigm of using contextualized representations from pretrained LMs for a variety of downstream tasks, integrating entity tracking capabilities into LMs presents an attractive proposition as it would allow for: (i) a wider application of entity tracking, and (ii) an easier adoption than a standalone entity tracking model.
Past work on incorporating explicit entity tracking as part of recurrent neural network (RNN) language model training has also shown benefits on a range of tasks~\citep{ji-etal-2017-dynamic, clark-etal-2018-neural}. 
One limitation of adding an external memory to a pretrained LM is that it changes the LM architecture, which in turn affects the ease of model adoption. 

\begin{figure}[t]
	
	\fbox{
		\begin{minipage}{0.92\textwidth}
			\emph{Yesterday I dropped my clothes off at the dry cleaner’s and I have yet to pick them up. Where are my clothes?} \textbf{I have a lot of clothes.}  
		\end{minipage}
	}
	\caption{Prompt (italicized) used by \href{https://www.technologyreview.com/2020/08/22/1007539/gpt3-openai-language-generator-artificial-intelligence-ai-opinion/}{Gary Marcus and Ernest Davis} to diagnose the entity tracking capability of GPT-3~\cite{brown2020language}. The model response (bold) evades answering the question.  
	}
	\label{fig:prompt_example}
\end{figure}

In the transformer era, \citet{ye-etal-2020-coreferential} demonstrated the success of incorporating a ``coreferential objective" in masked language model training.
However, the ``coreferential objective" uses just the simple distant supervision of exact matching mention spans, and it's not clear if the model learns the general coreference function.  
\citet{wu2021infusing} explore infusing predicate-argument structure in pretrained LMs and show improvements on natural language understanding tasks. However, as in \citet{ye-etal-2020-coreferential}, the resulting model still lacks interpretability i.e.\ it's not clear if the \emph{infused} model necessarily has a better understanding of semantics.

With the goal of interpretability and preserving the transformer LM architecture, we propose a data augmentation finetuning strategy where we train the LMs on entity state augmented training sequences. 
We first explore these ideas for \emph{simple} domains where we have access to the exact entity states, including our proposal to use chess as a testbed for entity state tracking.  
On these simple domains, we show that training LMs with state-augmented instances improves the language modeling  and state tracking performance and also allows probing for entity state at inference time simply via prompting. Note that diagnosing LMs via prompting without any finetuning can run into issues as shown in Figure~\ref{fig:prompt_example}. Finally, we adapt the state-augmented training for baking coreference knowledge into pretrained LMs and show improvements on a popular cloze task which requires entity tracking.

\section{Thesis Contributions}
\label{sec:thesis_contributions}

We make the following contributions in this thesis:
\begin{itemize}
    \item \textit{Chapter 3}: We propose a bounded memory model trained with sparse supervision for the pronoun resolution task. 
    The proposed model outperforms prior work on the pronoun resolution task and interpretabilty measures with fewer parameters and a simpler architecture. 
    \item \textit{Chapter 4}: We develop memory models for the coreference resolution task which can scale to long documents. In particular, we propose a bounded memory model which offers a linear runtime complexity in document length while being competitive with the state-of-the-art models. We establish a new state-of-the-art for LitBank~\cite{bamman2019annotated}, a long document coreference resolution task.
    \item \textit{Chapter 4}: We propose an evaluation suite consolidating eight popular coreference benchmarks to test generalization capability of coreference models via zero-shot evaluation. 
    We propose joint training and data augmentation strategies which aid the generalization performance. 
    We establish a new state-of-the-art for two coreference resolution benchmarks, namely PreCo~\cite{chen-etal-2018-preco} and WikiCoref~\cite{ghaddar-langlais-2016-wikicoref}.
    \item \textit{Chapter 5}: We propose the task of language modeling for the game of chess to evaluate the entity tracking capability of transformer language models. 
    We show that the use of appropriate chess notation allows for directly probing the entity state, without requiring any additional probing-related machinery. 
    Our results show that with enough training data, transformer LMs can learn to track pieces and predict legal moves with high accuracy. But for small training sets, providing access to board state information during training  yields significant improvement. 
    Finally, we have integrated the chess state tracking task into a popular LM benchmark~\cite{bigbench2022}.
    \item  \textit{Chapter 6}: We propose methods to integrate entity tracking capability into LMs by training them on text augmented with entity tracking related information represented as text tokens.  
    Training LMs in this way allows for entity state probing via prompting. 
    We first experiment in a closed domain with the assumption of access to the true entity states during training. 
    We show that integrating entity tracking into LMs improves both the state tracking performance and the text generation quality in this closed domain. 
    We extend these ideas to integrating coreference resolution into LMs, where we rely on model predictions rather than ground truth annotation for coreference structures. 
    Our results show that integrating coreference into LMs improves results on a popular cloze task.

\end{itemize}

\newpage
\chapter{Background}

In this chapter, 
we first discuss prior work for the coreference resolution task, a special class of the entity tracking problem. 
We then discuss prior work for the entity tracking problem at large. 
Next, we discuss prior work on analysis of NLP models with \emph{probing} to contextualize the interpretability aspect of our work on integrating entity tracking in language models.   
Finally, we discuss memory models and language models which form the backbone of all the models presented in this thesis.

\label{sec:preliminaries}

\section{Coreference Resolution}
\label{sec:coref_background}
In this section, we first discuss prior work in coreference resolution, and then the evaluation metrics used for the task.

\subsection{Prior Approaches}
\label{sec:coref_prior}
Coreference resolution has been one of the central problems in natural language understanding and  computational linguistics~\citep{Winograd1972Understanding, charniak72thesis}. Most of the earliest work used linguistically and cognitively motivated heuristics for the task~\citep{Hobbs1978ResolvingPR, brennan87centering, Carter1987InterpretingAI}. 
With the availability of large annotated resources, supervised learning, which is the focus of our work, has been the dominant paradigm for the past couple of decades~\citep{mccarthy95decision, ge-etal-1998-statistical, soon-etal-2001-machine, bengtson-roth-2008-understanding, durrett2013easy, wiseman-etal-2015-learning, lee-etal-2017-end, joshi-etal-2019-bert}. 

\begin{figure}
\resizebox{\textwidth}{!}{%
\large

\begin{tikzpicture}[node distance=1.3cm,
	start chain=going right,inner sep=0cm,outer sep=0cm,>=stealth']
	\Large
	\node[on chain,text width=7cm,](document) {
		\begin{tcolorbox}[enhanced,
			,colback=white]
			\centering{
				\texttt{Rafael Nadal is the champion at Roland-Garros for an unprecedented 13th time, his victory over world No.1 Novak Djokovic elevating him level with Roger Federer’s all-time mark of 20 major titles. The Spaniard delivered one of his finest performances against arguably his toughest rival to prevail 6-0, 6-2, 7-5.
	}}\end{tcolorbox}};

\node[on chain, punktchain, join, fill=white, text width=5cm] (encoder) {\huge{Pretrained\\[0.3em] Transformer}};

	\node[on chain, join, minimum height=10cm, inner sep=0cm, outer sep=0cm] (detector){
		
			\renewcommand{\arraystretch}{0.75}
			\setlength{\tabcolsep}{10pt}
			\begin{tabular}{l l l l d{3.2}}
				\texttt{Rafael} & \cellcolor[HTML]{FCA5A5}{} & \cellcolor[HTML]{407F7F}{}  & \cellcolor[HTML]{A5C663}{}  &   1.0\\\\
				\textcolor{ForestGreen}{\texttt{Rafael Nadal}} & \cellcolor[HTML]{45857E}{} & \cellcolor[HTML]{D6B26F}{}  & \cellcolor[HTML]{D16C73}{}  & 5.0 \\\\
				\multicolumn{1}{c}{\vdots} & \multicolumn{3}{c}{\vdots} & \vdots\\\\
				
				\texttt{is} & \cellcolor[HTML]{CB6672}{} & \cellcolor[HTML]{42847D}{}  & \cellcolor[HTML]{86A137}{}   & -10.0 \\\\
				
				\texttt{is the} & \cellcolor[HTML]{9173A9}{} & \cellcolor[HTML]{404040}{}  & \cellcolor[HTML]{D4936B}{}  & -15.0 \\\\
				
				\multicolumn{1}{c}{\vdots} & \multicolumn{3}{c}{\vdots} & \vdots\\\\
				
				\textcolor{ForestGreen}{\texttt{his}} & \cellcolor[HTML]{26547C}{} &
				\cellcolor[HTML]{D6B26F}{}   &
				\cellcolor[HTML]{EF476F}{}  &   3.0 \\\\
				
				\texttt{his toughest} & \cellcolor[HTML]{D4A56B}{} & \cellcolor[HTML]{505F8F}{}  & \cellcolor[HTML]{468C77}{}   &  -1.0 \\\\
				
				\textcolor{ForestGreen}{\texttt{his toughest rival}} & \cellcolor[HTML]{ca0020}{} & \cellcolor[HTML]{f4a582}{}  & \cellcolor[HTML]{404040}{}   &  2.0 \\\\
				
				\multicolumn{1}{c}{\vdots} & \multicolumn{3}{c}{\vdots} & \vdots\\
			\end{tabular}
	};

            \node[above=of detector, yshift=-5mm] (detector_title){
					
					\renewcommand{\arraystretch}{0.75}
					\setlength{\tabcolsep}{10pt}
					\begin{tabular}{l}
						\hspace{0.1in}\Huge{Mention Detector}\\\\
					\end{tabular}
					
				};

	\node[on chain, join=by {->,"\Large{top-$K$ mentions}", text width=3cm, align=center, above, midway}, xshift=20mm, right=of detector, minimum height=3cm, minimum width=3.2cm,  inner sep=0cm, outer sep=0cm] (clustering)
	{
			
			\begin{tikzpicture}[every node/.style={inner sep=0,outer sep=0}]
				\node[circle,
				draw=black,
				text=brown,
				fill=white, minimum size=6cm] (c) at (-0.5,2){};
				\node [font={\ttfamily\color{brown}}] (ment_1) at (-0.2cm, 2cm) {Rafael Nadal}; 
				\node [font={\ttfamily\color{brown}}] (ment_2) at (0.8cm, 3.5cm) {The Spaniard}; 
				\node [font={\ttfamily\color{brown}}] (ment_3) at (0cm, 0.5cm) {him}; 
				\node [font={\ttfamily\color{brown}}] (ment_4) at (0.5cm, 1cm) {his};
				\node [font={\ttfamily\color{brown}}] (ment_5) at (2.5cm, 1.5cm) {his};
				\node [font={\ttfamily\color{brown}}] (ment_6) at (2cm, 0.5cm) {his};

				\node[ellipse, draw=black, minimum width = 10.5cm, minimum height=2cm] (e) at (0.8,-2.1) {};
				
				\node [font={\ttfamily\color{blue}}] (ment_21) at (2.5cm, -2.5cm) {his toughest rival}; 
				\node [font={\ttfamily\color{blue}}] (ment_22) at (1.5cm, -2cm) {world No.\ 1 Noval Djokovic}; 
			\end{tikzpicture}

	};

		\node[above=of clustering, right=of detector_title, rectangle,  inner sep=5pt, rounded corners, color=black, very thick] (detector){
			
			\renewcommand{\arraystretch}{0.75}
			\setlength{\tabcolsep}{10pt}
			\begin{tabular}{l}
				\Huge{Mention Clustering}\\\\
			\end{tabular}
			
		};
	
\end{tikzpicture}
}
\caption{A typical end-to-end coreference pipeline. }
\label{fig:end_to_end_coref}
\end{figure}
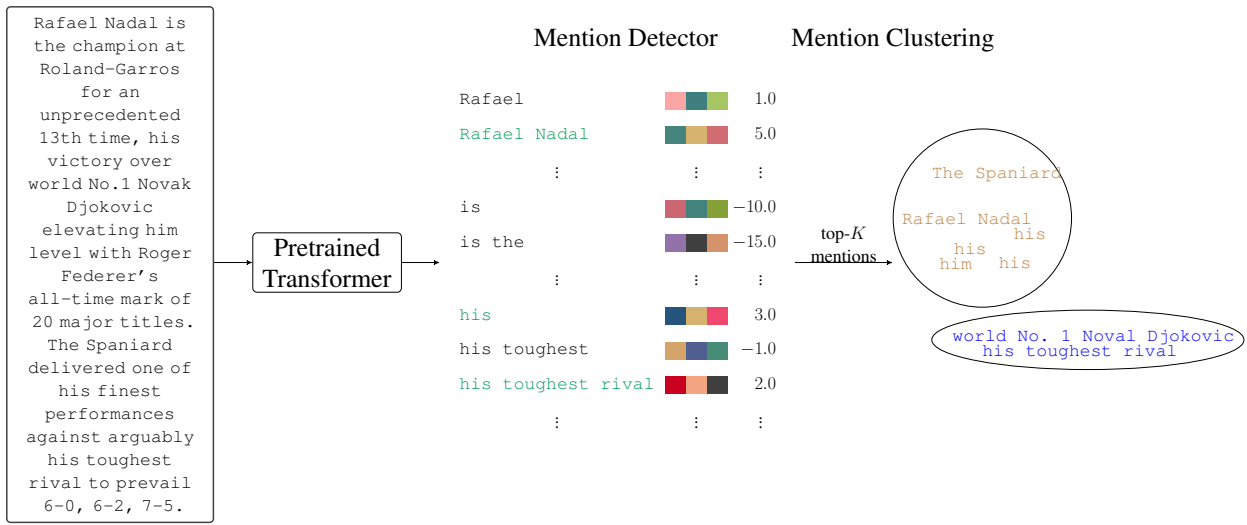

Current learning approaches for coreference resolution typically break down the task into two steps, namely \textbf{mention detection} and \textbf{mention clustering} (see Figure~\ref{fig:end_to_end_coref}). 
The mention detection step outputs a pruned set of candidate spans which serve as input to the mention clustering step which finally outputs the entity clusters. 
While earlier state of the art models used a pipeline approach i.e.\ separately learned the two steps~\citep{durrett2013easy,wiseman-etal-2015-learning},  \citet{lee-etal-2017-end} showed the benefits of jointly training the two, which remains a feature in current state of the art models. Since the major algorithmic difference among the approaches lies in the mention clustering step, and is relevant for situating our work, we will briefly overview the four popular clustering approaches (for a more detailed overview, see~\citet{rahman2011narrowing}). 

\begin{enumerate}
	
	\item \emph{Mention-Pair Models}: In this approach, a binary classifier is trained to predict if two candidate mentions are coreferent or not~\citep{mccarthy95decision}. While this approach has been very influential in the past~\citep{soon-etal-2001-machine, ng-cardie-2002-improving, bengtson-roth-2008-understanding}, it has some well-known drawbacks, the most important being the inability to enforce the transitivity property of clustering. 
	For example, even if the classifier predicts mentions $x_A$ and $x_B$ to be coreferent, mentions $x_B$ and $x_C$ to be coreferent, it can still predict mentions $x_A$ and $x_C$ to not be coreferent. The lack of transitivity means that a cluster-decoding algorithm is required by such approaches. Also, the number of binary predictions scales quadratically with the number of mentions which is again undesirable. 
	Finally, since the negative coreference pairs far outnumber the coreferent ones, and the positive pairs which are far apart can be quite hard, under-sampling the negative pairs can be key to performance~\citep{ng-cardie-2002-improving}.  
	
	\item \emph{Mention-Ranking Models}:
	Rather than making pairwise predictions for each mention, mention-ranking models rank \emph{antecedent} mentions for each mention~\citep{durrett2013easy, wiseman-etal-2015-learning, clark-manning-2016-deep, lee-etal-2017-end}. 	This avoids the transitivity issue of mention-pair models as clustering naturally follows from these ranking predictions by chaining together mentions based on their top antecedent picks. 
	Most of the recent state of the art models are based on this paradigm~\citep{joshi-etal-2020-spanbert, xu-choi-2020-revealing}.
	Without any heuristics, the mention-ranking model would keep around all the past mentions in memory, and thus, the runtime would scale quadratically in the number of mentions. Since the number of antecedent mentions can be quite large for long documents, in practice models often use heuristics such as capping the number of candidate antecedents~\citep{lee-etal-2017-end, lee-etal-2018-higher}.

	\item \emph{Entity-Mention Models}: The mention-based approaches use only mention-level features which may not be sufficient for accurate classification. The more expressive entity-level features can mitigate this bottleneck. The entity-mention models extend the mention-pair models by  training binary classifiers to predict the linking probability of a mention and a preceding partially formed entity cluster~\citep{luo-etal-2004-mention, yang-etal-2008-entity}. Mentions are merged with the entity-cluster with the highest linking probability.
	
	\item \emph{Entity/Cluster-Ranking Models}:
	While the mention-ranking models lack the entity level features used by entity-mention models, the entity-mention models inherit the drawbacks of pairwise predictions from mention-pair models (except for transitivity). 
	A fix to both of these problems is to incrementally build entity-clusters by  ranking incrementally built entity-clusters for merging a new mention~\citep{rahman2011narrowing, stoyanov-eisner-2012-easy, websterC14, clark-manning-2016-improving}. While entity-ranking models can, in theory, reduce the memory footprint by keeping around only cluster-level representation/features, typical entity-ranking approaches keep around the constituent mentions as well~\citep{rahman2011narrowing, stoyanov-eisner-2012-easy, clark-manning-2015-entity}. Finally, cluster-level features have also been used by mention-ranking approaches which use these  \emph{higher-order} (cluster-level) features along with mention-based features~\citep{wiseman-etal-2016-learning, lee-etal-2017-end, lee-etal-2018-higher, joshi-etal-2019-bert, joshi-etal-2020-spanbert}, though, recent work has questioned the utility of these higher-order features in current models~\citep{xu-choi-2020-revealing}.
	
\end{enumerate}

Some of the recent work in coreference resolution doesn't fit in the above categorization. The current state-of-the-art model by~\citet{wu2019coreference} frames coreference resolution as a question answering (QA) task. The model performs a QA query for each mention and the answer corresponds to all the coreferent spans. Since the number of mentions can be linear in document length, this model scales poorly with the length of document. 
\citet{paolini2021structured} model coreference resolution, and other structured prediction tasks, as a translation task. 
\citet{kirstain-etal-2021-coreference} and \citet{dobrovolskii-2021-word} propose methods which avoid an explicit mention detection step i.e.\ filtering of top candidate spans.

In this thesis, we propose memory models (Section~\ref{sec:memory_model}) which adopt the entity-ranking paradigm  (Chapters 3 and 4). The external memory in these models tracks the entities where entities are represented via a fixed-dimensional vector representation.

\subsection{Evaluation Metrics}
Evaluating the performance of an individual coreference link, which is just a binary classification problem, can be done by standard measures such as F-score. However, the full coreference resolution task, which we view as a clustering task, has no such clear evaluation. 
For example, it's not clear if adding/deleting a mention to/from a larger cluster should incur more penalty than the same action performed on a smaller cluster. Similarly,  how should penalty for a mention missed in a cluster compare with that of a spurious mention added to a cluster. 
In fact, the evaluation is even more challenging than a typical clustering task because the system output won't necessarily be perfect on mention detection, and thus, the system output and ground truth would differ even on the clustered elements.

Below we briefly discuss the three metrics, namely MUC~\citep{vilain-etal-1995-model}, B\textsuperscript{3}~\citep{Bagga98algorithmsfor}, and $\text{CEAF}_{\phi_4}$~\citep{luo-2005-coreference},  used in the coreference resolution literature. All three metrics define precision, recall, and F-score based on a quantity of interest. 
While these three metrics have known shortcomings~\citep{Bagga98algorithmsfor, luo-2005-coreference,  moosavi-strube-2016-coreference}, and the agreement between the them is low~\citep{holen-2013-critical}, we'll follow the proposal by \citet{denis09global}  of averaging the F-score of the three metrics, which is the popular metric of choice and is referred to as CoNLL F-score in the literature. We describe the three metrics next.

In the following discussion, assume $\mathcal{G}$ to represent the gold truth entity set and $\mathcal{S}$ to represent the predicted entity set. Each element in $\mathcal{G}$ and $\mathcal{S}$ represents an entity which is represented as a mention set. Finally, let $|x|$ denote the size of set $x$. 

\paragraph{MUC} is a link-based metric. It uses the number of links preserved for entity sets. MUC recall is defined as:
$$\textrm{Recall} = \frac{\sum_{g_i \in \mathcal{G}} (|g_i| - |p(g_i)|)}{\sum_{g_i \in \mathcal{G}} (|g_i| -1 )}$$
where $p(g_i)$ denotes the entity sets in $\mathcal{S}$ across which mentions of $g_i$ are present. Thus, the recall is lower, if entity $g_i$ is split across multiple predicted clusters in $\mathcal{S}$. 
MUC precision is computed by reversing the roles of $\mathcal{G}$ and $\mathcal{S}$. 
The MUC metric is known to prefer outputs with over-merged entities as they lead to higher recall ~\cite{luo-2005-coreference}.  

\paragraph{B\textsuperscript{3}} is a mention-based metric. For an entity $g \in \mathcal{G}$, it uses the fraction of mentions in $g$ present in predicted entities in $\mathcal{S}$. Formally, B\textsuperscript{3} recall is defined as: 

$$ \textrm{Recall} = \frac{\sum_{g_i \in \mathcal{G}} \sum_{s_j \in \mathcal{S}} \frac{|g_i \cap s_j|^2}{|g_i|}}{\sum_{g_i \in \mathcal{G}} |g_i|}$$

As in MUC, B\textsuperscript{3} precision is computed by reversing the roles of $\mathcal{G}$ and $\mathcal{S}$. Two known shortcomings of B\textsuperscript{3} are that:  (a)  the  B\textsuperscript{3} recall is 100\% if the system output merges all the gold mentions, and (b) the  B\textsuperscript{3} precision is 100\% if the system predicts every gold mention to be a singleton. 

\paragraph{$\text{CEAF}_{\phi_4}$} metric assumes that one gold entity should map to one system entity and vice-versa. It uses the similarity metric $\phi_4$ to calculate a one-to-one alignment between $\mathcal{G}$ and $\mathcal{S}$. The similarity for a gold entity $g_i$ and system entity $s_j$ is given by:
$$\phi_4(g_i, s_j) = \frac{2 \times |g_i \cap s_j|}{|g_i| + |s_j|}$$
Assuming $\mathcal{G}^*$ to be the set of gold entities included in the optimal entity mapping OPT, the $\text{CEAF}_{\phi_4}$ recall is given by:
$$ \textrm{Recall} = \frac{\sum_{g_i \in \mathcal{G}^*} \phi_4(g_i, \textrm{OPT}(g_i))}{\sum_{g_i \in \mathcal{G}} \phi_4(g_i, g_i)}$$
where OPT$(g_i) \in \mathcal{S}$ represents the optimal mapping of entity $g_i$ in $\mathcal{S}$. For calculating precision, the denominator is changed to  $\sum_{s_i \in \mathcal{S}} \phi_4(s_i, s_i)$.

\section{Entity Tracking}
Entity tracking refers to the broad umbrella of tasks concerning maintaining the entities and their attributes. Typically what constitutes an entity state is dataset-specific which means that the task is arguably less standardized than a typical linguistic task.

The usual approach to an entity tracking task is by training a supervised model on datasets annotated with entity states. The supervision can be in the form of question answers as in the bAbI tasks~\cite{weston2015aicomplete} and ProPara task~\cite{dalvi-etal-2018-tracking}, or in the form of classification task as in the Recipes task~\cite{kiddon-etal-2016-globally, bosselut-18}. 
Recent work by \citet{tandon-etal-2020-dataset} explores a generative task where given a sentence in the context of a procedural text, the task is to generate the state changes for all the entities involved. 
\citet{henaff2016tracking} proposed EntNet, a model which is equipped with an external memory to track entities mentioned in the discourse. 
\citet{bosselut-18} proposed Neural Process Networks (NPN), a memory model, where the model learns to simulate the action dynamics i.e.\ how the entity states change when an action is applied. The setup assumes that the entities mentioned in a discourse are given and the set of actions is known a priori while training. These simplifying assumptions severely limit the applicability of NPN beyond the Recipes task used in the work. \citet{gupta-durrett-2019-tracking} propose a structured architecture for tracking the observable discrete attributes, such as location, and the unobserved implicit entity state. Moving away from explicit entity-centric memory representations, \citet{gupta-durrett-2019-effective} explore the use of pretrained transformers for entity tracking in  the ProPara and Recipes datasets. While transformers outperform prior work, their error analysis shows that transformer-based models mostly rely on surface clues and don't form accurate intermediate entity representations.   

Entity tracking has also been explored for the cloze task LAMBADA~\cite{paperno-etal-2016-lambada}, where  the task is to predict the last word of a passage (\emph{cloze tasks} refers to tasks where the goal is to fill in the missing language item). The task instances were filtered such that succeeding on the task requires understanding the whole passage  instead of just the local context. 
\citet{chu-etal-2017-broad} conducted an manual analysis of the LAMBADA validation set which showed that roughly 20\% of the instances require coreference resolution. This analysis led to work, such as \citet{dhingra-etal-2018-neural} and \citet{hoang-etal-2018-entity}, where they utilize entity tracking related information in their models.  \citet{cheng2020entity} introduce a loss to promote attending to  coreferential mentions while processing mentions of the same coreference chain. 

There has been a rich line of work exploring entity tracking in tandem with language models. \citet{ji-etal-2017-dynamic} propose the EntityNLM language model which augments the LSTM network with an entity-centric external memory. \citet{clark-etal-2018-neural} extend the EntityNLM model for story generation. 
\citet{liu2019referential} propose the Referential Reader which is trained on both a sparsely annotated coreference resolution dataset and the language modeling task. Ablation studies for the Referential Reader show that the language modeling task aids entity tracking. 

There has been an ongoing debate on how much \emph{meaning} language models can learn given that they are just trained on text \emph{symbols}~\cite{bender-koller-2020-climbing, Bender2021OnTD}. The literature is divided with regards to entity tracking capabilities of pretrained language models.
\citet{li-etal-2021-implicit} demonstrate, via the use of probing classifiers (Section~\ref{sec:probing_classifier}), that pretrained transformer LMs implicitly learn to track entities. 
On the contrary, \citet{schuster-linzen-2022-sentence} show that even models at the scale of GPT-3 struggle with basic entity tracking capabilities, though scaling up does seem to aid these capabilities.
\citet{sorodoc-etal-2020-probing} show via probing analysis that transformer LMs lack a global notion of entities. \citet{tenney2019probing} use the ``edge probing'' framework (see Section~\ref{sec:probing_intro}) to diagnose linguistic capabilities of different pretrained LMs and find that models are better at capturing syntax than entity-centric information like coreference.

\section{Probing}
\label{sec:probing_intro}
The black box nature of neural models has given rise to a rich area of work on interpretability and analysis of neural NLP models. 
Interpretability is an important aspect of this thesis, especially for our work on integrating entity tracking into language models.
In this section, we discuss probing methods which refers to a broad umbrella of analysis techniques. 
For this discussion, we first discuss two prominent probing methods in the literature, namely \emph{diagnostic probing} and \emph{behavioral probing}. Finally, we discuss the emerging area of \emph{probing via prompting} which has the most relevance to this thesis.  

\subsection{Diagnostic Probing}
\label{sec:probing_classifier}
In diagnostic probing, a supervised probing classifier is trained on the internal representations of the model of interest to predict an external property. 
A high-performing probing classifier is assumed to imply that the external property is represented in the model and vice-versa. 
Typical design decisions in this probing mechanism involve the choice of: (a) probing classifier (linear, multilayer perceptron, etc.), (b) internal representation (which layer to use, how to reduce the representation to a fixed dimension), and (c) the amount of supervision used. 

Diagnostic probing has been one of the most popular probing methodologies in the literature, with an extensive use for probing for linguistic properties. 
\citet{ettinger-etal-2016-probing} and  ~\citet{adi17probing} use probing classifiers for analysis of sentence embeddings. \citet{belinkov-etal-2017-neural} analyze the representation of machine translation models for morphological tasks. \citet{liu-etal-2019-linguistic, tenney-etal-2019-bert} and ~\citet{tenney2019probing} use probing classifiers to diagnose the transformer layers of various pretrained LMs for various linguistic tasks.

Despite its popularity, the diagnostic probing methodology has some critical shortcomings. 
One of the most important shortcoming is what to attribute the performance of a probing classifier to? 
A high-performing probing classifier could be because the model representation indeed captures the property of interest, but it could also be because the probing classifier has learned the probing task.  
\citet{hewitt-liang-2019-designing} argue for probes which are more ``selective'' i.e. ones which are high-performing for the task of interest but have a low performance for a control task constructed by randomizing  input-output pairs from the task of interest.  
In contrast, \citet{pimentel-etal-2020-information} argue on information-theoretic principles that the best performing probing model, even if it's more complex, should be used.  
Another shortcoming of probing classifiers is that while they can reveal what information is present in the model's representation, they don't say anything about whether the model is using this information in actual downstream tasks. 
Ongoing work is trying to address these shortcomings. 
For a detailed overview of probing classifiers, we refer the reader to the recently published survey by \citet{belinkov2022probing}.

\subsection{Behavioral Probing}
 \nocite{goldberg2019assessing}  
In behavioral probing, analysis is carried out typically by observing the model's predictions i.e.\ behavior. 
To this end, evaluation sets are created to tease apart the model's performance. 
The evaluation datasets can be hand-crafted~\cite{ettinger-2020-bert}, automatically created~\cite{gulordava-etal-2018-colorless}, or obtained by filtering from a naturally occurring corpus~\cite{linzen-etal-2016-assessing}. 
In \citet{linzen-etal-2016-assessing}, the authors study the subject-verb agreement for a LSTM language model for sentences filtered from Wikipedia. 
\citet{marvin-linzen-2018-targeted} construct minimally different sentence pairs, consisting of an ungrammatical and grammatical sentence, to evaluate the grammatical preference of language models. \citet{ettinger-2020-bert} propose and perform a suite of psycholinguistic evaluations for diagnosing BERT, with one of the surprising findings being BERT's insensitivity to negation~\cite{Fischler1983BrainPR}.
 
While behavioral probing avoids introducing the confound of a probing classifier, as in diagnostic probing, it is limited by what can be probed by observing the model's behavior. Moreover, in the case of hand-curated datasets, constructing the dataset can be costly, which often limits the size of these tests to merely hundreds of examples~\cite{Fischler1983BrainPR, chow2015bag}. Finally, because the output space of model predictions may not be constrained, models can make reasonable predictions that are not covered by the hand-curated datasets and thus will be wrongly penalized by evaluations~\cite{lialin-etal-2022-life}.

\subsection{Probing via Prompting}
Prompting has emerged as a new modeling methodology which employs pretrained generative LMs~\cite{brown2020language, Liu2021PretrainPA}. 
Recent work has started exploring prompting for probing as well~\cite{toshniwal-etal-2022-chess, li-etal-2022-probing-via}. 
In this methodology, the probing task of interest is verbalized i.e.\ converted to text tokens, and trained along with the task of language modeling. 
The formulation assumes: (a) the probing task can be verbalized, and (b) the language model can be finetuned (which can be computationally infeasible for some of the big language models like GPT-3).  
Compared to diagnostic probing, a benefit of this formulation is that it avoids the hassle of selecting a probing classifier. And since the LM is finetuned for the probing task, the LM outputs tend to be more constrained in comparison to behavioral probing. 
We use the probing via prompting framework to diagnose the entity tracking capability of transformer LMs in Chapters 5 and 6 of this thesis.

\section{Memory Models}
\label{sec:memory_model}
Memory models refer to end-to-end differentiable neural networks coupled with an external memory~\cite{graves2014neural,graves2016hybrid}. 
The external memory can be thought of like a Turing tape~\cite{turing1950mind}, or when viewed through the lens of cognitive psychology, as \emph{working memory}~\cite{baddeley1986} (see \citet{Nematzadeh2020ONMI} for a discussion on the different memory types). 
The decoupling of the model parameters from the external memory means that a memory model can have an arbitrarily large memory without increasing the model parameters.\footnote{There are variants of memory models where the external memory is parameterized.}  
This differentiates memory models from recurrent neural networks such as  Long Short-Term Memory (LSTM)~\cite{hochreiter1997long} where the parameters grow quadratically with the memory size (memory cell size). 

\begin{figure}[t]
    \centering
    \includegraphics[scale=0.35]{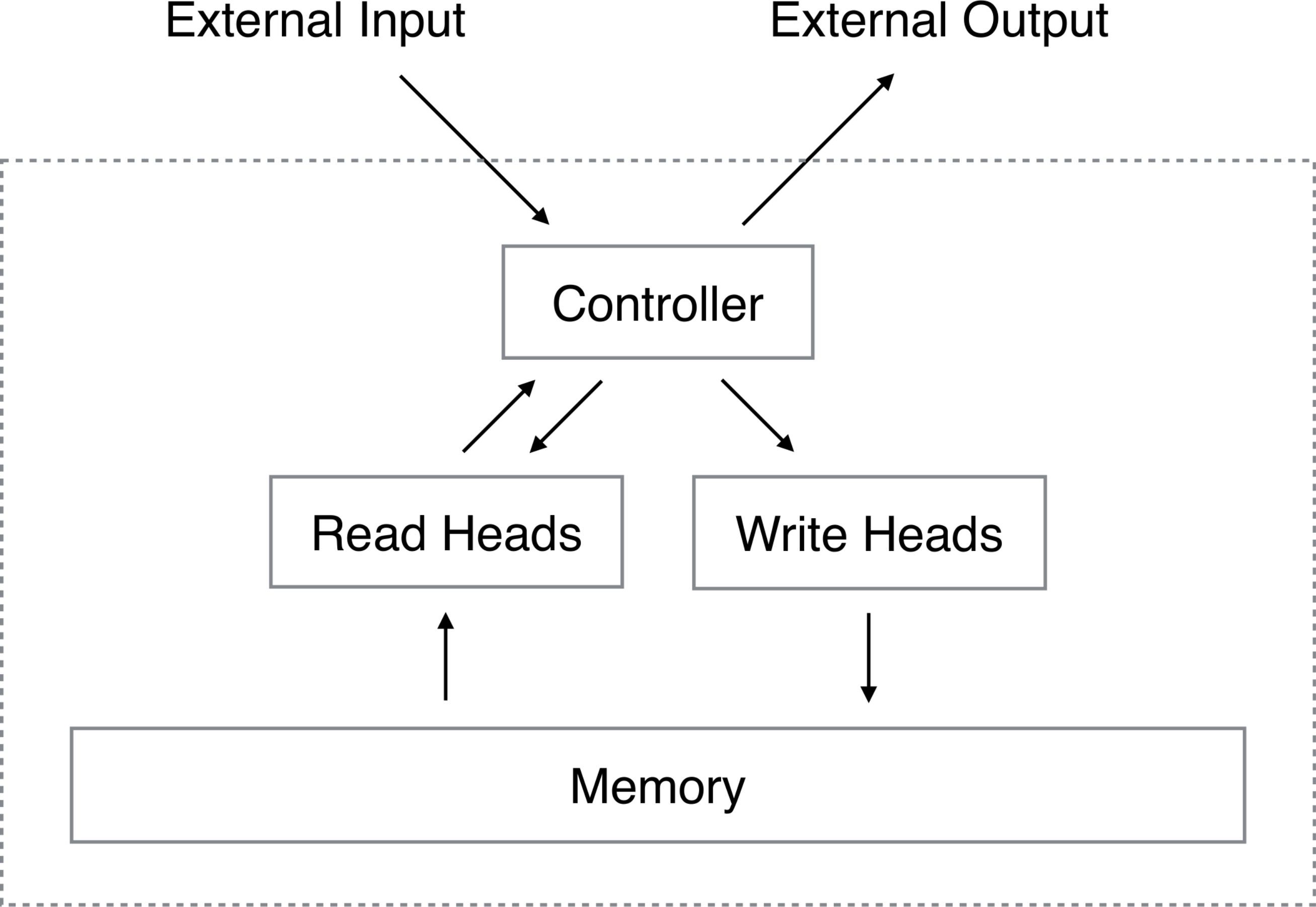}
    \caption{Schematic figure of the Neural Turing Machine. Source \citet{graves2014neural}.}
    \label{fig:ntm}
\end{figure}

Neural Turing Machine (NTM), shown in Figure~\ref{fig:ntm}, represents one of the first memory models proposed in the literature~\cite{graves2014neural}. 
The NTM model proposed two key abstractions, namely controller and memory. 
The controller receives an external input, such as the next word or the next sentence. Conditioned on this external input, the controller reads the memory state,  then combines the external input and the read memory to produce: (a) a write update for the memory and (b) an external output. 
While the exact implementation details of the model have been modified in follow-up work, the abstractions presented in the NTM work have largely persisted. We discuss the memory model components and design choices next.  

\subsection{Memory Model Details}
In this section, we discuss the key memory model components and associated design choices. 

\paragraph{Memory} There are various design choices associated with memory, such as the memory structure, the memory size, the memory initialization, etc. For a typical memory model, the memory can represented via a matrix $\vec{M} \in \mathbb{R}^{N \times d}$ where $N$ represents the number of memory cells and $d$ represents the size of each memory cell. 
The memory size $N$ can be unbounded~\cite{weston2014memory, graves2014neural} or bounded~\cite{sukhbaatar-15, henaff2016tracking}. 
Each memory cell can simply be a $d$-dimensional vector~\cite{weston2014memory, sukhbaatar-15} or have a key-value structure where the $d$-dimensional vector is split into two fixed-dimensional parts, namely key and value~\cite{miller-etal-2016-key, henaff2016tracking, liu2019referential}. 
At inference time, the memory cells could be initialized to a fixed/random vector~\cite{graves2014neural, sukhbaatar-15, graves2016hybrid} or the initialization could be learned~\cite{henaff2016tracking, liu2019referential}.

\paragraph{Memory Reading and Writing}
Given the input $\vec{x}_t$ at time step $t$, the controller uses read head(s) to attend to the memory $\vec{M}_{t} \in \mathbb{R}^{N \times d}$. Typically the memory models use content-based addressing where the read head(s) use a  similarity measure between the memory units and the input to retrieve  memories ``similar'' to an input. Models such as Differentiable Neural Controller (DNC)~\cite{graves2016hybrid} also support spatial and temporal memory access. For the content-based reading mechanism, a typical implementation does the following:
\begin{align*}
    \alpha_{it} &= \textrm{sim}(\vec{m}_{it}, \vec{x}_t)\\
    w_{it} &= \frac{\textrm{exp}(\alpha_{it})}{\sum_{j=1}^{N}\textrm{exp} (\alpha_{jt})}\\
    \vec{r}_t &= \sum_{i=1}^N w_{it} \cdot \vec{m}_{it}
\end{align*}
where $\vec{m}_{it}$ represents the $i$th memory cell at timestep $t$ and the function $\textrm{sim}(\cdot, \cdot)$ represents a similarity function. This \emph{soft} reading mechanism has been commonly used in prior work, including NTM and DNC. A simple tweak to the \emph{soft} reading mechanism is to use the top-$k$ scoring memory units~\cite{weston2014memory}.  
To support scalable memory reads, prior work has also proposed sparse read mechanisms based on approximate nearest neighbor search~\cite{weston2014memory, rae2016scaling}.

Memory write operations are required to store information which can be retrieved in later read operations. The memory update function has the form: 
$$\vec{M}_{t+1} = f(\vec{x}_t, \vec{r}_t, \vec{M}_{t})$$
The write operation, similar to the read operation, could have a \emph{soft} update where all the memories are overwritten~\cite{graves2014neural, graves2016hybrid, liu2019referential, henaff2016tracking} or a \emph{hard} update where a select few memory units are updated~\cite{rae2016scaling}. The criteria for selecting which memory unit to write to could be based on content-based similarity or based on the least recently used memory units~\cite{rae2016scaling}. If there are no memory size constraints, another simple write operation is to keep appending to the memory for every input~\cite{weston2014memory}.

\subsection{Application of Memory Models in NLP}

Some of the earliest applications of memory networks in NLP were for question answering, where the external memory simply stored all of the word/sentence embeddings for a document~\cite{weston2014memory, sukhbaatar-15, kumar2016ask, miller-etal-2016-key, henaff2016tracking}. 
Since then, memory models have been applied to several other NLP tasks in addition to question answering, including aspect-based sentiment analysis~\citep{liu-etal-2018-recurrent}, machine translation~\citep{maruf-haffari-2018-document}, narrative modeling~\citep{liu-etal-2018-narrative}, coreference resolution~\cite{liu2019referential}, and dialog state tracking~\citep{perez-liu-2017-dialog}. 
Story generation models have augmented language models with external memory which is meant to track the evolving entity state~\cite{clark-etal-2018-neural, rashkin-etal-2020-plotmachines}. 

In this thesis we explore memory models for the coreference resolution task in Chapters 3 and 4. 
The memory cells in these models are used for tracking entity cluster representations. 
In Chapter 3 we propose a \emph{soft} read and \emph{soft} write memory model. 
We show that learning memory initialization in our setup leads to worse performance and more model parameters.   
In Chapter 4 we propose a \emph{hard} read and \emph{hard} write memory model. In Chapter 3 the proposed memory model only has bounded memory while in Chapter 4 we explore both unbounded and bounded memory models.

\section{Pretrained Language Models}
In this section, we briefly overview pretrained language models. Pretrained LMs have lately become ubiquitous in NLP pipelines and are extensively used in this thesis as well. 
These LMs are trained on massive corpora, such as C4~\cite{raffel-etal-2020-exploring}, with self-supervised losses, such as masked LM loss~\cite{devlin-etal-2019-bert, joshi-etal-2020-spanbert}, autoregressive LM loss~\cite{radford2019language}, next sentence prediction loss~\cite{devlin2019bert}, sentence permutation loss~\cite{lewis-etal-2020-bart}. 
We limit our discussion to the model architecture of different pretrained LMs since that has the most relevance for this thesis. For a detailed discussion on pretrained LMs, we refer the reader to \citet{qiu2020ptmsurvey}.

\paragraph{Model Architecture} 

The pretrained LM paradigm has largely coincided with the emergence of the Transformer~\cite{vaswani2017attention} as the \emph{prima} network in NLP. Hence, since some of the earlier pretrained LMs, such as ELMo~\cite{peters-etal-2018-deep} and CoVe~\cite{mccann2017cove}, which used LSTM~\cite{hochreiter1997long}, almost all the popular pretrained LMs are based on the transformer network. The three main transformer-based LM architectures proposed by prior work are:
\begin{itemize}
    \item \emph{Encoder-only}: 
    This class of LMs includes the BERT model~\cite{devlin-etal-2019-bert} and its follow-ups~\citep[\emph{inter alia}]{joshi-etal-2019-bert, liu-etal-2019-roberta}. The LMs in this category are used as feature extractors for NLP tasks like question answering~\cite{karpukhin-etal-2020-dense}, text classification~\cite{sun2019fine}, coreference resolution~\cite{joshi-etal-2019-bert}. 
    \item \emph{Decoder-only}: The most popular class of LMs from this category is the GPT family of models~\cite{radford2018improving, radford2019language, brown2020language}. The decoder-only LMs have previously been used for text generation tasks, such as story generation~\cite{rashkin-etal-2020-plotmachines}, multiple choice question answering~\cite{radford2018improving}. Recently an interesting avenue of prompt-based learning has emerged with GPT-3 scale models~\cite{lester-etal-2021-power, li-liang-2021-prefix}.
    \item \emph{Encoder-Decoder}: The encoder-decoder LMs generalize the encoder-only and decoder-only architectures. Some of the popular models in this family include BART~\cite{lewis-etal-2020-bart} and T5~\cite{raffel-etal-2020-exploring}. These models are commonly used for conditional generation tasks, such as summarization, machine translation, etc.  
\end{itemize}

In this thesis we use: (a) the BERT model~\cite{devlin-etal-2019-bert} in Chapter 3, (b) the Longformer model~\cite{beltagy2020longformer} in Chapter 4, (c) the GPT-2 model~\cite{radford2019language} in Chapter 5 and 6 (in Chapter 5 we only use the architecture of GPT-2), and (d) the BART model~\cite{lewis-etal-2020-bart} in Chapter 6. In most of the cases, we have finetuned the pretrained models. We have released the trained LMs from Chapter 4 and 5 on Hugging Face model hub.\footnote{\url{https://huggingface.co/shtoshni}}

\chapter{PeTra: A Sparsely Supervised Memory Model for People Tracking}
\epigraph{
	\footnotesize{
		Consider a device designed to read a text in some natural language, interpret it, and store the content in some manner, say, for the purpose of being able to answer questions about it. To accomplish this task, the machine will have to fulfill at least the following basic requirement. It has to be able to build a file that consists of records of all the individuals, that is, events, objects, etc., mentioned in the text and for each individual record whatever is said about it.
}}{\textit{Lauri Karttunen, 1976}}

\begin{figure*}[ht]
    \centering
     \includegraphics[width=\textwidth]{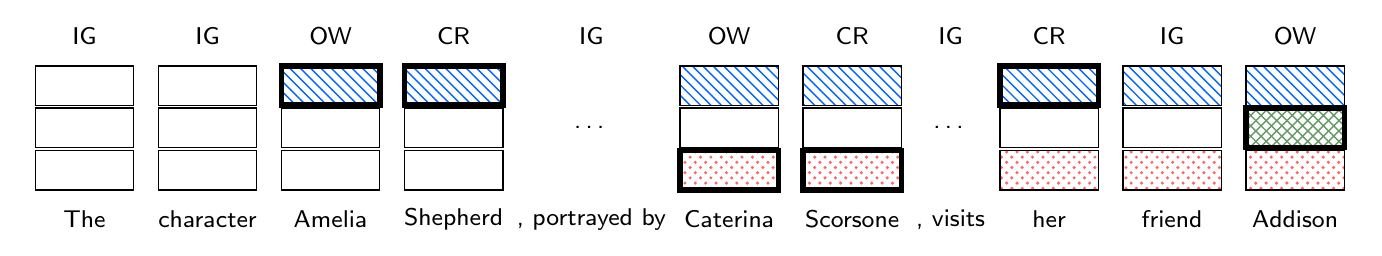}
     \captionof{figure}{Illustration of memory cell updates in an example sentence where IG = ignore, OW = overwrite, CR = coref. Different patterns indicate the different entities, and an empty pattern indicates that the cell has not been used. The updated memory cells at each time step are highlighted.
     }
     \label{fig:ideal_run}
\end{figure*}

\section{Introduction}
In this chapter, we introduce PeTra, a bounded memory model, which is trained with the sparse supervision of the GAP (Gendered Ambiguous Pronouns) pronoun resolution task~\citep{webster2018gap}. 
Since we train the model on the pronoun resolution task, the task of entity tracking is reduced to just people tracking which is where Pe(ople)Tra(cking) derives its name from. 
PeTra is inspired by the Referential Reader model from \citet{liu2019referential} but is substantially simpler. In terms of evaluation, apart from the end-task evaluation of pronoun resolution, we also investigate the interpretability and generalization capabilities of memory models by testing if they are indeed tracking people in their memories. For this end, we: (a) propose a new diagnostic evaluation based on counting the number of unique entities in text, and (b) conduct a small scale human evaluation to compare evidence of people tracking in the memory logs of PeTra relative to the Referential Reader. %

Despite using  a simpler architecture, PeTra outperforms the Referential Reader on the end task of pronoun resolution.  
Importantly, while Referential Reader performance degrades with larger memory, PeTra improves with increase in memory capacity (before saturation).
Moreover,  we find PeTra to be highly effective in the interpretability evaluations, especially the human evaluation where annotators overwhelmingly favor PeTra in comparison to the Referential Reader.\footnote{The material in this section is adapted from \citet{toshniwal2020petra}.}
\footnote{Code available at \url{https://github.com/shtoshni/petra}.}

\section{Model}

\begin{figure}[th]
	\centering
	\includegraphics[width=0.4\textwidth]{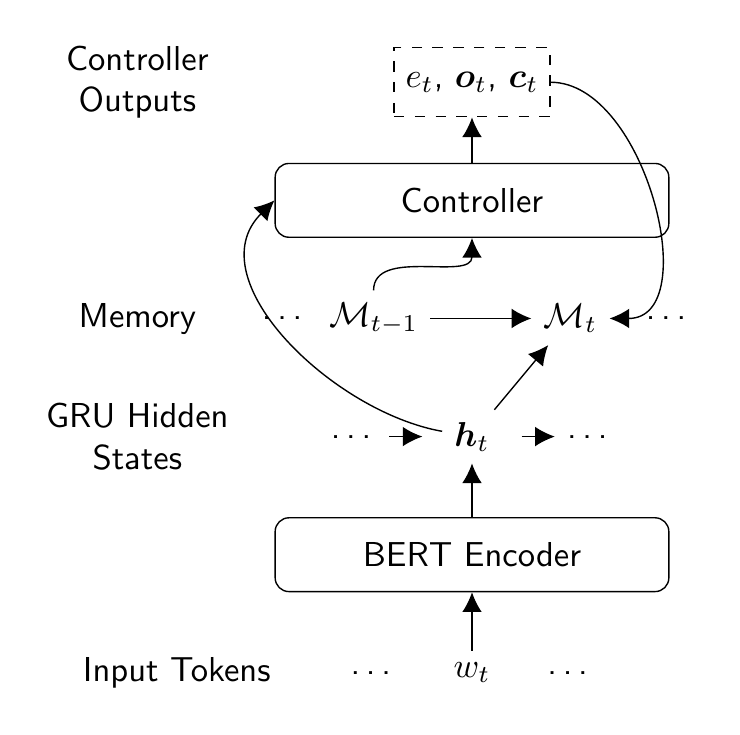}
	\captionof{figure}{Proposed model.}
	\label{fig:model_sch}
\end{figure}

Figure~\ref{fig:model_sch} depicts PeTra, which
consists of three components: an {\it input encoder} that given the tokens generates the token embeddings, a {\it memory module} that tracks information about the entities present in the text, and a {\it controller network} that acts as an interface between the encoder and the memory.

\subsection{Input Encoder}
\label{sec:input_encoder}
Given a document consisting of a sequence of tokens $\{w_1, \cdots, w_T\}$, we first pass the document through a fixed pretrained BERT model~\cite{devlin2019bert} to extract contextual token embeddings.
Next, the BERT-based token embeddings are fed into a single-layer unidirectional Gated Recurrent Unit (GRU) ~\cite{cho2014learning} running left-to-right to get task-specific token embeddings $\{\vec{h}_1, \cdots, \vec{h}_T\}$.

\subsection{Memory}
The memory $\mathcal{M}_t$ consists of $N$ memory cells. The $i^{\textrm{th}}$ memory cell state at time step $t$ consists of a tuple $(\vec{m}^i_t, u^i_t)$
where the vector $\vec{m}^i_t$ represents the content of the memory cell, and the scalar $u^i_t \in [0, 1]$ represents
its recency of usage. %
A high value of $u^i_t$ is intended to mean that the cell is tracking an entity that has been recently mentioned. %

\paragraph{Initialization}
Memory cells are initialized to the null tuple, i.e.\ (\vec{0}, 0); thus, our memory is parameter-free. This is in contrast with previous %
entity tracking models such as EntNet~\cite{henaff2016tracking} and the Referential Reader~\cite{liu2019referential} where memory initialization is learned and the cells are represented with separate {\it key} and {\it value} vectors.
We will later discuss variants of our memory with some of these changes.

\subsection{Controller}
At each time step $t$ the controller network determines whether token $t$ is part of an entity span and, if so, whether the token is coreferent with any of the entities already being tracked by the memory. Depending on these two
variables, there are three possible actions: %
\begin{enumerate}[label=(\roman*)]
	\itemsep0em
	\item \actignore: The token is not part of any entity span, in which case we simply ignore it.
	\item \actoverwrite: The token is part of an entity span \emph{but} is not already being tracked in the memory.
	\item \actcoref: The token is part of an entity span and the entity is being tracked in the memory.
\end{enumerate}
Therefore, the two ways of updating the memory are \actoverwrite and \actcoref.
There is a strict ordering constraint to the two operations: \actoverwrite precedes \actcoref, because it is not possible to corefer with a memory cell that is not yet tracking anything. %
That is, the \actcoref operation cannot be applied to a previously unwritten memory cell, i.e.~one with $u^i_t = 0$. Figure~\ref{fig:ideal_run} illustrates an idealized version of this process.

Next we describe in detail the computation of the probabilities of the two operations for each memory cell at each time step $t$.

First, the {\bf entity mention probability} $e_t$, which reflects the probability that the current token $w_t$ is part of an entity mention, is computed by:
\begin{equation} \label{ent_eqn}
	e_t = \sigma(\mlp_1(\vec{h}_t))
\end{equation}
where $\mlp_1$ is a multi-layer perceptron and $\sigma$ is the logistic function.

\paragraph{Overwrite and Coref}
If the current token $w_t$ is part of an entity mention, we need to determine whether it corresponds to an entity being currently tracked by the memory or not.
For this we compute the similarity between the token embedding $\vec{h}_t$ and the contents of the memory cells currently tracking entities.
For the $i^\textrm{th}$ memory cell with memory vector $\vec{m}_{t-1}^i$ the similarity with $\vec{h}_t$ is given by:%
\begin{equation} \label{sim_eqn}
	\mysim_{t}^{i} = \mlp_2([\vec{h}_t; \vec{m}_{t-1}^{i};
	\vec{h}_t \odot \vec{m}_{t-1}^{i}; u_{t-1}^i])
\end{equation}
where $\mlp_2$ is a second MLP and $\odot$ is the Hadamard (elementwise) product.
The usage scalar $u_{t-1}^i$ in the above expression provides a notion of distance between the last mention of the entity in cell $i$ and the potential current mention.
The higher the value of $u_{t-1}^i$, the more likely
there was a recent mention of the entity being tracked by the cell.
Thus %
$u_{t-1}^i$ provides an alternative to distance-based features commonly used in pairwise scores for spans
\citep{lee-etal-2017-end}.

Given the entity mention probability $e_t$ and similarity score $\mysim_{t}^{i}$, we define the \textbf{coref score} $\cs_{t}^{i}$ as:
\begin{equation} \label{coref_score_eqn}
	\cs_{t}^{i} = \mysim_{t}^{i} - \infty \cdot \mathds{1} [u^i_{t-1} = 0]
\end{equation}
where the second term ensures that the model does not predict coreference with a memory cell that has not been previously used, something not enforced by \citet{liu2019referential}.\footnote{A threshold higher than 0 can also be used to limit coreference to only more recent mentions.} Assuming the coref score for a new entity to be 0,\footnote{The new entity coref score is a free variable that can be assigned any value, since only the relative value %
	matters.} we compute the \textbf{coref probability} $c_{t}^{i}$ and \textbf{new entity probability} $n_t$ as follows:
\begin{equation}
	\label{coref_over_eqn}
	\begin{pmatrix}
		c_t^1 \\
		\vdots \\
		c_t^N \\
		n_t
	\end{pmatrix} = e_t \cdot \text{ softmax}
	\begin{pmatrix}
		\cs_{t}^{1} \\
		\vdots \\
		\cs_{t}^{N} \\
		0
	\end{pmatrix}
\end{equation}
Based on the memory usage scalars $u^i_t$ and the new entity probability $n_t$, the \textbf{overwrite probability} for each memory cell is determined as follows:
\begin{equation}\label{over_eqn_inf}
	o_t^{i} = n_t \cdot \mathbbm{1}_{i = \arg\min_j u^j_{t-1}}
\end{equation}
Thus we pick the cell with the lowest usage scalar $u^j_{t-1}$ to \actoverwrite. In case of a tie, a cell is picked randomly among the ones with the lowest usage scalar.
The above operation is non-differentiable,
so during training we instead use %
\begin{equation}\label{over_eqn_train}
	o_t^{i} = n_t \cdot \text{GS}\left(\frac{1 - u_{t-1}^{i}}{\tau}\right)_i
\end{equation}
where $\text{GS}(.)$ refers to Gumbel-Softmax~\cite{jang2017categorical}, which makes overwrites differentiable. %

For each memory cell, the memory vector is updated based on the three possibilities of ignoring the current token, being coreferent with the token, or considering the token to represent a new entity (causing an overwrite):
\begin{equation}
	\begin{aligned}\label{memory_up}
		\vec{m}_t^{i}  = &\ \overbrace{(1 - (o_t^{i} + c_t^{i})) \vec{m}_{t-1}^{i}}^{\mbox{\actignore}}  \,+\!\!\!\!\!\! \overbrace{o_t^{i}\cdot \vec{h}_t}^{\mbox{\actoverwrite}} \\[3pt]
		& +\, \underbrace{c_t^{i} \cdot \mlp_3([\vec{h}_t; \vec{m}_{t-1}^i])}_{\mbox{\actcoref}}
	\end{aligned}
\end{equation}
In this expression, the coreference term takes into account both the previous cell vector $\vec{m}_{t-1}^{i}$ and the current token representation $\vec{h}_t$, while the overwrite term is based only on $\vec{h}_t$.  In contrast to a similar
memory update equation in the Referential Reader which employs a pair of GRUs and MLPs for each memory cell, our update parameter uses just $\mathrm{MLP}_3$ which is memory cell-agnostic. 

Finally, the memory usage scalar is
updated as
\begin{equation}
	u_t^{i} = \min(1, o_t^{i} + c_t^{i} + \gamma \cdot u_{t-1}^{i})
\end{equation}
where $\gamma \in (0, 1)$ is the decay rate for the usage scalar.
Thus the usage scalar $u_t^{i}$ keeps decaying with time unless the memory is updated via \actoverwrite or \actcoref in which case the value is increased to reflect the memory cell's recent use.

\paragraph{Memory Variants}
In vanilla PeTra, each memory cell is represented as a single vector and the memory is parameter-free,
so the total number of model parameters is independent of memory size.  This is a property that is shared with, for example, %
differentiable neural computers \citep{graves2016hybrid}.
On the other hand, recent models for entity tracking, such as the EntNet~\cite{henaff2016tracking} and the %
Referential Reader~\cite{liu2019referential}, learn memory initialization parameters and separate the memory cell into key-value pairs.
To compare these memory cell architectures, we investigate the following two variants of PeTra:
\begin{enumerate}
	\item \emph{PeTra + Learned Initialization}: memory cells are initialized at $t=0$ to learned parameter vectors.
	\item \emph{PeTra + Fixed Key}: a fixed dimensions of each memory cell are initialized with  learned parameters and kept fixed throughout the document read, as in EntNet~\cite{henaff2016tracking}.
\end{enumerate}
Apart from initialization, %
the initial cell vectors are also used to break ties for overwrites in Eqs.~\eqref{over_eqn_inf} and \eqref{over_eqn_train} when deciding among unused cells (with $u^i_t = 0$). The criterion for breaking the tie is the similarity score computed using Eq.~\eqref{sim_eqn}.

\subsection{Coreference Link Probability}
\label{sec:coref_link_prob}
The probability that the tokens $w_{t_1}$ and $w_{t_2}$ are coreferential according to, say, cell $i$ of the memory depends on three things:
\begin{enumerate*}[label=(\alph*)]
	\item $w_{t_1}$ is identified as part of an entity mention and is either overwritten to cell $i$ or is part of an earlier coreference chain for an entity tracked by cell $i$,
	\item Cell $i$ is not overwritten by any other entity mention from $t = {t_1} + 1$ to $t = {t_2}$, and
	\item $w_{t_2}$ is also predicted to be part of an entity mention and  is coreferential with cell $i$.
\end{enumerate*}
Combining these %
factors and marginalizing over the cell index results in the
following expression for the {\bf coreference link probability}: %
\begin{align}\label{prob_eqn}
	P_{\mathrm{CL}}&(w_{t_1}, w_{t_2})  \nonumber\\
	&= \sum_{i=1}^{N} (o_{t_1}^{i} + c_{t_1}^{i}) \cdot \! \prod_{j=t_1 + 1}^{t_2} (1 - o_{j} ^ {i}) \cdot c_{t_2}^{i}
\end{align}

\subsection{Losses}
The GAP~\cite{webster2018gap} training dataset is small and provides sparse supervision with labels for only two coreference links per instance.
In order to compensate for this lack of supervision, we use a heuristic loss $\mathcal{L}_{\mathit{ent}}$ over entity mention probabilities  %
in combination with the end task loss $\mathcal{L}_{\mathit{coref}}$ for coreference. The two losses are combined with a tunable hyperparameter $\lambda$ resulting in the following total loss: $\mathcal{L} = \mathcal{L}_{\mathit{coref}} + \lambda \mathcal{L}_{\mathit{ent}}$. 

\subsubsection{Coreference Loss}
\label{sec:coref_loss}
The coreference loss is the binary cross entropy between the ground truth labels for mention pairs and the coreference link probability $P_{\mathrm{CL}}$ in 
Eq.~\eqref{prob_eqn}. 
Eq.~\eqref{prob_eqn} expects 
a pair of tokens while the annotations are on pairs of spans, %
so we compute the loss for all ground truth token pairs: $\mathcal{L}_{\mathit{coref}} =$
\begin{align*}
	\sum_{(s_a, s_b, y_{ab}) \in {\mathrm{G}}} \left(\sum_{w_a \in s_a} \sum_{w_b \in s_b} H(y_{ab}, P_{\mathrm{CL}}(w_a, w_b))\right)
\end{align*}
where $\mathrm{G}$ is the set of annotated span pairs and $H(p, q)$ represents the cross entropy of the distribution $q$ relative to distribution $p$.

Apart from the ground truth labels, we use 
``implied labels" %
in the coreference loss calculation.
For handling multi-token spans, we assume that all tokens following the head token are coreferential %
with the head token (self-links). 
We infer more supervision based on knowledge %
of the %
setup of the GAP task. Each GAP instance has two candidate names and a pronoun mention with supervision provided for the \{name, pronoun\} pairs. By design the two names are different, and therefore we use them as a negative coreference pair.

Even after the addition of this implied %
supervision, 
our coreference loss calculation is restricted to the three mention spans in each training instance; therefore, the running time is $\mathcal{O}(T)$ for finite-sized mention spans. 
In contrast, \citet{liu2019referential} compute the above coreference loss for all token pairs (assuming a negative label for all %
pairs outside of the mentions), which results in a runtime of $\mathcal{O}(T^3)$ due to the   $\mathcal{O}(T^2)$ pairs and $\mathcal{O}(T)$ computation per pair, and thus will scale poorly to long documents.

\subsubsection{Entity Mention Loss}
\label{sec:ent_pred_loss}
We use the inductive bias that most tokens do not correspond to entities by imposing a loss on the average of the entity mention probabilities predicted across time steps, after masking out the labeled entity spans. 
For a %
training instance where spans $s_A$ and $s_B$ correspond to the %
person mentions and span $s_P$ is a pronoun, the entity mention loss is %
\vspace{-0.05in}
$$\mathcal{L}_{\mathit{ent}} = \frac{\sum_{t=1}^T e_t \cdot m_t}{\sum_{t=1}^T m_t}$$ 
where $m_t = 0$ if $w_t \in s_A \cup s_B \cup s_P$ and $m_t = 1$ otherwise. 

Each GAP instance has only 3 labeled entity mention spans, but the text typically has other entity mentions that are not labeled. 
Unlabeled entity mentions will be 
inhibited by this loss. However, on average there are far more tokens outside entity spans than inside the spans. 
In experiments without this loss, we observed that the model is susceptible to 
predicting a high entity probability for all tokens while still performing well on the end task of pronoun resolution. %
We are interested in tracking people beyond just the entities that are labeled in the GAP task, 
for which this loss is very helpful. %

\section{Experimental Setup}
\subsection{Data}
GAP is a gender-balanced pronoun resolution dataset
introduced by \citet{webster2018gap}.
Each instance consists of a small snippet of text from Wikipedia, two spans corresponding to candidate names along with a pronoun span, and two binary labels indicating the coreference relationship between the pronoun and the two candidate names. Relative to other popular coreference datasets~\cite{weischedel2013ontonotes, chen-etal-2018-preco},
GAP is comparatively small and sparsely annotated. We choose GAP because its small size allows us to do extensive experiments.

\subsection{Model Details}
For the input BERT embeddings, we concatenate either the last four layers of \bertbase, %
or layers 19--22 of \bertlarge since those layers have been found to carry the most information related to coreference~\cite{liu2019linguistic}. The BERT embeddings are fed to a 300-dimensional GRU model, which matches the dimensionality of the memory vectors.

We vary the number of memory cells $N$
from 2 to 20. The decay rate for the memory usage scalar $\gamma$ is 0.98. The MLPs used for predicting the entity probability and similarity score consist of two 300-dimensional ReLU hidden layers.
For the \emph{Fixed Key} variant of PeTra we use 20 dimensions for the learned key vector and the remaining 280 dimensions as the value vector.

\subsection{Training}
All models are trained for a maximum of 100 epochs with the Adam optimizer~\cite{Kingma2015AdamAM}.
The learning rate is initialized to $10^{-3}$
and is reduced by half, until a minimum of $10^{-4}$,
whenever there is no improvement on the validation performance for the last 5 epochs.
Training stops when there is no improvement in validation performance for the last 15 epochs.
The temperature $\tau$ of the Gumbel-Softmax distribution used in the \actoverwrite operation is initialized to $1$ and halved every 10 epochs.
The coreference loss terms in Section~\ref{sec:coref_loss} are weighted differently for different coreference links: \begin{enumerate*}[label=(\alph*)]
    \item self-link losses for multi-token spans are given a weight of 1,
    \item positive coreference link losses are weighted by 5, and
    \item negative coreference link losses are multiplied by 50.
\end{enumerate*}
To prevent overfitting: \begin{enumerate*}[label=(\alph*)]
    \item we use early stopping based on validation performance, and
    \item apply dropout at a rate of 0.5 on the output of the GRU model.
\end{enumerate*}
Finally, we choose $\lambda=0.1$ to weight the entity prediction loss described in Section~\ref{sec:ent_pred_loss}.

\subsection{People Tracking Evaluation}
\begin{figure}[!h]
\begin{mdframed}%
    \begin{itemize}
    \item In this user study we will be comparing memory models at tracking people.
    \item What are memory models? Memory models are neural networks coupled with an external memory which can be used for reading/writing.
    \item \textcolor{red}{(IMPORTANT)} What does it mean to track people for memory models?
    \begin{itemize}
        \item Detect all references to people which includes pronouns.
        \item A 1-to-1 correspondence between people and memory cells i.e. all references corresponding to a person should be associated with the same memory cell AND each memory cell should be associated with at most 1 person.
    \end{itemize}

    \item The memory models use the following scores (which are visualized) to indicate the tracking decisions:
    \begin{itemize}
    \item New Person Probability (Cell $i$): Probability that the token refers to a new person (not introduced in the text till now) and we start tracking it in cell $i$.
    \item Coreference Probability (Cell $i$): Probability that the token refers to a person already being tracked in cell $i$.
    \end{itemize}
    \item The objective of this study is to compare the models on the interpretability of their memory logs i.e.\ are the models actually tracking entities or not. You can choose how you weigh the different requirements for tracking people (from 3).
    \item For this study, you will compare two memory models with 8 memory cells (represented via 8 rows). The models are ordered randomly for each instance.
    \item For each document, you can choose model A or model B, or stay neutral in case both the models perform similarly.
    \end{itemize}
\end{mdframed}
\caption{Instructions for the human evaluation study. We simplified certain memory model specific terms such as ``overwrite" to ``new person" since the study was really about people tracking.}
\label{fig:instructions}
\end{figure}

\label{sec:exp_people_tracking_eval}
One of the goals of PeTra was to develop memory models that not only do well on the coreference resolution task, but also are interpretable in the sense that the memory cells actually track entities. Hence in addition to reporting the standard metrics on GAP, we consider two other ways to evaluate memory models. %

  As our first task, we propose an auxiliary entity-counting task. %
We take 100 examples from the GAP validation set and annotate them
with the number of unique people mentioned in them.\footnote{In the GAP dataset, the only relevant entities are people.}
We test the models by predicting the number of people from their memory logs as explained in Section~\ref{sec:inference}.
The motivation behind this exercise is that if a memory model is truly tracking entities, then its memory usage logs should allow us to recover  this information.

To assess the people tracking performance more holistically, we conduct a human evaluation in which we ask annotators to assess the memory models on people tracking performance, defined as:%
(a) detecting references to people including pronouns, and (b) maintaining a 1-to-1 correspondence between people and memory cells.
For this study, we pick the best run (among 5 runs) of PeTra and the Referential Reader for the 8-cell configuration using \bertbase (PeTra: 81 F1; Referential Reader: 79 F1).
Next we randomly pick 50 documents (without replacement) from the GAP dev set and split those into groups of 10 to get 5 evaluation sets.
We shuffle the original 50 documents and follow the same steps to get another 5 evaluation sets.
In the end, we have a total of 10 evaluation sets with 10 documents each, where each unique document belongs to exactly 2 evaluation sets.

We recruit 10 annotators for the 10 evaluation sets.
The annotators are shown memory log visualizations as in Figure~\ref{fig:visualize}, and instructed to compare the models on their people tracking performance (detailed instructions in Figure~\ref{fig:instructions}). %
For each document the annotators are presented memory logs of the two models (ordered randomly) and asked whether they prefer the first model, prefer the second model, or have no preference
(neutral). %

\subsection{Inference}
\label{sec:inference}
\paragraph{GAP} %
Given a pronoun span $s_P$ and two candidate name spans $s_A$ \&  $s_B$, we
have to predict binary labels for potential coreference links between ($s_A$, $s_P$) and ($s_B$, $s_P$). %
Thus, for a pair of entity spans, say $s_A$ and $s_P$, we predict the coreference link probability as:
\begin{align*}
    P_{\mathrm{CL}}(s_A, s_P) = \max_{w_A \in s_A, w_P \in s_P} P_{\mathrm{CL}}(w_A, w_P)
\end{align*}
where $P_{\mathrm{CL}}(w_A, w_P)$ is calculated using the procedure described in Section~\ref{sec:coref_link_prob}\footnote{The computation of this probability includes the mention detection steps required by\citet{webster2018gap}.}. The final binary prediction is made by comparing the probability against a threshold.

\paragraph{Counting unique people}
For the test of unique
people counting, we discretize the overwrite operation, which corresponds to new entities, against a threshold $\alpha$ and sum over all
tokens and all
memory cells to predict the count as follows:\vspace{-0.02in}
$$\text{\# unique people} = \sum_{t=1}^T\sum_{i=1}^N \mathbbm{1}[o^i_t \geq \alpha] $$

\subsection{Evaluation Metrics}
For GAP we evaluate models using F-score.\footnote{GAP also includes evaluation related to gender bias, but this is not a focus of this paper so we do not report it.}
First, we pick a threshold from the set \{0.01, 0.02, $\cdots$, 1.00\} which maximizes the validation F-score.
This threshold is then used to evaluate performance on the GAP test set.

For the interpretability task of counting unique people, we
choose a threshold that minimizes the absolute difference between ground truth count and predicted count summed over the 100 annotated examples.
We select the best threshold from the set \{0.01, 0.02, $\cdots$, 1.00\}.
The metric is then the number of errors corresponding to the best threshold.\footnote{Note that the error we report is therefore a best-case result.  We are not proposing a way of counting unique people in new test data, but rather using this task for analysis.}

\subsection{Baselines}
The Referential Reader~\cite{liu2019referential} is the most relevant baseline in the literature, and the most similar to PeTra.
The numbers reported by~\citet{liu2019referential} are obtained by a version of the model using \bertbase, with only two memory cells.
To compare against PeTra for other configurations, we retrain the Referential Reader using the code made available by the authors.\footnote{\url{https://github.com/liufly/refreader}}

We also report the results of \citet{joshi-etal-2019-bert} and \citet{wu2019coreference}, %
although these numbers are not comparable since both of them train on the much larger OntoNotes corpus and just test on GAP.

\section{Results}

\begin{figure}[t]
\centering
\begin{subfigure}[b]{0.47\textwidth}
        \centering
        \includegraphics[width=\textwidth]{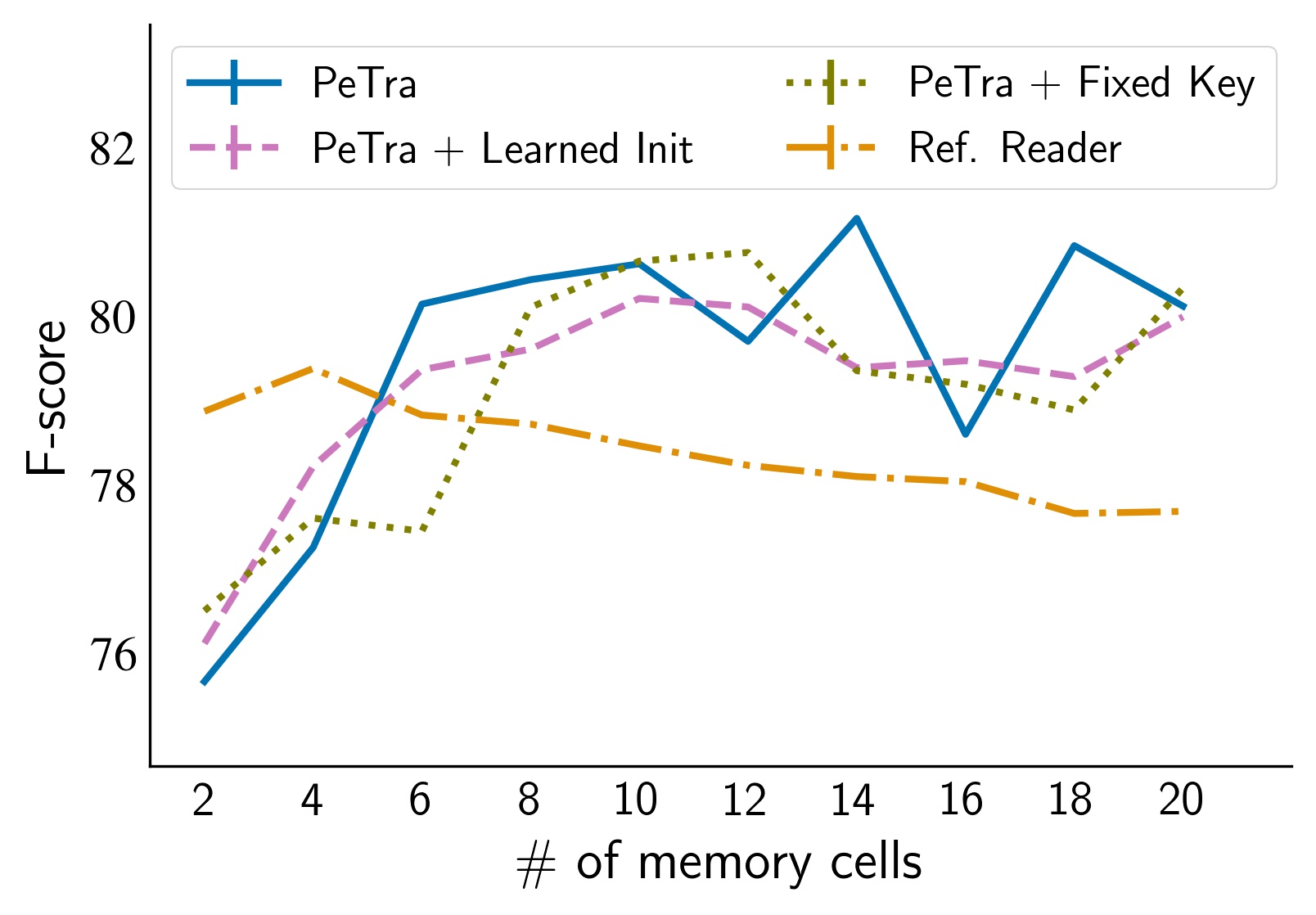}
        \caption{\bertbase}
        \label{fig:gap_val_small}
    \end{subfigure}%
    ~
    \begin{subfigure}[b]{0.47\textwidth}
        \centering
        \includegraphics[width=\textwidth]{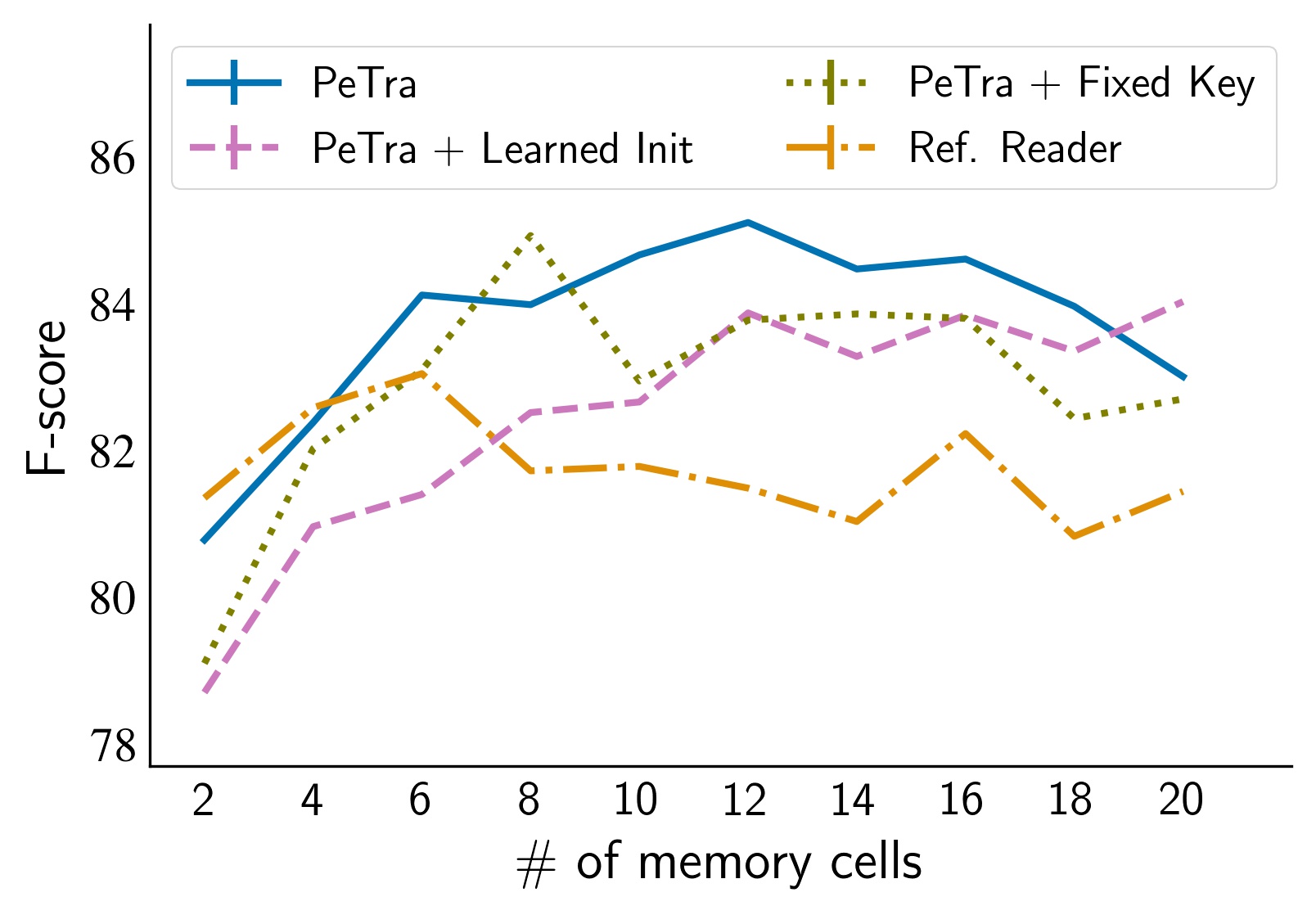}
        \caption{\bertlarge}
        \label{fig:gap_val_large}
    \end{subfigure}
    \caption{Mean F1 score on the GAP validation set as a function of the number of memory cells.}
    \label{fig:gap_val}
\end{figure}

\begin{table}[t]
\centering{
\small{
\begin{tabular}{lll}
\toprule
 & \bertbase  & \bertlarge \\\midrule
PeTra                   & {\bf 81.5 $\pm$ 0.6} &  {\bf 85.3 $\pm$ 0.6} \\
\hspace{0.1in} + Learned Init. & 80.9 $\pm$ 0.7 &  84.4 $\pm$ 1.2\\
\hspace{0.1in} + Fixed Key & 81.1 $\pm$ 0.7 & 85.1 $\pm$ 0.8\\
Ref. Reader & 78.9 $\pm$ 1.3 & 83.7 $\pm$ 0.8\\
Ref. Reader ~\cite{liu2019referential} & 78.8 & - \\\midrule
\citet{joshi-etal-2019-bert} & 82.8 & 85.0 \\
\citet{wu2019coreference} & - & 87.5 (SpanBERT)\\
\bottomrule
\end{tabular}
}
}
\caption{Results (\%F1) on the GAP test set.}
\label{tab:gap_res}
\end{table}
\subsection{GAP results}
\label{sec:gap_result}

We train all the memory models, including the Referential Reader, with memory size varying from \{2, 4, $\cdots$, 20\} memory cells for both \bertbase and \bertlarge, with each configuration being trained 5 times.
Figure~\ref{fig:gap_val} shows the performance of the
models on the GAP validation set as a function of memory size.
The Referential Reader outperforms
PeTra (and its memory variants) when using a small number of memory cells, but its performance starts degrading after 4 and 6 memory cells for \bertbase and \bertlarge respectively. PeTra and its memory variants, in contrast, keep improving with increased memory size (before saturation at a higher number of cells) and outperform the best Referential Reader performance for all memory sizes $\ge$ 6 cells. With larger numbers of memory cells, we see a higher variance, but the curves for PeTra and its memory variants are still consistently higher than those of the Referential Reader.

Among different memory variants of PeTra  %
, when using \bertbase the performances are comparable with no clear advantage for any particular choice. %
For \bertlarge, however, vanilla PeTra has a clear edge for almost all memory sizes, suggesting the limited utility of initialization.
The results show that PeTra works well without learning vectors for initializing the key or memory cell contents. Rather, we can remove the key/value distinction and simply initialize all memory cells with the zero vector.

To evaluate on the GAP test set, we pick the memory size corresponding to the best validation performance for all memory models. Table~\ref{tab:gap_res} shows that the trends from validation hold true for test as well, with PeTra outperforming the Referential Reader and the other memory variants of PeTra.

\begin{figure}[ht]
\centering
\begin{subfigure}[b]{0.47\textwidth}
        \centering
        \includegraphics[width=\textwidth]{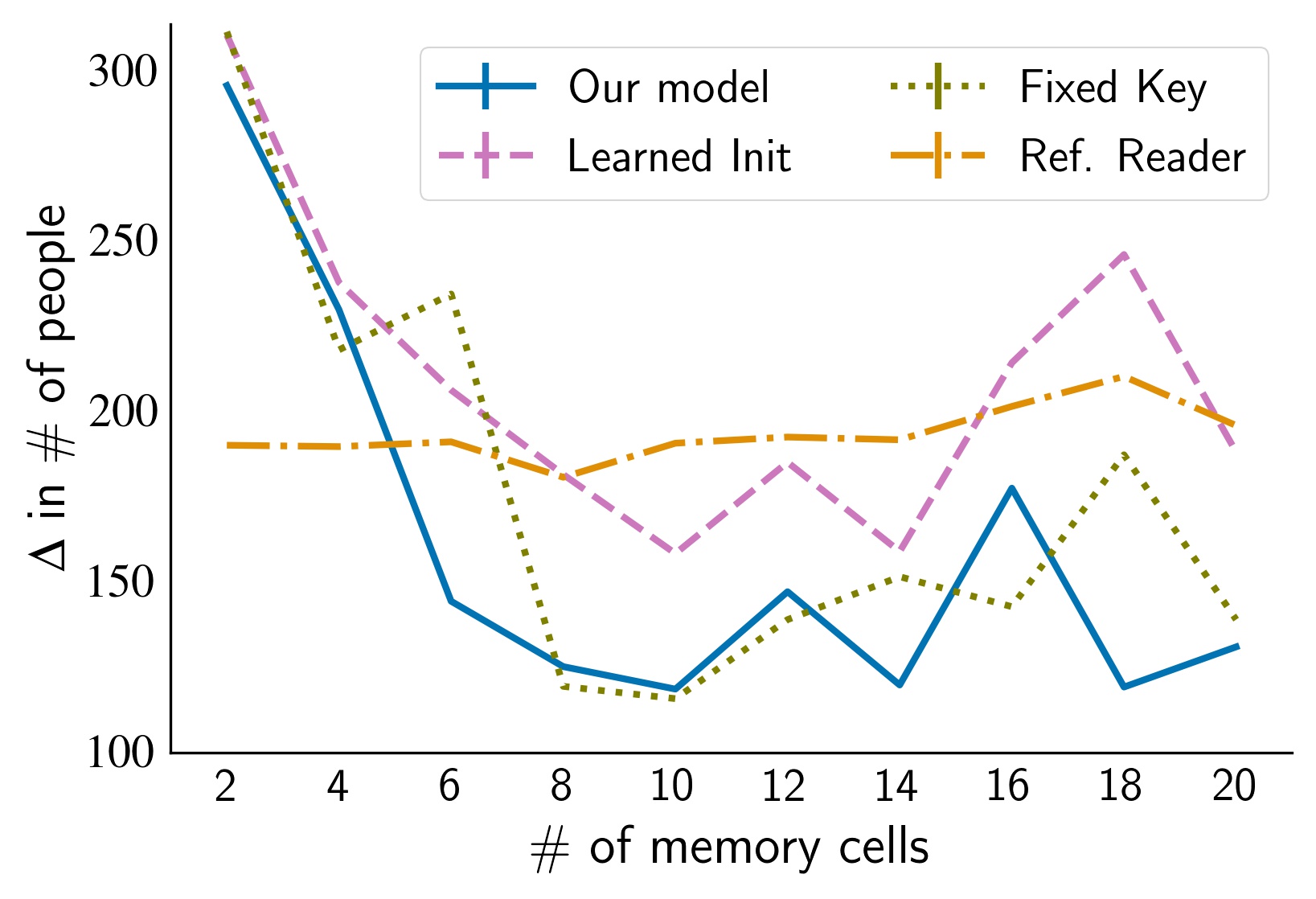}
        \caption{\bertbase}
        \label{fig:count_uniq_people_small}
    \end{subfigure}%
    ~
    \begin{subfigure}[b]{0.47\textwidth}
        \centering
        \includegraphics[width=\textwidth]{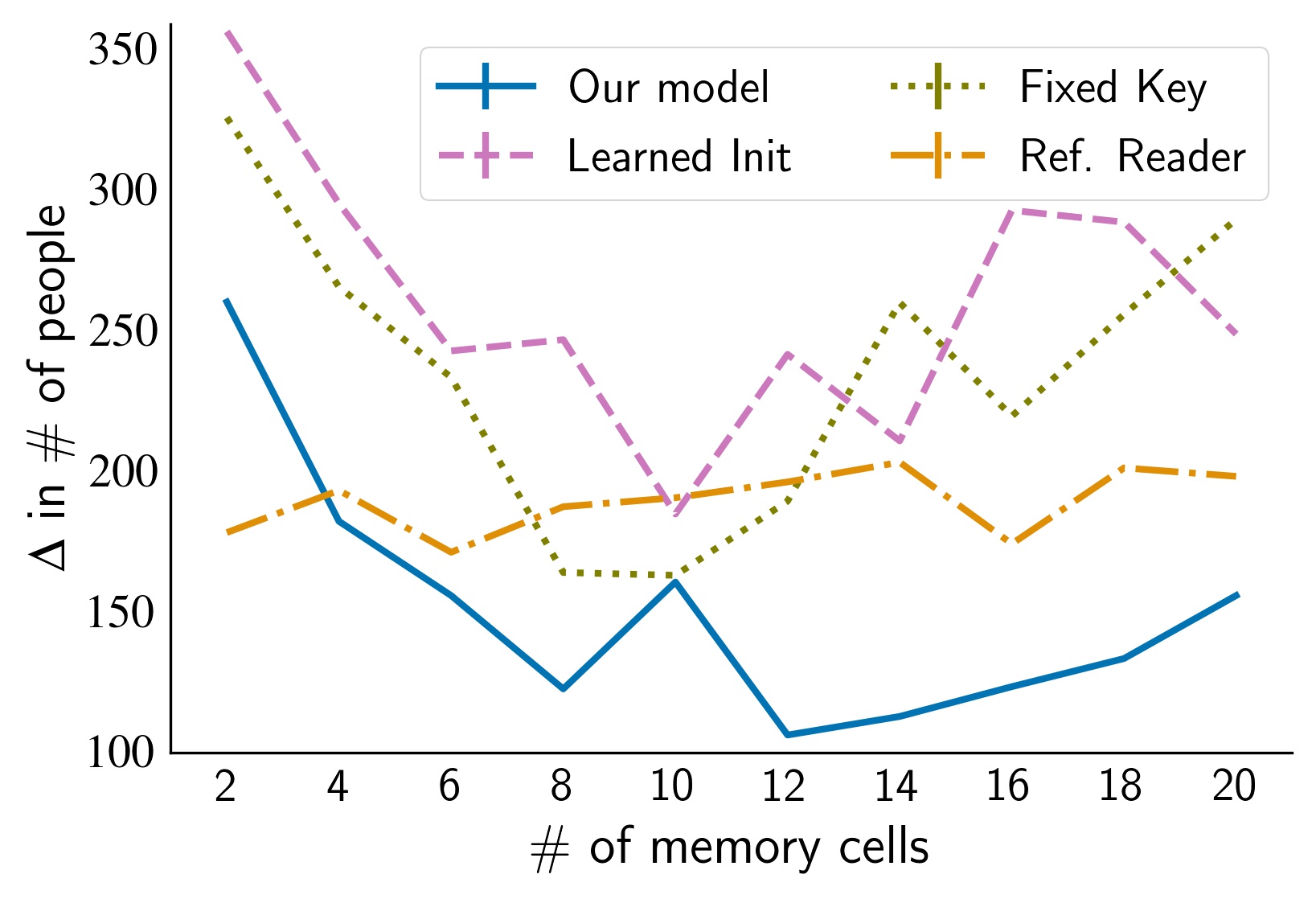}
        \caption{\bertlarge}
        \label{fig:count_uniq_people_large}
    \end{subfigure}
    \caption{Error in counting unique people as a function of number of memory cells; lower is better. }
    \label{fig:count_uniq_people}
\end{figure}

\begin{table}[ht]
\setlength{\tabcolsep}{5pt}
\centering{
    \begin{tabular}{lcc}
    \toprule
         & \bertbase & \bertlarge \\
        \toprule
        PeTra & \textbf{0.76} & \textbf{0.69} \\
        \hspace{0.1in} + Learned Init & 0.72 & 0.60 \\
        \hspace{0.1in} + Fixed Key & 0.72 & 0.65 \\
        Ref. Reader & 0.49 &  0.54 \\
        \bottomrule
    \end{tabular}
}
\caption{Spearman's correlation between GAP validation F1 and negative error count for unique people.} %
\label{tab:correlation}
\end{table}

\subsection{Counting unique people}
\label{sec:count_result}
Figure~\ref{fig:count_uniq_people} shows the results for the proposed interpretability task of counting unique people.
For both \bertbase and \bertlarge, PeTra achieves the lowest error count.
Interestingly, from Figure~\ref{fig:count_uniq_people_large} we can see that for $\ge 14$ memory cells, the other memory variants of PeTra perform worse than the Referential Reader while being better at the GAP validation task (see Figure~\ref{fig:gap_val_large}). This shows that a better performing model is not necessarily better at tracking people.

To test the relationship between the GAP task and the proposed interpretability task, we compute the correlation between the GAP F-score and the negative count of unique people for each model separately.\footnote{Each correlation is computed over 50 runs (5 runs each for 10 memory sizes).}
Table \ref{tab:correlation} shows the Spearman's correlation between these measures. For all models we see a positive correlation, indicating that a dip in coreference performance corresponds to an increase in error on counting unique people. The correlations for PeTra are especially high, again suggesting it's greater interpretability.

\begin{table}[t]
\centering{
\small{
\begin{tabular}{lc}
\toprule
Model & Preference (in \%) \\
\midrule
PeTra        & \textbf{74} \\
Ref. Reader &  08\\
Neutral & 18 \\\bottomrule
\end{tabular}
}
}
\caption{Human Evaluation results for people tracking.}
\label{tab:hum_eval}
\end{table}

\begin{figure}[t]
    \centering
    \includegraphics[width=0.45\textwidth]{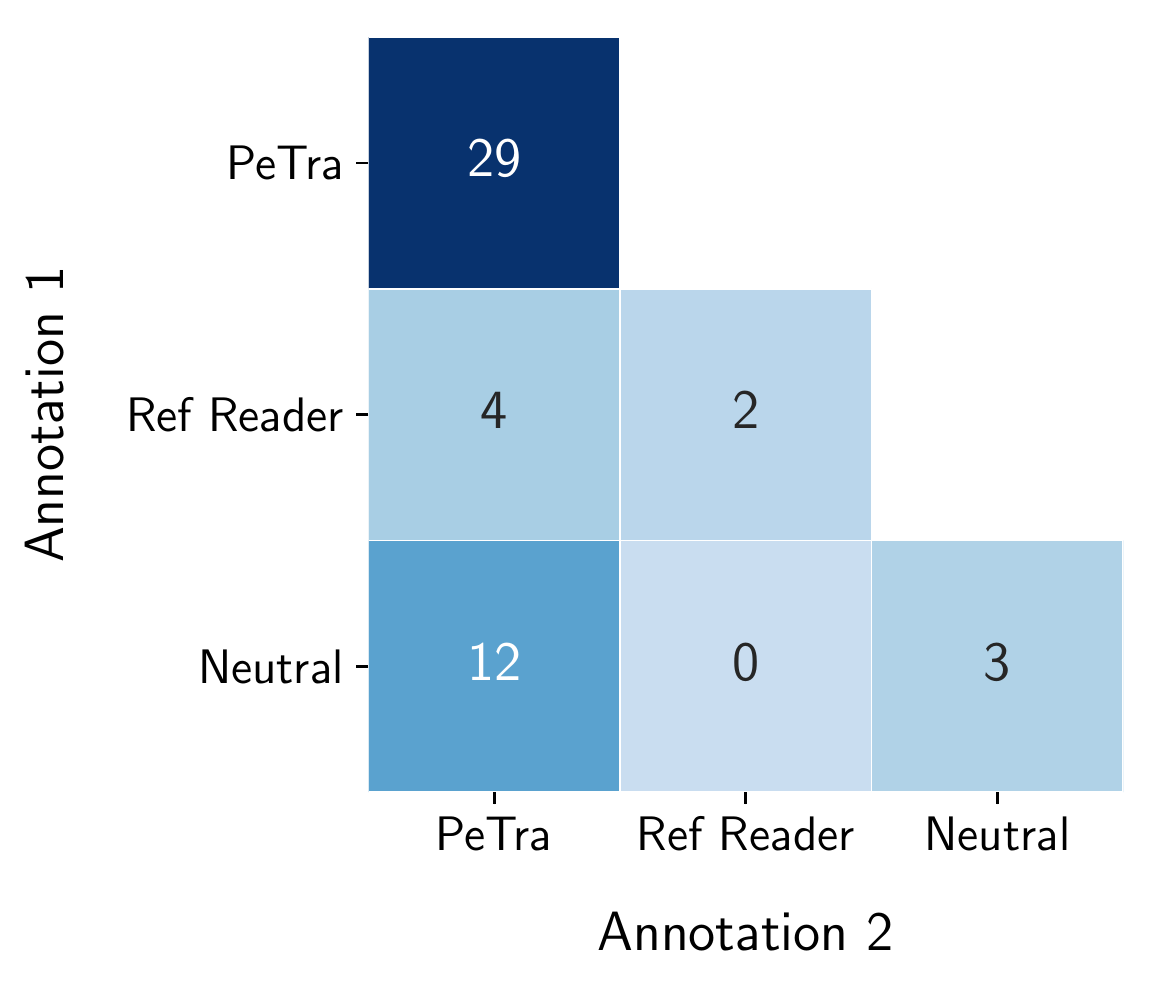}
    \caption{Agreement matrix for human evaluation study.}
    \label{fig:agreement}
\end{figure}

\subsection{Human Evaluation for People Tracking}
\label{sec:hum_eval}
\begin{figure}[t]
    \begin{subfigure}[t]{\textwidth}
    \begin{mdframed}
        \hlent{\textcolor{aqua}{\bf Neef}}{1} took an individual silver medal at the 1994 European Cup behind Russia's \hlent{\textcolor{red}{\it Svetlana Goncharenko}}{2} and returned the following year to win gold. \hlent{\textcolor{aqua}{\bf She}}{1} was a finalist individually at the 1994 European Championships and came sixth for Scotland at the 1994 Commonwealth Games.
    \end{mdframed}
    \caption{GAP validation instance 293. The ground truth GAP annotation is indicated via colors.}
    \end{subfigure}
    ~\vspace{0.2in}
    \centering
    \begin{subfigure}[t]{\textwidth}
    \includegraphics[width=\textwidth]{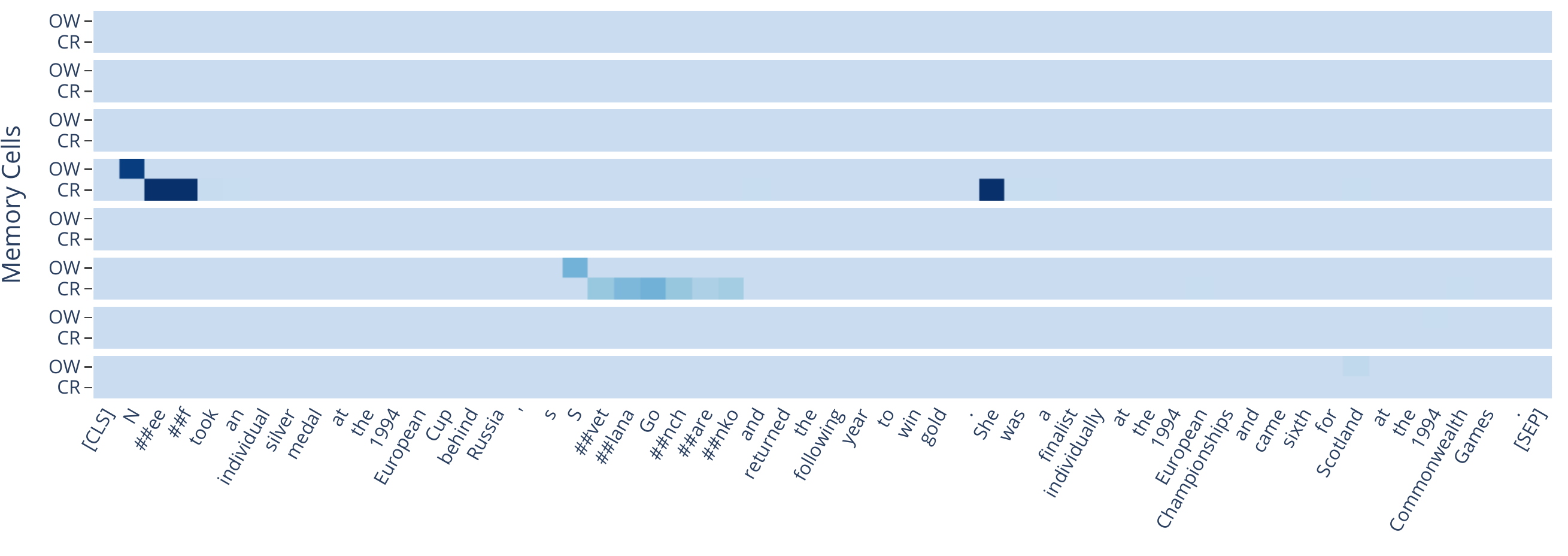}
    \caption{Memory log of PeTra with 8 memory cells. PeTra uses only 2 memory cells for the 2 unique people, namely Neef and Svetlana Goncharenko, and correctly resolves the pronoun.}
    \end{subfigure}
    \begin{subfigure}[t]{\textwidth}
    \includegraphics[width=\textwidth]{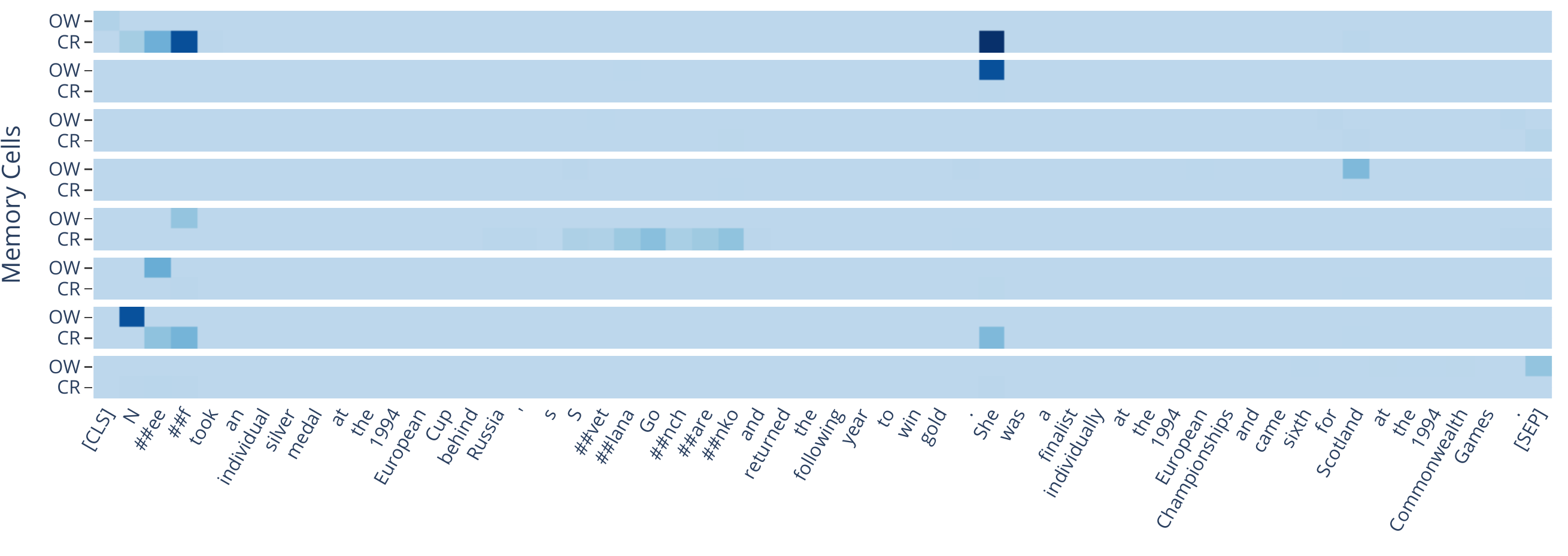}
    \caption{Memory log of the Referential Reader with 8-memory cells. The Referential Reader does successfully resolve the pronoun in the topmost memory cell but it ends up tracking Neef in as many as 4 memory cells.}
    \end{subfigure}
    \caption{
    Both the models only weakly detect ``Svetlana Goncharenko" which could be due to lack of span modeling.}
    \label{fig:petra_vs_ref_1}
\end{figure}

\begin{figure}[t]
    \begin{subfigure}[t]{\textwidth}
    \begin{mdframed}
        \hlent{Fripp}{1} has performed Soundscapes in several situations: * \hlent{Fripp}{1} has featured Soundscapes on various King Crimson albums. \hlent{He}{1} has also released pure Soundscape recordings as well: * On May 4, 2006, \hlent{\textcolor{red}{\it Steve Ball}}{2} invited \hlent{\textcolor{aqua}{\bf Robert Fripp}}{1} back to the Microsoft campus for a second full day of work on Windows Vista following up on \hlent{\textcolor{aqua}{\bf his}}{1} first visit in the Fall of 2005.
    \end{mdframed}
    \caption{GAP validation instance 17. The ground truth GAP annotation is indicated via colors.}
    \end{subfigure}
    \centering
    \begin{subfigure}[t]{\textwidth}
    \includegraphics[width=\textwidth]{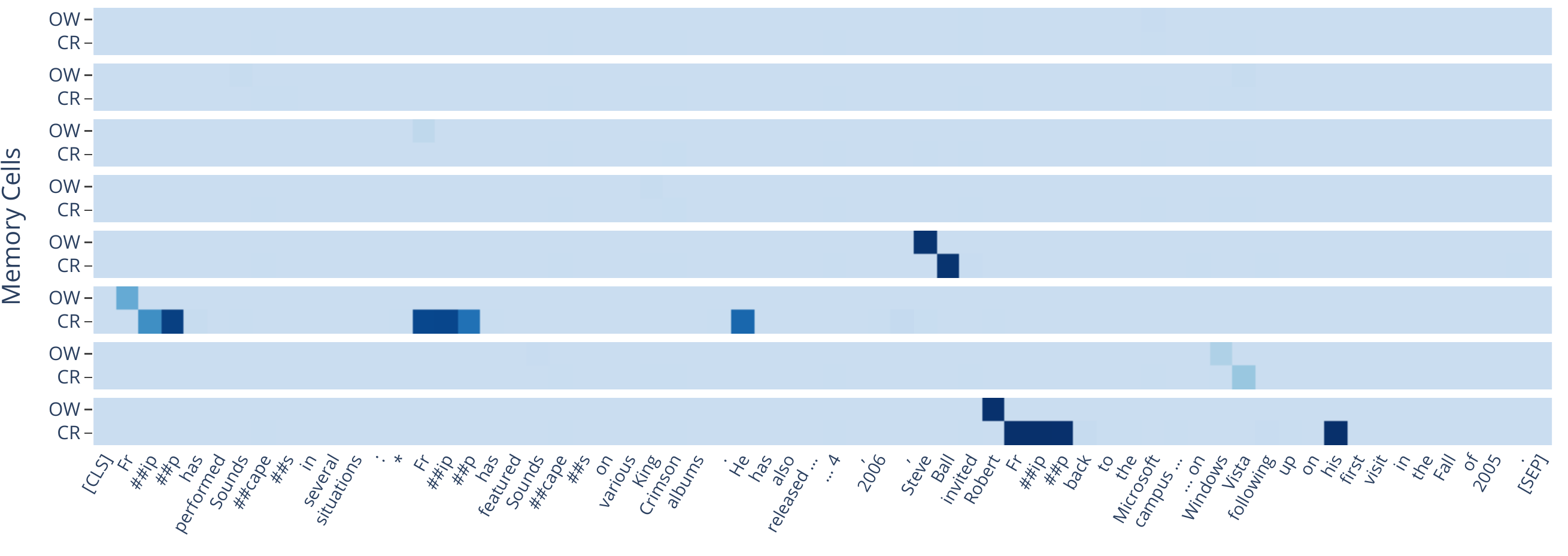}
    \caption{Memory log of PeTra with 8-memory cells. PeTra is pretty accurate at tracking Robert Fripp but it misses out on connecting ``Fripp" from the earlier part of the document to ``Robert Fripp". }
    \end{subfigure}
    \begin{subfigure}[t]{\textwidth}
    \includegraphics[width=\textwidth]{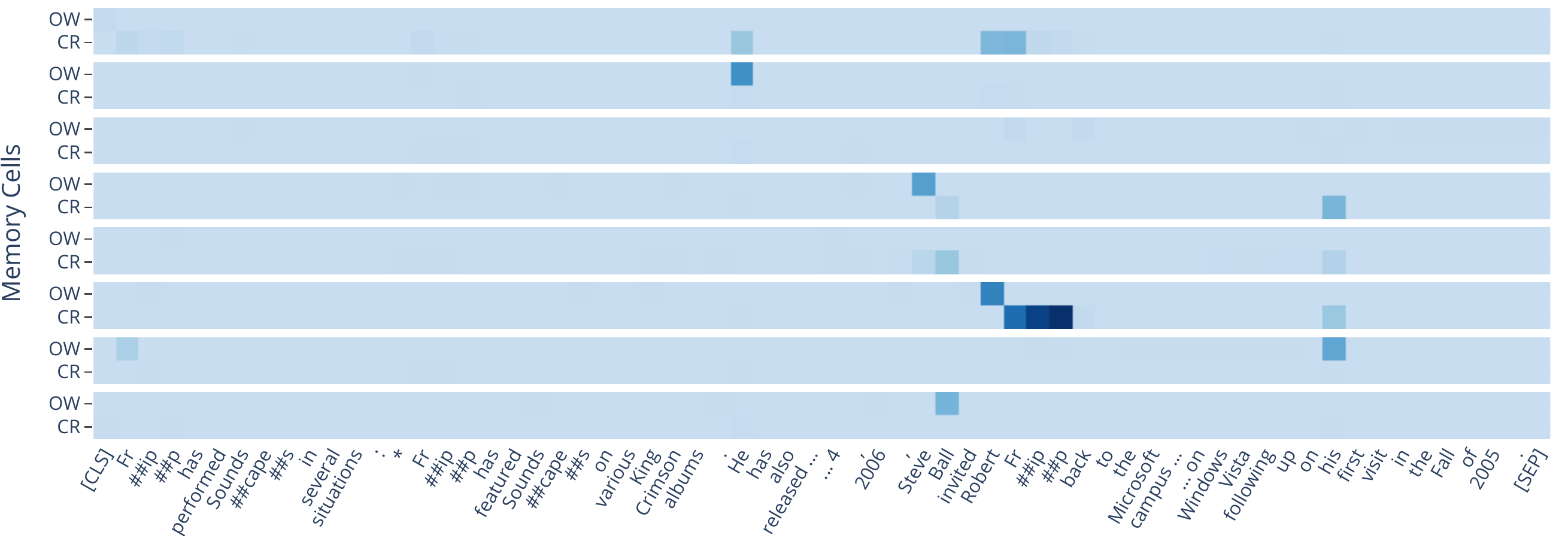}
    \caption{Memory log of the Referential Reader with 8-memory cells. The Referential Reader completely misses out on all the mentions in the first half of the document (which is not penalized in GAP evaluations where the relevant annotations are typically towards the end of the document). Apart from this, the model ends up tracking Robert Fripp in as many as 6 memory cells, and Steve Ball in 3 memory cells.}
    \end{subfigure}
    \caption{PeTra clearly performs better than the Referential Reader at people tracking for this instance. PeTra's output is more sparse, detects more relevant mentions, and is better at maintaining a 1-to-1 correspondence between memory cells and people.}
    \label{fig:petra_vs_ref_2}
\end{figure}

Table~\ref{tab:hum_eval} summarizes the results of the human evaluation for people tracking. %
The annotators prefer PeTra in 74\% cases while the Referential Reader for only 8\% instances (sample comparisons shown in Figure~\ref{fig:petra_vs_ref_1} and ~\ref{fig:petra_vs_ref_2}). %
Thus, PeTra easily outperforms the Referential Reader on this task even though they are quite close on the GAP evaluation.

The agreement matrix for the study is shown in Figure~\ref{fig:agreement}.
This agreement matrix is a result of the two annotations per document that we get as per the setup described in Section~\ref{sec:exp_people_tracking_eval}.
The annotators agree on 68\% of the documents, disagree between PeTra and Neutral for 24\% of the documents, and disagree between PeTra and the Referential Reader for  the remaining 8\%  documents. 
Note that the annotations are coming from two sets of annotators rather than two individual annotators. This is also the reason why we don't report standard inter-annotator agreement coefficients.

\begin{figure*}[!ht]
    \centering

    \begin{subfigure}[t]{\textwidth}
    \begin{mdframed}
        \footnotesize{
        \hlent{Amelia Shepherd}{1}, M.D. is a fictional character on the ABC American television medical drama Private Practice, and the spinoff series' progenitor show, Grey's Anatomy, portrayed by \hlent{\textcolor{red}{\it Caterina Scorsone}}{2}. In \hlent{\textcolor{aqua}{\bf her}}{1} debut appearance in season three, \hlent{\textcolor{aqua}{\bf Amelia}}{1} visited her former sister-in-law, \hlent{Addison Montgomery}{3}, and became a partner at the Oceanside Wellness Group.
        }
    \end{mdframed}
    \end{subfigure}
    \begin{subfigure}[t]{\textwidth}
    \includegraphics[width=\textwidth]{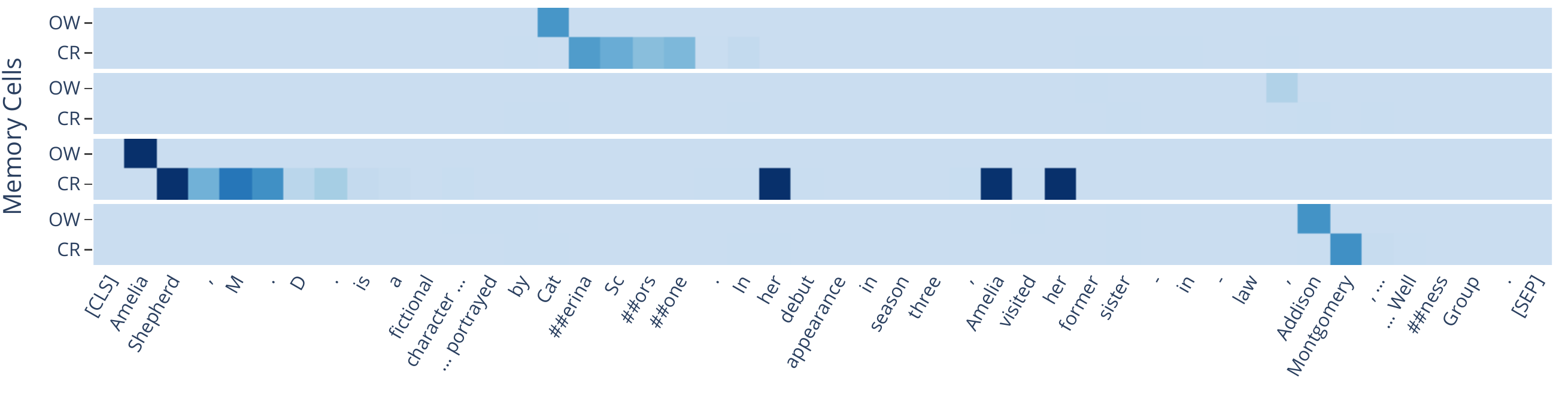}
    \caption{A successful run of PeTra with 4 memory cells. The model accurately links all the mentions of ``Amelia" to the same memory cell while also detecting other people in the discourse.}
    \label{fig:success}
    \end{subfigure}
    \begin{subfigure}[t]{\textwidth}
    \begin{mdframed}
        \footnotesize{
        \hlent{Bethenny}{1} calls a meeting to get everyone on the same page, but \hlent{Jason}{2} is hostile with the group, making things worse and forcing \hlent{Bethenny}{1} to play referee. Emotions are running high with \hlent{\textcolor{red}{\it Bethenny}}{1}'s assistant, \hlent{\textcolor{aqua}{\bf Julie}}{3}, who breaks down at a lunch meeting when asked if \hlent{\textcolor{aqua}{\bf she}}{3} is committed to the company for the long haul.
        }
    \end{mdframed}
    \end{subfigure}
    \begin{subfigure}[t]{\textwidth}
    \includegraphics[width=\textwidth]{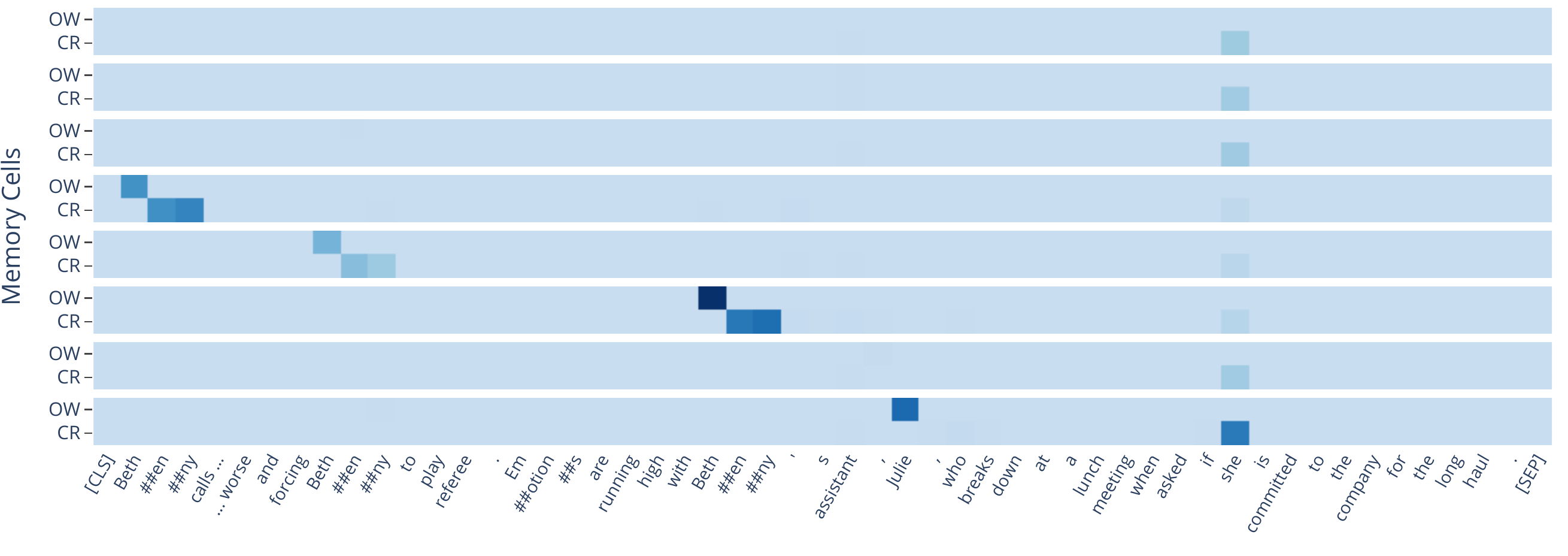}
    \caption{Memory log of PeTra with 8 memory cells. The model correctly links ``she" and ``Julie" but fails at linking the three ``Bethenny" mentions, and also fails at detecting ``Jason".}
    \label{fig:failure}
    \end{subfigure}
    \caption{Visualization of memory logs for different configurations of PeTra.
    The documents have their GAP annotations highlighted in red (italics) and blue (bold), with blue (bold) corresponding to the right answer. For illustration purposes only, we highlight all the spans corresponding to mentions of people and mark cluster indices as subscript.
    In the plot, X-axis corresponds to document tokens, and Y-axis corresponds to memory cells. Each memory cell has the OW=\actoverwrite and CR=\actcoref labels. Darker color implies higher value.
    We skip text, indicated via ellipsis, when the model doesn't detect people for extended lengths of text.}
    \label{fig:visualize}
\end{figure*}

\subsection{Model Runs}
We visualize two runs of PeTra with different configurations in Figure~\ref{fig:visualize}. For both instances the model gets the right pronoun resolution, but clearly in Figure~\ref{fig:failure} the model fails at correctly tracking repeated mentions of ``Bethenny". We believe these errors happen because (a) GAP supervision is limited to pronoun-proper name pairs, so the model is never explicitly supervised to link proper names, and (b) there is a lack of span-level features, which hurts the model when a name is split across multiple tokens. %

\section{Conclusion}

In this chapter we proposed PeTra, a bounded memory model, for entity tracking, which is trained using sparse coreference resolution supervision.
PeTra outperforms a previous approach with fewer parameters and a simpler architecture.
We conduct a human evaluation to test the interpretability of PeTra, and find that
our model again does better on this evaluation.

In the experiments presented in this chapter, PeTra deals with just the pronoun resolution task. %
There are a few hurdles to using PeTra for the full coreference resolution task. 
Firstly, PeTra makes coreference link predictions at token-level which works for GAP, but general coreference resolution benchmarks evaluate on exact mention boundaries which can span multiple tokens. To overcome this limitation, the model needs to make \emph{span-level predictions
} which means moving away from the token-level incremental predictions made by PeTra. 

Secondly, even though PeTra, in terms of how it operates, is an entity-ranking model, the scoring scheme described in Section~\ref{sec:coref_link_prob} only allows for making pairwise predictions between mentions, which as we have discussed in Section~\ref{sec:coref_prior} has well-known drawbacks. 
Moreover, the ``soft" updates used in PeTra mean that every token is an entity mention and is coreferent with all memory cells with non-zero probability. Generalizing PeTra's ``soft" updates to span-level predictions would mean that every span would be a mention which is not scalable for long documents~\cite{lee-etal-2017-end}. One way of addressing these shortcomings is by \emph{performing ``hard" updates} for both the mention detection and memory update steps. We explore such a (bounded) memory model in the next chapter.

\chapter{Scalability and Generalization in Coreference Resolution}

In this chapter, we focus on two key issues relating to applicability of coreference resolution models, namely scalability to long documents and generalizability.
We first present an adaptation of the PeTra model from the previous chapter which can scale to long documents. 
The presented models achieved state-of-the-art performance on a long document dataset
when they were first introduced. 
The results presented in this chapter are with retraining these models with a newer document encoder which reestablishes their state-of-the-performance as of this writing.  
Next, we test the presented models for their generalization capability to help benchmark how robust the models are when used in the wild. 
To this end, we form an evaluation suite formed by consolidating existing coreference benchmarks to measure the generalization capability of coreference models.
The results suggest that domain shift is a challenge in coreference resolution though annotation differences across datasets partly exaggerate this challenge. 
We also find that joint training on multiple datasets moderately alleviates the domain shift challenge.
\footnote{The material for this chapter has been adapted from \citet{toshniwal-etal-2020-learning} and \citet{toshniwal-etal-2021-generalization}. Code is available at \url{https://github.com/shtoshni/fast-coref}}

\section{Long Document Coreference Resolution}
We noted at the end of the previous chapter, that extending PeTra for the ``full" coreference resolution task and for scaling it to long documents would require moving from: (a) token to span-level predictions, and (b) probabilistic to discrete clustering actions. 
In particular, we explore such memory models.  
We explore variants where the number of entities in memory can be either bounded, as in PeTra, or unbounded. 
The unbounded memory models add any new entity to the memory and are essentially the vanilla entity-ranking model~\cite{rahman2011narrowing, stoyanov-eisner-2012-easy, websterC14} viewed from the lens of memory models.
The bounded memory models require a memory management scheme, similar to PeTra, once the number of entities in the memory reaches the capacity.

We're particularly interested in bounded memory models because they guarantee a linear runtime in length of document (assuming the number of mentions scales linearly with the document length). 
While the computational benefits of bounded memory models make them attractive, but can a bounded memory model even work for long documents, say book-length documents with potentially hundreds of entities. For comparison, PeTra was only tested on relatively short documents in GAP (average of 95 words per document) and while focusing only on entities of type people.
But for long documents, \emph{Is it necessary to maintain an unbounded number of mentions or entities?}  
Psycholinguistic evidence suggests it is not, as human language processing is incremental \citep{Tanenhaus1632, keller2010cognitively} and has limited working memory~\citep{baddeley1986}. 
We find that in practice it is not necessary to keep all the entities in memory all the time as well, and we formalize the empirical arguments to support this claim in the next section.

\begin{figure*}[t]
\centering
\begin{subfigure}[b]{0.45\textwidth}
        \centering
        \includegraphics[width=\textwidth]{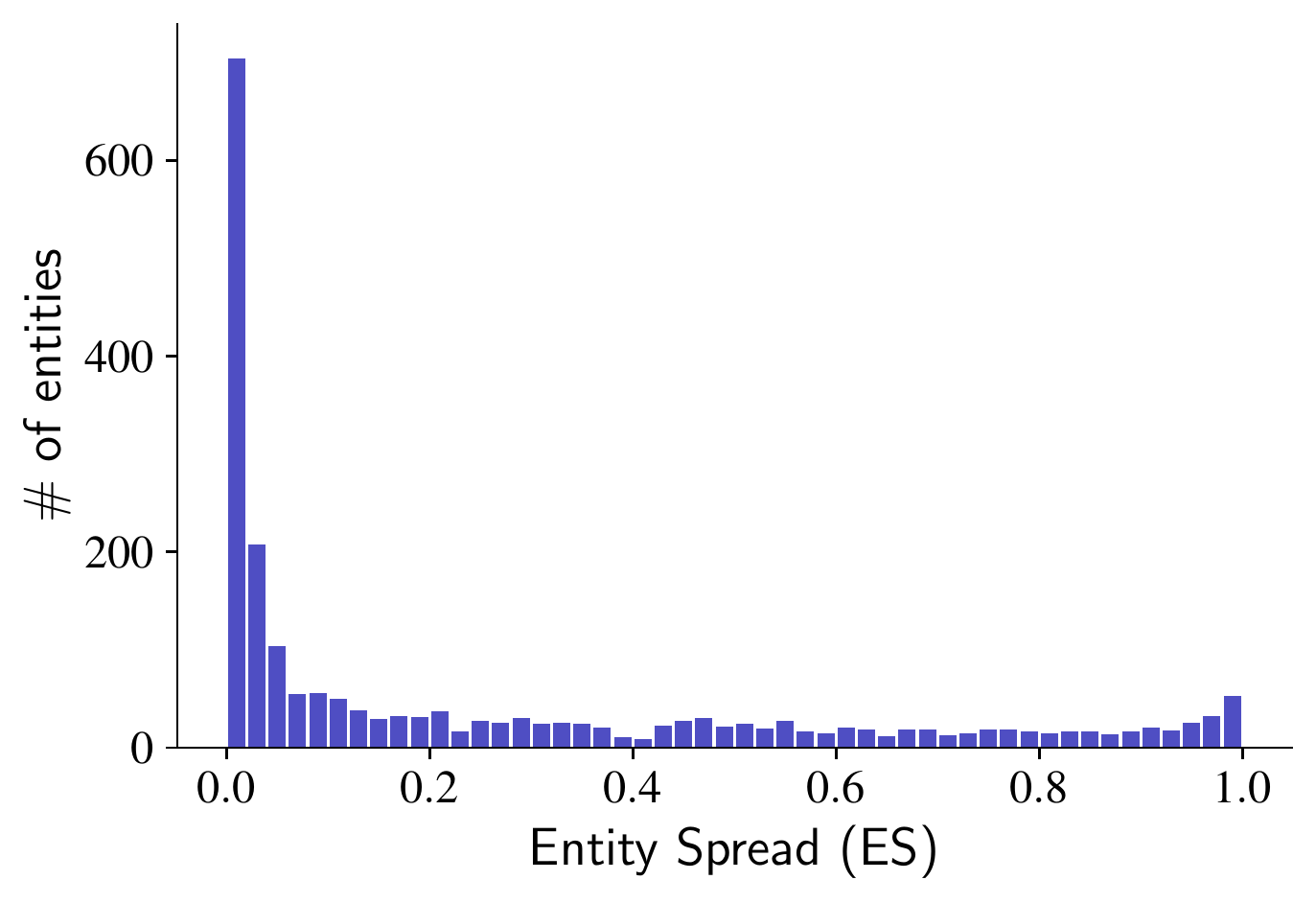}
        \caption{LitBank.}
        \label{fig:litbank_spread}
    \end{subfigure}%
    ~
    \begin{subfigure}[b]{0.45\textwidth}
        \centering
        \includegraphics[width=\textwidth]{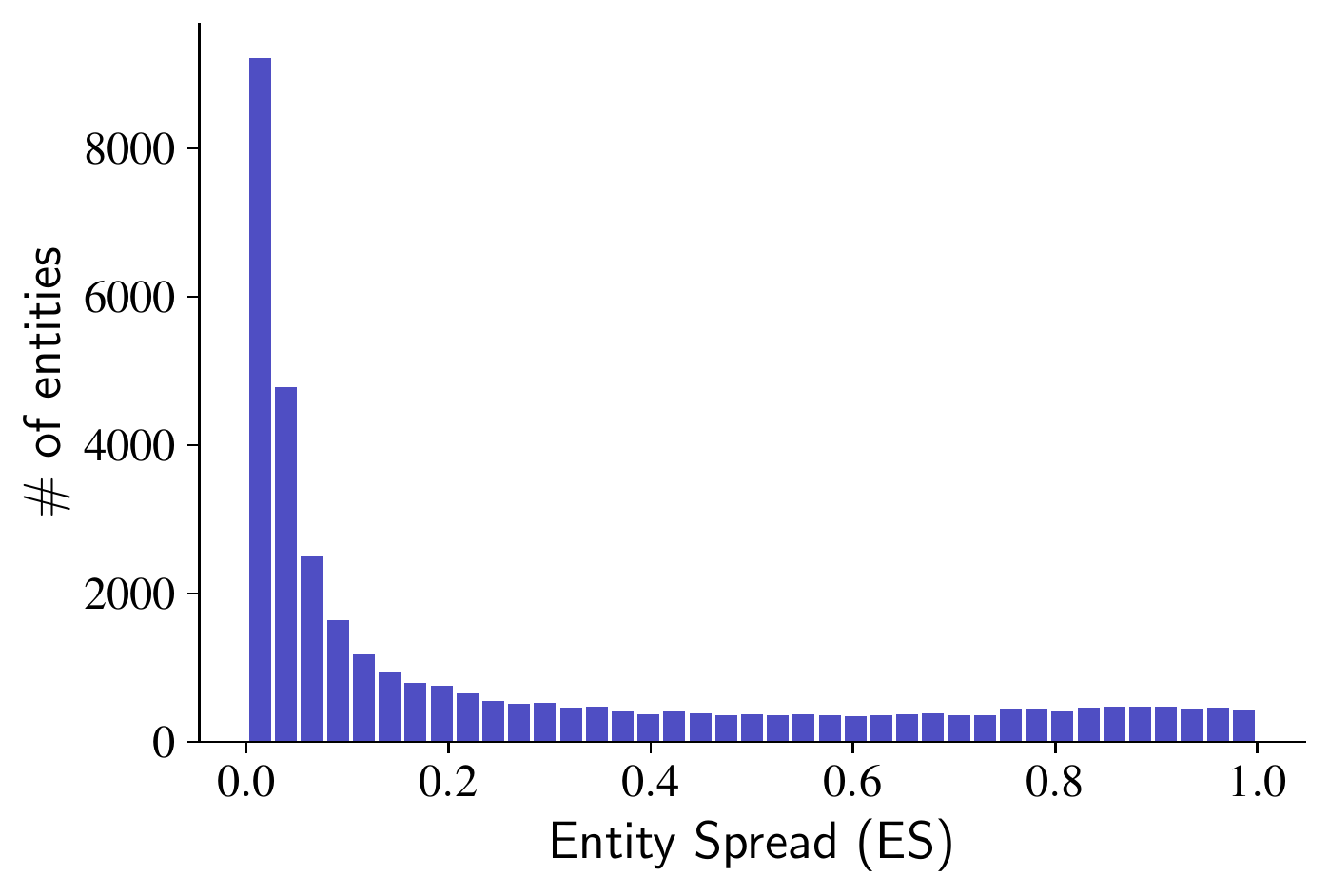}
        \caption{OntoNotes Training Set.}
        \label{fig:ontonotes_spread}
    \end{subfigure}
    \caption{Histograms of Entity Spread as fraction of document length for LitBank and OntoNotes. For LitBank we only visualize the entity spread of non-singleton clusters because the histogram is even more skewed towards zero with singletons included. }
    \label{fig:entity_spread}
\end{figure*}

\subsection{Entities are Transient: A Case for Bounded Memory Models}
\begin{table}[h]
    \centering{
    \small{

    \begin{tabular}{lcc}
    \toprule
        & LitBank & OntoNotes \\\midrule
    Max.\ Total Entity Count &  199 & 94 \\
    Max.\ Active Entity Count & $\phantom{1}$18 & 24 \\\bottomrule
    \end{tabular}
    }
    }
    \caption{Max.\ Total Entity Count vs.\ Max.\ Active Entity Count.}
    \label{sec:tab_active_entities}

\end{table}

\label{sec:active}
Given input document $\mathcal{D}$, let $(x_n)_{n=1}^N$ represent the $N$ mention spans corresponding to $M$ underlying entities $(e_m)_{m=1}^M$. %
Let $\textrm{START}(x_i)$ and $\textrm{END}(x_i)$ denote the start and end token indices of the mention span $x_i$ in document $\mathcal{D}$.
Let $\textrm{ENT}(x_i)$ %
 denote the entity of which $x_i$ is a mention.
Given this notation we next define the following concepts.

\paragraph{Entity Spread} Entity spread denotes the interval of token indices from the first mention to the last mention of an entity. %
The entity spread $\textrm{ES}(e)$ of entity $e$ is given by: %
$$\textrm{ES}(e) = [\min_{\textrm{ENT}(x) = e}\textrm{START}(x), \max_{\textrm{ENT}(x) = e}\textrm{END}(x)]$$

Figure~\ref{fig:entity_spread} shows the histogram of Entity Spread normalized by the document length for documents in LitBank and OntoNotes (training set).    

\begin{figure*}[t]
\centering
\begin{subfigure}[b]{0.45\textwidth}
        \centering
        \includegraphics[width=\textwidth]{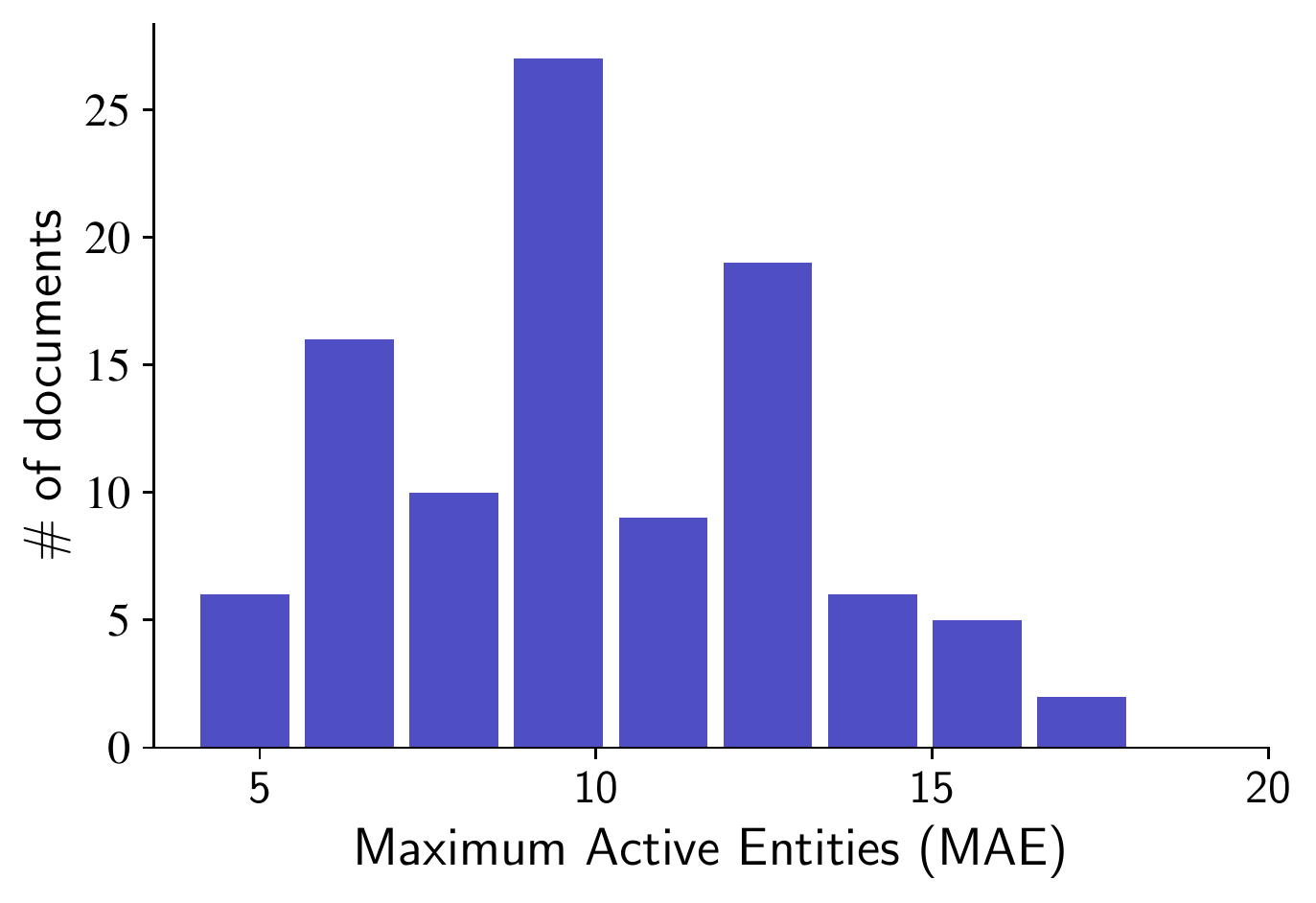}
        \caption{LitBank.}
        \label{fig:litbank_active}
    \end{subfigure}%
    ~
    \begin{subfigure}[b]{0.45\textwidth}
        \centering
        \includegraphics[width=\textwidth]{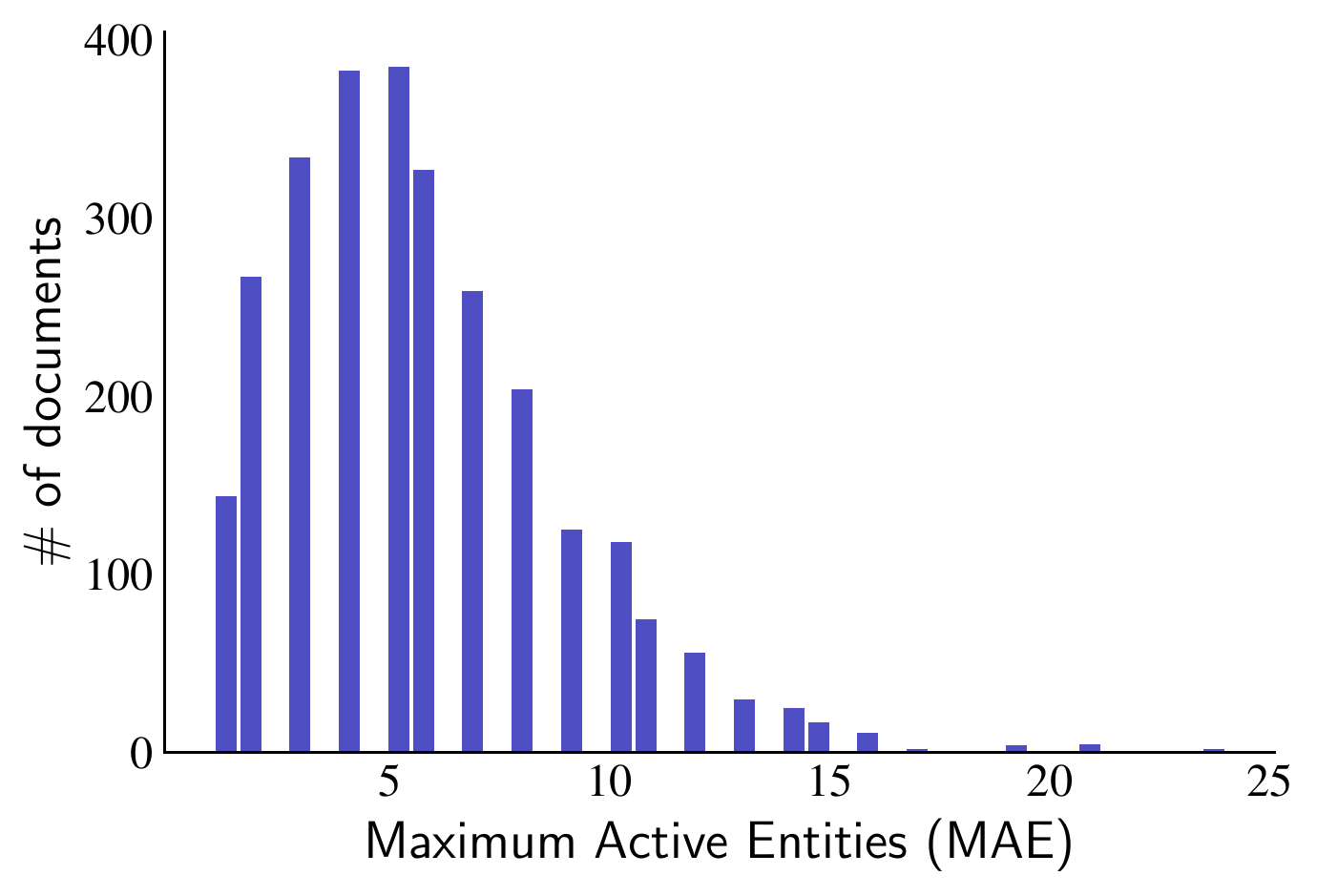}
        \caption{OntoNotes Training Set.}
        \label{fig:ontonotes_active}
    \end{subfigure}
    \caption{Histograms of Maximum Active Entities for documents in LitBank and OntoNotes.}
    \label{fig:active_chains}
\end{figure*}

\paragraph{Active Entity Count}
Active entity count $\textrm{AE}(t)$ at token index $t$  denotes the number of unique entities whose spread covers the token $t$, i.e.,
$$\textrm{AE}(t) = |\{e \;|\; t \in \textrm{ES}(e) \}|$$

\paragraph{Maximum Active Entity Count}
Maximum active entity count $\textrm{MAE}(\mathcal{D})$ for a document $\mathcal{D}$ denotes the maximum number of active entities at any token index in $\mathcal{D}$, i.e.,
$$\textrm{MAE}(\mathcal{D}) = \max_{t \in [|\mathcal{D}|]} \textrm{AE}(t)$$ 
This measure can be simply extended to a corpus $\mathcal{C}$ as: $$\textrm{MAE}(\mathcal{C}) = \max_{\mathcal{D} \in \mathcal{C}} \textrm{MAE}(\mathcal{D})$$ %

 Figure~\ref{fig:active_chains} visualizes the histograms of Maximum Active Entity Count (MAE) for documents in LitBank and OntoNotes (training set). 
To illustrate the difference between MAE and maximum total entity count i.e.\ maximum clusters in a single document, Table~\ref{sec:tab_active_entities} presents this comparison %
 for LitBank and OntoNotes. %
For both datasets the maximum active entity count is much smaller than the maximum total entity count.
Thus, rather than keeping all the entities in memory at all times, models can in principle simply focus on the far fewer active entities at any given time.

\subsection{Model Details}
\label{sec:longdoc_model}
We next describe memory models which track entity cluster representations in their external memory.  
Based on the findings from the previous section, we also propose model variants that require tracking only a small, bounded number of entities at any time.

The models are composed of three modules: (a) Document Encoder which encodes the document, (b) Mention Detector which proposes candidate mentions i.e.\ text spans which reference an entity, and the (c) Mention Clustering module which clusters the candidate spans proposed by the detector module. 
The mention clustering module processes the candidate mentions sequentially and performs one of the three actions: (a) adds the mention to an existing entity cluster, (b) creates a new cluster, %
or (c) ignores the mention due to limited memory capacity (for bounded memory models).
The three modules are tied together by a shared, differentiable mention representation obtained from the encoded document as in \citet{lee-etal-2017-end}, and this also makes the model end-to-end trainable. 
Next we describe the three modules in detail.  %

\subsubsection{Document Encoder}
\label{sec:doc_encoder}
Document encoding is done using the Longformer\textsubscript{LARGE}~\cite{beltagy2020longformer} model which is finetuned jointly along with the rest of the model parameters.\footnote{The finetuned document encoders are available on Huggingface model hub. For example, here's the link to the encoder finetuned for OntoNotes \url{https://huggingface.co/shtoshni/longformer_coreference_ontonotes}} 
To encoder documents beyond the window size of the Longformer\textsubscript{LARGE} encoder,  we chunk the document into maximal possible independent chunks while segmenting at the sentence boundary. 
Unless otherwise stated, we train all models with the publicly released Longformer\textsubscript{LARGE} model's maximum chunk length of 4096 tokens.\footnote{\url{https://huggingface.co/allenai/longformer-large-4096}} 

Note that when we first presented the work in \citet{toshniwal-etal-2020-learning}, we used the SpanBERT\textsubscript{LARGE} model finetuned for OntoNotes and released as part of the coreference model of \citet{joshi-etal-2020-spanbert}. In subsequent experiments we found switching to Longformer\textsubscript{LARGE} improved performance. One of the reasons for improvement in performance with Longformer\textsubscript{LARGE} is the reduction in \emph{context fragmentation}~\cite{dai-etal-2019-transformer}. SpanBERT\textsubscript{LARGE} has a maximum input size of 512 tokens while for Longformer\textsubscript{LARGE} it is 4096 tokens. Comparing the input size in terms of subword tokens is not quite precise since the vocabularies for the two models are different.  For LitBank, on average Longformer tokenized documents are 2\% shorter than using the SpanBERT tokenizer (2342 vs 2392 tokens) which implies that the difference in vocabulary only exacerbates the context fragmentation issue for SpanBERT. 
We also empirically validate the impact of context fragmentation for LitBank in Section~\ref{sec:context_frag} where we show that for the same encoder, Longformer\textsubscript{LARGE}, encoding documents with longer contexts results in better performance.  

\subsubsection{Mention Detector}
\label{sec:ment_proposal}
Given a document $\mathcal{D}$, we score all candidate mentions $\mathcal{X}$ of some maximum length $L$ ($L = 20$) subword tokens using a learned scoring function $s_m(.)$. 
For a span $x \in \mathcal{X}$, $s_m(x)$ represents how likely is span $x$ an entity mention and is trained to assign positive score to gold mentions (any mention in a gold cluster), and negative score otherwise.

\paragraph{Mention/Span Representation} We use the span representation proposed by \citet{lee-etal-2017-end} which we describe next. Given a span $x$, let $\vec{x}_s$ and $\vec{x}_e$ represent the contextual representation of START($x$) and END($x$) respectively. Let $\hat{\vec{x}}$ represent the weighted combination of contextual representation of tokens in span   $x$ calculated as follows:

\begin{align*}
    a_t &= \frac{\textrm{exp}(\vec{w}_a \cdot \vec{x}_t)}{\sum\limits_{i=\textrm{START(}x\textrm{)}}^{\textrm{END(}x\textrm{)}} \textrm{exp}(\vec{w}_a \cdot \vec{x}_i)}\\
    \hat{\vec{x}} &= \sum_{i=\textrm{START(}x\textrm{)}}^{\textrm{END(}x\textrm{)}} a_i \cdot \vec{x}_i
\end{align*}

Finally, let $\phi(x)$ represent the feature vector for length of span $x$. The span representation $r(x)$ for span $x$ is then given by:
$$r(x) = [\vec{x}_s; \vec{x}_e; \hat{\vec{x}}; \phi(x)]$$

While most follow up work since \citet{lee-etal-2017-end}, including ours, has largely stuck by the above presented span representation, some recent work has explored alternate span representations. \citet{kirstain-etal-2021-coreference} proposed using the just the end points representations i.e.\ $\vec{x}_s$ and $\vec{x}_e$ to represent the span (even if they claim not using span representations!) while \citet{dobrovolskii-2021-word} get rid of span representations altogether and make coreference predictions at word-level (and only predict the mention span given the word after clustering). Our own work  \citet{toshniwal2020cross} explores the suitability of different span representations for different linguistic tasks and concludes that the choice of span representation is more important when the document encoder is fixed/frozen than when it is finetuned.

\paragraph{Mention Selection} During training, we use teacher forcing and only pass the gold mentions which are among the top-$K$ mentions where $K = 0.4 \times |\mathcal{D}|$ to the clustering step.
During inference, we only pass mentions with a positive mention score i.e. $\{x: x \in \mathcal{X}, s_m(x) \geq 0\}$ to the clustering step.
This difference in mention selection between training and inference is because during initial phases of training, the scoring function can be quite unreliable and using it to filter mentions could lead to noisy mentions being passed to the clustering module. Rather than training the clustering module to deal with noisy mentions, we prefer a high precision mention detector as argued by \citet{wu2020understanding}, and only train the clustering module with gold mentions. Future work can explore strategies on exposing the clustering module to high-scoring negative mentions.

\paragraph{Iterative Mention Detector Training: Augmenting Pseudo-Singletons}
OntoNotes~\cite{weischedel2013ontonotes} famously does not annotate singletons.\footnote{Singletons are annotated in some of the more recent coreference datasets such as   Preco~\cite{chen-etal-2018-preco} and LitBank~\cite{bamman2019annotated}.
}  Thus, a perfect mention detector model for OntoNotes   
needs to predict a mention span's anaphoricity to decide if a span is a mention or not. 
This leads to a chicken and egg situation where a mention detector module needs to perform clustering to identify mentions which ultimately can be clustered.
\citet{wu2020understanding} show that mention detectors struggle with taking anaphoricity into account while predicting mentions. Prior work by \cite{lee-etal-2017-end, lee-etal-2018-higher, joshi-etal-2019-bert} circumvent this issue by taking a high recall approach to predicting mentions which has the side effect of a drop in precision and as a result the clustering module needs to deal with a large number of noisy mentions. However, \citet{wu2020understanding} argue that high precision mention detectors can have a more significant impact on the overall performance than high recall mention detectors. 

\emph{How to address the lack of singleton annotation in OntoNotes and its resulting impact on mention detectors?}
We exploit the observation from \citet{wu2020understanding} that  mention detectors struggle with taking anaphoricity into account while predicting mentions.
Thus, high-scoring spans that are not part of the gold annotation can be treated as singletons or rather \textit{pseudo-singletons}.  
Our recipe for training a mention detector is then to: (a) first train a standalone mention detector, (b) select the top-scoring mentions outside the gold mentions and treat them as singleton clusters for the next iteration, and (c) train the full coreference model, including the mention detector, from scratch with the augmented data from previous step.   
In particular, for OntoNotes we experiment with adding 0, 30K, and 60K 
pseudo-singletons (in total, there are 156K gold mentions) in the second and final iteration of training the full coreference model (result in Section~\ref{sec:singleton_ontonotes}). In all our experiments with OntoNotes, we find it beneficial to add pseudo-singletons.\footnote{Pseudo-singletons can be downloaded from \href{https://drive.google.com/drive/folders/1OdkX4xdhaKlkDM97U4rMNIxtzGi_BnNB?usp=sharing}{this link}} Later in our discussion on generalization of coreference models in Section~\ref{sec:gen_results}, we will see that pseudo-singletons also aid the out-of-domain performance of OntoNotes trained coreference models.   

\emph{Note on Subtleties of Subword Tokenizers:} 
 I (like many others working on coreference resolution) had adapted my model from Kenton Lee's model~\cite{lee-etal-2017-end, lee-etal-2018-higher}.\footnote{\url{https://github.com/kentonl/e2e-coref}}   In Kenton's model, due to the choice of ElMo \cite{peters-etal-2018-deep} as an encoder, the candidate spans  always adhered to word boundaries. While porting the model to the era of subword tokenization based encoders, I didn't constrain the mention detector to respect word boundaries. Because by the end of the training the model rarely predicted mentions violating the word boundary mistake, I never caught the ``bug", and it remained in the codebase for a long time until Andreas van Cranenburgh raised an issue of duplicate mentions in the prediction.\footnote{\url{https://github.com/shtoshni/fast-coref/issues/3}}
 At first glance, the problem sounded absurd because it suggests that the mention detector is predicting duplicate candidate mentions which I was sure of as being impossible given the implementation. But then I realized that while the mention detector is predicting unique candidate spans in the space of subword boundaries, these spans can be duplicates when mapped to word boundaries. 
  By simply adding the constraint of word boundaries, there's a significant reduction in the number of candidate mentions which ultimately lead to a significant improvement in performance. 
  All the results presented in this chapter are with the addition of the word boundary constraint. 
  Ending this discussion on a speculative note, I believe the success of the  recent word-level coreference model by \citet{dobrovolskii-2021-word} can be partially attributed to the model baking in the word boundary constraint by design.

\subsubsection{Mention Clustering}
Let $(x_i)_{i=1}^K$ represent the top-$K$ candidate mention spans from the mention proposal step. %
Assume that the mentions are already ordered based on their position in the document and are processed sequentially in that order.\footnote{Specifically, they are ordered based on START($\cdot$) index with ties broken using END($\cdot$).}
Let $E = (e_m)_{m=1}^M$ represent the $M$ entities currently being tracked by the model (initially $M = 0$).
For ease of discussion, we will overload the terms $x_i$ and $e_j$ to also correspond to their respective representations.

In the \emph{first} step, the model decides whether the span $x_i$ refers to any of the entities in $E$ as follows:
\begin{align*}
s_c(x_i, e_j) &\!=\! f_c([x_i; e_j; x_i \odot e_j; g(x_i, e_j)])\\ %
s_c^{\mathit{top}} &\!=\! \max_{j=1 \dotsc M} s_c(x_i, e_j)\\
e^{\mathit{top}} &\!=\! \underset{{j=1 \dotsc M}}{\arg\max}\; s_c(x_i, e_j)
\end{align*} %
where $\odot$ represents the element-wise product, and $f_c(\cdot)$ corresponds to a learned feedforward neural network.
The term $g(x_i, e_j)$ correponds to a concatenation of feature embeddings that includes embeddings for (a) number of mentions in $e_j$, (b) number of tokens between $x_i$ and last mention of $e_j$, and optionally (c) metadata information such as document genre in OntoNotes. %

Now if $s_c^{\mathit{top}} > 0$ then $x_i$ is considered to refer to $e^{\mathit{top}}$, and $e^{\mathit{top}}$ is updated accordingly.\footnote{We use weighted averaging where the weight for $e^{\mathit{top}}$ corresponds to the number of previous mentions seen for  $e^{\mathit{top}}$.}
Otherwise, $x_i$ does not refer to any entity in $E$ and a \emph{second} step is executed, which will depend on the choice of memory architecture.
We test three
 memory architectures, described below.

\begin{enumerate}%
    \item \textbf{Unbounded Memory (\unbounded)}: We create a new entity $e_{M + 1} = x_i$ and append it to $E$.

    \item \textbf{Bounded Memory}: Suppose the model has a capacity of tracking $C$ entities at a time.
    If $C > M$, i.e., the memory capacity has not been fully utilized, then the model behaves like U-MEM. %
    Otherwise, the bounded memory models must decide between: (a) evicting an entity already being tracked, or (b) ignoring $x_i$ due to limited capacity. %
    We test two bounded memory variants that are described below.

    \begin{enumerate}%
        \item \textbf{Learned Bounded Memory (\learned)}:
        The proposed \learned architecture tries to predict a score $f_r(.)$ corresponding to the anticipated number of remaining mentions for any entity or mention%
        as follows:
        \begin{align*}
            &d = \arg\min [f_r(e_1), \dotsc, f_r(e_M), f_r(x_i)]
        \end{align*}
        where $f_r(\cdot)$ is a learned feedforward neural network. %
        If $1 \le d \le M$ then then the model evicts the previous entity $e_d$ and reinitialize it to $x_i$.
        Otherwise if $d = M + 1$ then the model ignores $x_i$ due to limited capacity.
        \item \textbf{Rule-based Bounded Memory (\lru)}
        The Least Recently Used (LRU) principle is a popular choice among memory models~\citep{rae2016scaling, santoro2016one}.
        While \learned  considers all potential entities for eviction, with \lru  this choice is restricted to just the LRU entity, i.e., the entity whose mention was least recently seen.
        The rest of the steps are similar to the \learned model.
    \end{enumerate}
\end{enumerate}

\paragraph{Training}
All the models are trained as an autoregressive model via teacher forcing where the oracle clustering actions are determined given the predicted mentions.
The oracle clustering actions for bounded memory models are chosen to maximize the number of mentions tracked by the model. %
For the bounded memory models, we keep growing the number of entities till we hit the memory ceiling.
For all the entities in memory, we maintain the number of mentions remaining in the ground truth cluster.
For example, a cluster with a total of five mentions, two of which have already been processed by the model, has three remaining mentions.

Suppose now a mention corresponding to a currently untracked entity comes in and the memory is already at full capacity.
Then for the \learned model, we compare the number of mentions of this new entity (along with the current mention) against the number of mentions remaining for all the entities currently being tracked. If there are entities in memory with number of remaining mentions less than or equal to the number of mentions of this currently untracked entity, then the untracked entity replaces the entity with the least number of remaining mentions. Ties among the entities with least number of remaining mentions are broken by the least recently seen entity. If there's no such entity in the memory, then the mention is ignored.
For the \lru model, the comparison is done in a similar way but is limited to just the least recently seen (LRU) entity.
Finally, the training loss is calculated via the addition of the cross-entropy losses for the two steps of mention clustering.

\subsection{Experimental Setup}

\subsection*{Datasets}
\label{sec:long_doc_datasets}

\paragraph{OntoNotes 5.0 (ON)} \cite{weischedel2013ontonotes} is a collection of news-like, web, and religious texts spanning seven distinct genres. 
Notably, it does not contain singleton annotations. %
Some genres are transcripts (phone conversations and news). We assume that transcripts are annotated with speaker information, and unless otherwise stated, we encode the speaker information directly in text via \texttt{[SPEAKER\_START] Speaker Name [SPEAKER\_END]} as in \citet{wu-etal-2020-corefqa}.

\paragraph{LitBank (LB)} \cite{bamman2019annotated} is a set of public domain works of literature drawn from Project Gutenberg. On average, coreference in the first 2,000 tokens of each work is fully annotated for six entity types.\footnote{They are people, facilities, locations, geopolitical entities, organizations, and vehicles.}
Singletons are marked and used for evaluation.  %
Evaluation is done via 10-fold cross-validation over 80/10/10 splits.\footnote{\url{https://github.com/dbamman/lrec2020-coref/tree/master/data}}

\begin{table}[t]
\parbox{.45\linewidth}{
\centering
\small

\begin{tabular}{lcc}
\toprule
Model & Dev F1 & Test F1 \\\midrule
\unbounded & 80.5 & 80.2 \\
\learned &  & \\
\hspace{0.1in} 5 cells & 74.4 & 73.8 \\
\hspace{0.1in} 10 cells & 78.5 & 77.9 \\
\hspace{0.1in} 20 cells & 79.6 & 78.8 \\
\lru & & \\
\hspace{0.1in} 5 cells & 64.8 & 63.5 \\
\hspace{0.1in} 10 cells & 72.6 & 71.6 \\
\hspace{0.1in} 20 cells & 78.0 & 77.1  \\\midrule
\citet{bamman2019annotated} & - & 68.1 \\
\citet{toshniwal-etal-2020-learning} & 76.5 & 75.9 \\
\citet{xia-van-durme-2021-moving} & - & 76.7 \\
\citet{thirukovalluru-etal-2021-scaling} & - & 78.4 \\
\bottomrule
\end{tabular}
\caption{Results for LitBank (CoNLL F1).}
\label{tab:res_litbank}
}
\hfill
\parbox{.48\linewidth}{
\centering
\small
\begin{tabular}{lcc}
\toprule
Model & Dev F1 & Test F1 \\\midrule
\unbounded & 80.5 & 80.9  \\
\learned &  & \\
\hspace{0.1in} 5 cells  & 72.6 & 71.8  \\
\hspace{0.1in} 10 cells  & 78.2 & 78.5  \\
\hspace{0.1in} 20 cells  & 79.6 & 79.9  \\
\lru & & \\
\hspace{0.1in} 5 cells   & 62.6 & 61.4  \\
\hspace{0.1in} 10 cells   & 74.9 & 74.3  \\
\hspace{0.1in} 20 cells   & 78.6 & 78.8  \\\midrule
\citet{joshi-etal-2020-spanbert} & 80.1 & 79.6 \\
\citet{wu2019coreference} & 83.4 & 83.1 \\
\citet{toshniwal-etal-2020-learning} & 79.6 & 79.6 \\
\citet{xia-etal-2020-incremental}) & 79.7 & 79.4 \\
\citet{xu-choi-2020-revealing} & 79.9 & 80.2 \\
\citet{kirstain-etal-2021-coreference} & - & 80.3 \\
\citet{dobrovolskii-2021-word} & 80.7 & 81.0 \\
\bottomrule
\end{tabular}
\caption{Results for OntoNotes (CoNLL F1).}
\label{tab:res_ontonotes}

}
\end{table}

\subsection{Results}

Tables \ref{tab:res_litbank} and~\ref{tab:res_ontonotes} show results of all the proposed models for LitBank and OntoNotes respectively. 
Detailed results with the performance on the constituent CoNLL metrics are presented in Table~\ref{tab:full_litbank} for LitBank, and Table~\ref{tab:full_ontonotes} for OntoNotes (Section~\ref{sec:additional_res}).
As expected, the bounded memory models improve with increase in memory.
For both the datasets, the \learned model with 20 memory cells is competitive with the \unbounded model though the gap is non-trivial. 
Among the bounded memory models, the \learned model is significantly better than \lru for lower numbers of memory cells. We analyze the reasons for this in the next section.

By default all the models are trained with a maximum context of 4096 tokens (maximum input size  of Longformer\textsubscript{LARGE} encoder), but we found that for LitBank truncating the document to 1024 tokens improved performance for all the \lru settings and also improved the performance of the \learned model with 5 memory cells. For higher memory sizes for the \learned model, the default maximum context size of 4096 tokens worked the best. The reason that document truncation benefits the bounded memory models, especially in low memory settings, is because a bounded memory model is simultaneously being trained for mention detection, mention clustering, and memory management. Without truncation, there are more entities to deal with, and hence, a higher memory management loss which lead to reduced mention detection performance and ultimately worse overall performance. 
A better training curriculum such as pretraining the mention detector could be helpful, but we leave this exploration for future work. 

For OntoNotes, we found the addition of \emph{pseudo-singletons}, as described in Section~\ref{sec:ment_proposal}, to benefit the \unbounded and \learned model. Addition of 60K pseudo-singletons worked the best for the \unbounded model while adding 30K pseudo-singletons gave the best result for the \learned model with 10 and 20 memory cells. For all other bounded models, adding no pseudo-singletons gave the best result. We further discuss the impact of pseudo-singletons in Section~\ref{sec:gen_results}.

Comparing our proposed models to prior work, we find that the \unbounded and the \learned model significantly improve over the previous state-of-the-art for LitBank.
For OntoNotes, the \unbounded model is competitive with most of the prior work with the notable exception of \citet{wu-etal-2020-corefqa}.  
The CorefQA model by \citet{wu-etal-2020-corefqa}, unlike all other prior results presented here, requires conditionally encoding the document for at least each entity cluster separately which makes it quite slow and ultimately unsuitable for long documents. Future work can look into bridging the performance gap between CorefQA and other models while being scalable to long documents.

\subsection{Analysis}
In this section we analyze the behavior of the three memory models on LitBank and OntoNotes.

\begin{table}[t!]

\centering{
\small{

\begin{tabular}{l ccc}
\toprule
\multirow{2}{*}{Model} & Peak training   & Peak inference  & Inference  \\
     & mem. (in GB) &  mem. (in GB) &   time (in s)\\\midrule
\unbounded              &  11.6   &    3.1 &  29.25 \\
\learned    &      &                &            \\
\hspace{0.1in} 5 cells &  \phantom{1}8.0     &  3.2   &  27.31 \\
\hspace{0.1in} 10 cells &  \phantom{1}8.4    &  3.2   &  27.44 \\
\hspace{0.1in} 20 cells &  \phantom{1}9.1   &   3.2   &  27.86 \\
\lru & &  &  \\
\hspace{0.1in} 5 cells &  \phantom{1}8.0   &  3.2   & 26.19   \\
\hspace{0.1in} 10 cells &  \phantom{1}8.3   & 3.2   & 26.50  \\
\hspace{0.1in} 20 cells &  \phantom{1}8.9   &  3.2  & 26.19  \\\bottomrule
\end{tabular}
}
}
\caption{Peak memory and inference time statistics for the LitBank cross-validation split $0$.}%
\label{tab:mem_stats}
\end{table}

\paragraph{Memory Utilization}
Table~\ref{tab:mem_stats} compares the memory and inference time statistics for the different memory models for the LitBank cross-validation split zero.\footnote{Peak memory usage estimated via \texttt{torch.cuda.max\_memory\_allocated()}}
For training, the bounded memory models are significantly less memory intensive than the \unbounded model.
The table also shows that the bounded memory models are faster than the \unbounded memory model during inference (inference time calculated by averaging over three runs).
This is because the number of entities tracked by the \unbounded memory model grows well beyond the maximum of 20 memory slots reserved for the bounded models as shown in Table~\ref{tab:num_ents}.

Surprisingly, for inference we see that the bounded models have a slightly larger memory footprint than the \unbounded model.
This is because the document encoder, Longformer, dominates the memory usage during inference (as also observed by \citet{xia-etal-2020-incremental}).
Thus the peak memory usage during inference is determined by the mention proposal stage rather than the mention clustering stage.
And during the mention proposal stage, the additional parameters of bounded memory models, which are loaded as part of the whole model, cause the slight uptick in peak inference memory.
Note that using a cheaper encoder or running on a sufficiently long document, such as a book,
can change these results.

\begin{table}[t]

\centering{
\small{

\begin{tabular}{l c c c c}
\toprule
\multirow{2}{*}{Model} & \multicolumn{2}{c}{LitBank} & \multicolumn{2}{c}{OntoNotes} \\
     & Avg & Max & Avg & Max\\\midrule
\unbounded  & 81.5  & 180  & 35.0   &  185 \\

\learned    &      &                &           & \\

\hspace{0.1in} 5 cells 
& \phantom{1}5.0  & \phantom{11}5   & \phantom{1}4.5       & \phantom{11}5 \\

\hspace{0.1in} 10 cells 
& 10.0 & \phantom{1}10 & \phantom{1}7.8 & \phantom{1}10 \\

\hspace{0.1in} 20 cells 
& 20.0 & \phantom{1}20 &  15.0       & \phantom{1}20 \\

\lru & &  & & \\

\hspace{0.1in} 5 cells 
& \phantom{1}5.0 & \phantom{11}5    & \phantom{1}4.5       & \phantom{11}5   \\

\hspace{0.1in} 10 cells 
& 10.0 & \phantom{1}10 & \phantom{1}8.3 & \phantom{1}10 \\

\hspace{0.1in} 20 cells 
& 20.0 & \phantom{1}20 & 13.9      & \phantom{1}20 \\\bottomrule
\end{tabular}
}
}
\caption{Comparison of number of entities in memory.}
\label{tab:num_ents}

\end{table}
\paragraph{Number of Entities in Memory}
Table~\ref{tab:num_ents} compares the maximum number of entities kept in memory by the different memory models for the LitBank cross-validation dev sets and the OntoNotes dev set.
As expected, on average the \unbounded model keeps far more entities in memory than the bounded memory models for both the datasets.
For LitBank the difference is especially stark with the  \unbounded model tracking about 4/9 times more entities in memory on average/worst case, respectively. 
While the maximum memory usage for both the datasets is comparable, on average models track fewer entities for OntoNotes documents. 
This is especially evident for bounded memory models where for some OntoNotes documents these models do not even use the full 5 memory cell capacity, all LitBank documents fully utilize even the 20 memory cell capacity. 
This is because %
LitBank documents are more than four times as long as OntoNotes documents on average, and LitBank has singletons marked. These results also justify our initial motivation that with long documents, the memory requirement will increase even if we only keep the entity representations.

\paragraph{\learned vs. \lru}

\begin{table}[t!]
\centering{
\small{

\addtolength{\tabcolsep}{-3pt}
\begin{tabular}{c cccc}
\toprule
Memory & \multicolumn{2}{c}{LitBank} & \multicolumn{2}{c}{OntoNotes} \\
size     & \learned & \lru & \learned & \lru\\\midrule
\phantom{1}5 
& \phantom{1}1.6  & 38.2        & 0.9  & 4.5 \\
10 
& \phantom{1}0.0 & 19.9         & 0.0 & 0.7 \\
20 
& \phantom{1}1.1 & \phantom{1}1.4 & 0.0  & 0.1 \\\bottomrule
\end{tabular}
}
}
\caption{Average number of mentions ignored by the two bounded memory models.}
\label{tab:num_ignored}

\end{table}

Table~\ref{tab:num_ignored} compares the number of mentions ignored by \learned and \lru.
The \learned model ignores far fewer mentions than \lru.
This is because while the \lru model can only evict the LRU entity, which might not be optimal,
the \learned model can choose any entity for eviction.
These statistics combined with the fact that the \learned model typically outperforms \lru mean
that the \learned model is able to anticipate which entities are important and which are not.

\begin{table}[t!]

\centering{
\small{
\begin{tabular}{l c c c c c c}
\toprule
Model &  CE  & EM & EE & DE & MM & ME \\\midrule
\unbounded 
& \phantom{1}742& 451& 518& \phantom{1}\textbf{794}& \phantom{1}\textbf{575}& \phantom{1}\textbf{547} \\
\learned    &      &     &      &    &   &  \\
\hspace{0.1in}  5 cells 
& \phantom{1}962& 465& 318& 1135& 1015& 1003 \\
\hspace{0.1in}  10 cells 
& \phantom{1}902& 576& 399& 1034& \phantom{1}674& \phantom{1}658 \\
\hspace{0.1in}  20 cells 
& \phantom{1}\textbf{707}& \textbf{394}& 445& \phantom{1}904& \phantom{1}645& \phantom{1}688 \\
\lru    &      &     &      &    &   & \\
\hspace{0.1in}  5 cells 
& 1527& 520& \textbf{250}& 1736& 1559& 1218 \\
\hspace{0.1in}  10 cells 
& 1069& 614& 402& 1145& \phantom{1}748& \phantom{1}713 \\
\hspace{0.1in}  20 cells 
& \phantom{1}816& 529& 388& \phantom{1}952& \phantom{1}618& \phantom{1}687 \\\bottomrule
\end{tabular}
}
}
\caption{Error Analysis for OntoNotes dev set. CE=Conflated Entities, DE=Divided Entity, EM=Extra Mention, EE=Extra Entity, MM=Missing Mention, ME=Missing Entity.}
\label{tab:err_analysis}

\end{table}

\paragraph{Error Analysis}
Table~\ref{tab:err_analysis} presents the results of automated error analysis done using the Berkeley Coreference Analyzer~\citep{kummerfeld2013error} for the OntoNotes dev set.
As the memory capacity of models increases, the errors shift from missing mention and missing entity, to extra mention and extra entity categories. The bounded memory models perform worse than the \unbounded model in divided entity, missing mention, and missing entity category which is because the bounded models are forced to track a limited number of entities.  
For limited memory settings, the \learned model outperforms \lru in terms of tracking more entities. %

\subsection{Conclusion}
We proposed a variety of memory models for coreference resolution which track entity cluster representations in their external memory. 
In particular, we proposed a bounded memory model with a learned memory management scheme which guarantees a linear runtime in document length in theory, and  in practice significantly reduces peak memory usage during training and inference time in comparison to its unbounded memory counterpart. 
The models are competitive with prior work for OntoNotes and LitBank, with the unbounded variant establishing the new state-of-the-art performance for LitBank.
Results for LitBank, a long document coreference dataset, in particular demonstrate that our models are scalable and high performing.  

Till this point, we've limited our focus to \emph{in-domain} evaluation i.e. training and testing the model on the same domain. Coreference resolution being a core component of the NLP pipeline, a coreference resolution model is expected to be used in a wide range of domains. In the next section, we focus on the problem on generalization in coreference resolution. We use the \unbounded model proposed in this section for conducting the generalization experiments.

\section{Generalization in Coreference Resolution}
\label{sec:generalization}
A key metric to establishing the usability of a machine learning model is how robust is the model in the wild. 
In this section, we test the generalization performance of coreference resolution models described in the previous section. Specifically, we will be testing the unbounded memory model (U-MEM) for its zero-shot capabilities i.e. performance on datasets unseen during training. 
To this end, we consolidate a set of 8 coreference resolution datasets targeting different domains to evaluate the off-the-shelf performance of models. 
We then mix three datasets for training; even though their domain, annotation guidelines, and metadata differ, we propose a method for jointly training a single model on this heterogeneous data mixture by using data augmentation to account for annotation differences and sampling to balance the data quantities. 
We find that in a zero-shot setting, models trained on a single dataset transfer poorly while joint training moderately improves the out-of-domain performance.  
We also find that annotation differences across datasets exaggerate the domain transfer challenge.  
Finally, at the time of this writing, the models presented here have the state-of-the-performance in three coreference benchmarks, namely PreCo, LitBank, and WikiCoref.   

\subsection{Introduction}
Coreference resolution is a core component of the NLP pipeline, as determining which mentions in text refer to the same entity is used %
for a wide variety of downstream tasks like knowledge extraction \cite{li-etal-2020-gaia}, question answering \cite{dhingra-etal-2018-neural}, and dialog systems \cite{gao-etal-2019-interconnected}. 
As these tasks span many domains, we need coreference models to generalize well. 

Meanwhile, 
models for coreference resolution have improved due to neural architectures with millions of parameters and the emergence of pretrained %
encoders. However, model generalization across domains has always been a challenge %
\cite{yang-etal-2012-domain, zhao-ng-2014-domain, poot-van-cranenburgh-2020-benchmark, aktas-etal-2020-adapting, moosavi-thesis}. Since these models are %
usually engineered for a single dataset, they capture idiosyncrasies inherent in that dataset. %
As an example, OntoNotes \cite{weischedel2013ontonotes}, a widely-used general-purpose dataset, provides metadata, like the document genre and speaker information. However, this assumption cannot be made more broadly, especially if the input is raw text~\cite{wiseman-etal-2016-antecedent}. %

Furthermore, while there are datasets aimed at capturing a broad set of genres \cite{weischedel2013ontonotes, poesio-etal-2018-anaphora, zhu-etal-2021-ontogum}, they are not mutually compatible due to differences in annotation guidelines. For example, some datasets do not annotate singleton clusters (clusters with a single mention). %
Ideally, we would like a coreference model to be robust to all the standard datasets. 
In this work, we consolidate 8 datasets spanning multiple domains, document lengths, and annotation guidelines. We use them to evaluate the off-the-shelf performance of models trained on a single dataset. While they perform well %
within-domain (e.g., a new state-of-the-art of 80.2 F1 on LitBank), they still perform poorly out-of-domain. 

To address poor out-of-domain performance, we propose joint training for coreference resolution, which is challenging due to the incompatible training procedues for different datasets.  Among other things, we need to address (unannotated) singleton clusters, as OntoNotes does not include singleton annotations. We propose adding  \textit{pseudo-singletons}, described in Section~\ref{sec:ment_proposal}, into the training data to match the other datasets which have gold singleton annotations.

Concretely, we contribute a benchmark for coreference to highlight the disparity in model performance and track generalization. 
We find joint training highly effective for modeling multiple datasets, though the improvement on out-of-domain evaluation is moderate.  
We find that our data augmentation method of adding pseudo-singletons is also effective.
We also find that the challenge of out-of-domain evaluation can be exaggerated by differences in annotation guidelines across datasets (e.g., possessives are part of span annotation in OntoNotes but not in Quiz Bowl). 
Finally, the presented models establish new state-of-the-art for three coreference benchmarks, namely PreCo, LitBank, and WikiCoref.

\subsection{Datasets}
\label{sec:datasets}

\begin{table*}[t]
    \centering
    \small
    \setlength\tabcolsep{4pt} %
        \begin{tabular}{lccrccccc}
    \toprule
         & \multicolumn{3}{c}{Num. Docs} & Words/ &  Mentions/ &Mention & Cluster & \% of singleton \\
         Dataset & Train & Dev. & Test  & Doc &Doc &  length &  size & mentions\\
    \midrule
         OntoNotes  & \phantom{1}2802 & 343 & 348 & \phantom{1}467 & \phantom{1}56 & 2.3 & 4.4 &  \phantom{1}0.0 \\
         LitBank\textsuperscript{k} & \phantom{11}80 & \phantom{1}10 & 10 & 2105 & 291 & 2.0 & 3.7 & 19.8 \\
         PreCo & 36120 & 500 & 500  & \phantom{1}337 & 105 & 2.7 & 1.6 & 52.0 \\
         Character Identification & \phantom{11}987 & 122 & 192  & \phantom{1}262 & \phantom{1}36 & 1.0 & 5.1 &  \phantom{1}6.4 \\
          WikiCoref & \phantom{1111}0 & \phantom{11}0 & 30 & 1996 & 230 & 2.6 & 5.0 & \phantom{1}0.0 \\
         Quiz Bowl Coreference\textsuperscript{k} & \phantom{1111}0 & \phantom{11}0 & 400 & \phantom{1}126 & \phantom{1}24 & 2.7 & 2.0 & 26.0 \\
         GAP \textsuperscript{p} & \phantom{1}2000 & 400 & 2000 &  \phantom{11}95 & \phantom{11}3 & 2.0 & - & - \\
         WSC\textsuperscript{p} & \phantom{1111}0 & \phantom{11}0 & 271 & \phantom{11}16 & \phantom{11}3 & 1.5 & - & -  \\
    \bottomrule
    \end{tabular}
    \caption{Statistics of datasets. Datasets with \textsuperscript{k} indicate that prior work uses \textit{k}-fold cross-validation; we record the splits used in this work. Datasets with \textsuperscript{p} are partially annotated, so we do not include cluster details.}
    
    \label{tab:data}
\end{table*}

We organize our datasets into three types. %
Training datasets (Sec.~\ref{sec:data:train}) are large in terms of number of tokens and clusters and more suitable for training. Evaluation datasets (Sec.~\ref{sec:data:eval}) are out-of-domain compared to our training sets and are entirely held out. Analysis datasets (Sec.~\ref{sec:data:an}) contain annotations aimed at probing specific phenomena. \autoref{tab:data} lists the full statistics.

\subsubsection{Training Datasets}
\label{sec:data:train}
For training, we reuse the OntoNotes and LitBank datasets from the previous section which we briefly describe again for completeness sake. 

\paragraph{OntoNotes 5.0 (ON)} \cite{weischedel2013ontonotes} is a collection of news-like, web, and religious texts spanning seven distinct genres. Some genres are transcripts (phone conversations and news). As the primary training and evaluation set for developing coreference resolution models, many features specific to this corpus are tightly integrated into publicly released models. For example, the metadata includes information on the document genre and the speaker of every token (for spoken transcripts). Notably, it does not contain singleton annotations. %

\paragraph{LitBank (LB)} \cite{bamman2019annotated} is a set of public domain works of literature drawn from Project Gutenberg. On average, coreference in the first 2,000 tokens of each work is fully annotated for six entity types.\footnote{They are people, facilities, locations, geopolitical entities, organizations, and vehicles.} %
We only use the first cross-validation fold of LitBank, which we call LB\textsubscript{0}.

\paragraph{PreCo (PC)} \cite{chen-etal-2018-preco} contains documents from reading comprehension examinations, each fully annotated for coreference resolution. Notably, the corpus is the largest such dataset released. %

\subsubsection{Evaluation Datasets}
\label{sec:data:eval}

\paragraph{Character Identification (CI)} \cite{zhou-choi-2018-exist} has multiparty conversations derived from TV show transcripts. Each scene in an episode is considered a separate document. 
This character-centric dataset only annotates mentions of people. %

\paragraph{WikiCoref (WC)} \cite{ghaddar-langlais-2016-wikicoref} contains documents from English Wikipedia. This corpus contains sampled and annotated documents of different lengths, from 209 to 9,869 tokens. 

\paragraph{Quiz Bowl Coreference (QBC)} \cite{guha-etal-2015-removing} contains questions from Quiz Bowl, a trivia competition. These paragraph-long questions are dense with entities. Only certain entity types (titles, authors, characters, and answers) are annotated.

\subsubsection{Analysis Datasets}
\label{sec:data:an}

\paragraph{Gendered Ambiguous Pronouns (GAP)} \cite{webster2018gap} is a corpus of ambiguous pronoun-name pairs derived from Wikipedia. While only pronoun-name pairs are annotated, they are provided alongside their full-document context. This corpus has been previously used to study gender bias in coreference resolution systems. %

\paragraph{Winograd Schema Challenge (WSC)} \cite{levasque2012wino} is a challenge dataset for measuring common sense in AI systems.\footnote{\url{https://cs.nyu.edu/~davise/papers/WinogradSchemas/WSCollection.html}} 
Unlike the other datasets, each document contains one or two sentences with a multiple-choice question.
We manually align the multiple choices to the text and remove 2 of the 273 examples due to plurals.
\subsection{Models}

In this section, we first introduce the models trained on single datasets, including a model from prior work by \citet{xu-choi-2020-revealing}. We then introduce  models which are jointly trained on ON, PC, and LB\textsubscript{0}. 
As our model of choice, we will be building on the U-MEM model described in Section~\ref{sec:longdoc_model}. We used the name U-MEM in the previous section to differentiate the model from the bounded memory model variants. Since the ultimate goal of previous section was scalability to long documents, we refer to the U-MEM model as \emph{longdoc} henceforth.

\subsubsection{Single Dataset Models}

\paragraph{longdoc} 
We train the longdoc model separately for the three training datasets, namely ON, PC, and LB\textsubscript{0} which results in three different models. For ON, we also experiment with using speaker information, genre information (as metadata embedding), and pseudo-singletons. We experiment with adding 0, 30K, and 60K 
pseudo-singletons (in total, there are 156K gold mentions).

\paragraph{\citet{xu-choi-2020-revealing}}
extends a mention-ranking model \cite{lee-etal-2018-higher} by making modifications in the decoding step.
We run the off-the-shelf model trained on OntoNotes and released by \citet{xu-choi-2020-revealing} on the test sets of ON, LB\textsubscript{0}, PC, and QBC. LB\textsubscript{0} requires a 24GB GPU, while WC runs out of memory even on that hardware. The model shows strong in-domain performance with 80.2 on ON. However, out-of-domain performance is weak: 57.2 on LB\textsubscript{0}, 49.3 on PC, and 37.6 on QBC. These are roughly on par with the ON longdoc models.

\subsubsection{Joint Training}

With copious amounts of text in OntoNotes, PreCo, and LitBank, we can train a joint model on the combined dataset. However, this is impractical as the annotation guidelines between the datasets are misaligned (OntoNotes does not annotate singletons and uses metadata) and because there are substantially more documents in PreCo. 

\paragraph{Augmenting Singletons} 
Since OntoNotes does not annotate for singletons, our training objective for OntoNotes is different from that of PreCo and LitBank. 
To address this, we experiment with adding pseudo-singletons in the joint training setup as well. We again experiment with adding 0, 30K, and 60K 
pseudo-singletons to OntoNotes in the joint training setup. 

\paragraph{Data Imbalance}
PreCo has 36K training documents, compared to 2.8K and 80 training documents for OntoNotes and 
LitBank respectively. A naive dataset-agnostic sampling strategy would mostly sample PreCo documents. To address this issue, we downsample OntoNotes and PreCo to 0.5K documents per epoch. Downsampling to 1K documents per epoch led to slightly worse performance (Section \ref{sec:downsampling joint}).

\paragraph{Metadata Embeddings}
For the joint model to be applicable to unknown domains, we avoid using any domain or dataset-identity embeddings, including the OntoNotes genre embedding.  
We do make use of speaker identity in the joint model because: (a) this is possible to obtain in conversational and dialog data, and (b) it does not affect other datasets that are known to be single-speaker at test time. %

\subsubsection{Training Details}
We train all the longdoc models for 100K gradient steps with a batch size of 1 document. Only the LB-only models are trained for 8K gradient steps which corresponds to 100 epochs for LB. The models are evaluated a total of 20 times (every 5K training steps) for all models except the LB-only models which are evaluated every 400 steps. 
We use early stopping and a patience of 5 i.e.\ training stops if the validation performance doesn't improve for 5 consecutive evaluations. 

For optimizer, we use AdamW with a weight decay of 0.01 and initial learning rate of 1e-5 for the Longformer encoder, and Adam with an initial rate of 3e-4 for the rest of the model parameters. The learning rate is linearly decayed throughout the training.

\begin{table*}[t]
    \centering
    \footnotesize
    \begin{tabular}{llcccccccccc}%
    \toprule
          Model & Training & ON & LB\textsubscript{0} & PC & CI & WC & QBC & GAP & WSC & All Avg. & OOD Avg.\\
   \midrule
        longdoc &  ON & 
        79.6 & 56.7 & 44.1 & 49.7 & 61.0 & 37.1 & 89.4 & 64.2 &  60.2& 60.3\\
        longdoc \textsuperscript{S} &  ON & 
        79.9 & 55.6 & 44.1 & \textbf{60.4} & 61.2 & 37.2 & 89.4 & 62.7 &  61.3 & 62.2\\
        longdoc \textsuperscript{S, G} &  ON  & 
        80.1 & 56.3 & 44.0 & 59.8 & 60.9 & 37.4 & 89.0 & 60.9 &  61.0 & 61.6\\
        longdoc \textsuperscript{S} &  ON + PS 60K &
        \textbf{80.9} & 58.2 & 48.9 & 57.1 & \textbf{63.8} & 39.0 & \textbf{89.9} & 63.1 &  \textbf{62.6} & \textbf{62.6}\\
        longdoc &  LB\textsubscript{0} &
        57.7 & 79.6 & 46.7 & 54.0 & 47.0 & \textbf{50.9} & 86.5 & 31.0 &  56.7 & 53.9\\
        longdoc &  PC &
        60.2 & 50.8 & \textbf{88.3} & 39.5 & 52.6 & 47.6 & 87.9 & 63.5 &  61.3 & 58.2\\
    \midrule
        longdoc \textsuperscript{S} & Joint &
         79.0 & 80.4 & 87.6 & 59.8 & 62.3 & 37.6 & 89.2 & 61.3 &  69.6 & 62.0\\
        longdoc \textsuperscript{S} & Joint + PS 60K      & 
        80.3 & \textbf{81.0} & 87.7 & 56.7 & 64.6 & 41.8 & 89.8 & 62.0 &  \textbf{70.5} & \textbf{63.0}\\
    \bottomrule
    \end{tabular}
    \caption{Performance of each model on 8 datasets measured by CoNLL F\textsubscript{1} \cite{weischedel2013ontonotes}, except for GAP (F\textsubscript{1}) and WSC (accuracy). Some models use speaker (\textsuperscript{S}) features, genre (\textsuperscript{G}) features, or pseudo-singletons (PS).} %
    \label{tab:gen_results}
\end{table*}

\subsection{Results}
\label{sec:gen_results}
Table~\ref{tab:gen_results} shows the results for all our models on all 8 datasets. We report each dataset's associated metric (e.g., CoNLL F\textsubscript{1}), the average across all eight datasets (All Average), and the average performance on the five evaluation and analysis datasets, namely CI, WC, QBC, GAP, and WSC -- Out-of-domain (OOD) average.

Among the longdoc baseline models trained on one of OntoNotes, PreCo, or LitBank, we observe a sharp drop in out-of-domain evaluations. 
The LitBank model is generally substantially worse than the models trained on OntoNotes and PreCo, likely due to both a smaller training set and a larger domain shift. Interestingly, the LitBank model performs the best among all models on QBC, which can be attributed to both LB and QBC being restricted to a similar set of markable entity types. 
Meanwhile, all OntoNotes-only models perform well on WC and GAP, possibly due to the more diverse set of genres within ON and because WC also does not contain singletons.

For models trained on OntoNotes, we find that the addition of speaker tokens leads to an almost 11 point increase on CI, which is a conversational dataset, but has little impact for non-conversational evaluations.
Surprisingly, the addition of genre embeddings has almost no impact on the overall evaluation.\footnote{We find that for the model trained with genre embeddings, modifying the genre value during inference has almost no impact on the final performance.}
In fact, the addition of genre embeddings hurts the OOD performance by 0.6 points which validates our motivation to avoid using metadata embeddings while training joint models. 
Finally, the addition of pseudo-singletons leads to consistent  significant gains across almost all the evaluations, including OntoNotes.

The joint models, which are trained on a combination of OntoNotes, LitBank, and PreCo, suffer only a small drop in performance on OntoNotes and PreCo, and achieve the best performance for LitBank. 
Like the results observed when training with only OntoNotes, we see a significant performance gain with pseudo-singletons in joint training as well, which justifies our intuition that they can bridge the annotation gap. 
In fact, the Joint + PS 60K model achieves the state-of-the-art for WC. 
Overall, the Joint + PS 60K improves the OOD performance by 0.4 points in comparison to the ON + PS 60K which achieves the best single dataset OOD performance. 
In comparison to the average increase of 7.9 points across all the datasets for the Joint + PS 60K model, the modest increase of 0.4 points for the OOD evaluation suggests that the gains with joint modeling are largely limited to the \emph{in-domain} datasets.

\begin{table}[t!]
	\centering
	\begin{tabular}{lccc}
		\toprule
		Data & Singleton & Non-singleton & Overall \\
		\midrule
		ON &  \phantom{1}0.7 & 44.6 & 37.2 \\
		ON + PS 60K &  \phantom{1}9.3  & 44.9 & 39.0 \\
		LB\textsubscript{0} & \textbf{44.9} & 41.8 & \textbf{50.9} \\
		PC & 29.2 & \textbf{52.0} & 47.6 \\
		Joint & \phantom{1}6.1  & 44.4 & 37.6 \\
		Joint + PS 60K  & 20.4  & 46.0 & 41.8 \\ 
		\bottomrule
	\end{tabular}
	\caption{Performance on singleton and non-singleton clusters for QBC. ON=longdoc\textsuperscript{S} and PS=pseudo-singletons.}
	\label{tab:singleton_ment}
\end{table}

\subsection{Analysis}
\paragraph{Impact of Singletons}

Singletons are known to artificially boost the coreference metrics~\cite{kubler-zhekova-2011-singletons}, and their utility for downstream applications is arguable. 
To determine the impact of singletons on final scores, we present separate results for singleton and non-singleton clusters in QBC in Table~\ref{tab:singleton_ment}. 
For non-singleton clusters we use the standard CoNLL F\textsubscript{1} but for singleton clusters the CoNLL score is undefined, and hence, we use the vanilla F\textsubscript{1}-score. 

The poor performance of ON models for singletons is expected, as singletons are not seen during training. Adding pseudo-singletons improves the performance of both the ON and the Joint model for singletons. Interestingly, adding pseudo-singletons also leads to a small improvement for non-singleton clusters. 

The PC model has the best performance for non-singleton clusters while the LB\textsubscript{0} model, which performs the best in the overall evaluation, has the worst performance for non-singleton clusters. This means that the gains for the LB\textsubscript{0} model can be all but attributed to the superior mention detection performance which can be explained by the fact that both LB and QBC are restricted to a similar set of markable entity types.

\begin{table}[t!]
	\centering
	\small{
	\begin{tabular}{lcccccc}
		\toprule
		 & \multicolumn{3}{c}{QBC} & \multicolumn{3}{c}{LB\textsubscript{0}} \\
		Data & Original & Post-Processed & $\Delta$ & Original & Post-Processed  & $\Delta$\\
		\midrule
		ON  &  
		37.2 & 40.7 & 3.5  & 55.6 & 56.0 & 0.4\\
		
		ON + PS 60K &  
		39.0 & 42.6 & 3.6  & 58.2 & 58.8 & 0.6\\
		
		LB\textsubscript{0} & 
		50.9 & 50.9 & 0.0 & 79.6 & 79.6 & 0.0\\
		
		PC & 
		47.6 & 50.9 & 3.3 & 50.8 & 51.2 & 0.4\\
		
		Joint & 
		37.6 & 41.0 & 3.4 & 80.4 & 80.4 & 0.0 \\
		
		Joint + PS 60K  & 
		41.8 & 45.2 & 3.4 & 81.0 & 81.0 & 0.0 \\ 
		
		\bottomrule
	\end{tabular}
	}
	\caption{Performance on Quiz Bowl and LitBank after possessive normalization.}
	\label{tab:possessive_normalization}
\end{table}
\paragraph{Impact of Span Annotation Differences}
Possessives are part of span annotation in OntoNotes and PreCo but not in LitBank and QuizBowl  (e.g. ``John's'' vs ``John'').  
Analyzing the output of the ON model for QBC  with the Berkeley coreference analyser \cite{kummerfeld2013error} revealed that around 48\% of span errors were due to the predicted mentions having possessives. 
Since these errors are due to a trivial annotation difference rather than the model's lack of coreferential understanding, we analyze the impact of post-hoc  possessive normalization of predicted cluster mentions in Table~\ref{tab:possessive_normalization}. 
We see a significant boost in QBC performance for all models, except LB\textsubscript{0}, which shares the possessive annotation artifact with QBC. 
For LB\textsubscript{0}, we see a mild improvement in performance of ON-only and PC-only models. 
Interestingly, joint models, which are trained on LB\textsubscript{0}, don't make any possessive ``errors" for LB. But the joint models predict mentions with possessives for QBC presumably because the majority of the training data (ON and PC) includes possessives. 
Our takeaway from this analysis is that out-of-domain evaluation in coreference needs to be carefully carried out. Otherwise, trivial annotation differences can lead to severe underestimation of model capabilities. Future work can do an in-depth study on other annotation differences, their impact on coreference evaluation, and try to bridge this gap whenever possible.

\begin{table*}
		\setlength{\tabcolsep}{0pt}

	\begin{tabular}{p{0.1\textwidth}p{0.9\textwidth}}
	\toprule
	Dataset & Instance \\
	\midrule
	(1) QBC & (\colorbox{blue!30}{\textbf{This poem}}) is often considered \colorbox{blue!30}{the counterpart of another poem} \dots name \colorbox{blue!30}{this poem about a creature ``burning bright, in the forests of the night,"} \dots\\\midrule
	(2) QBC & This author's non fiction works \dots \colorbox{blue!30}{another work}, a plague strikes secluded valley where teenage boys have been evacuated \dots name this author of \colorbox{blue!30}{Nip the Buds, Shoot the Kids}
	\dots\\\midrule

	(3) QBC & This poet of ``(\textbf{I}) felt a Funeral in (\textbf{my}) Brain" and ``I'm Nobody, Who are you?"  wrote about a speaker who hears a Blue, uncertain, stumbling buzz before expiring in ``(\textbf{I}) heard a fly buzz when (\textbf{I}) died". For 10 points, name this female American poet of Because (\textbf{I}) could not stop for Death.\\\midrule
		(4) CI & \textit{Chandler Bing:} Okay, I don't sound like that. (\textbf{That}) is so not true. (\textbf{That}) is so not ... (\textbf{That}) is so not ... That ... Oh , shut up !\\
	\bottomrule
	\end{tabular}
\caption{Joint + PS 60K error analysis for zero-shot evaluation sets. Each row highlights one cluster where spans in parenthesis are predicted by the model while the blue-colored spans represent ground truth annotations. Thus, in (2) the model misses out on the ground truth cluster entirely while in  (3) and (4) the model predicts an additional cluster.
}

\label{tab:error}
\end{table*}

\paragraph{Impact of Domain Shift}
Table~\ref{tab:error} presents instances where the Joint + PS 60K model makes mistakes. 
In examples (1) and (2), the model misses out on mentions referring to literary works which is because references to literary texts are rare in the joint training data. 
Example (2) also requires world knowledge to make a connection between the description of the work and its title. 
In example (3) the model introduces an extraneous cluster consisting of first person pronouns mentioned in titles of different works. The model lacks the domain knowledge that narrators across different works are not necessarily related.  
Apart from the language shift, there are annotation differences across datasets as well. For example (4) drawn from CI, the model predicts a valid cluster (for Chandler Bing's speaking style) according to the ON annotation guidelines but the CI dataset doesn't annotate such clusters.

\subsection{Related work}
Joint training is commonly used in NLP for training robust models,
usually aided by learning dataset, language, or domain embeddings
(e.g.,~\cite{stymne-etal-2018-parser} for parsing;~\cite{kobus-etal-2017-domain, tan-etal-2019-multilingual} for machine translation). %
This is essentially what models for OntoNotes already do with genre embeddings \cite{lee-etal-2017-end}. Unlike prior work, our test domains are 
unseen, so we cannot 
learn test-domain embeddings. 

For coreference resolution, \citet{aralikatte-etal-2019-rewarding} %
augment annotations using relation extraction systems to better incorporate world knowledge, a step towards generalization. \citet{subramanian-roth-2019-improving} use adversarial training to target names, %
with improvements on GAP. 
\citet{moosavi-strube-2018-using} incorporate linguistic features to improve generalization to WC.
Recently, \citet{zhu-etal-2021-ontogum} proposed the OntoGUM dataset which consists of multiple genres. However, compared to the datasets used in our work, OntoGUM is much smaller, and is also restricted to a single annotation scheme.
To the best of our knowledge, our work is the first to evaluate generalization at scale. 

Missing singletons in OntoNotes has been previously addressed through new data annotations, leading to the creation of the ARRAU \cite{poesio-etal-2018-anaphora} and PreCo \cite{chen-etal-2018-preco} corpora. While we include PreCo in this work, ARRAU contains additional challenges, like split-antecedents, that further increase the heterogeneity, and its domain overlaps with OntoNotes. Pipeline models for coreference resolution 
that first detect mentions naturally leave behind unclustered mentions as singletons, although understanding singletons can also 
improve performance \cite{recasens-etal-2013-life}. 

Recent end-to-end neural models have been evaluated on OntoNotes, and therefore conflate ``not a mention'' with ``is a singleton'' \cite{lee-etal-2017-end, lee-etal-2018-higher, kantor-globerson-2019-coreference, wu-etal-2020-corefqa}.
For datasets with singletons, this has been addressed explicitly through a cluster-based model \cite{toshniwal-etal-2020-learning, yu-etal-2020-cluster}. For those without, they can be implicitly accounted for with auxiliary objectives \cite{zhang-etal-2018-neural, swayamdipta-etal-2018-syntactic}. We go one step further 
by augmenting with \textit{pseudo-singletons}, so that the training objective is identical regardless of whether the training set contains annotated singletons. %

\subsection{Conclusion}
Our eight-dataset benchmark highlights disparities in coreference resolution model performance and tracks cross-domain generalization.
Our work begins to address cross-domain gaps, first by handling differences in singleton annotation via data augmentation with pseudo-singletons, and second by training a single model jointly on multiple datasets.  
This approach produces promising improvements in generalization, as well as new state-of-the-art results on multiple datasets.  
We also highlight that trivial annotation differences can lead to significant cross-domain performance drop.  
Overall, we find that generalization in coreference resolution remains a big challenge.

\section{Additional Results}
\label{sec:additional_res}
\subsection{Detailed Results for LitBank and OntoNotes}
Table~\ref{tab:full_litbank} and ~\ref{tab:full_ontonotes} present the detailed results for the LitBank and OntoNotes test sets.  
\begin{table*}[!ht]

\centering{

\begin{tabular}{l c c c c c c c c c c}
\toprule
Model & \multicolumn{3}{c}{MUC} & \multicolumn{3}{c}{$\text{B}^3$} & \multicolumn{3}{c}{$\text{CEAF}_{\phi_4}$} & \\
 & Prec. & Rec. & F1 & Prec. & Rec. & F1 & Prec. & Rec. & F1 & Avg.\ F1\\\midrule
 \unbounded   & 90.9 & 88.8 & 89.9  & 81.2 & 79.8 & 80.5  & 69.4 & 71.1 & 70.2 & 80.2\\
\learned \\
\hspace{0.1in} 5 cells 
    & 89.5 & 81.8 & 85.5  & 75.8 & 69.2 & 72.3  & 57.4 & 71.2 & 63.6 & 73.8 \\
\hspace{0.1in} 10 cells 
& 90.7 & 86.2 & 88.4  & 79.9 & 76.5 & 78.1  & 64.8 & 69.5 & 67.1 
& 77.9 \\
\hspace{0.1in} 20 cells 
& 90.6 & 87.4 & 88.9  & 80.6 & 77.6 & 79.1  & 65.9 & 70.9 & 68.3 &  78.8\\
\lru \\
\hspace{0.1in} 5 cells & 85.8 & 74.2 & 79.6  & 62.6 & 54.0 & 58.0  & 47.2 & 59.9 & 52.8 & 63.5\\
\hspace{0.1in} 10 cells & 89.4 & 80.0 & 84.4  & 74.2 & 65.6 & 69.7  & 54.3 & 68.5 & 60.6 &  71.6\\
\hspace{0.1in} 20 cells & 90.5 & 86.0 & 88.2  & 80.8 & 72.1 & 76.2  & 63.8 & 70.5 & 67.0 & 77.1\\
\bottomrule
\end{tabular}
}

\caption{Detailed results of the proposed models on the aggregated LitBank cross-validation test set.}
\label{tab:full_litbank}
\end{table*}
\begin{table*}[!ht]

\centering{

\begin{tabular}{l c c c c c c c c c c}
\toprule
Model & \multicolumn{3}{c}{MUC} & \multicolumn{3}{c}{$\text{B}^3$} & \multicolumn{3}{c}{$\text{CEAF}_{\phi_4}$} & \\
 & Prec. & Rec. & F1 & Prec. & Rec. & F1 & Prec. & Rec. & F1 & Avg.\ F1\\\midrule
 \unbounded   
 & 86.1 & 86.2 & 86.2  & 79.5 & 80.3 & 79.9  & 78.2 & 75.0 & 76.6 
&  80.9\\
\learned \\
\hspace{0.1in} 5 cells 
    & 84.0 & 76.1 & 79.9  & 72.6 & 66.4 & 69.4  & 74.8 & 59.4 & 66.2
&  71.8 \\
\hspace{0.1in} 10 cells 
 & 86.8 & 82.6 & 84.6  & 80.5 & 75.0 & 77.6  & 76.3 & 70.7 & 73.4 &  78.5 \\
\hspace{0.1in} 20 cells 
& 85.8 & 85.3 & 85.5  & 79.2 & 79.0 & 79.1  & 76.8 & 73.6 & 75.2 &  79.9\\
\lru \\
\hspace{0.1in} 5 cells & 77.9 & 64.3 & 70.5  & 63.2 & 53.8 & 58.1  & 67.1 & 47.4 & 55.5 &  61.4\\
\hspace{0.1in} 10 cells & 82.3 & 80.5 & 81.4  & 71.8 & 72.2 & 72.0  & 73.6 & 65.7 & 69.4 
&  74.3 \\
\hspace{0.1in} 20 cells  & 85.3 & 84.3 & 84.8  & 78.0 & 77.5 & 77.7  & 77.6 & 70.7 & 74.0 
&  78.8 \\
\bottomrule
\end{tabular}
}
\caption{Detailed results of the proposed models on the OntoNotes test set.}
\label{tab:full_ontonotes}
\end{table*}

\subsection{LitBank Cross-Validation Results}
\label{sec:appendix:other:lb}
\begin{table}[h]
    \centering
    \small
    \begin{tabular}{ccc}%
    \toprule
Cross-val split & Dev & Test \\
\midrule
0                    & 79.2 & 79.6 \\
1                    & 79.5 & 81   \\
2                    & 81.8 & 79   \\
3                    & 79.5 & 80.5 \\
4                    & 80.8 & 79.3 \\
5                    & 80.2 & 78.7 \\
6                    & 79.3 & 82.4 \\
7                    & 82.2 & 79.7 \\
8                    & 79.3 & 81.1 \\
9                    & 82.4 & 79.9 \\
\midrule
Total & 80.5 & 80.2 \\
\bottomrule
\end{tabular}
\caption{LitBank cross-validation results.}
\label{table:litbank_all}
\end{table}

Table~\ref{table:litbank_all} presents the results for all the cross-validation splits of LitBank. The overall performance of 80.2 CoNLL F1 is state-of-the-art for LitBank. Note that in this work, the joint model outperformed (81.0 vs. 79.6) this baseline model on split 0 (LB\textsubscript{0}). However, training 10 joint models contradicts the purpose of this work, which is to create a single, generalizable model. Realistically, we recommend jointly training with the entirely of LitBank.

\subsection{Impact of Context Fragmentation}
\label{sec:context_frag}
\begin{table*}[!ht]

\footnotesize
\centering{

\begin{tabular}{l l c c c c c c c c c c}
\toprule
Max segment length & \# of segments  & \multicolumn{3}{c}{MUC} & \multicolumn{3}{c}{$\text{B}^3$} & \multicolumn{3}{c}{$\text{CEAF}_{\phi_4}$} & \\
& & Prec. & Rec. & F1 & Prec. & Rec. & F1 & Prec. & Rec. & F1 & Avg.\ F1\\\midrule
512 & 8 
& 90.8 & 88.5 & 89.7  & 80.8 & 79.7 & 80.2  & 69.5 & 71.3 & 70.3 
&  80.1 \\
1024 & 4 
& 91.0 & 88.8 & 89.9  & 80.7 & 80.1 & 80.4  & 68.8 & 72.2 & 70.4 &  80.2 \\
2048 & 2 & 91.4 & 88.5 & 89.9  & 81.7 & 79.5 & 80.5  & 69.0 & 71.9 & 70.5 &  80.3 \\
4096 & 1 & 91.0 & 88.9 & 90.0  & 81.2 & 80.3 & 80.8  & 69.8 & 71.7 & 70.7 & \textbf{80.5} \\
\bottomrule
\end{tabular}
}
\caption{Detailed results of context fragmentation for the LitBank validation set.}
\label{tab:litbank_encoder_len}
\end{table*}

\begin{table*}[t]
    \centering
\begin{tabular}{lrr}
\toprule
Max Segment Length  & 512  & 4096 \\
\midrule
Span Error         & 290  & 279  \\
Conflated Entities & \textbf{1318} & 1266 \\
Extra Mention      & 420  & 420  \\
Extra Entity       & 42   & 43   \\
Divided Entity     & \textbf{1706} & 1633 \\
Missing Mention    & 520  & 528  \\
Missing Entity     & 55   & 51  \\\bottomrule
\end{tabular}
\caption{Comparing error categories for longdoc models trained with varying maximum segments length for the LitBank cross-validation set. }
\label{tab:context_frag_error}
\end{table*}

We had speculated in Section~\ref{sec:doc_encoder} that the performance gains with the switch from SpanBERT to Longformer had something to do with the larger context offered by Longformer.
We can't easily test this hypothesis by extending SpanBERT's context since that would require retraining the SpanBERT model on longer text chunks and with additional positional embeddings. 
But we can easily truncate Longformer's context window to test if coreference performance degrades with increase in context fragmentation. 

Table~\ref{tab:litbank_encoder_len} presents the validation set results of longdoc models trained with maximum segment lengths varying from 512 to 4096 subword tokens. 
Note that for all the settings, the rest of the model parameters are still trained on the entire LitBank documents (increased number of segments). 
The only difference across the model rows is that the document encoder is fed documents chunked to different maximum lengths. 
From the table, we see a (moderate) drop in validation performance from 80.5 F1 for the 4096 segment length model to 80.1 F1 for the 512 segment length model which validates our context fragmentation hypothesis. 
Table~\ref{tab:context_frag_error} presents the result of the error analysis using the Berkeley coreference analyzer~\cite{kummerfeld2013error} for the 512 segment length model and the 4096 segment length model. 
Comparing the error categories across the two models, we conclude that most of the error reduction is for the two error types which represent clustering errors, namely conflated entities, and divided entity. 
Thus, with longer context, the coreference models are able to better capture the entity identity.

\subsection{Singleton Results for OntoNotes}
\label{sec:singleton_ontonotes}

\begin{table}[h]
    \centering
    \small
    \begin{tabular}{ccc}%
    \toprule
    Num PS & Val. & Test \\
    \midrule
    \phantom{11}0 & 79.8 & 79.9 \\
    30K & 80.4  & 80.7 \\
    60K & \textbf{80.5} & \textbf{80.9} \\
    \bottomrule
    \end{tabular}
    \caption{Validation and test set results for the longdoc\textsuperscript{S} ON-only model trained with varying amount of pseudo-singletons (PS). }
    \label{tab:ontonotes_singletons}
\end{table}

For ON-only models, we tune over the number of pseudo-singletons sampled among \{0, 30K, 60K\}. Table~\ref{tab:ontonotes_singletons} shows that adding 
60K pseudo-singletons yields the best result for OntoNotes validation set.

\subsection{Downsampling and Singleton Tuning Results for Joint Training}
\label{sec:downsampling joint}
\begin{table*}[h]
    \centering
    \small
    \begin{tabular}{ccccccc}
    \toprule
        \multicolumn{3}{c}{Training} & \multicolumn{4}{c}{Evaluation} \\
        Num ON & Num PC & Num PS & ON & LB & PC & Train Avg.\\
    \midrule
        
        \phantom{1}500 & \phantom{1}500 & \phantom{11}0 & 
        79.1 & 80.6 & 85.1 &  81.6 \\
        
        \phantom{1}500 & \phantom{1}500 & 30K  & 
        79.7 & 79.8 & 85.1 &  81.5 \\
        
        \phantom{1}500 & \phantom{1}500 & 60K  & 
        80.3 & 80.1 & 85.2 &  81.9 \\
        
        1000 & 1000 & \phantom{1}0 & 
        79.7 & 80.0 & 85.3 &  81.7 \\
        
        1000 & 1000 & 30K & 
        79.7 & 79.6 & 85.0 &  81.4 \\
        
        1000 & 1000 & 60K & 
        80.1 & 79.6 & 84.8 &  81.5 \\
        
    \bottomrule
    
    \end{tabular}
    \caption{Validation set performance for the training set datasets when downsampling OntoNotes (ON) and PreCo (PC) in joint training.}
    \label{tab:downsample}
\end{table*}

For sampling PreCo and OntoNotes, we tune over dowsampling both the datasets to \{500, 1000\} documents per epoch. For adding pseudo-singletons, we tune over adding  \{0, 30K, 60K\} pseudo-singletons. For hyperparameter selection, the average of the validation set performance over the three training sets, namely OntoNotes, PreCo, and LitBank(0), is used. The results of the hyperparameter tuning experiment are shown in Table~\ref{tab:downsample}. Based on the validation set results, we downsample PreCo and OntoNotes to 500 documents per epoch and add 60K pseudo-singletons to OntoNotes.

\chapter{Learning Chess Blindfolded: Evaluating Language Models on State Tracking}
\epigraph{
	\footnotesize{
		The Chess pieces are the block alphabet which shapes thoughts; and these thoughts, although making a visual design on the chessboard, express their beauty abstractly, like a poem.
}}{\textit{Marcel Duchamp}}

Transformer language models have made tremendous strides in natural language understanding tasks. 
In this chapter, we evaluate the \emph{implicit} entity tracking capability of transformer language models. 
We refer to entity tracking via language models as ``implicit" because unlike the the memory models based coreference models discussed in Chapters 3 and 4, language models don't necessarily maintain an explicit mapping between hidden states (working memory of transformer LMs) and entities. 
Since measuring the entity tracking capability of natural language models is challenging due to the complexity of natural language and the underlying world they describe, we propose the task of language modeling for the game of chess.
We observe that the appropriate choice of chess notation allows for directly probing the world state, without requiring any additional probing-related machinery. 
Our empirical results show that: (a) With enough training data, transformer language
models can learn to track pieces and predict legal moves with high accuracy when trained solely on move sequences. (b) For small training sets providing access to board state information during training can yield significant improvement. 
\footnote{The material for this chapter has been adapted from \citet{toshniwal-etal-2022-chess}. Code is available at \url{https://github.com/shtoshni/learning-chess-blindfolded}}

\section{Introduction}
Recently, transformer-based language models 
have stretched notions of what is possible with the simple self-supervised objective of language modeling, becoming a fixture in state-of-the-art language technologies
\citep{vaswani2017attention, devlin-etal-2019-bert, brown2020language}.
However, the black box nature of these models combined with the complexity of natural language makes it
challenging to measure how accurately they
represent the world state underlying the text.

In order to better measure the extent to which these models can capture the world state underlying the symbolic data they consume, we propose training and studying
transformer language models for the game of chess.
Chess provides a simple, constrained, and deterministic domain where the exact world state is known.
Chess games can also be transcribed exactly and unambiguously using chess notations (Section~\ref{sec:chess}).
Most importantly, the form of chess notations allows us to probe our language models for aspects of the board state using simple prompts (Section~\ref{sec:probing}) and without changing the language modeling objective or introducing any new classifiers.\footnote{Code and data available at - \url{https://github.com/shtoshni/learning-chess-blindfolded}}

Due to the simplicity and precision of chess, we can evaluate language model predictions at a more fine-grained level than merely comparing them to the ground truth.
For example, even if the next move prediction doesn't match the ground truth move, we can still evaluate whether the move is legal given the board state, and if it is illegal, the error can be automatically analyzed (Section~\ref{sec:error_analysis}).
Moreover, since world state transitions are deterministic and known, we can  evaluate models using counterfactual queries as well.
Our proposed evaluation sets and metrics are described in Section~\ref{sec:cloze}.

\begin{figure*}
\begin{subfigure}{0.3\textwidth}
   \includegraphics[width=\linewidth]{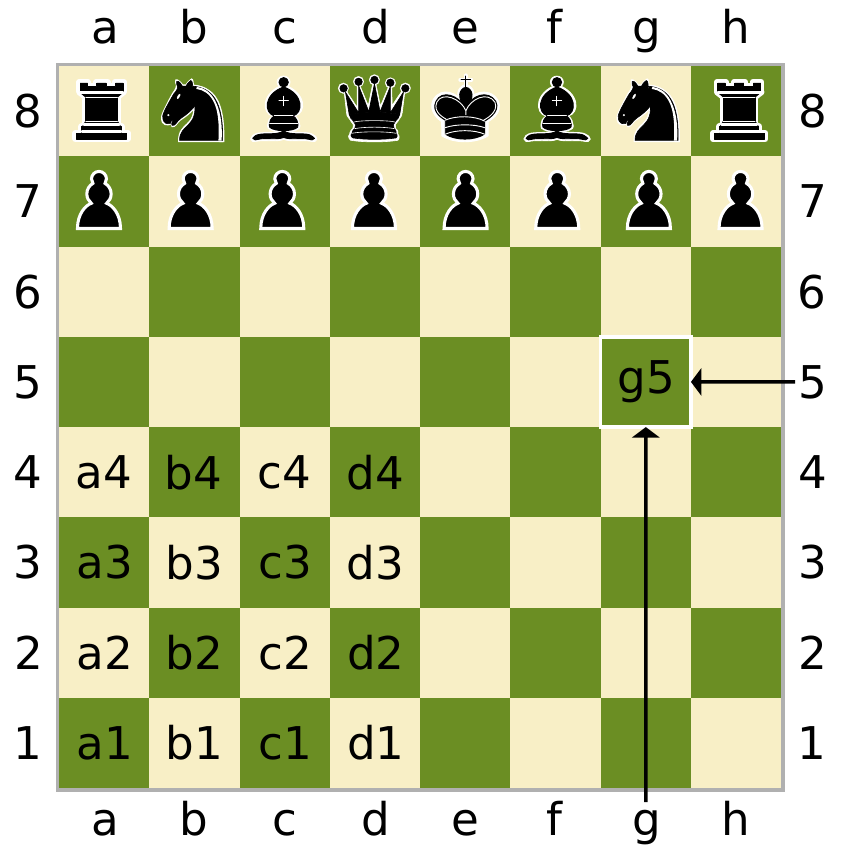}
   \caption{Square naming} \label{fig:algebraic_notation}
\end{subfigure}
\hspace*{\fill}
\begin{subfigure}{0.63\textwidth}
\begin{subfigure}{0.43\textwidth}
   \vspace{.2in}\includegraphics[width=\linewidth]{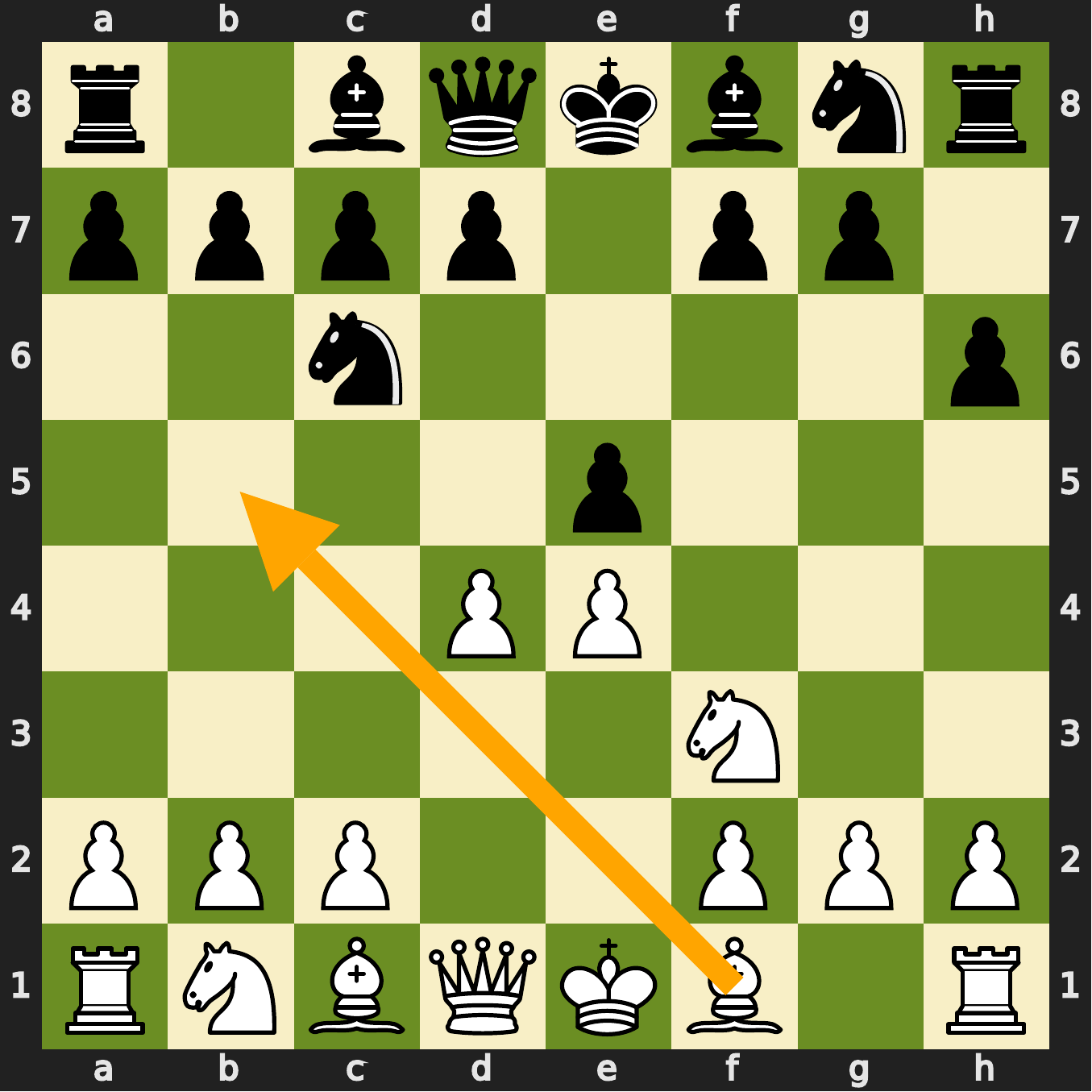}
\end{subfigure}
\hspace*{\fill}
\begin{subfigure}{0.43\textwidth}
   \vspace{.2in}\includegraphics[width=\linewidth]{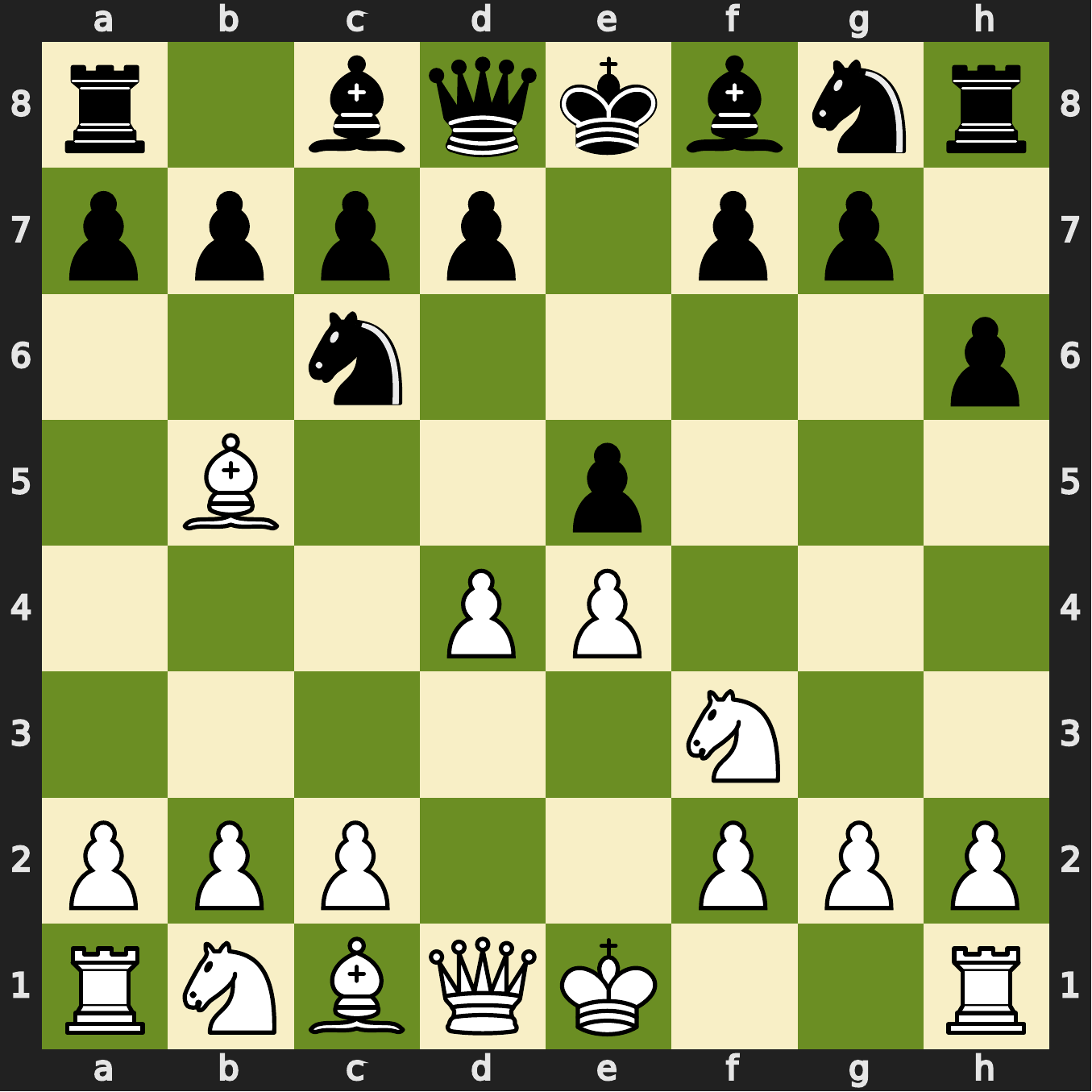}
\end{subfigure}
\vspace{.1in}
\caption{Board state before (left) and after (right) the bishop at \pos{f1} is moved to \pos{b5}. UCI notation represents the move as \pos{f1b5}.}
\label{fig:move_notation}
\end{subfigure}
\vspace{-0.05in}
\caption{
Chess Notation %
}
\end{figure*}

While chess represents a controlled domain,
it is by no means trivial for a language model.
To illustrate the challenges of language modeling for chess,
consider the left board shown in Figure~\ref{fig:move_notation}, where white is next to move.
In order to generate a valid next move, the language model needs to (a) infer that it is white's turn, (b) represent the locations of all pieces, both white and black, (c) select one of the white pieces which can be legally moved, and finally (d) make a legal move with the selected piece.
Thus, a language model has to learn to track the board state, learn to generate moves according to the rules of chess, and on top of that learn chess strategies to predict the actual move. %

We find that when given enough training data, transformers can learn to both track piece locations and predict legal moves with high accuracy.
However, when trained on small training sets, predictive ability suffers. 
In this more challenging setting, introducing parts of the board state as tokens in the training sequences (Section~\ref{sec:rap_board})  improves piece tracking significantly (Section~\ref{sec:error_analysis}). 
Our results also provide some key insights on transformer language models:
\begin{enumerate*}[label=(\roman*)]
	\item They are robust to changes in input distribution where additional tokens, related to board state, are added to input sequence \emph{only during training} (Section~\ref{sec:rap_board}). 
	In contrast to LSTMs, transformers achieve this robustness even with smaller training sets (Section~\ref{sec:other_models}). 
	\item Even though chess is Markovian, the model relies on having access to the whole history, and the performance drops when limiting this access (Section~\ref{sec:limited_history}). 
\end{enumerate*}

\section{Chess Preliminaries}
\label{sec:chess}

\paragraph{Standard Algebraic Notation (SAN) notation}
combines the piece type moved and the destination square to denote a move.\footnote{For more details see \url{https://en.wikipedia.org/wiki/Algebraic_notation_(chess)}}
For example, the move in Figure~\ref{fig:move_notation} is represented as \texttt{Bb5} in SAN where \texttt{B} represents the piece type bishop and \texttt{b5} represents the destination square.

\paragraph{Universal Chess Interface (UCI) notation}
combines the starting square and the destination square to represent a move.\footnote{For more details see \url{https://en.wikipedia.org/wiki/Universal_Chess_Interface}} 
The move in Figure~\ref{fig:move_notation} is represented as \texttt{f1b5} in UCI where \texttt{f1} indicates the starting square and \texttt{b5} denotes the ending square.

While the SAN notation is the standard choice for game play, SAN notation doesn't use the starting square of the piece in its move representation which limits the ability to prompt a SAN-based language model with specific piece type instances.
For example, given the prompt ``\texttt{\underline{e4 e5 Nf3 Nc6 d4 h6} B}'' (the underlined move sequence leads to the left board state in Figure~\ref{fig:move_notation}), it's not clear whether the token \texttt{B} refers to the bishop at \texttt{f1} or \texttt{c1}. Due to this limitation on the specificity of probing queries, we use the UCI notation rather than SAN for our experiments.

For training language models, we first tokenize games represented in UCI notation using a simple regular expression based tokenizer, which considers a board square symbol such as \texttt{b1} as a single token.
This gives us a vocabulary of 77 token types, 
which includes the 64 squares, piece type symbols, and other special symbols (see Table~\ref{tab:model_vocab}). 
No delimiter token is used to denote the move boundary. 
\footnote{In initial experiments we used a delimiter token to indicate move boundary. However, removing it did not degrade performance and made training faster due to reduced sequence length.}
For example, the move sequence ``\pos{e2e4 e7e5 g1f3}" is tokenized to ``\pos{e2}, \pos{e4}, \pos{e7}, \pos{e5}, \pos{g1}, \pos{f3}". We then train an autoregressive language model on these move sequences, using the standard maximum likelihood objective.

\begin{table}[t]
\centering{
\begin{tabular}{llc}
    \toprule
    Type & Examples & Count \\
    \midrule
    Square names & \pos{e4}, \pos{d1} & 64 \\
    Piece type & \pos{P}, \pos{K}, \pos{Q}, \pos{R}, \pos{B}, \pos{N} & \phantom{1}6\\
    Promoted Pawn Piece type & q, r, b, n & \phantom{1}4 \\
    Special symbols & BOS, EOS, PAD & \phantom{1}3 \\
    \midrule
    Total & & 77\\
    \bottomrule
\end{tabular}
}
\caption{Model Vocabulary}
\label{tab:model_vocab}

\end{table}

\begin{table*}
	\centering{
		\begin{tabular}{lll}
			\toprule
			Notation 		& Training 					& Inference \\\midrule
			UCI		 		& \pos{e2}, \pos{e4}, \pos{e7}, \pos{e5}, \pos{g1}, \pos{f3}     						& \pos{e2}, \pos{e4}, \pos{e7}, \pos{e5}, \pos{g1}, \pos{f3} 		\\
			UCI + RAP 15 	& \pos{e2}, \pos{e4}, \pos{P}, \pos{e7}, \pos{e5}, \pos{g1}, \pos{f3} 							& \pos{e2}, \pos{e4}, \pos{e7}, \pos{e5}, \pos{g1}, \pos{f3} 		\\
			UCI + RAP 100 	& \pos{P}, \pos{e2}, \pos{e4}, \pos{P}, \pos{e7}, \pos{e5}, \pos{N}, \pos{g1}, \pos{f3}							& \pos{e2}, \pos{e4}, \pos{e7}, \pos{e5}, \pos{g1}, \pos{f3} 		\\
			UCI + \piecetype 	& \pos{P}, \pos{e2}, \pos{e4}, \pos{P}, \pos{e7}, \pos{e5}, \pos{N}, \pos{g1}, \pos{f3}							& \pos{P}, \pos{e2}, \pos{e4}, \pos{P}, \pos{e7}, \pos{e5}, \pos{N}, \pos{g1}, \pos{f3}		\\
			\bottomrule
		\end{tabular}
	}
		\caption{Token sequences corresponding to the move sequence \pos{e2e4 e7e5 g1f3} for different notations during training and inference. Notice that regardless of the RAP probability used during training, at inference time the token sequences have no piece types.}
	\label{tab:token_seq}

\end{table*}

\section{Language Model Prompts as Board State Probes}
\label{sec:probing}
One attractive property of having a language model trained on chess games represented in UCI notation (as described in the previous section) is that the notation \textit{itself} allows us to probe the trained model's state tracking abilities. In particular, by feeding the trained language model a prefix of a game as a prompt, we can determine --- using the language model's next-token predictions --- what the model understands about the board state implied by this prefix.
For example, consider the prompt ``\pos{\underline{e2e4 e7e5 g1f3 b8c6 d2d4 h7h6} f1},''  where the underlined move sequence leads to the left board state in Figure~\ref{fig:move_notation}. A language model's next-token prediction (after consuming the prompt) can be interpreted as the ending square predicted %
for the bishop at \pos{f1}, which can be used to determine the level of board state awareness of the model. %
If, for instance, the model predicts \pos{g1}, this may indicate that the model does not recognize that the piece type at \pos{f1} is a bishop, as such a move is not possible for a bishop.
If, on the other hand, the model predicts \pos{g2}, that may indicate that the model is not aware that another piece is currently  at \pos{g2}.

\subsection{Randomly Annotated Piece type (RAP)}
\label{sec:rap_board}
While predicting the token representing the ending-square of a move given a prompt allows us to assess the model's state tracking abilities, it also to some extent conflates the model's understanding of the board state with its understanding of chess strategy. If we could easily probe for where the model thinks a piece \textit{currently} is (rather than where it is likely to end up) given a game prefix, this would allow us to more directly probe the model's state tracking abilities. 
In particular, we would like to give a language model a prompt such as ``\pos{e2e4 e7e5 g1f3 b8c6 d2d4 h7h6 \underline{N}}", where \pos{N} represents knight, and expect it to generate a valid starting position for a knight of the correct color. 
While UCI notation does not ordinarily include these piece type tokens, to allow for testing the model with such prompts, 
we propose to randomly include these piece types tokens in moves during training with some fixed probability $p$.
We refer to this strategy as ``randomly annotated piece type'' (RAP) and 
use the nomenclature ``UCI + RAP $p$'' to indicate that with $p\pct$ probability, piece type is part of the move notation during training.
Note that for $p = 0$, the notation reduces to UCI. 

When \emph{testing} with these starting square prediction prompts, we only include piece type for the prompt, not for any moves in the history.
Thus, using RAP during training allows us to probe, at test time, where the model thinks each piece is, given any game history's prefix; by simply providing the desired piece type (e.g., \pos{N}) the model outputs the predicted starting square for a piece of that type.
For example, given the prompt ``\pos{e2e4 e7e5 g1f3 b8c6 d2d4 h7h6 N}", a prediction of \pos{f3} or \pos{b1} shows that the model is aware of where the knights are.%

We also experiment with an ``oracle" variant of RAP where piece types are added both during training and testing. We refer to this notation as ``UCI + \piecetype" where AP stands for ``always piece type".
For our running example the equivalent prompt in this notation would be ``\pos{Pe2e4 Pe7e5 Ng1f3 Nb8c6 Pd2d4 Ph7h6 N}".

In terms of the language modeling training objective, addition of RAP represents a distribution change between training and inference.
Table~\ref{tab:token_seq} illustrates how the use of RAP changes the token sequence during training but not during inference.  
While there's a distribution mismatch, we hypothesize that addition of RAP can aid the model in learning to track the pieces by providing additional supervision which, in turn, can improve language modeling performance as well.

\begin{table*}[t]
\centering
\begin{tabular}{lccc}
\toprule
Task & Prompt Token & Correct Answers (\exactmove) & Correct Answers (\legalmove) \\
\midrule
 End-Actual & \pos{f1} & \{\pos{b5}\} & \{\pos{e2, d3, c4, b5 ,a6} \} \\
End-Other & \pos{f3} & N/A & \{\pos{d2, g1, h4, g5, e5}\} \\
  \midrule
Start-Actual & \pos{B} & \{\pos{f1}\} & \{\pos{f1, c1}\} \\
Start-Other & \pos{N} & N/A & \{\pos{f3, b1}\} \\
\bottomrule
\end{tabular}
\caption{Examples of each probing task, as well as the corresponding exact move (\exactmove) and legal move (\legalmove) correct answers, are shown below. All examples assume the language model was fed the prefix \pos{e2e4 e7e5 g1f3 b8c6 d2d4 h7h6} (see Figure~\ref{fig:move_notation}), and that the actual next move was \pos{f1b5}. While there is only one valid prompt token for both End-Actual and Start-Actual tasks, there are many valid prompt tokens for the other tasks, and we show just one possibility for each. Start-tasks (bottom sub-table) assume the model was trained on games described in UCI+RAP notation.}
\label{tab:tasks}

\end{table*}

\subsection{Board State Probing Tasks}
\label{sec:cloze}
In this subsection we describe the probing tasks introduced above more concretely. %
In each probing task we feed the model a prefix of a game followed by a single prompt token, and the model is evaluated based on the highest probability next-token under the model given this context. We show an example of each probing task in Table~\ref{tab:tasks} (which we further describe below), assuming the model has been fed the move sequence prefix \pos{e2e4 e7e5 g1f3 b8c6 d2d4 h7h6}, %
which is visualized as the left board in Figure~\ref{fig:move_notation}. The actual next move played in the game is \pos{f1b5}, which takes the white bishop at square \pos{f1} to square \pos{b5}, as shown in the right board of Figure~\ref{fig:move_notation}.

\subsection{Ending Square Tasks}
In this set of tasks, the model is given a game prefix and prompted with the starting square of the next move (\pos{f1} in the example of Table~\ref{tab:tasks}). The model's next-token prediction represents its prediction for the ending square of this move,
which
tests the model's ability to track the board state and follow
the rules of chess,
as well as strategic awareness.\footnote{Strategic capabilities of a chess language model are strongly tied to the quality of training games.}  We consider two task variants: %
\begin{enumerate}
	\item \textbf{End-Actual}: Given a move sequence prefix, the model is prompted with the starting square of the actual piece moved next in the game. %
	\item \textbf{End-Other}: Given a move sequence prefix, the model is prompted with the starting square of any piece on the board that can be legally moved according to the rules of chess. 
\end{enumerate}
We evaluate End-Actual predictions in terms of both exact move (\exactmove) accuracy (whether the model predicted the true ending square, \pos{b5} in our running example) and legal move (\legalmove) accuracy (whether the model predicted a legal ending square for the piece starting at the square in the prompt). 
For \legalmove evaluation, we also calculate the R-Precision which is the Precision@R where R is the total number of legal ending squares~\cite{ir-book}. In our running example, there are 5 legal ending squares, and R-Precision will be calculated for the model's top-5 predictions.
\exactmove accuracy evaluation is similar to the typical evaluation of language models on natural language data, while \legalmove is less stringent and focuses on testing just the model's understanding of chess rules and the board state. Note that for End-Other, only \legalmove evaluation is available. See Table~\ref{tab:tasks} for examples.

\subsection{Starting Square Tasks}
In this category of task, the model is again given a game prefix, but prompted with just the piece type of the next move, such as \pos{B} for bishop in the example in Table~\ref{tab:tasks}. The model's next-token prediction thus represents its prediction for where the prompted piece type currently is on the board. This task tests the model's ability to track pieces.\footnote{In certain cases, this task also tests understanding of chess rules. For example, in Figure~\ref{fig:move_notation} only the rook at \pos{h1} can be moved.}
Note that only models which have seen piece types during training, i.e.\ ``UCI + RAP'' models, can actually be tested on this task.
Also, no piece types are used in the game prefix. %
We again have two variants of this task:
\begin{enumerate}
	\item \textbf{Start-Actual}: Given a move sequence prefix, the model is prompted with the piece type of the actual piece moved next in the game. 
	\item \textbf{Start-Other}: Given a move sequence prefix, the model is prompted with the piece type of any piece on the board that can be legally moved according to the rules of chess. %
\end{enumerate}
We again evaluate Start-Actual %
both in terms of \exactmove accuracy (whether the model predicts the starting square of the piece actually moved next in the game), as well as in terms of \legalmove accuracy (whether the model predicts the starting square of a legally movable piece of the given piece type) and \legalmove R-Precision (precision of the model's top-R predictions with respect to all of the R starting squares of legally movable pieces of the given piece type). For Start-Other, only \legalmove evaluation is applicable; see Table~\ref{tab:tasks} for examples.

\section{Experimental Setup}
\label{sec:setup}

\paragraph{Data}
We use the Millionbase dataset which is freely available and has close to 2.9 million quality chess games.\footnote{Download link available at \url{https://rebel13.nl/rebel13/rebel\%2013.html}}
After filtering out duplicate games, games with
fewer than 10 moves, and games with
more than 150 moves (for the complete game to fit into one transformer window), we are left with around 2.5 million games.
From this filtered set we randomly select 200K games for training, 15K games each for dev and test, and another 50K games to create board state probing evaluation sets described in Section~\ref{sec:cloze}.
The dev and test sets are used for perplexity evaluations. 
The dev set perplexity is used for choosing hyperparameters.
From the 200K training set, we create subsets of size 15K and 50K which we refer to as ``Train-S'' and ``Train-M'', while the full training set is referred to as ``Train-L''.
For detailed statistics, see Table~\ref{tab:data_stats} in Section~\ref{sec:data_stats}.
All the data processing steps requiring chess knowledge, including parsing chess databases, are carried out using python-chess~\citep{python-chess}.

To create the board state probing evaluation sets, we use the 50K games reserved for this task. %
We only consider prompts for non-pawn pieces since the dynamics of pawns are fairly limited.
We ensure that the game prefixes selected are never seen in the training data.
The final evaluation set consists of 1000 instances with prefix length (in number of moves) in the range $51 \le l \le 100$.

\begin{figure*}
	\begin{minipage}{\textwidth}
		\begin{minipage}[b]{0.48\textwidth}
			\centering
			\begin{tabular}{llcc}
				\toprule
				Training Set & Model   & Dev set & Test set \\
				\midrule
				\multirow{2}{*}{Train-S} 
				& UCI 				& 23.6 & 23.6\\
				& UCI + RAP 		& 15.9 & 15.9\\
				& UCI + \piecetype 	& 16.1 & 16.2 \\
				\midrule
				\multirow{2}{*}{Train-M} 
				& UCI 				& 11.6 & 11.6\\
				& UCI + RAP 		& 10.4 & 10.4\\
				& UCI + \piecetype 	& 10.1 & 10.0 \\
				\midrule
				\multirow{2}{*}{Train-L} 
				& UCI 				& \phantom{1}7.7 & \phantom{1}7.7\\
				& UCI + RAP 		& \phantom{1}7.4 & \phantom{1}7.4\\
				& UCI + \piecetype 	& \phantom{1}7.2 & \phantom{1}7.2 \\
				\bottomrule
				
			\end{tabular}
			\captionof{table}{Canonical validation and test set perplexity. By canonical we mean that one move, say \pos{f1b5}, counts as one token.}
			\label{tab:perplexity}
		\end{minipage}
		\hfill
		\begin{minipage}[b]{0.48\textwidth}
			\centering
			\includegraphics[width=\textwidth]{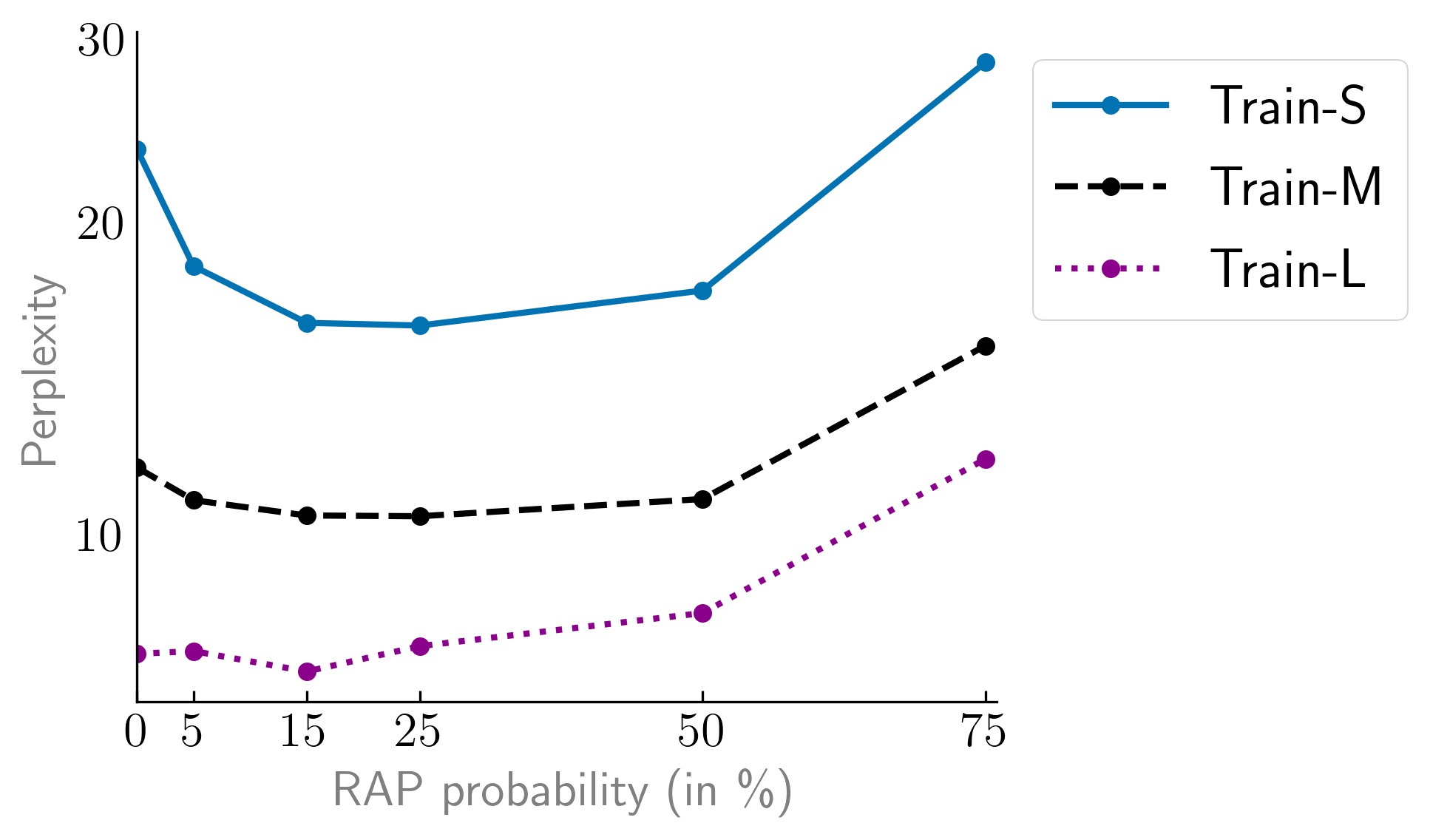}
			\captionof{figure}{Validation set perplexities as a function of RAP probabilities for the different training set sizes. RAP $0$ %
				is %
				the standard UCI notation. 
				RAP $100$ is not shown as perplexities are too high. }
			\label{fig:rap_vals}
		\end{minipage}
	\end{minipage}
\end{figure*}

\paragraph{Model Details}
We use the GPT2-small architecture for our base language model \citep{vaswani2017attention,radford2019language}.
GPT2-small is a 12-layer transformer model with 12 attention heads and an embedding size of 768 dimensions.
The context size of the model is limited to 512, which is sufficient to cover the longest game in our training set.
Note that we only borrow the model architecture; the models themselves are \emph{trained from scratch}.
\footnote{Colab notebook to play chess against the base language model \url{https://github.com/shtoshni/learning-chess-blindfolded/blob/master/GPT2_Chess_Model.ipynb}}

For the UCI + RAP $p$ models, we tune over $p \in \{5, 15, 25, 50, 75, 100\}$ based on %
perplexity on the validation set.
Note that for perplexity evaluation, logits corresponding to piece type tokens are masked out since piece type tokens are only available during training.
We find that $p=25$ performs the best for Train-S and Train-M, while $p=15$ is best for Train-L (Figure~\ref{fig:rap_vals}). %
Larger values of $p$ lead to greater mismatch between training and inference, while smaller values likely do not provide enough training signal.

We also experiment with other transformer and non-transformer models in Section~\ref{sec:other_models}.
Among the transformer models, we experiment with two ``approximate" attention models (i.e., models which approximate the full attention of vanilla transformer models), namely, Reformer \cite{kitaev2020reformer} and Performer \cite{choromanski2021rethinking}.  
We set the number of layers and attention heads to 12 for both 
architectures, as in GPT2-small.
We also train LSTM language models with and without RAP. 
For details on hyperparameters and tuning, see Section~\ref{sec:hyperparams}.

\paragraph{Training Details}
Models are trained for 10 epochs with a batch size of 60. Validation is performed %
 at the end of every epoch and training stops whenever the validation loss starts increasing.
For optimization we use Adam \citep{kingma2014adam} with learning rate of $5\times10^{-4}$ and L2 weight decay of $0.01$.
The learning rate is warmed up linearly over the first 10\% of training followed by a linear decay.
To accelerate training, we use mixed precision training~\citep{micikevicius2018mixed}. %
All experiments are carried out using the PyTorch Lightning framework %
built on top of PyTorch \citep{falcon2019pytorch, pytorch}.
We use the transformers library \citep{Wolf2019HuggingFacesTS} for all models\footnote{Reformer implementation in \pos{transformers} library is was still a work in progress when we did our experiments. The presented results are with the 4.2.2 version.} %
except for the Performer model %
for which we use a popular unofficial implementation.
\footnote{\url{https://github.com/lucidrains/performer-pytorch}}

\section{Results}
We first present language modeling results, where we show significant 
improvements with the addition of RAP (Section~\ref{sec:perplexity_res}). 
Next, we show results on the board state probing tasks for the base language model, where we demonstrate that the 
model trained on the large training set can learn to track pieces and predict legal moves with high accuracy (Section~\ref{sec:state_tracking_res}).
Finally, we present results on the probing task 
with approximate attention transformer architectures and LSTMs,  where we show a performance drop in comparison to the base model with full attention (Section~\ref{sec:other_models}).

\subsection{Language Modeling}
\label{sec:perplexity_res}
Table~\ref{tab:perplexity} presents the perplexity results on the validation and test sets.  
Figure \ref{fig:rap_vals} plots the validation set perplexities as a function of RAP probability for different training set sizes. 
The addition of RAP and \piecetype leads to a decrease in perplexity for all training sizes, particularly for small training sets.
For small training sets, RAP probabilities as high as 50\% can improve the validation perplexity, but for larger training sets, lower RAP probabilities are preferred. 
The reductions in perplexity for RAP are surprising given that the extra tokens added via RAP are not present in the validation and test sets, and thus there is a data distribution shift. %
Models trained with UCI + \piecetype achieve the lowest perplexities on larger training sets. 
Both RAP and \piecetype aid the model in piece tracking, as we will see in later results, and in the case of chess this can significantly improve the language modeling results as well.
Note that for calculating the perplexity of UCI + RAP models, we mask out the logits corresponding to piece type tokens since they are never present during inference.

\subsection{Board State Tracking}
\label{sec:state_tracking_res}
Tables~\ref{tab:results-starting} and~\ref{tab:results-ending} show results when predicting starting squares and ending squares, respectively. 
There are several observations to note. First,  \textbf{transformers can learn to identify where pieces are located}.
This is shown by the \legalmove accuracies in Table~\ref{tab:results-starting}.
UCI + RAP can predict legal starting positions with perfect accuracy and R-Precision. 
However, this capability requires Train-L, and the accuracy drops to 91.3\% for Train-S. 
The gap between UCI + RAP and its ``oracle" counterpart, UCI + \piecetype, also reduces with an increase in training set size with UCI + RAP achieving parity for Train-L.
When asked to identify the location of a piece other than the one selected to be moved next, this accuracy drops only slightly to 99.6\%. 
Typically, the piece location tracking is slightly better for the piece type that is actually moved 
than for other piece types.

The difference between the location of the piece in the exact move (\exactmove) and the location of either piece of the given type (\legalmove) is substantial, at more than 8\% absolute.  
However, this difference relates to chess strategy rather than board state tracking.

\begin{table}[t]
	\setlength\tabcolsep{4pt}
	\centering{
		\begin{tabular}{llccccc}
			\toprule
			&  Notation 	&  \multicolumn{4}{c}{\legalmove} & \exactmove\\
			
			&  &  \multicolumn{2}{c}{Actual} &   \multicolumn{2}{c}{Other} &  \\
			&  &  Acc. & R-Prec. & Acc. & R-Prec. & Acc.\\
			\midrule
			\multirow{2}{*}{S} 	& UCI + RAP& \phantom{1}91.3 & \phantom{1}90.2 & \phantom{1}89.3 & \phantom{1}89.2  & 78.8 \\
			& UCI + AP & \phantom{1}99.2 & \phantom{1}99.1 & \phantom{1}98.8 & \phantom{1}98.8  & 86.9 \\
			\midrule
			\multirow{2}{*}{M} 	& UCI + RAP & \phantom{1}98.2 & \phantom{1}98.0 & \phantom{1}98.6 & \phantom{1}98.7  & 88.0 \\
			& UCI + AP  & \phantom{1}99.9 & \phantom{1}99.8 & 100.0 & 100.0  & 90.2 \\
			\midrule
			\multirow{2}{*}{L} 	& UCI + RAP& 100.0 & 100.0 & \phantom{1}99.6 & \phantom{1}99.5  & 91.8 \\
			& UCI + AP & \phantom{1}99.9 & \phantom{1}99.9 & \phantom{1}99.7 & \phantom{1}99.7  & 91.1 \\
			\midrule
			\multicolumn{2}{l}{Random Legal}  & - & - & - & - & 86.0\\
			\bottomrule
		\end{tabular}
		
	\caption{Accuracies and R-Precisions (\%) for predicting starting squares (``Start-Actual'' and ``Start-Other'' tasks). S, M, L in the first column refer to the training set sizes. 
	}
	\label{tab:results-starting}
	
}
\end{table}
\begin{table}
	\setlength\tabcolsep{4pt}
	\centering{
		\begin{tabular}{llccccc}
			\toprule
			&  Notation 	&  \multicolumn{4}{c}{\legalmove} & \exactmove\\
			
			&  &  \multicolumn{2}{c}{Actual} &   \multicolumn{2}{c}{Other} &  \\
			&  &  Acc. & R-Prec. & Acc. & R-Prec. & Acc.\\
			\midrule
			\multirow{3}{*}{S}
			& UCI  				&  74.0 & 61.1 & 65.5 & 57.7  & 26.7 \\
			& UCI + RAP  		&  88.4 & 75.5 & 80.4 & 72.1  & 33.3  \\
			& UCI + \piecetype 	&  87.0 & 77.0 & 78.8 & 72.3  & 36.1   \\
			\midrule
			\multirow{3}{*}{M}
			& UCI  				&  92.9 & 80.6 & 85.8 & 78.5  & 42.2  \\
			& UCI + RAP  		&  94.9 & 82.2 & 87.9 & 78.0  & 45.9  \\
			& UCI + \piecetype 	&  94.7 & 82.4 & 88.3 & 79.1  & 47.3   \\ 
			\midrule
			\multirow{3}{*}{L}
			& UCI  				&  97.7 & 85.6 & 91.9 & 83.8  & 52.0   \\
			& UCI + RAP  		&  97.0 & 86.1 & 93.1 & 83.9  & 54.7  \\
			& UCI + \piecetype 	&  98.2 & 87.3 & 95.2 & 86.3  & 56.7  \\
			\midrule
			\multicolumn{2}{l}{Random Legal} & - & - & - & - & 19.6 \\
			\bottomrule
			
		\end{tabular}
	\caption{Accuracies and R-Precisions (\%) for predicting ending squares (``End-Actual'' and ``End-Other'' tasks). S, M, L in the first column refer to the training set sizes.
	}
	\label{tab:results-ending}
	
	}
\end{table}

Second, \textbf{transformers can learn to predict legal moves}.
This is shown by the \legalmove accuracies in  Table~\ref{tab:results-ending}, for which both UCI and UCI + RAP exceed 97\% accuracy. 
However, while the top predictions of the models have high accuracy, their ability to predict all legal moves is significantly lower, with R-precision of about 85\%. 
This is to be expected, since the model is trained on only actual games, where the emphasis is on ``meaningful" moves rather than any legal move. 
Due to similar reasons, there's a significant drop in performance when predicting ending squares for starting squares other than the one in the actual game. 
	The ``other" starting square would, by design, have legal continuations, but lack any ``meaningful" ones 	(see examples in Section \ref{sec:app_error_analysis}).

We find consistent gains in almost all metrics with the addition of RAP during training, with the gains being particularly impressive for small training sets. Thus, not only are the transformers robust to distribution shift due to RAP (available only during training), they are in fact able to utilize this additional information. Error analysis of illegal predictions shows that the addition of RAP improves piece tracking related errors (Section~\ref{sec:error_analysis}).  

The relatively low ExM accuracies of the models can be attributed to the inherent difficulty of the task.   
Randomly selecting an ending square from all legal ending squares 
has an accuracy of only around 20\%, implying that on average there are roughly 5 legal choices, which might explain the difficulty of the task.  

\begin{table}[t]
\centering{
		\setlength\tabcolsep{3.1pt}
		\begin{tabular}{llccccc}
			\toprule
			&  Model 	&  \multicolumn{4}{c}{\legalmove} & \exactmove\\
			
			&  &  \multicolumn{2}{c}{Actual} &   \multicolumn{2}{c}{Other} &  \\
			&  &  Acc. & R-Prec. & Acc. & R-Prec. & Acc.\\
			\midrule
			\multirow{6}{*}{S}
			& GPT2  				&  74.0 & 61.1 & 65.5 & 57.7  & 26.7  \\
			& GPT2 ($w=50$)    			& 69.5 & 57.4 & 60.4 & 53.2  & 23.1  \\
			
			& Reformer				&  71.0 & 57.2 & 61.5 & 53.5  & 24.8 \\
			& Performer				&  65.4 & 54.3 & 57.9 & 49.5  & 20.5  \\
			& LSTM 					&  60.2 & 51.0 & 52.5 & 46.4  & 20.9  \\
			& LSTM + RAP 			&  59.5 & 50.5 & 52.4 & 46.0  & 21.9 \\
			
			\midrule
			\multirow{6}{*}{M}
			& GPT2  				&  92.9 & 80.6 & 85.8 & 78.5  & 42.2   	\\
			& GPT2 ($w=50$)  			&  86.0 & 74.9 & 80.9 & 71.3  & 35.8  	\\
			& Reformer 				&  86.4 & 73.2 & 76.6 & 68.6  & 32.4 	\\
			& Performer 			&  89.2 & 76.3 & 80.5 & 71.5  & 36.0   	\\
			& LSTM  				&  73.8 & 61.6 & 67.2 & 59.8  & 32.0   	\\
			& LSTM + RAP 			&  77.5 & 64.9 & 69.7 & 61.7  & 32.1  	\\ 
			\midrule
			\multirow{6}{*}{L}
			& GPT2  				&  97.7 & 85.6 & 91.9 & 83.8  & 52.0  	\\
			& GPT2 ($w=50$)    		&  95.8 & 84.5 & 90.5 & 82.7  & 51.6  	\\
			& Reformer 				&  88.0 & 74.9 & 77.0 & 68.1  & 33.5 	\\
			& Performer				&  95.8 & 84.5 & 90.5 & 82.7  & 51.6  	\\
			
			& LSTM  				&  93.4 & 79.5 & 86.1 & 76.0  & 45.2  	\\
			& LSTM + RAP 			&  92.8 & 80.4 & 87.3 & 77.1  & 46.0	\\
			\bottomrule
			
		\end{tabular}
		
	\caption{Accuracy and R-Precision (\%) for predicting ending squares (``End-Actual'' and ``End-Other'' tasks)  with varying attention window sizes. 
		LSTM + RAP refers to LSTM  trained with UCI + RAP.
	}
	\label{tab:results-ending-window}
	
	}
\end{table}

\subsection{Compressing the Game History}
\label{sec:other_models}
The base transformer language model, based on GPT2, attends to the entire history (i.e., it uses ``full attention"), which results in complexity quadratic in the length of the sequence. We might wonder whether attending to this entire history is necessary for the impressive state tracking performance observed in the %
previous section.
We accordingly 
explore models that do not attend to the entire history in Table \ref{tab:results-ending-window}. %

We first experiment with a variant of the GPT2 model that limits its attention to a window of only the 50 most recent tokens (``GPT2 $(w=50)$''). In Table \ref{tab:results-ending-window} we see worse performance for this model across data sizes, but especially for small- and medium-sized datasets. 

In Table~\ref{tab:results-ending-window} we also consider a language model based on the LSTM~\citep{hochreiter1997long}, which considers only its current hidden state and cell state in making its predictions, and does not explicitly attend to the history. %
Here we find an even more significant drop in performance, in all settings. (Interestingly, we also find that training LSTM language models on sequences with RAP improves performance, but only for larger training sets; transformer language models generally improve when trained with RAP data). 

The results of GPT2 $(w = 50)$ and of the LSTM language model suggest that attending to the full game history is, unsurprisingly, useful for board state tracking in chess. This finding further suggests that the task of board state tracking in chess can serve as an excellent testbed for recently proposed transformer variants~\citep[\textit{inter alia}]{kitaev2020reformer,katharopoulos20,choromanski2021rethinking} that attempt to make use of long histories or contexts, but \textit{without} incurring a quadratic runtime.

\subsubsection{Approximate Attention Transformers}
\label{sec:limited_history}

We experiment with the recently proposed Reformer~\cite{kitaev2020reformer} and Performer~\cite{choromanski2021rethinking} architectures. Reformer replaces the ``full attention" with attention based on locality-sensitive hashing, while Performer approximates the ``full attention" with random features.\footnote{In practice, these models often use a combination of the proposed approximate global attention and simple local attention (for details see Section~\ref{sec:hyperparams}).}

The results, in Table~\ref{tab:results-ending-window}, suggest that the Performer generally outperforms the Reformer, except in the small dataset-setting. Furthermore, we find that neither of these architectures significantly outperforms the GPT2 $(w = 50)$ baseline, except for Performer in the medium-sized data setting. 
These models do, however, typically outperform the LSTM models. 
These results demonstrate the challenge of modeling chess with an approximate attention. 
To make it easier for future work to use chess state tracking as a way to benchmark future transformer architectures, we have added our task to the BIG-Bench benchmark~\cite{bigbench2022}.\footnote{\url{https://github.com/google/BIG-bench/tree/main/bigbench/benchmark_tasks/chess_state_tracking}}

\begin{figure*}[t]
\begin{subfigure}{0.3\textwidth}
   \includegraphics[width=\linewidth]{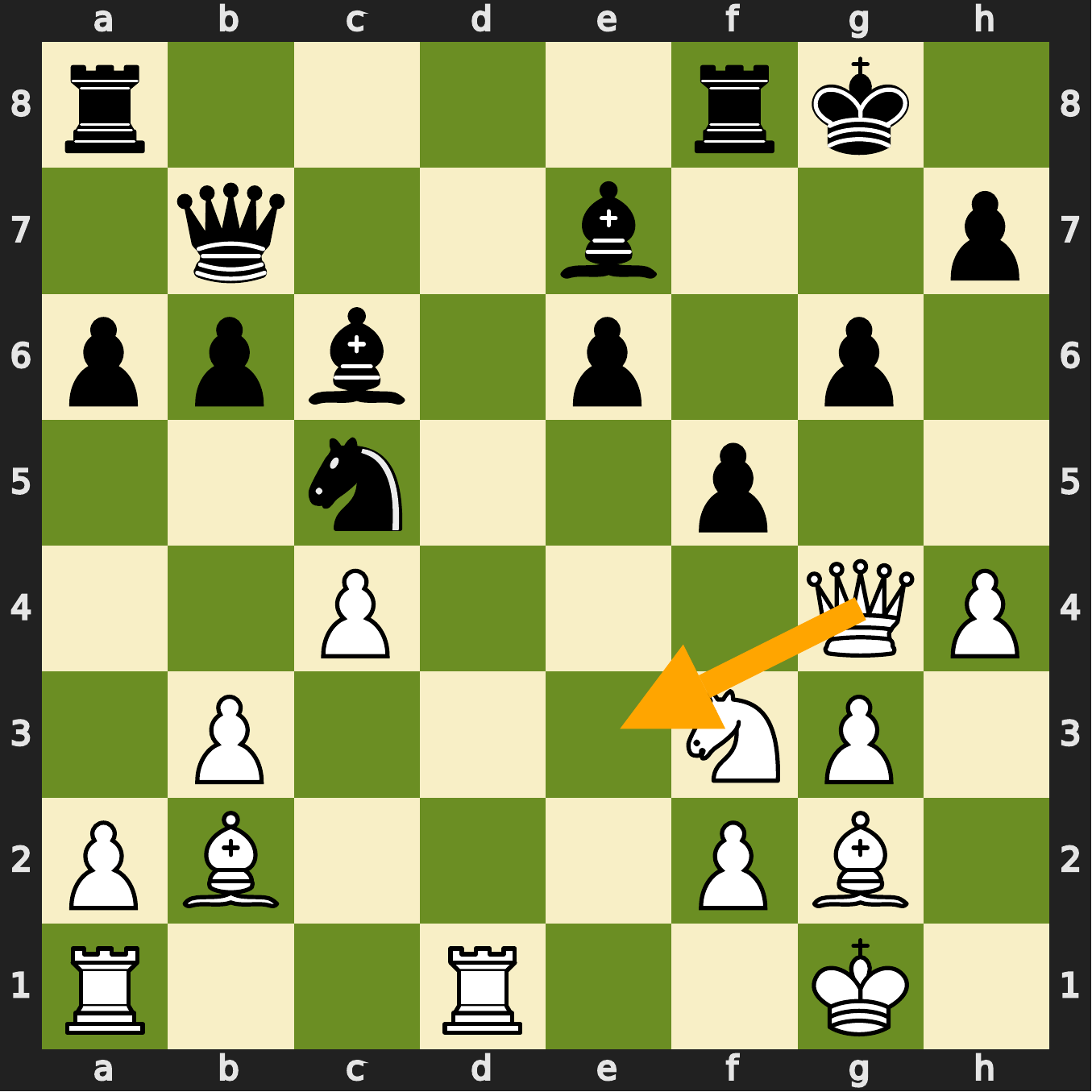}
   \caption{\emph{Syntax}: Queen can move like all other piece types except for knight.} \label{fig:error_syntax}
\end{subfigure}
\hspace*{\fill}
\begin{subfigure}{0.3\textwidth}
   \includegraphics[width=\linewidth]{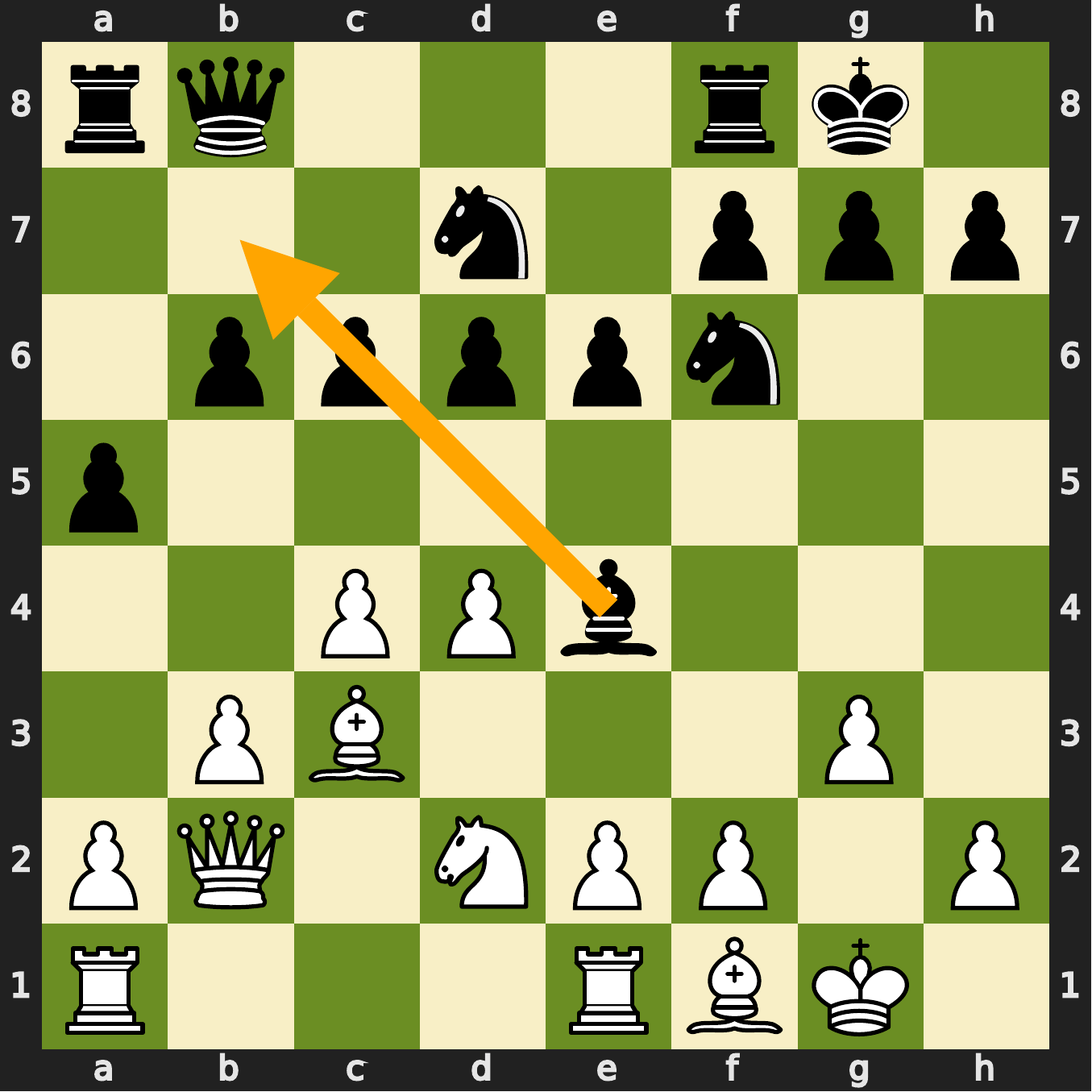}
   \caption{\emph{Path Obstruction}: The pawn at \pos{c6} is blocking the bishop.} \label{fig:error_path}
\end{subfigure}
\hspace*{\fill}
\begin{subfigure}{0.3\textwidth}
   \includegraphics[width=\linewidth]{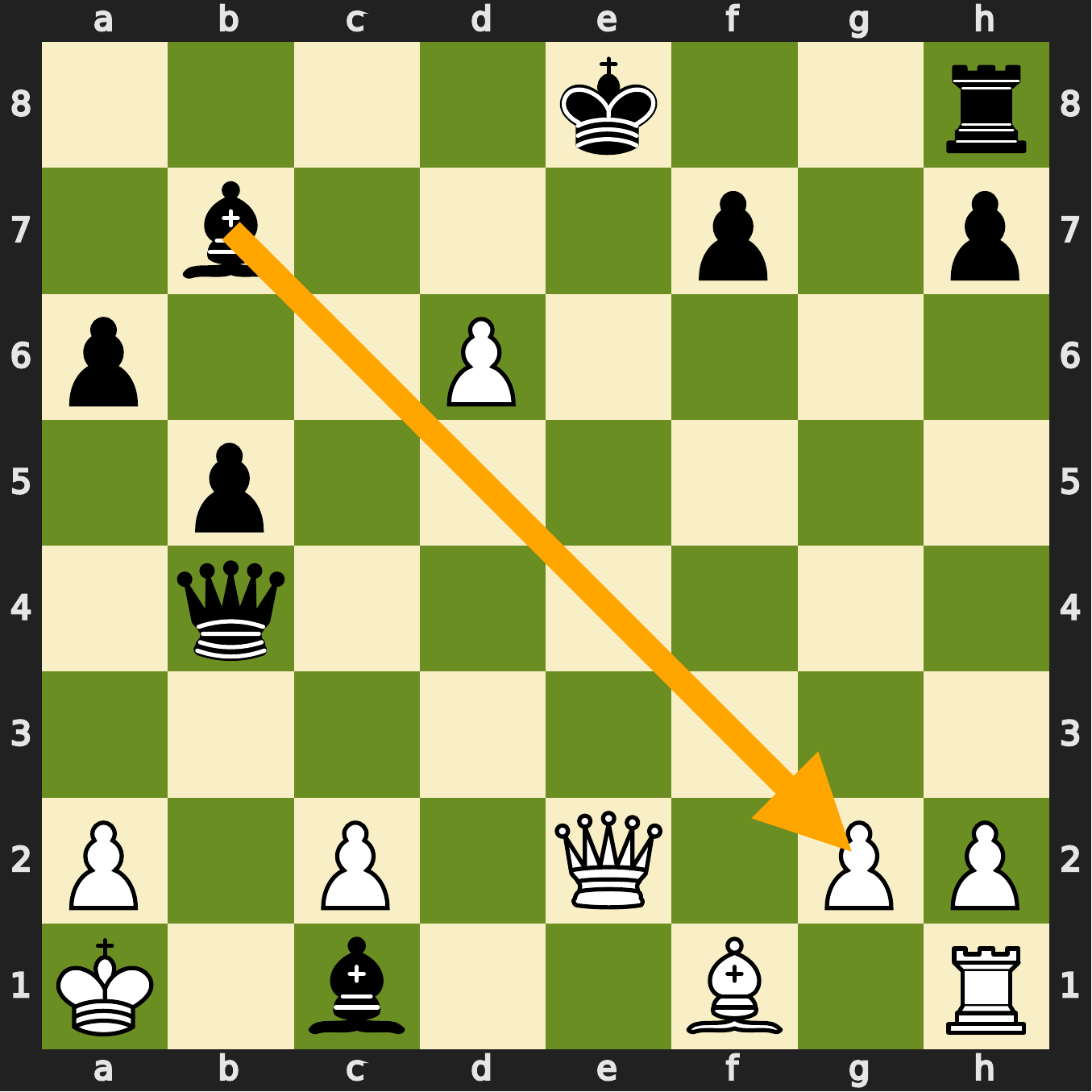}
   \caption{\emph{Pseudo Legal}: The black king remains in check.} \label{fig:error_pseudo}
\end{subfigure}
\caption{Instances of the three prominent categories of illegal ending square predictions.}
\label{fig:error_categories}
\end{figure*}

\begin{table*}[ht]
	\centering{
		\caption{Error counts for ending square prediction.}
		\label{tab:error_analysis}
\begin{tabular}{llcccccc}
	\toprule
	& \multirow{2.5}{*}{Model}   &  \multicolumn{3}{c}{Actual}      &   \multicolumn{3}{c}{Other}    \\
	\cmidrule(lr){3-5}\cmidrule(lr){6-8}
	&        & Syntax & Path Obst. & Pseudo Leg.  & Syntax & Path Obst. & Pseudo Leg.\\
	
	\midrule
	
	\multirow{4}{*}{Train-S} 
	& UCI 				& 168 & 48 & 40 & 173 & \phantom{1}90 & \phantom{1}80 \\ 
	& UCI + RAP			& \phantom{1}20 & 58 & 38 & \phantom{1}17 & \phantom{1}96 & \phantom{1}81\\
	& UCI + \piecetype	& \phantom{11}1 & 99 & 29 & \phantom{11}3 & 126 & \phantom{1}81\\
	& Performer 		& 235 			& 56 & 53 & 243 & \phantom{1}70 & 106\\
	
	\midrule
	
	\multirow{4}{*}{Train-M} 
	& UCI 				& \phantom{1}16 & 30 & 25 & \phantom{1}15 & \phantom{1}54 & \phantom{1}72 \\ 
	& UCI + RAP			& \phantom{11}3 & 30 & 18 & \phantom{11}7 & \phantom{1}56 & \phantom{1}55\\
	& UCI + \piecetype	& \phantom{11}0 & 36 & 17 & \phantom{11}3 & \phantom{1}59 & \phantom{1}53 \\
	& Performer			& \phantom{1}41 & 27 & 40 			& \phantom{1}42 	& \phantom{1}45 & 108\\

	\midrule
	
	\multirow{4}{*}{Train-L} 
	&  UCI 				& \phantom{11}1 & 10 & 12 & \phantom{11}4 & \phantom{1}26 & \phantom{1}49 \\ 
	& UCI + RAP			& \phantom{11}0 & 19 & 11 & \phantom{11}3 & \phantom{1}29 & \phantom{1}36 \\
	& UCI + \piecetype 	& \phantom{11}0 & 13 & \phantom{1}5 & \phantom{11}3 & \phantom{1}13 & \phantom{1}31\\
	& Performer			& \phantom{11}8 & 18 & 16 & \phantom{11}9 & \phantom{1}23 & \phantom{1}63 \\
	\bottomrule
\end{tabular}
	}
\end{table*}

\section{Error Analysis}
\label{sec:error_analysis}

In this section we analyze errors %
on the ending square prediction task. 
Incorrect predictions for this task %
can be (exhaustively) categorized into four categories:

\begin{itemizesquish}
	\itemsep0em 
	\item {\em Unreachable}: The predicted ending square cannot be reached %
	by any possible piece type %
	at the starting square regardless of the board state. 
	\item {\em Syntax}: The predicted ending square cannot be reached %
	by the piece type present at the starting square regardless of the board state. This error indicates failure at tracking the piece type present at the starting square. 
	\item {\em Path Obstruction}: The predicted ending square cannot be reached 
	because there are other pieces obstructing the %
	path. This error indicates failure at tracking other pieces on the board or a lack of %
	understanding that for all piece types except the knight, the path %
	must be clear. %
	For example, in Figure~\ref{fig:error_path}, the pawn at \pos{c6} blocks the bishop's move from \pos{e4} to \pos{b7}.
	\item {\em Pseudo Legal}: 
	The move is illegal because the moving player's king is in check at the end of the move. 
\end{itemizesquish}
Table~\ref{tab:error_analysis} shows error counts for the ending square prediction task. 
For brevity we omit unreachable errors since they are rare ($< 5$ for all models).

Errors across all categories decrease with more training data. For syntax errors this reduction is particularly dramatic, decreasing by roughly an order of magnitude when moving from Train-S to Train-M. %
In contrast, both path obstruction and pseudo legal errors decline more gradually.
Determining whether a path is blocked or if the king is in check requires a computation involving multiple piece locations which all need to be computed from the move  history. 
These trends suggest that identifying the piece type at a starting square 
requires data but is learnable, while keeping track of all \emph{other} pieces  on the board remains challenging even with large training sets.

UCI + RAP %
consistently outperforms %
UCI in syntax errors, %
the differences being largest for the small training sets. This validates our hypothesis that RAP can aid the model in piece tracking (Section~\ref{sec:rap_board}). Across other error categories we don't see consistent trends, suggesting piece tracking improvements do not necessarily translate to other error categories. The Performer generally makes more errors than the transformers, especially in the syntax category. The partial attention in the Performer may be limiting its ability to attend to the most relevant prior positions to determine the piece type at the given starting square. 

Predicting ending squares for the actual move made (``Actual'') is easier than for a randomly chosen legal move (``Other''). However, the syntax errors are comparable between the two settings, while there are many more path obstruction and pseudo legal errors for the Other instances. 
	The higher error rate for these categories could be because:
	\begin{itemizesquish}
		\item Avoiding path obstruction and check are difficult functions to learn and may therefore be being ``mimicked'' from training data rather than being learned as a general algorithmic function.
		\item The model is trained on only actual games with emphasis on meaningful moves rather than legal moves. We observe that some of the Other instances lack any ``meaningful" continuations (Section~\ref{sec:app_error_analysis}).
		\item There's a distribution shift between piece types moved in actual moves vs randomly chosen legal moves. 
		For example, the End-Actual task has only about 13\% prompts for moves made by king in comparison to the 33\% for the End-Other task (Section~\ref{sec:data_stats}).  We find that moves made by king have a higher chance of resulting in pseudo legal errors in comparison to other piece types (Section~\ref{sec:pseudo_legal}).  
	\end{itemizesquish}

\section{Related Work}

In this section we discuss related work not included in our discussion of prior work in Chapter 2. Specifically, we skip discussion of probing methodologies which we discuss in detail in Chapter 2. 

\paragraph{Simulated Worlds.} %
There have been several prior efforts in relating simulated worlds to natural language. 
The bAbI framework simulates a world modeled via templates to generate question answering tasks \citep{weston2015aicomplete}. 
The recent TextWorld framework facilitates generating, training, and evaluating interactive text-based games \citep{cote18textworld}. 
\citet{hermann17grounded} and \citet{hill17understanding} develop and use 3D world simulations for learning grounded language.
These efforts are similar to our proposed testbed of chess in the sense that the true world state is, by construction, available, but our setup differs in that it provides a natural way of probing the state tracking of a model trained with an LM objective.

\paragraph{Cloze Tasks for Natural Language Models.}
There has been a great deal of work on cloze tasks for evaluating natural language models~\citep{hermann2015cnn, hill2016cbt}. 
These tasks range from testing general text understanding~\citep{paperno-etal-2016-lambada} to targeting particular aspects of natural language, such as commonsense/pragmatics \citep{mostafazadeh-etal-2016-corpus, ettinger2020bert}, narrative understanding \citep{mostafazadeh-etal-2017-lsdsem}, and factual knowledge \citep{petroni-etal-2019-language}.
Creating these tasks often requires human curation, and the evaluation is typically limited to exact match.\footnote{Automated cloze tasks without human filtering can yield instances which even humans can't answer \citep{hill2016cbt}.}  
Our proposed tasks are a form of cloze tasks, but can be precisely 
automated so that they require no human curation, and can be evaluated at a fine-grained level.

\paragraph{Deep Learning for Chess.}
Deep networks have been used in prior work to predict the next move given the true game state~\cite{david16deepchess, Oshri2015PredictingMI}.
For example, using only self-play and the rules of chess, AlphaZero achieves superhuman performance starting from random play~\citep{silver18general}.
The focus of this prior work is the quality of game play given the true board state, while we use chess as a testbed for evaluating a language model's board state tracking capability.
Recently there has also been work focusing on transformer language models for chess \citep{presser2020chess,cheng2020chess,noever2020chess}. 
This work is similar to ours in the sense that the input is limited to the move sequence without the true board state, but the focus is again the quality of game play rather than the model's awareness of the underlying state.

\section{Additional Details}

\subsection{Effect of Model Size}

In this section, we present results for training larger transformer models to evaluate the impact of increase in model size with increase in training set size.

\begin{table}[t]
\caption{Accuracy and R-Precision (\%) for predicting ending squares (``End-Actual'' and ``End-Other'' tasks)  for different model sizes.
	S, M, L in the first column refer to the training set sizes.
	 GPT2-small = \{12 layers, 12 heads, 768 embedding size\}; GPT2-intermediate = \{16 layers, 12 heads, 768 embedding size\}; and GPT2-medium = \{24 layers, 16 heads, 1024 embedding size\}.
}
\label{tab:results-ending-size}
    \centering{
\setlength\tabcolsep{4pt}
\begin{tabular}{llccccc}
	\toprule
    	&  Model 	&  \multicolumn{4}{c}{\legalmove} & \exactmove\\
    
    &  &  \multicolumn{2}{c}{Actual} &   \multicolumn{2}{c}{Other} &  \\
    &  &  Acc. & R-Prec. & Acc. & R-Prec. & Acc.\\
    \midrule
\multirow{3}{*}{S}
& GPT2-small  		&  74.0 & 61.1 & 65.5 & 57.7  & 26.7  \\
& GPT2-inter.  	&  72.3 & 60.7 & 64.5 & 58.6  & 24.8  \\
& GPT2-med.  		&  67.8 & 58.2 & 62.5 & 55.7  & 24.5  \\
\midrule
\multirow{3}{*}{M}
& GPT2-small  		&  92.9 & 80.6 & 85.8 & 78.5  & 42.2  \\
& GPT2-inter.  	&  92.9 & 81.8 & 84.8 & 77.8  & 41.5  \\
& GPT2-med.  		&  93.7 & 81.8 & 86.2 & 77.1  & 41.7  \\
\midrule
\multirow{3}{*}{L}
& GPT2-small  		&  97.7 & 85.6 & 91.9 & 83.8  & 52.0  \\
& GPT2-inter.  	&  97.5 & 86.6 & 94.7 & 85.2  & 54.0  \\
& GPT2-med.  		&  98.2 & 87.4 & 94.6 & 85.8  & 57.0  \\
\bottomrule

\end{tabular}
}
\end{table}

Table~\ref{tab:results-ending-size} presents results with transformer models of sizes varying from GPT2-small to GPT2-medium.
We also introduce a new configuration, referred to as GPT2-intermediate, which serves as an intermediate between GPT2-small and GPT2-medium.
For Train-S, GPT2-small outperforms both GPT2-intermediate and GPT2-medium on almost all evaluations.
However, with increasing in training data,  GPT2-intermediate and GPT2-medium are are able to
outperform GPT2-small on most evaluations. 

These results are along the expected lines of larger training sets alleviating the overfitting problem with larger models~\citep{kaplan2020scaling}.  
Note that we stick with the default GPT2 configuration for all our experiments. Tuning the regularization hyperparameters such as dropout, can further improve results for bigger models trained with small training sets.

\begin{figure*}
\begin{minipage}{\textwidth}
	\begin{minipage}[b]{0.5\textwidth}
		\setlength\tabcolsep{4pt}
		\begin{tabular}{lcc}
			\toprule
			Split & \# of games (in $10^3$) & \# of moves (in $10^6$)\\ %
			\midrule
			Train-S & 	\phantom{1}15 	& \phantom{1}1.1\\ %
			Train-M & 	\phantom{1}50 	& \phantom{1}3.7\\ %
			Train-L & 	200 			& 15.0\\ 			 %
			Dev 	& 	\phantom{1}15 	& \phantom{1}1.1\\ %
			Test 	& 	\phantom{1}15 	& \phantom{1}1.1\\ %
			\bottomrule
		\end{tabular}
		\captionof{table}{Statistics of the language modeling data.}
		\label{tab:data_stats}
	\end{minipage}
	\hfill
	\begin{minipage}[b]{0.48\textwidth}
		\centering
		\includegraphics[width=0.8\textwidth]{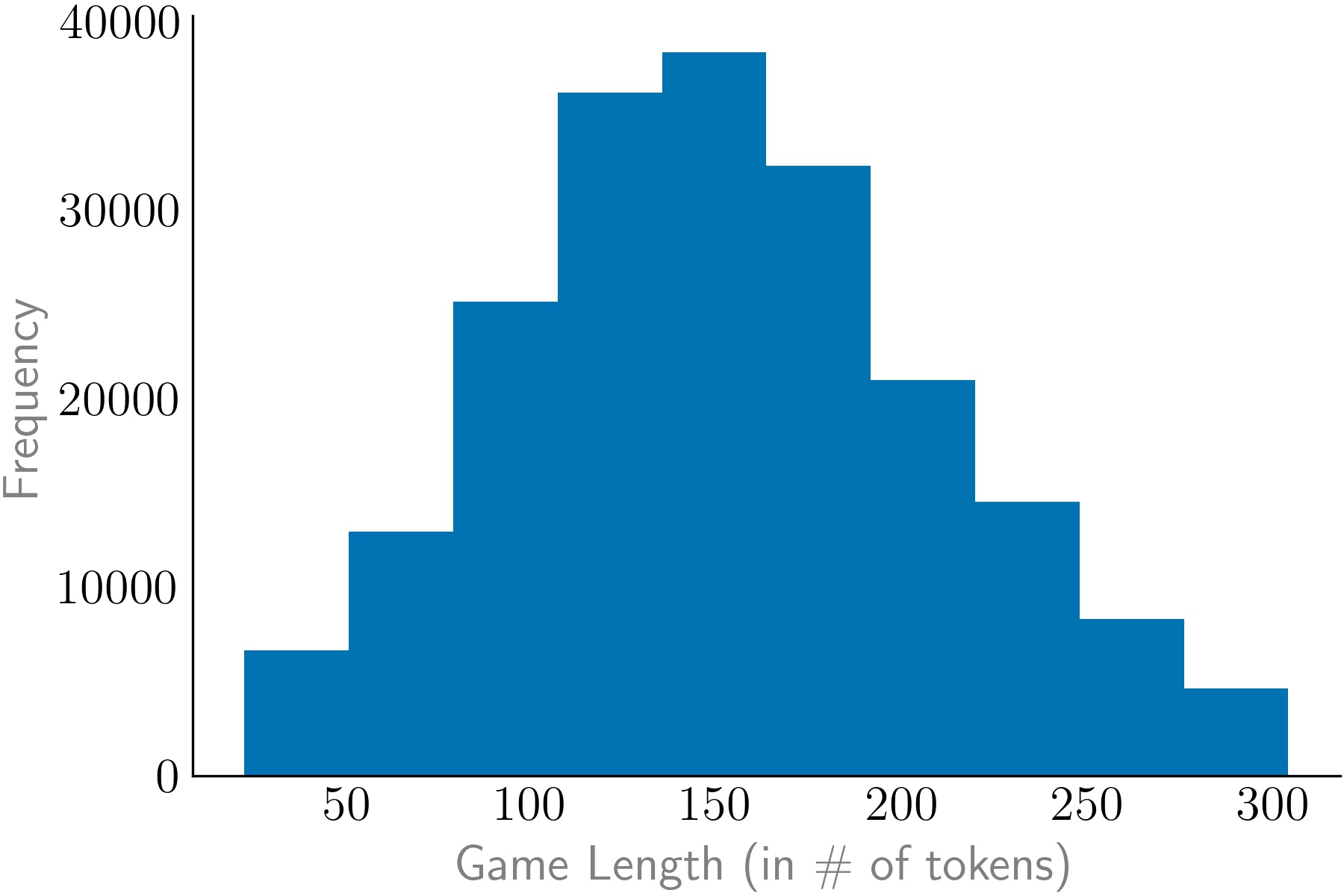}
		\vspace{-0.1in}
		\captionof{figure}{Histogram of tokenized game lengths for Train-L.}
		\label{fig:hist_len}
	\end{minipage}
\end{minipage}
\end{figure*}

\begin{figure*}
	\begin{minipage}{\textwidth}
		\begin{minipage}[b]{0.48\textwidth}
			\centering
			\setlength\tabcolsep{4pt}
			\begin{tabular}{lccc}
				\toprule
				Piece type & End/Start-Actual & End-Other & Start-Other\\ %
				\midrule
				Rook (\pos{R}) & 	358 	& 273     & 197\\ 
				Knight (\pos{N}) & 	144 	& 136 & 126\\  
				Bishop (\pos{B}) & 	164 			& 170 & 161\\
				Queen (\pos{Q}) 	& 	204 	& 103  & 129 \\ 
				
				King (\pos{K}) & 	130 	& 318  & 387\\
				\midrule
				Total & 1000 & 1000 & 1000 \\\bottomrule
			\end{tabular}
			\captionof{table}{Piece type counts for ending square prediction prompts.}
			\label{tab:probe_data_stats}
		\end{minipage}
		\hfill
		\begin{minipage}[b]{0.48\textwidth}
			\centering
			\includegraphics[width=0.8\textwidth]{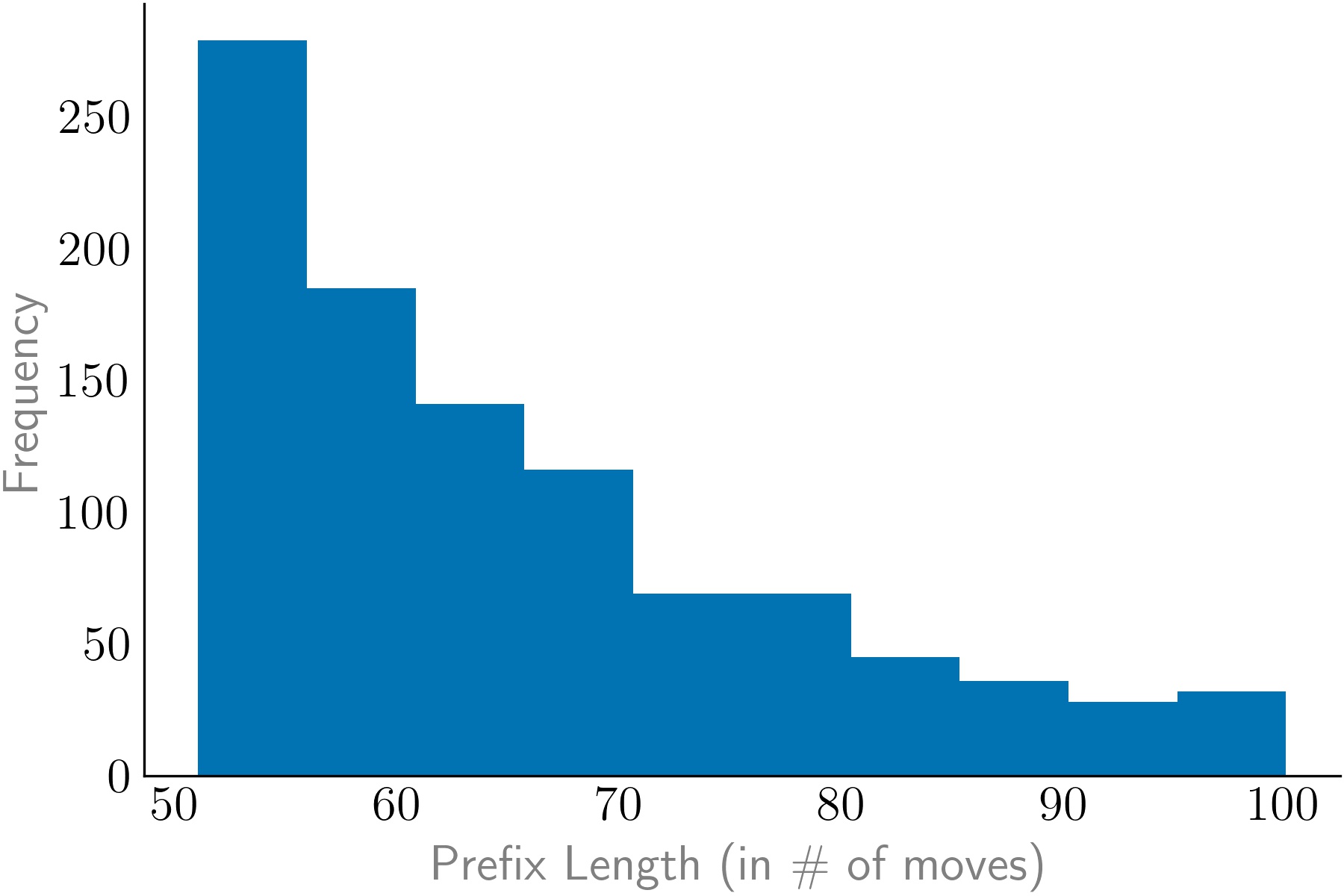}
			\vspace{-0.1in}
			\captionof{figure}{Histogram of prefix lengths of board state prompts.}
			\label{fig:hist_prompt}
		\end{minipage}
		\end{minipage}
\end{figure*}

\subsection{Data Statistics}
\label{sec:data_stats}
Table~\ref{tab:data_stats} presents the statistics of the language modeling dataset used. The average game length for all splits is around 75 moves.
Figure \ref{fig:hist_len} presents the histogram of lengths of tokenized UCI games in Train-L. 

Table~\ref{tab:probe_data_stats} presents the piece type counts for the different board state prompts. All the prompts have the same game prefix i.e. the previous moves, though, the move prefix is different - starting square of the move is used for the ending square predictions while the piece type used for the move is used for the starting square prediction. As the game prefix is the same, End-Actual and Start-Actual use the same piece type for each prompt. 
For the End-Other task, we pick a random starting square among all starting squares from which a legal move can be made, except the starting square used for the actual move.
For the Start-Other task, we pick a random piece type among all piece types which can be legally moved, except the piece type which is actually moved in the game.
The different strategies for picking the random starting square and random piece type explains the different piece type distributions for End-Other and Start-Other. Figure~\ref{fig:hist_prompt} shows the histogram of length of game prefixes (in number of moves) used in board state prompts.

\subsection{Model Hyperparameters and Training time}
\label{sec:hyperparams}
Table~\ref{tab:hyperparam} presents the hyperparameters used for the different models. For the base language model based on GPT2-small we use the default hyperparameters. For other baselines we perform separate hyperparameter grid search for Train-S and Train-M, and use the Train-M hyperparameters for Train-L. 
Only exception to this rule is the Reformer model, which we found particularly difficult to train, for which we explain the details next.

Reformer model uses a combination of local and LSH-based self attention layers. We borrow the attention layer configuration used for enwiki8 experiments in the original paper. \footnote{\url{https://cdn.huggingface.co/google/reformer-enwik8/config.json}} 
For both the local and LSH attention, we use a chunk length of 50 tokens - the model divides the sequence into chunks with the causal attention limited to tokens within a chunk and one before.
The transformers library implementation suggests not pre-specifying the number of hash buckets. The implementation sets the number of buckets on the fly based on the sequence length, which in this case it sets to 8 hash buckets. The original paper experiments with the number of hashing rounds and shows consistent improvement with more hashing rounds. However, we didn't find that to be the case, and hyperparameter tuning sometimes preferred lower number of hashing rounds. 
We found it particularly difficult to train the model on Train-L where the training loss started increasing after only a couple of epochs which triggered early stopping. To alleviate this: (a) we experimented with a different learning rate decay mechanism, namely, the inverse square root decay schedule which lead to slightly better final results  \footnote{\url{https://fairseq.readthedocs.io/en/latest/_modules/fairseq/optim/lr_scheduler/inverse_square_root_schedule.html}}, and (b) perform a separate hyperparameter tuning for Train-L.  
Note that all other experiments use the learning rate schedule described in Section~\ref{sec:setup} and use the hyperparameters for Train-M.  

\paragraph{Training Time}
Experiments with transformers take around 4 hrs for Train-S, less than 10 hrs for Train-M, and less than 24 hrs for Train-L on a single GeForce RTX 2080 Ti. For LSTMs it takes less than 2 hrs for Train-S, less than 4 hrs for Train-M, and less than 8 hrs for Train-L on a single GeForce RTX 2080 Ti.

\begin{table*}
	
\centering{
	\caption{Hyperparameters used for the different models.
		Bold values are selected for all the training set sizes, otherwise, training set specific hyperparameter values are specified via parenthesis.
	}
	\label{tab:hyperparam}
	\begin{tabular}{lllll}
		\toprule
	  Hyperparameters&	GPT2	 		& 	LSTM	  	& 	Reformer		&	 Performer \\\midrule
\# of layers  &  	12		& 	3 (S), 4 (M, L), 5				& 12 & 12 \\
\# of attention heads  & 12		& 	0	&  12 & 12\\
Embedding size 	& 768	& \textbf{768}, 1024 	& 768 & 768 \\
Hidden size		& 768	& 768, \textbf{1024}	& 768 & 768 \\
Dropout probability & 0.1	& 0, 0.1, \textbf{0.2}, 0.5 & 0.05 (0 for LSH attn.) & 0.1\\
\# of hash buckets	  & - & - & 8 & -\\
\# rounds of hashing  & - & - & 1 (L), 2 (S), 4 (M) & - \\
Axial position shape  & - & - & [14, 25] &- \\
Axial position embedding size  & - & - & [256, 512] &- \\
Generalized attention & - & - & - & \textbf{Yes}, No \\
Feature redraw frequency  			& - & - & - & 1000 \\
\# of local attention heads 		& - & - & -  & 0 (M, L), 6 (S) \\
Local/LSH attn chunk size 				& - & - & 50 & - 				\\
Local attn window size 				& - & - & - & 50 					\\\midrule
\# of parameters (in millions) 	   	& 85	& 24 (S)/32 (M, L)  & 83 & 86\\
				
\bottomrule

\end{tabular}
}
\end{table*}

\subsection{Detailed Error Analysis}
\label{sec:app_error_analysis}
\begin{figure*}[!ht]
	\centering{
		\hspace*{\fill}
\begin{subfigure}{0.43\textwidth}
   \includegraphics[width=\linewidth]{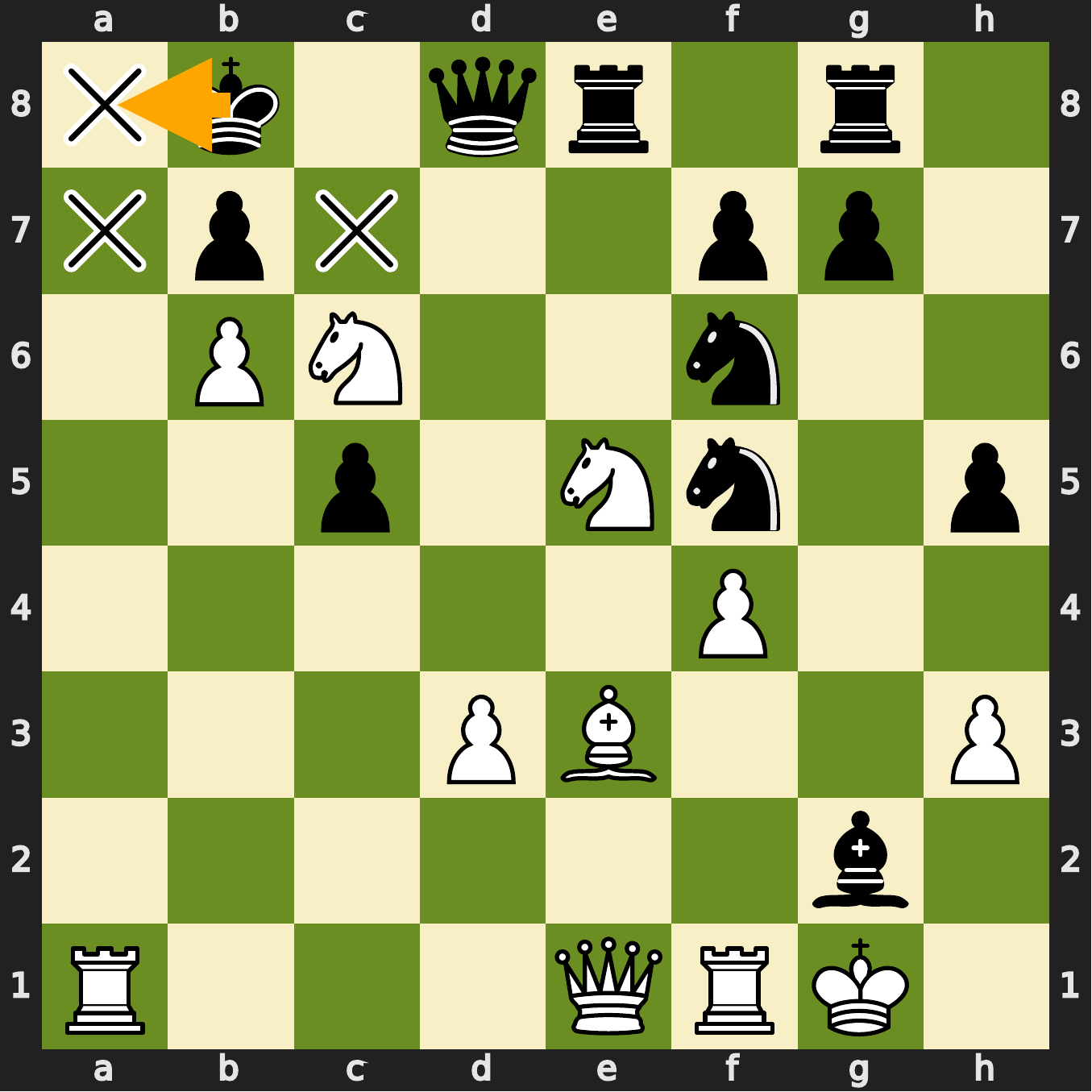}
   \caption{\emph{Check + King}: Black king is in check and the predicted ending square is already covered by the white rook on \pos{a1}.} 
   \label{fig:error_check_king}
\end{subfigure}
\hspace*{\fill}
\begin{subfigure}{0.43\textwidth}
	\includegraphics[width=\linewidth]{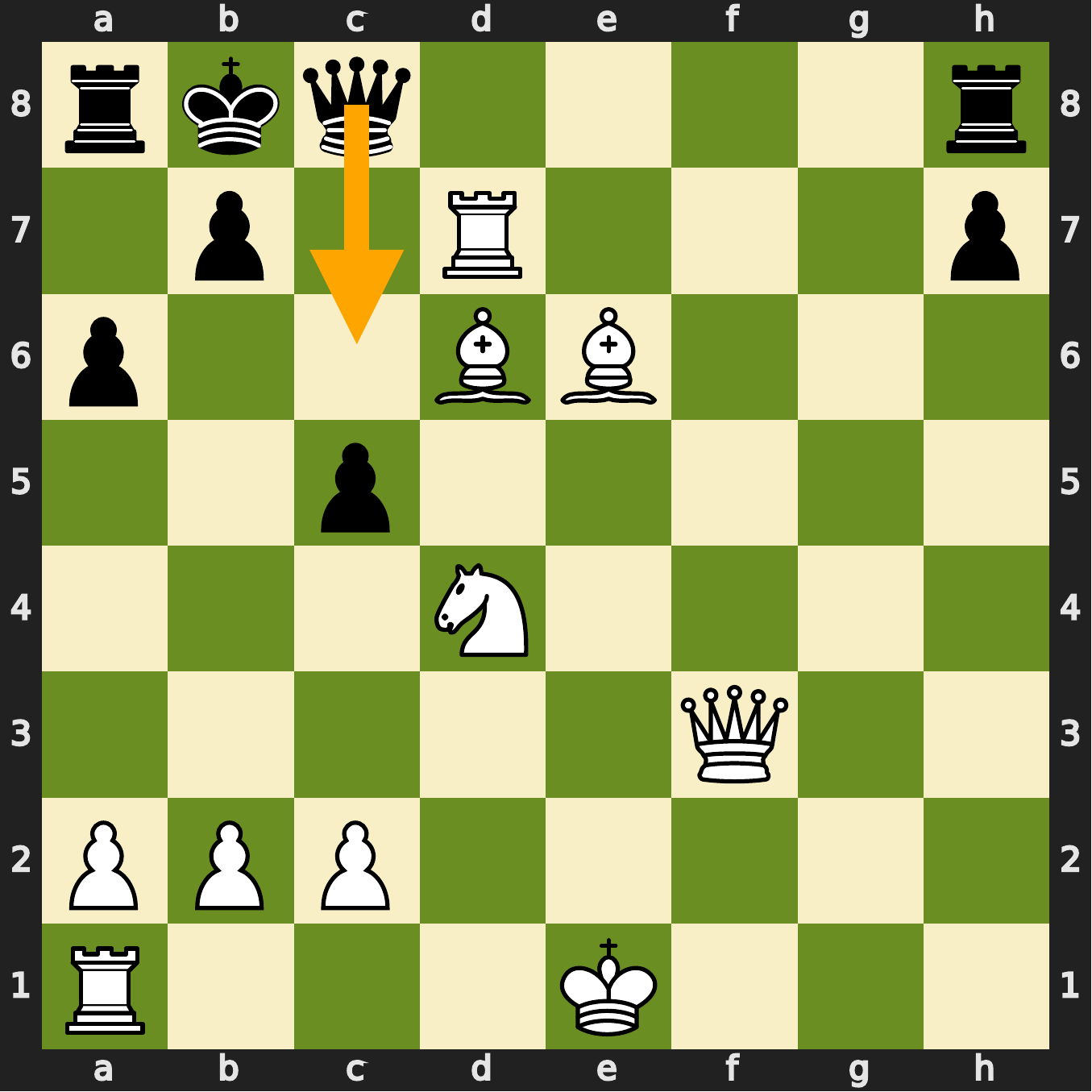}
	\caption{\emph{Check + Other}: Black king is in check and the only legal move for the black queen is \pos{c7} but the model predicts \pos{c6}.} 
	\label{fig:error_check_no_king}
\end{subfigure}
\hspace*{\fill}
\\[1em]
\hspace*{\fill}
\begin{subfigure}{0.43\textwidth}
   \includegraphics[width=\linewidth]{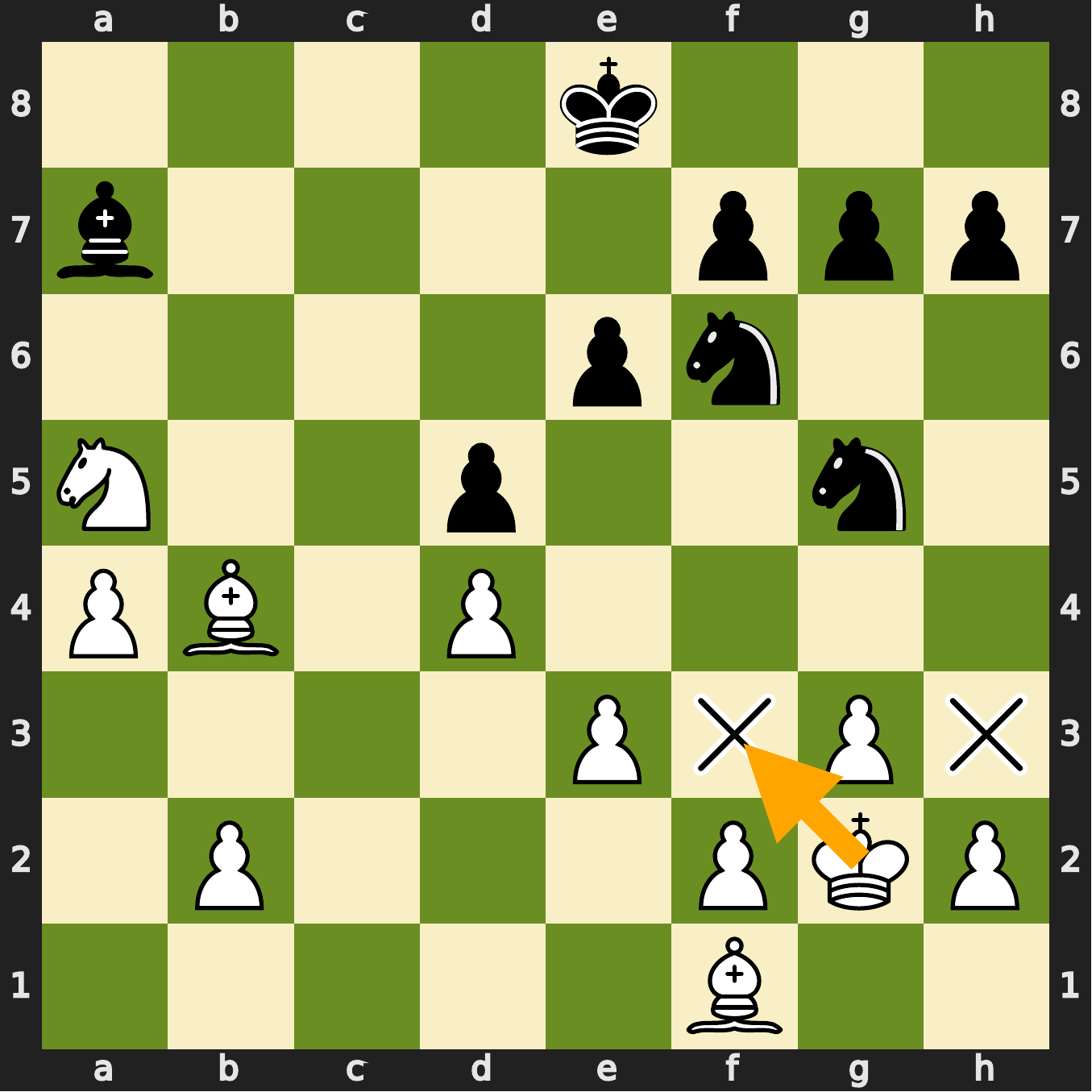}
   \caption{\emph{No Check + King}: The predicted ending square \pos{f3} for the white king is guarded by the black knight at \pos{g5}.} 
   \label{fig:error_no_check_king}
\end{subfigure}
\hspace*{\fill}
\begin{subfigure}{0.43\textwidth}
	\includegraphics[width=\linewidth]{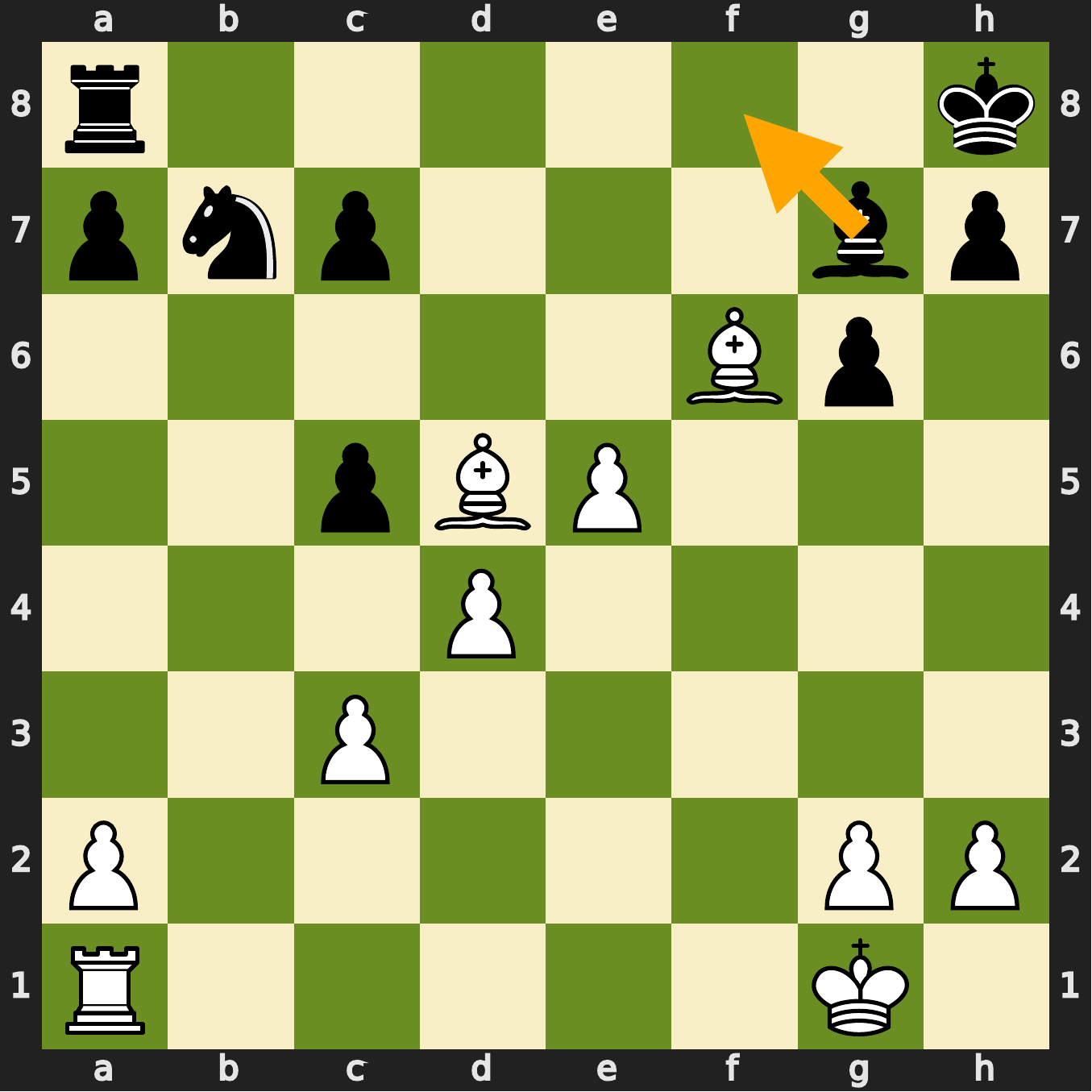}
	\caption{\emph{No Check + Other}: The predicted ending square \pos{f8} for the black bishop exposes its king to the white bishop at \pos{f6}. } 
	\label{fig:error_no_check_no_king}
\end{subfigure}
\hspace*{\fill}
\caption{Four combinations of the king being in check or not, and if the king is moved or not, that can result in Pseudo Legal errors.}
\label{fig:pseudo_legal_errors}
}
\end{figure*}

In this section we conduct a more in-depth analysis of errors made by the UCI model trained with Train-L for the ending square prediction task. We limit our focus to the two main error categories, namely, Pseudo Legal and Path Obstruction.

\begin{table}[t]
	\centering{
		\caption{Pseudo Legal error counts for different categories. For the total column we remove instances with errors of other category.}
		\label{tab:pseudo_legal}
		\begin{tabular}{lcccc}
			\toprule
			Category & \multicolumn{2}{c}{End-Actual} & \multicolumn{2}{c}{End-Other}\\
			& Errors & Total & Errors & Total \\
			\midrule
			Check + King 		& 1	& \phantom{1}27  & \phantom{1}2 & \phantom{1}20	\\
			Check + Other 		& 7	& \phantom{1}26  & 16			& \phantom{1}33 \\ 			 
			No Check + King 	& 4	& 101 			& 31 			& 296	\\ 
			No Check + Other 	& 0 & 835 &	\phantom{1}0  			& 619	\\ 
			\midrule
			Total 				& 12 & 989  	& 49 			& 968\\\bottomrule
		\end{tabular}
	}
	
\end{table}

\begin{table}[ht]
	\caption{Piece type counts for Path Obstruction error category. For the total column we remove instances with errors of other category.}
	\label{tab:path_obs}
	\centering
	\begin{tabular}{lcccc}
		\toprule
		Piece type & \multicolumn{2}{c}{End-Actual} & \multicolumn{2}{c}{End-Other} \\ %
		& Errors & Total & Errors & Total \\
		\midrule
		Rook (\pos{R}) 	& 	3 	& 355	& 17  			& 267    		\\ 
		Knight (\pos{N}) 	& 	1  	& 144	& \phantom{1}1 	& 131 			\\  
		Bishop (\pos{B}) 	& 	1 	& 162	& \phantom{1}3 	& 164 			\\
		Queen (\pos{Q}) 	& 	4  	& 202	& \phantom{1}4  & \phantom{1}99	\\ 
		King (\pos{K}) 	& 	1  	& 124 	& \phantom{1}1  & 284			\\
		\midrule
		Total 		& 	10 	& 987  	& 26 			& 945\\\bottomrule
	\end{tabular}
\end{table}

\subsection{Pseudo Legal Errors}
\label{sec:pseudo_legal}
We conduct our analysis by categorizing instances according to: (a) if the king was in check before the current move, and (b) if the king is being moved in the current move. 
Figure~\ref{fig:pseudo_legal_errors} presents one instance for each of these four categories.
Table~\ref{tab:pseudo_legal} presents the breakdown of errors for the End-Actual and End-Other instances. The key takeaways from the error categorization are: (a) Error counts for ``Check + King" and ``No Check + Other" are relatively low and similar across the two classes of prompts. (b) ``Check + Other" i.e.\ the king is in check and some other piece is moved, has high count for both the splits. The particularly high count for End-Other could be explained by the lack of ``meaningful" moves for certain prompts of this kind. For example, in figure~\ref{fig:error_check_no_king} the prompt asks for the queen at \pos{c8} to move, and the only legal continuation is for the queen to bring itself to the firing line at \pos{c7}. (c) ``No Check + King"  is another common error category. 
The significantly higher error count for End-Other could be due to a combination of the higher frequency of such prompts and the out-of-distribution prompts.

\begin{figure*}[!ht]
	\centering{
		\hspace*{\fill}
		\begin{subfigure}{0.4\textwidth}
			\includegraphics[width=\linewidth]{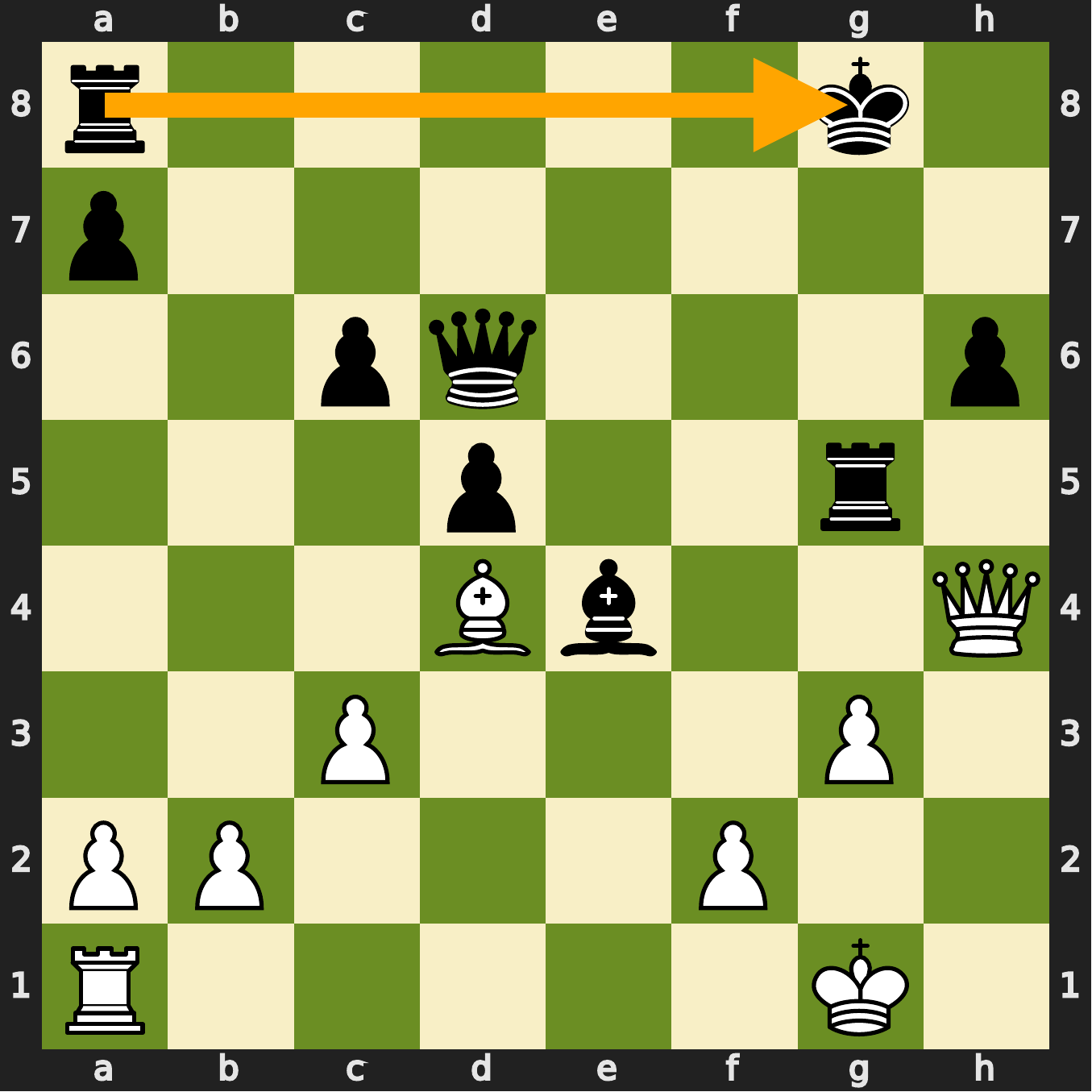}
			\caption{Rook forgets about its own king at \pos{g8}!} 
			\label{fig:rook_path}
		\end{subfigure}
		\hspace*{\fill}
		\begin{subfigure}{0.4\textwidth}
			\includegraphics[width=\linewidth]{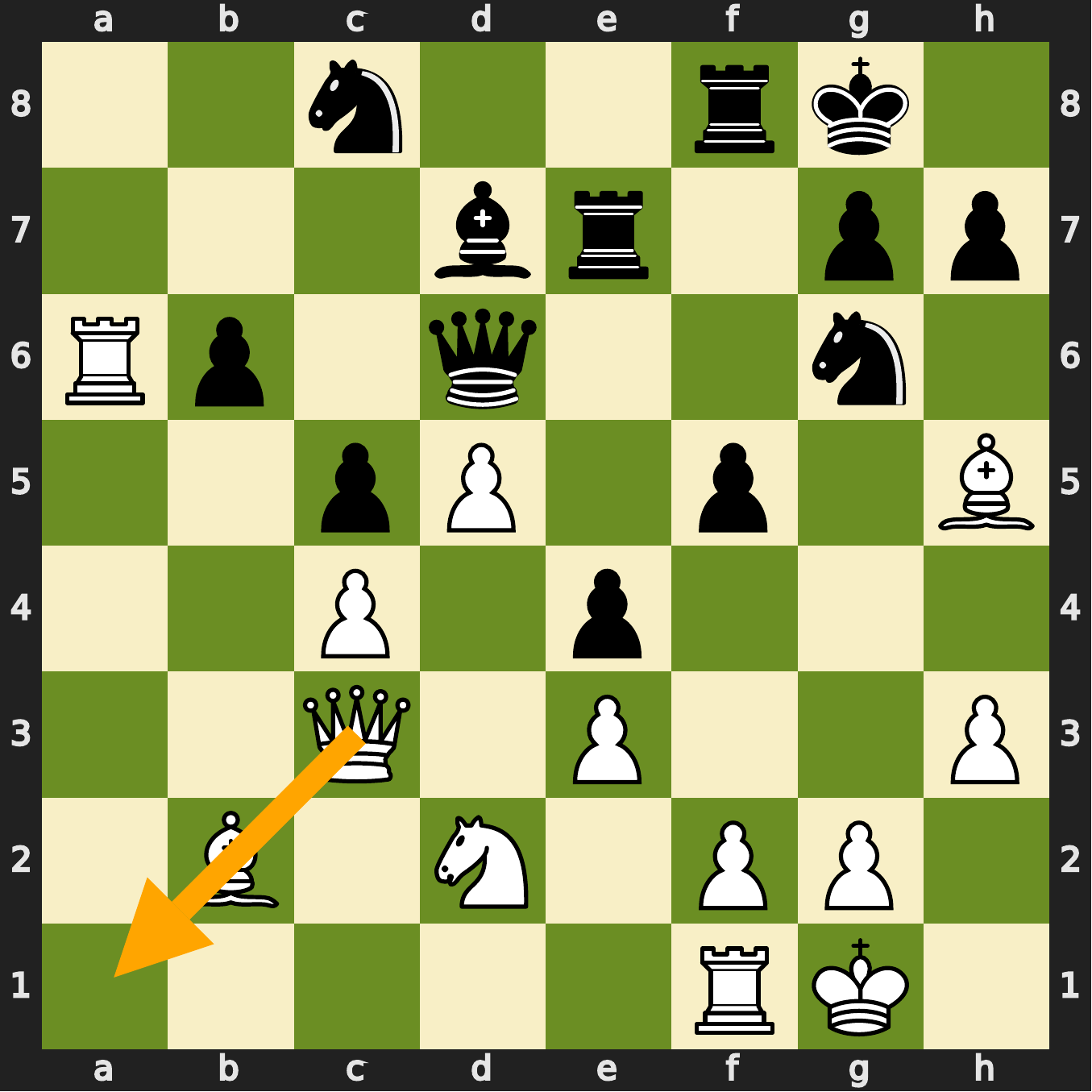}
			\caption{Bishop at \pos{b2} stands in the way of the queen.} 
			\label{fig:queen_path}
		\end{subfigure}
		\hspace*{\fill}
		\\[1em]
		\hspace*{\fill}
		\begin{subfigure}{0.4\textwidth}
			\includegraphics[width=\linewidth]{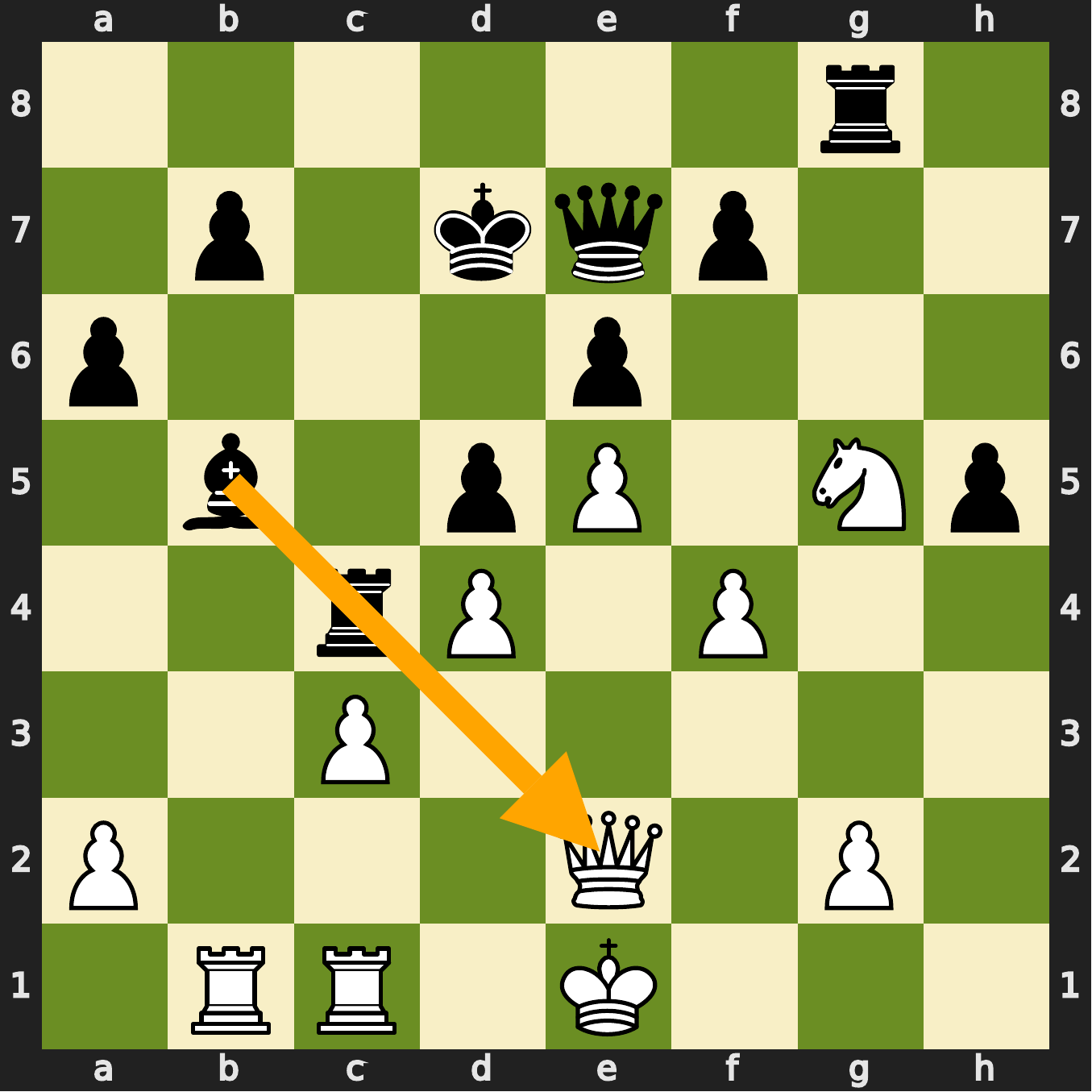}
			\caption{Bishop forgets reality in pursuit of fantasy queen kill!} 
			\label{fig:bishop_path}
		\end{subfigure}
		\hspace*{\fill}
		\begin{subfigure}{0.4\textwidth}
			\includegraphics[width=\linewidth]{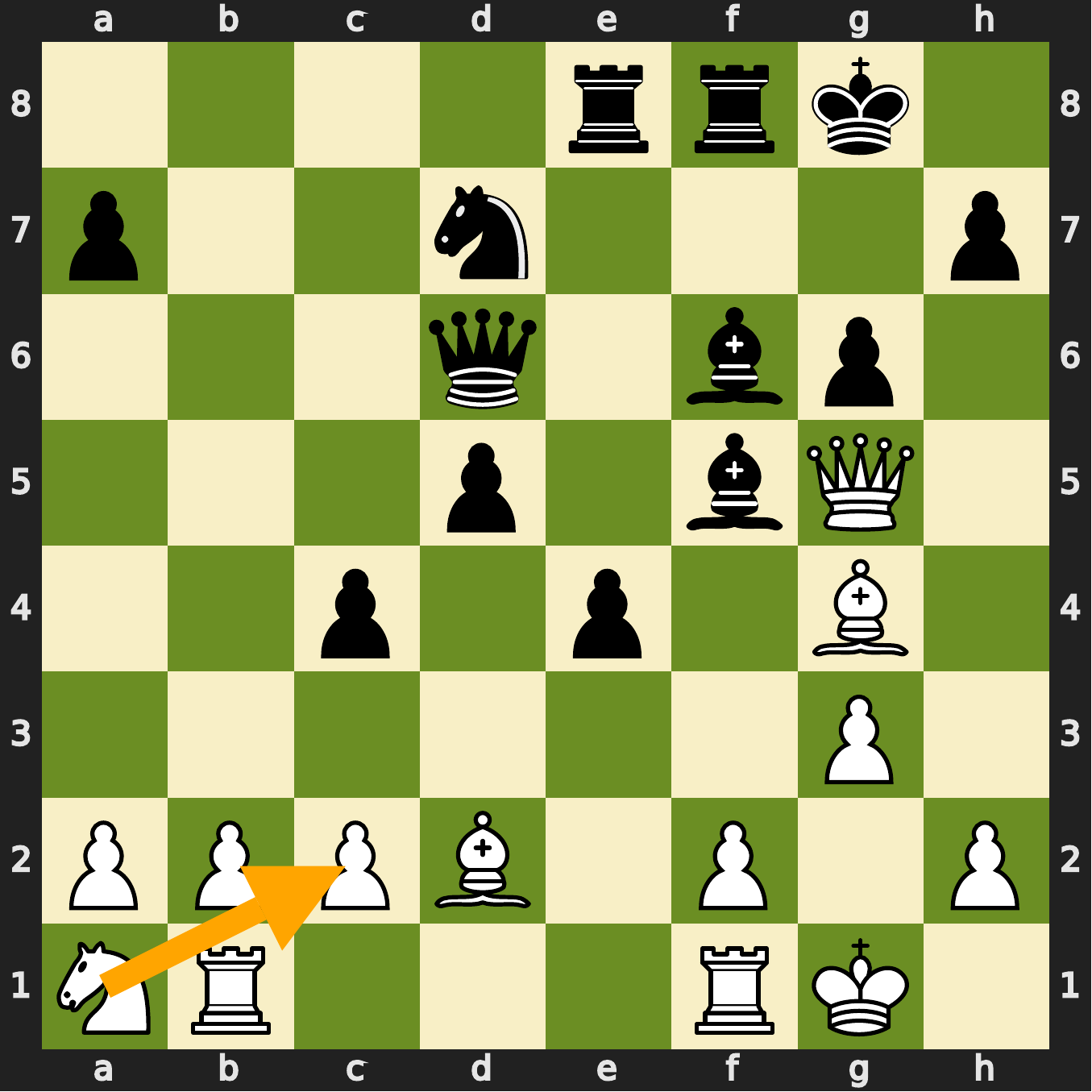}
			\caption{A trapped, frustrated knight is out to kill its own pawn!} 
			\label{fig:knight_path}
		\end{subfigure}
		\hspace*{\fill}
		\caption{Instances of Path Obstruction errors with different piece types.}
		\label{fig:path_obstruction}
	}
\end{figure*}

\begin{figure*}[!ht]
	\centering{
		\hspace*{\fill}
		\begin{subfigure}{0.35\textwidth}
			\includegraphics[width=\linewidth]{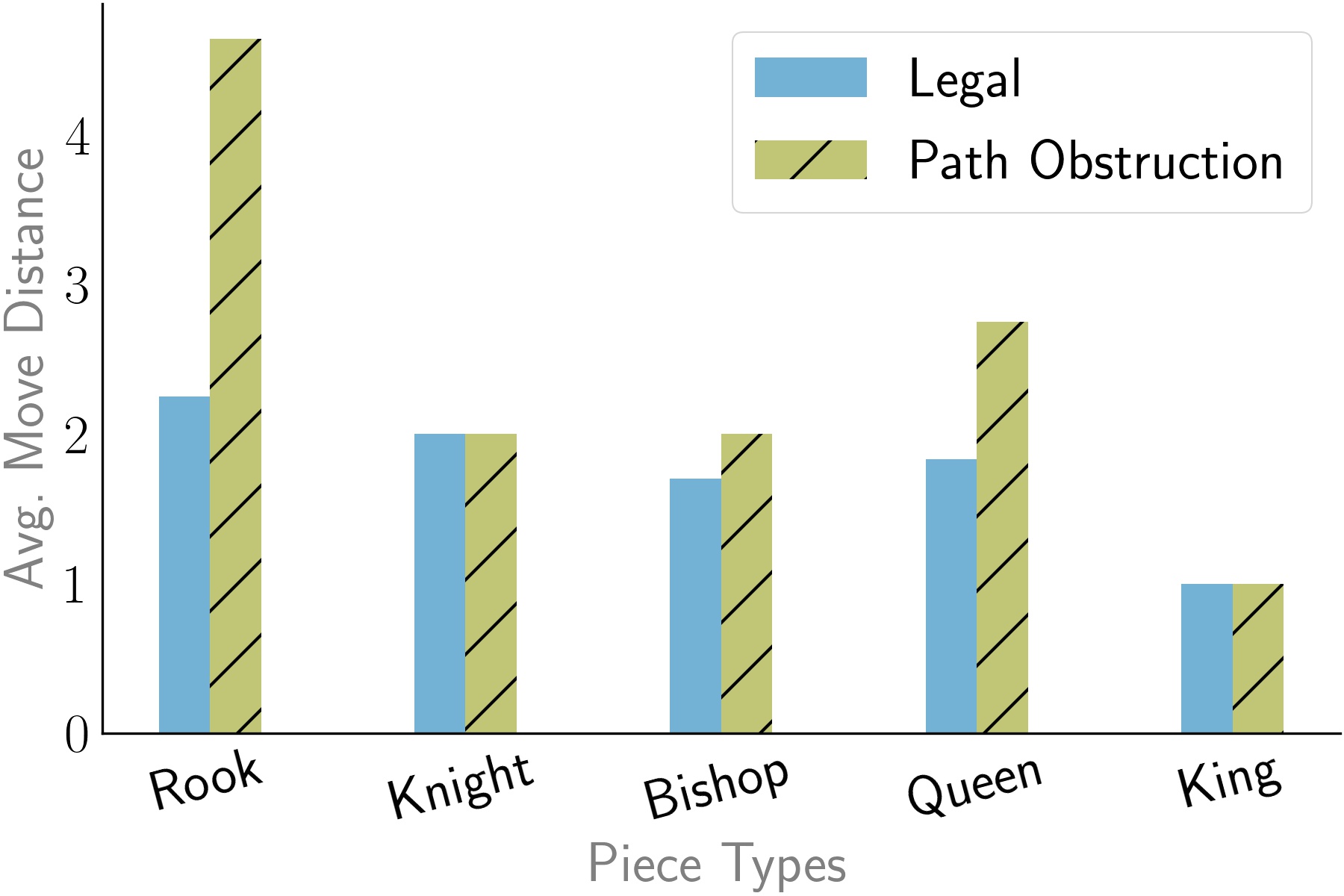}
			\caption{End-Actual} 
			\label{fig:path_length_actual}
		\end{subfigure}
		\hspace*{\fill}
		\begin{subfigure}{0.35\textwidth}
			\includegraphics[width=\linewidth]{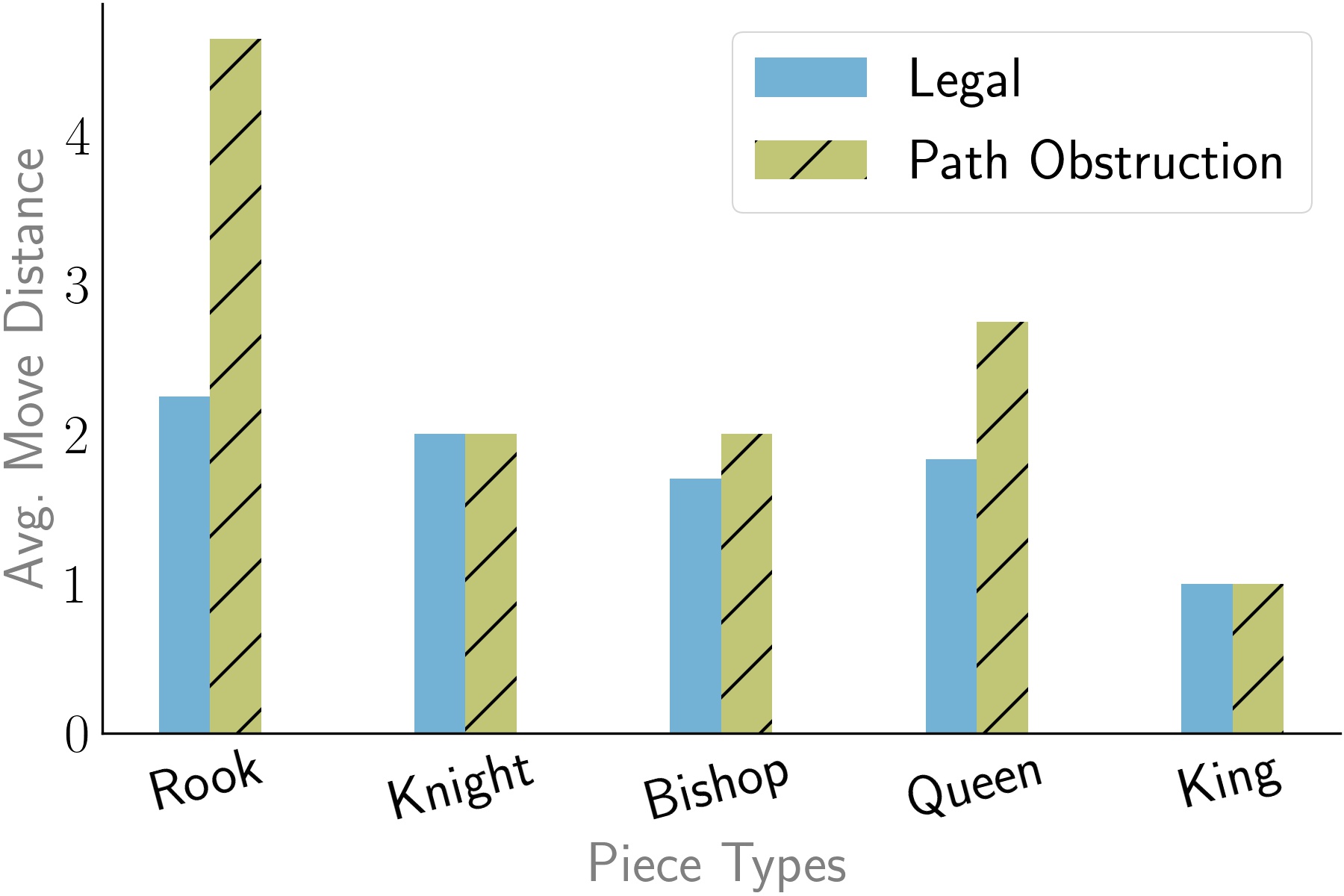}\\
			\caption{End-Other} 
			\label{fig:path_length_other}
		\end{subfigure}
	\hspace*{\fill}
\caption{Comparison of average path length of predicted moves for different piece types when the move is legal vs ones with path obstruction error.}
\label{fig:path_length}
}
\end{figure*}

\subsection{Path Obstruction}
Table~\ref{tab:path_obs} presents the path obstruction errors for different piece types for the End-Actual and End-Other task. Figure~\ref{fig:path_obstruction} represents instances of path obstruction error for different piece types.  
The error counts show that piece types with more dynamic movement except knight i.e. rook, bishop, and queen, have more path obstruction errors  (a knight just needs to avoid landing on its own piece to avoid path obstruction errors). 
These piece types also show a significant increase in frequency of path obstruction errors for End-Other in comparison to End-Actual. As in pseudo legal errors, this could again be due to the out-of-distribution nature of these prompts. Figure~\ref{fig:path_length} shows that the average path length for predicted moves with path obstruction error is significantly higher than the legal predictions for both the kinds of prompts (knight and king have a constant path length). \footnote{Path length is measured in number of king moves i.e. number of moves it will take a king to go from starting to ending square.}

\section{Conclusion}
We proposed the game of chess as a testbed for evaluating how well language models capture the underlying world state. 
We showed that with an appropriate choice of chess notation, a language model can be probed for different aspects of the board state via simple prompts.
The simple and precise dynamics of chess allowed for (a) training models with varying amount of explicit state, and (b) evaluating 
model predictions at a fine-grained level.
Results showed that transformer language models are able to track the board state when given enough
data, but with limited data, %
providing access to board state information during training can yield consistent improvement. 
In the next chapter, we build on the idea of using RAP-like tokens for  injecting world state knowledge and coreference knowledge during pretraining and finetuning of transformers.

\chapter{Integrating Entity Tracking into Language Models}

In this chapter, we build on the findings from our previous chapter on chess language modeling to integrate entity tracking into natural language models. 
Specifically, we finetune pretrained transformer language models such as BART~\cite{lewis-etal-2020-bart} and GPT-2~\cite{radford2019language} on text augmented with entity tracking related information represented as text tokens. 
This chapter is divided into two parts. 
In the first part, we work in a closed domain where a sequence-to-sequence language model, BART, is trained to predict the next action given the initial state and previous actions. 
Previous work by \citet{li-etal-2021-implicit} claims that language models trained with just this self-supervision learn to implicitly capture the world state (a collection of entity states). We propose a new evaluation and find results to the contrary. 
Assuming access to entity states during training time, we propose methods of baking in the state tracking information in language models. 
Results suggest that baking in the state knowledge during training leads to significant improvements in state tracking performance and text generation quality. 
In the second half, we return to the task of coreference resolution with the goal of integrating coreference knowledge into a pretrained autoregressive language model (GPT-2). We finetune models on text augmented with coreference structures represented in text -- the coreference structures are generated using coreference resolution models from Chapter 4. We show  improvements on a popular cloze task~\cite{paperno-etal-2016-lambada} with most of the gains attributable to improved entity tracking.

\section{Baked-in State Probing}
\label{sec:bake_state}
Neural language models have been analyzed for their linguistic and extra-linguistic knowledge via probing.  
Of particular interest has been the following question: how much can a language model trained only on \emph{form} learn about \emph{meaning}? 
Recent work has demonstrated via probing classifiers that in the setting of simple procedural text, where by ``meaning" we mean the underlying world state, language models have a non-trivial performance on world state tracking~\cite{li-etal-2021-implicit, toshniwal-etal-2022-chess}. 
However, our proposed evaluation based on model predictions shows differing results, suggesting that these models are either not capturing the world state or not using it.
\emph{How do these results change if the model has access to the world state?}
We explore this alternate setting with access to the underlying world state only during training and investigate ways of ``baking in'' the state knowledge along with the primary task of language modeling. 
Our proposed approaches allow for state probing during inference simply via text prompts, avoiding any probing classifier machinery.  
In terms of performance, we show that baking in the state knowledge during training leads to significant improvements in state tracking performance and text generation quality.

\subsection{Introduction}
There has been recent interest in the extent to which models trained solely on text (\emph{form}) capture aspects of the underlying world state (\emph{meaning})~\cite{bender-koller-2020-climbing, bisk-etal-2020-experience, wu2021infusing, Bender2021OnTD, li-etal-2021-implicit,toshniwal-etal-2022-chess}. 
\citet{li-etal-2021-implicit} show via the use of probing classifiers that language models trained for simple semantic domains learn to represent the state of the world described by the text without any explicit state supervision. 
However, one of the key limitations of probing classifiers is that the extractability of any information, such as world state, doesn't necessarily mean the model is using this information (See \citet{belinkov2022probing} for a detailed discussion on limitations of probing classifiers). 
Our proposed evaluation based on model predictions shows results contrary to \citet{li-etal-2021-implicit}, wherein we find that the model performance is close to chance in capturing/utilizing the state knowledge (see Table~\ref{tab:results}).    

\emph{How do these results change if the language model has access to the ground truth world state knowledge?} 
We explore this alternate setting where we assume that: (a) the language model has access to the world state \emph{only during training}, and (b) the world state (or parts of it) can be translated to text. 
Access to such oracle annotations during training has also been explored in recent work~\cite{nye2021show, lampinen2022explanation}.
Concretely, we build on the setup of \citet{li-etal-2021-implicit} where 
we explore simple approaches to explicitly adding the state to the language model training sequences.
We show that this \emph{baking-in} of the world state knowledge allows for state probing during inference simply via prompting. 
Our experiments on the Alchemy dataset~\cite{long-etal-2016-simpler} show that our proposed approaches, particularly a multitask learning approach, significantly improve over a baseline language model in terms of both state prediction  and language modeling. 
More broadly speaking, the proposed approaches could present an alternate way for injecting linguistic knowledge during end task training for NLP tasks~\cite{wu2021infusing}.

\begin{table*}[t]
  \centering
  \footnotesize
  \renewcommand{\arraystretch}{1.7}
  \begin{tabular}{ p{0.47\textwidth} p{0.47\textwidth} }
    \begin{minipage}{.47\textwidth}
      \includegraphics[width=\linewidth]{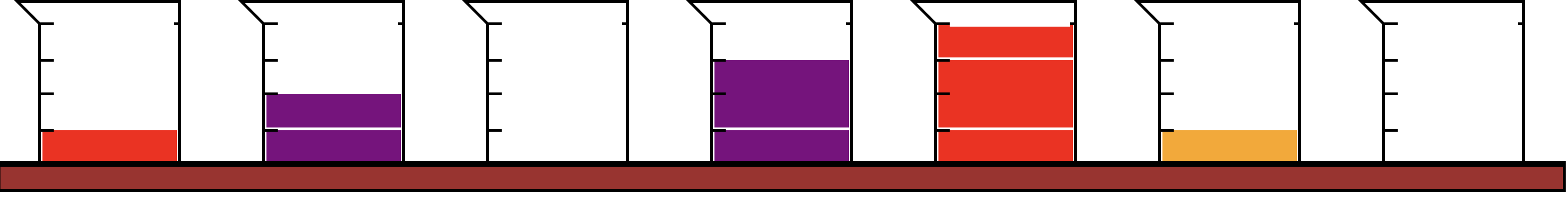}
        \caption*{$W_0$}

    \end{minipage}
    &
    \begin{minipage}{.47\textwidth}
      \includegraphics[width=\linewidth]{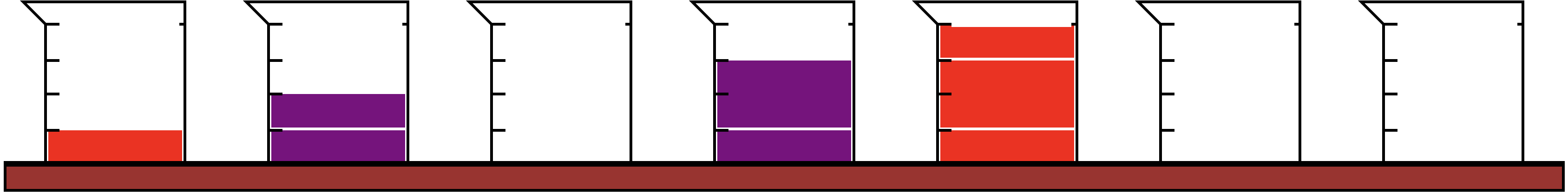}
      \caption*{$W_1$}
    \end{minipage}
    \\ 
    & $s_1  \rightarrow $ \texttt{throw out the orange beaker}  \\
    $\mathcal{T}(W_0) \rightarrow $  \texttt{the first beaker has 1 red, the second beaker has 2 purple, 
    $\cdots$, 
    \textbf{the second to last beaker has 1 orange}, the last beaker is empty} & 
    $\mathcal{T}(W_1) \rightarrow$   \texttt{the first beaker has 1 red, the second beaker has 2 purple,
    $\cdots$, 
    \textbf{the second to last beaker is empty}, the last beaker is empty} \\ 
    $\mathcal{T}(W_0, \textrm{fifth beaker}) \rightarrow$  \texttt{fifth beaker has 4 red} & 
    $\mathcal{T}(W_1, \textrm{fifth beaker}) \rightarrow$  \texttt{fifth beaker has 4 red}
    \\ 
    \end{tabular}
    \caption{Sample Alchemy instance. $W_0$ represents the initial world state while $W_1$ represents the world state at the end of sentence $s_1$. $\mathcal{T}(.)$ represents the state translator which given the world state and optionally an entity, outputs a natural language description of the world/entity state. The Alchemy world consists of beakers (entities) and the state of a beaker is represented by the volume of the colored liquids residing in it. }
    \label{fig:definitions}
\end{table*}
\subsection{Task Setup}
\label{sec:classifier_setup}

\subsubsection{Task Description}
Let $(s)_{i=1}^n$ represent the segmentation of a text discourse into a sequence of sentences.
Suppose the world described by this text consists of entities $(e)_{j=1}^m$.
Let $(W)_{i=0}^n$ represent the sequence of world states where $W_0$ represents the initial world state and $W_i$ represents the world state at the end of sentences $s_{1:i}$.
 We assume that the world state is simply the union of the state of all the entities in the world.
Formally speaking, let $\mathcal{S}(e_j, s_{1:i})$ represent the state of the entity $e_j$ at the end of sentences $s_{1:i}$. Then $W_i = [\mathcal{S}(e_1, s_{1:i}) \cdot \mathcal{S}(e_2, s_{1:i}) \cdots \mathcal{S}(e_m, s_{1:i})]$. Finally, we assume access to a state translator $\mathcal{T}$ which takes as input the entity or world state and outputs a corresponding natural language description.
Figure~\ref{fig:definitions} illustrates these concepts in the context of Alchemy. 
We next describe the language modeling and state probing task avoiding any model-specific details.  

\paragraph{Language Modeling} 
We finetune the sequence-to-sequence (seq2seq) language model  BART~\cite{lewis-etal-2020-bart} in all our experiments. 
The input to the seq2seq language model is $\mathcal{T}(W_0) \cdot s_{1:i}$ i.e.\ the concatenation of the initial state translated to text and the initial $i$ sentences of the discourse.
The language model is (primarily) trained to predict the next sentence but the exact output sequence differs across models (see Table~\ref{tab:output_seq}). 
After training the language model, we probe the model for its state tracking capabilities, which we define next. 

\paragraph{State Probing} tasks require predicting the world state $W_i$ given the initial state $W_0$ and the first $i$ sentences of the discourse $s_{1:i}$. Since the world state is simply a union of entity states, the world state probing task can be decomposed into predicting the states of all the entities involved in the discourse.   

Given these generic descriptions of the language modeling and state probing tasks, we next describe the model variants.
The input sequence to the seq2seq language model remains the same across all the model variants. Hence, we limit  the model details to: (a) the output sequence with which the language model is trained, and (b) how state probing is done after language model training.

\subsection{Baseline Model}
For our baseline, we use the setup of \citet{li-etal-2021-implicit}. 
The baseline model is the language model finetuned to predict the next sentence of the discourse $s_{i+1}$.

\paragraph{State Probing} 
\citet{li-etal-2021-implicit} predict the state of each entity individually with a linear probing classifier. 
Predicting the state of each entity requires extracting its representation from the discourse representation, for which \citet{li-etal-2021-implicit} employ 
heuristics.

\begin{table*}
    \centering
    \small{
          \renewcommand{\arraystretch}{1.1}
    \begin{tabular}{
    l|ll|c|c
    }
    \toprule
        \multirow{2}{*}{\textbf{Model}}  &  \multicolumn{2}{c}{\textbf{Training}} & \textbf{Inference} & \textbf{Probing via Prompt} \\
                 &  \textbf{Prob.} & \textbf{Output Seq.} & \textbf{Output Seq.} & \\\midrule 
        Baseline &  1 & $s_{i+1}$ & $s_{i+1}$ & \xmark\\
        \midrule
        \multirow{2}{*}{Multitask Learning} & $p$  & \state $\mathcal{T}(W_i)$ \state &   \multirow{2}{*}{$s_{i+1}$} & \multirow{2}{*}{\cmark}\\
        & $(1 - p)$  & $s_{i+1}$  &   \\
        \midrule
        \multirow{2}{*}{RAS} &  $p$ & 
        \state  $\mathcal{T}(W_i)$ \state  $s_{i+1}$ 
           &  \multirow{2}{*}{$s_{i+1}$} & \multirow{2}{*}{\cmark}\\
          & $(1 - p)$  & $s_{i+1}$ &   \\
        \bottomrule

    \end{tabular}
    }
    \caption{Summary of Output Sequences during Training and Inference for all the language model variants. $W_i$ denotes the world state at the end of the first $i$ sentences $s_{1:i}$ of the discourse. $\mathcal{T}(W_i)$ denotes the translation of world state to natural language text.  
    \state\ denotes the special token used to mark the state description boundary. }
    \label{tab:output_seq}
    \vspace{-0.1in}
\end{table*}

\subsection{Baking in State Knowledge}
\label{sec:baked_probing}
In this section, we assume access to the ground truth world state $W_i$, or equivalently $\mathcal{T}(W_i)$, during language model training.
We propose two approaches to explicitly adding the state sequence $\mathcal{T}(W_i)$ to language model training sequences. 
We first describe the language model training details for the proposed models, and then describe the shared \emph{baked-in probing} methodology.

\subsubsection{Multitask Learning}
\label{sec:multitask}
In multitask learning, we train the language model to predict the next sentence $s_{i+1}$ with probability $(1 - p)$ or the current world state \state $\mathcal{T}(W_i)$ \state with probability $p$, where the probability $p$ is a hyperparameter, and the delimiter  token \state denotes that the text sequence is about world state. In practice, we use the probabilities of the tasks to scale the prediction losses i.e.\ 
$$\mathcal{L}_{{mult}} = p\mathcal{L}_{{state}}  + (1 - p) \mathcal{L}_{next} $$
where $\mathcal{L}_{{state}}$ is the loss for predicting the state sequence, and $\mathcal{L}_{next}$ is the loss for predicting the next sentence. 

\subsubsection{Randomly Added State (RAS)}
\label{sec:ras}
In the multitask learning approach, the current state prediction and next sentence prediction are treated as separate prediction tasks. In RAS, we combine the two tasks stochastically. Specifically, for some chosen probability $p$, the RAS($p$) language model is trained to predict the concatenated sequence \state  $\mathcal{T}(W_i)$ \state  $s_{i+1}$ with probability $p$, and with probability $(1 - p)$ it is trained on the canonical task of predicting the next sentence $s_{i+1}$. 
The RAS model allows for the flexibility of the Multitask Learning model in the sense that both the state sequence (see discussion on State Probing later in this Section) and next sentence can be independently predicted. At the same time, during training it allows the model to learn the relationship between the two tasks.

Note that the RAS ($p=1$) model is always trained to predict the next sentence with the state sequence. Hence, during inference we first predict the state sequence, and then predict the next sentence.

\subsubsection{State Probing} In the above proposed models, either the language model is explicitly trained to predict the world state or the world state is a prefix of the predicted sequence. Moreover, the state sequence $\mathcal{T}(W_i)$ is delimited by the special token \state. Thus, prompting the decoder of the trained seq2seq language model  with the \state\ token conditions the model to generate the state.\footnote{This bears similarity with the use of control tokens in text generation~\cite{keskar2019ctrl, see-etal-2019-makes}.} 

For Alchemy, we can enumerate all the possible states for any entity (beaker) and score them with the language model, the predicted entity state being the one with the highest probability.\footnote{Any beaker in our Alchemy setup has one of 210 states.}    
We predict the world state in Alchemy by individually predicting the state for each entity (beaker). \footnote{During training, we 
shuffle the order in which the beaker states are presented in $\mathcal{T}(W_i)$. We do this to avoid bias towards any particular order in which the language model is used to predicting the beaker states.}

\subsubsection{Next Sentence Probability}
For the baseline model, the probability of the next sentence $P(s_{i+1})$ is trivially calculated using the autoregressive decoder of the seq2seq language model (for ease of notation we hide the conditioning on $W_0$ and $s_{1:i}$). For the proposed model variants, we additionally mask out the probability assigned to the \state\ token at all timesteps i.e.\ renormalize the distribution after zeroing out the probability assigned to the \state\ token
as the special token is only used while predicting the state. 

For the RAS ($p=1$) model, calculating the exact $P(s_{i+1})$ is non-trivial because the model conditions the next sentence prediction on the state prediction. 
To explain this challenge, suppose the world state space is $\mathcal{Z}$. 
Under the RAS($p=1$) model, we would be required marginalize over the entire world state space i.e.\ 
$$P(s_{i+1}) = \sum_{z \in \mathcal{Z}} P(s_{i+1} | z) P(z)$$
Since this marginalization is impractical for Alchemy, we instead report:
$$P(s_{i+1}) = \max_{z \in \mathcal{Z}} P(s_{i+1} | z)$$
Thus, the perplexity results for the RAS($p=1$) model are not directly comparable to other variants.

\subsection{Experimental Details}
\subsubsection{Hyperparameter Details}
For all our experiments, we use the BART-base language model, as used by~\citet{li-etal-2021-implicit}.  
The model is trained for a maximum of 100 epochs with a batch size of 24.
 Validation set perplexity is computed at the end of every epoch and training stops when there is no improvement for 5 consecutive epochs. 
We use the Adam optimizer where the learning rate is warmed up linearly for the first 10 epochs to $10^{-5}$ followed by a linear decay. For the Multitask and RAS models, we tune the probability $p$ of the auxiliary task over $\{0.1, 0.2, 0.3, \cdots, 0.8, 0.9, 1.0\}$.

\subsubsection{Data and Evaluation Details} We borrow the Alchemy setup from \citet{li-etal-2021-implicit}. For the validation split, we report: (a) \emph{Perplexity}, and (b) \emph{World/Entity State Accuracy}: The world state accuracy is the proportion of instances for which the model predicts the correct state for all the entities, while the entity state accuracy is the proportion of predicting the correct entity states. 

\paragraph{Valid Next Sentence Evaluation}
Additionally, we create an artificial evaluation set of 100 Alchemy language modeling instances where for each input we also have the exact set of valid outputs, i.e.\ the next sentences which can be ``executed" given the world state. For each input, the model selects the most probable next sentence among all the candidate next sentences, where the candidate next sentences consist of both valid and invalid continuations. 
We then measure the \emph{Valid Next Sentence Accuracy}, which is the proportion of instances for which a model selects a valid next sentence as the top choice.  
A model doing poorly on this evaluation is either failing at state tracking or unable to use its state knowledge while predicting the next sentence. 
For context, a naive baseline of selecting the next sentence among the 100 candidate next sentences using a uniform distribution gets an accuracy of $45.56 \pm 4.65$.

\begin{table*}
    \centering
      \renewcommand{\arraystretch}{1.1}

    \begin{tabular}{p{0.25 \linewidth}  c c c c 
    }
    \toprule
        \textbf{Model}  &   \textbf{Perplexity} & \textbf{World State} & \textbf{Entity State} & \textbf{Valid Next Sentence}  
        \\\midrule
        Baseline \citep{li-etal-2021-implicit}  & 2.98          & \phantom{1}7.6  &   75.0 &  48
        \\\midrule
        Mulitask Learning   & \textbf{2.91} & \textbf{57.8}  & \textbf{92.2}  & 70  
        \\
        RAS                 & \textbf{2.91}         & 49.3  & 90.1  & \textbf{74}
        \\\bottomrule

    \end{tabular}
    \caption{Comparison of performance of Language Model variants on the proposed evaluations. The state tracking results for the Baseline model are from \citet{li-etal-2021-implicit}.}
    \label{tab:results}
\end{table*}

\subsection{Results}
\vspace{-0.05in}
Table~\ref{tab:results} presents the results for the baseline model and our proposed language model variants. 
Across all evaluations we see a clear benefit of having access to state knowledge during training. 
In particular, the Multitask Learning model improves over the baseline model on the world state prediction task by about 50 points absolute (and also avoids training a separate probing classifier). For the language modeling task, we see a drop in perplexity for both the proposed models in comparison to the baseline model suggesting that the state knowledge also aids the language modeling task. 
This is even more clearly reflected in the Valid Next Sentence evaluation where the proposed variants improve over the baseline language model by about 40\% relative. Interestingly, the baseline language model gets an accuracy of 48 on this task, which is within one standard deviation of the random baseline's score. 
This suggests that 
even if the baseline language model trained on just the text discourse (form) has implicitly learned the state (meaning), it most likely has not learned to use the state knowledge while predicting the next sentence. 
Among our proposed variants the Multitask Learning variant excels in predicting the state while the RAS model is the best at utilizing the state knowledge in next sentence prediction.

\subsection{Conclusion}
We show via our proposed evaluation of Valid Next Sentence prediction that a baseline language model is on par with chance performance, suggesting that the model struggles to capture the state knowledge. 
Based on this evidence, we explore language model variants which assume access to the world state during training.   
The proposed language model variants can be easily probed for world state via a text prompt, and substantially improve over the baseline language model for both state tracking and text generation.
A key limitation of the setup used in the above experiments is the  assumption of access to perfect ground truth state knowledge. 
In the next section, we explore a setting where we rely on the (imperfect) output of a linguistic analyzer.

\section{Baking in Coreference Knowledge into Language Models}
\label{sec:coref_lm}
In Chapter 5 and Section~\ref{sec:bake_state} we experiment with integrating entity tracking into transformer LMs by training them on state-augmented training sequences. During training we assumed access to the perfect entity state information. But in general settings we lack the ground truth entity state knowledge. 
We explore this general setting in this section.  
Specifically, we experiment with \emph{baking-in} coreference knowledge into pretrained LMs, using models from Chapter 4 to predict the coreference structure. 
We then train the GPT-2~\cite{radford2019language} model on coreference-augmented training sequences. 
The language model is tested on the LAMBADA cloze-prediction task~\cite{paperno-etal-2016-lambada}, which has been shown by prior work to require coreference resolution~\cite{chu-etal-2017-broad}. 

\subsection{Models}
For all our experiments we use the GPT-2 medium language model, which is a autoregressive transformer-based language model proposed by~\citet{radford2019language}. The baseline model is the GPT-2 model finetuned for the target domain (filtered instances from BookCorpus~\cite{Zhu2015AligningBA}).  We use the coreference models from Chapter 4 to predict coreference structures for the training sequences. From anecdotal coreference evaluation, we find the OntoNotes-only model to be a better fit for the LAMBADA/BookCorpus. We explore different formats to linearize the coreference structure. 
The full set of hyperparameters explored is shown in Table~\ref{tab:lambada_hyperparam}. 
For example, we experiment with whether to use the canonical coreference mention (first mention) to denote the chain or the antecedent mention (most recent), and whether to truncate the coreference mention or not (truncation is used to cap the length of the mention used for denoting the coreference chain). 
Sample training instances with the chosen format (last column of Table~\ref{tab:lambada_hyperparam}) are shown in Table~\ref{tab:coref_aug_sequences}.

\begin{table}[t]
    \centering
    \begin{tabular}{l l l }
    \toprule
    \textbf{Hyperparameter} & \textbf{Choices} & \textbf{Best} \\\midrule 
    Singleton & \cmark \xmark & \xmark \\
    Coreference mention & Antecedent, Canonical  & Antecedent \\
    Coreference mention length truncation  & 2, 5, No truncation & 5 \\
    Coreference mention probability & 0.15, 0.25 & 0.25 \\
    \bottomrule
    \end{tabular}
    \caption{Hyperparameter choices explored for training the GPT-2 + Coref model. Canonical refers to the first mention in the coreference chain. Best hyperparameters correspond to lowest validation set perplexity. }
    \label{tab:lambada_hyperparam}
\end{table}

\begin{table}[t]
\small
    \centering
    \begin{tabular}{p{0.95\textwidth} }
    \toprule
    i grab a towel from the linen closet and head off for a shower . chapter 3 the water is cold and dark and suffocating . i lift my face to the surface , watching the light disappear as i sink . i [ i  ] 'm holding my [ i ] breath , my [ i  ] arms flailing out as i try to swim , but it 's like i 'm weighed down with stones . fear crawls up my throat , and i [ i  ] have my first convulsion as i try not to breathe in water \\\midrule
they keep coming until they surround us ; dozens of clactures of all sizes . one of them [ they  ] swipes his bat wing , knocking a group of ten men aside . i have the tavin out and am just twisting the knob for the blade when pentaim flies through the air , a clacture carrying her . fear and adrenaline mix in my gut as i light the tavin . it attaches to my hand as i [ i  ] run after pentaim \\
    \bottomrule
    \end{tabular}
    \caption{Coreference-augmented training sequences.}
    \label{tab:coref_aug_sequences}
\end{table}

\subsection{Experimental Setup}
\begin{table}[t]
\footnotesize
    \centering
    \begin{tabular}{p{0.6\textwidth} c c c }
    \toprule
    \textbf{Context} & \textbf{Ground truth} & \textbf{GPT-2} & \textbf{GPT-2 + Coref} \\
    \midrule
    we need to get fawn and stella over here. stella? you there?' 'where else would i be?' 'right. get suited up to come over here. we need you two to undo the lines so we can put it in orbit. you have both space walked?'  'i'm strictly a girl who keeps both feet on the ground,' said              &  stella & fawn & \textcolor{dkgreen}{stella} \\\midrule
    gemma had never been introduced to any of them before, but she knew of them. she had heard her parents and hawke discuss mr. percival at great length. he was next in line for the earldom of worcester, and one of the royal duke's favorite cousins. no doubt they had designs on him as a match for & gemma & mr. & \textcolor{dkgreen}{gemma} \\\midrule
     don't think so. who's the owner?  " massachusetts man. banker, i think. has a young lady, jenny, live in the little house and care for them. " " some pretty expensive pasturage. " " some pretty expensive horses. " " yeah? i don't know too much about & horses & \textcolor{dkgreen}{horses} & jenny \\ \bottomrule
    \end{tabular}
    \caption{Sample LAMBADA instances with predictions from GPT-2 models trained with and without coreference-augmented sequences. }
    \label{tab:lambada_pred}
\end{table}

\paragraph{Data}
We use the training data created by \citet{chu-etal-2017-broad} from the BookCorpus~\cite{Zhu2015AligningBA}. For evaluation, we use the LAMBADA corpus~\cite{paperno-etal-2016-lambada}. The LAMBADA dataset consists of passages with length of around 4-5 sentences, where the task is to predict the last word of the passage given the prior context. The passages are filtered by annotators such that predicting the last word is not possible given just the last sentence, but is possible given the full passage (see Table~\ref{tab:lambada_pred} for sample LAMBADA instances).

\subsection{Results}
\begin{table}[t]
    \centering
    \begin{tabular}{c c c }
    \toprule
    \textbf{Model} & \textbf{Accuracy} & \textbf{Perplexity}  \\ \midrule
    GPT-2 & 59.1 & \phantom{1}4.59 \\
    GPT-2 + Coref & 61.3 & \phantom{1}4.58 \\ 
    \midrule
    GPT-2 (Zero-shot)~\cite{radford2019language} & 55.5 & 15.60 \\\bottomrule
    
    \end{tabular}
    \caption{Results on the LAMBADA test set for the GPT-2 medium model. Note that results reported by \citet{radford2019language} are with a different format where casing is preserved.}
    \label{tab:lambada_res}
\end{table}

\begin{figure}[t]
    \centering
    \includegraphics[width=0.6\textwidth]{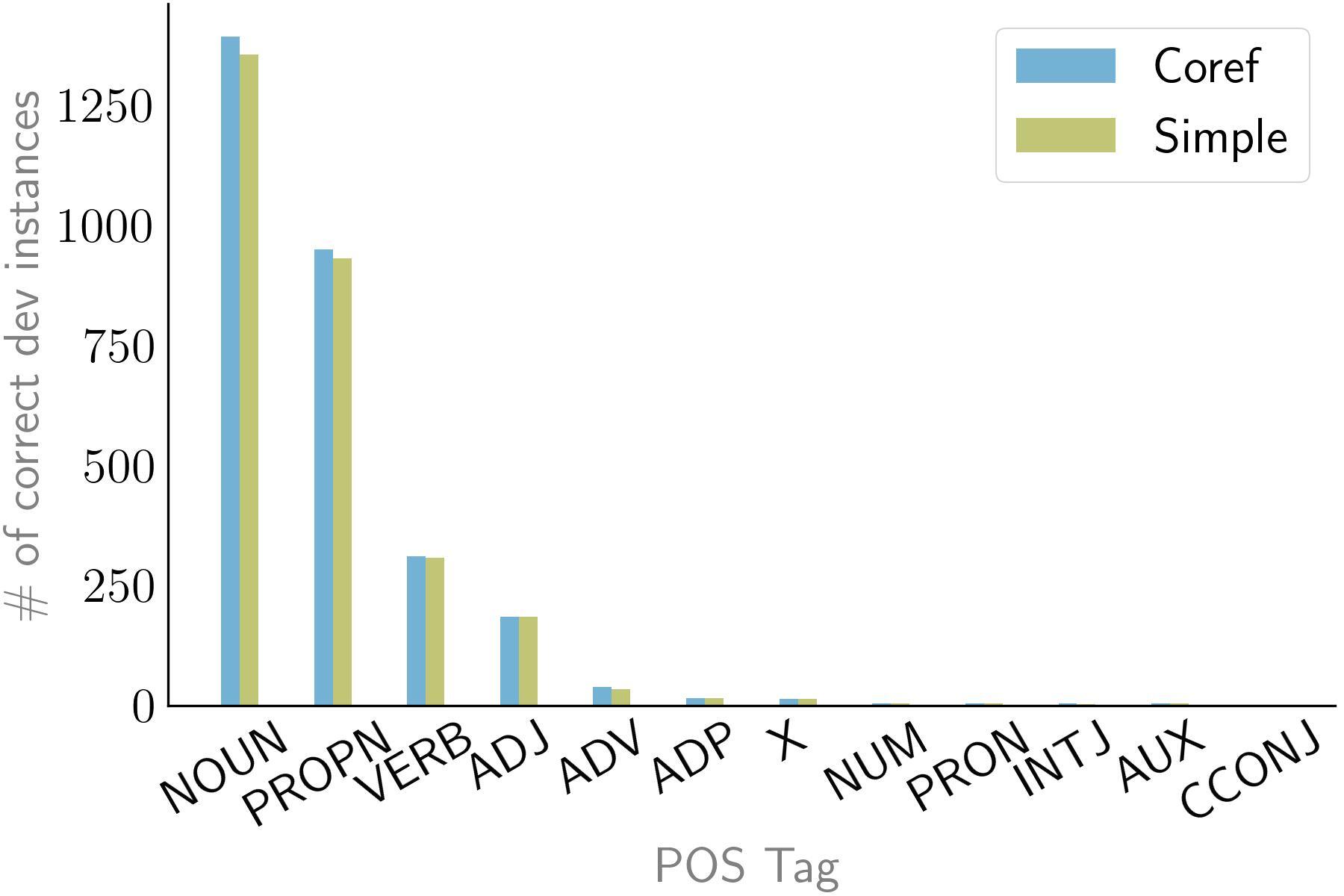}
    \caption{Comparison of models on the LAMBADA validation set categorized by part-of-speech tags.}
    \label{fig:pos_comp}
\end{figure}

Table~\ref{tab:lambada_res} compares the GPT-2 language model results for  different settings. We see accuracy gains with training on coreference-augmented sequences, which suggests that this improves the model's entity tracking capability. 
Figure~\ref{fig:pos_comp} shows that training with coreference-augmented sequence results in a clear performance gain for prediction of nouns and proper nouns, which further illustrates evidence of improvement in entity tracking capability. 
Finally, we highlight a few LAMBADA validation instances with noun and proper noun predictions in Table~\ref{tab:lambada_pred}.

\subsection{Conclusion}
We show that by training the GPT-2 language model on coreference-augmented text sequences, the performance on the LAMBADA cloze-prediction task improves. We believe this recipe of integrating linguistic structures can be extended to other linguistic tasks as well. On a side note, the successful use of coreference structures predicted by the coreference models from Chapter 4 also points to the generalizability of the coreference models, since the LAMBADA/BookCorpus has a clear domain shift from OntoNotes.

\chapter{Conclusion}

\section{Thesis Summary}
The thesis begins by motivating the importance of entity tracking in developing NLP models. 
Chapter 1 motivates the two entity tracking problems that form this thesis's basis: (i) efficient entity tracking models for long documents and (ii) integrating entity tracking into language models.

Chapters 3 and 4 focus on developing efficient models for the task of coreference resolution, a particular class of entity tracking problem. 
We propose memory models for the coreference resolution task where the external memory tracks the entity representation. 
Specifically, we use a fixed-dimensional entity representation derived from pretrained LMs. This presents a significant departure from the popular paradigm of mention-ranking models in coreference resolution, which requires keeping around the representation of individual mentions of the entities. The use of a compact entity representation aids the scalability of the proposed models. 

In Chapter 3 we present PeTra, a bounded memory model trained with sparse supervision for the pronoun resolution task. We also propose interpretability measures, including human evaluation, to capture the entity tracking capability beyond the pronoun resolution evaluation. 
Results show that the proposed model outperformed prior work on the pronoun resolution task and interpretability measures with fewer parameters and a simpler architecture. 

Chapter 4 extends the proposed model from Chapter 3 to the full coreference resolution task, focusing on two key issues relating to the applicability of coreference resolution models, namely scalability to long documents and generalizability. Section 4.1 explores memory models for the coreference resolution task, which can scale to long documents. In particular, we propose a bounded memory model with a learned memory management scheme that guarantees a linear runtime in document length in theory and, in practice, significantly reduced peak memory usage during training and inference time compared to its unbounded memory counterpart. The proposed memory models are competitive with prior work, with the unbounded variant establishing a new state-of-the-art performance for LitBank, a recently proposed long document coreference resolution task. 

In Section 4.2, we test the generalization performance of coreference resolution models described in Section 4.1 via zero-shot evaluation i.e., performance on datasets unseen during training. To this end, we consolidate a set of eight coreference resolution datasets targeting different domains. 
We find that in a zero-shot setting, models trained on a single dataset transfer poorly while joint training moderately improves the out-of-domain performance.  
We propose joint training and data augmentation strategies which moderately aid the generalization performance. 
Overall, we find that generalization remains a big challenge for coreference resolution. 
We also establish a new state-of-the-art for two coreference resolution benchmarks, PreCo and WikiCoref.

In the second half of this thesis, we shift our attention to the task of integrating entity tracking into LMs. But as a first step, we need to ascertain the entity tracking capabilities of transformer LMs. 
To this end, we propose the task of language modeling for the game of chess in Chapter 5.  
We show that with an appropriate choice of chess notation, a language model can be probed for different aspects of the board state via simple prompts.
The simple and precise dynamics of chess allowed for: (a) training models with varying amounts of board state information and (b) evaluating model predictions at a fine-grained level.
Results show that transformer language models can track the board state when given enough
data, but with limited data, providing access to board state information during training yields consistent improvement. 

In Chapter 6, we extend our findings from the chess tested in the previous chapter to natural language. 
We propose methods to integrate entity tracking capability into LMs by training them on entity state augmented training sequences.  
We first experiment in a closed domain, assuming access to the true entity states during training. 
We show that integrating entity tracking into LMs improves state tracking performance and text generation quality in this closed domain. 
We extend these ideas to integrate coreference resolution into LMs. 
Specifically, we experiment with integrating the coreference resolution predictions made by models developed in Chapter 4 into pretrained LMs. 
Our results show that integrating coreference into LMs improves results on the LAMBADA cloze prediction task.

Reflecting on the two problems which are central to this thesis: (a) efficient entity tracking models for long documents and (b) integrating entity tracking into language models, we believe the models and approaches proposed in this thesis have made progress on both fronts. Our proposed models for coreference resolution can efficiently scale to long, book-length documents. However, the lack of annotated book-length coreference datasets means that the models haven't been tested. Preliminary analysis of model predictions for books suggests that the models tend to split entity clusters when there's a relatively long gap between consecutive entity mentions. We believe testing NLP models for intermediate annotations, such as  coreference, would aid the development of long-text NLP models. 

Regarding integrating entity tracking into language models, our proposed recipe of training language models with state-augmented text sequences preserves the transformer language model architecture. Also, it endows the ability to probe the model for entity tracking simply via prompting. While this thesis applies the recipe only for entity tracking in limited settings, we believe practitioners can use this approach in a broad array of linguistic tasks and domains, some of which we describe in the next section describing potential future work directions.

\section{Future Work}

In this section, we discuss potential future work directions that can build on the findings of this thesis.

\paragraph{Entity Tracking for Downstream Tasks.} 
In recent work, \citet{ye-etal-2020-coreferential} have demonstrated the success of incorporating a ``coreferential objective"  in masked language model training.
However, their ``coreferential objective" uses just the simple distant supervision of exact matching mention spans, and it's not clear if the model learns the general coreference function. 
In section~\ref{sec:coref_lm} we incorporate predictions of an actual coreference resolution model into the GPT-2 language model and show improvements on the LAMBADA cloze task. 
We believe that given a high-quality coreference model, incorporating its prediction in pretrained LMs should be better than relying on a limited distant supervision signal. 
Future work should test this hypothesis, and 
 can extend our proposed entity tracking integration idea with other language model architectures, such as encoder-only and encoder-decoder pretrained language models, and test it on a variety of downstream tasks such as question answering, summarization.

\paragraph{Discourse Modeling Beyond Entity Tracking.} 
Diagnostic evaluations based on cloze tasks have shown a lack of event understanding in pretrained LMs~\cite{ettinger-2020-bert, lialin-etal-2022-life}. 
\citet{mou-etal-2021-narrative} show that question answering models struggle with event-centric questions in a  Book QA task. 
How to bake in event-related knowledge, such as temporal order of events, causality of events, in NLP models remains an open question. 
As a first attempt, we believe that similarly to our work on integrating coreference knowledge into LMs, future work can look into infusing the knowledge of models trained on supervised event-based tasks into LMs.

\paragraph{Long-context NLP models.} 
One of the biggest challenges facing NLP is scaling language technologies to long contexts, such as book-length texts. We propose future directions for the following two ways of scaling up NLP models to long contexts. 

\emph{Memory-based Long-Context Models}: Similarly to our bounded memory coreference models in Chapters 3 and 4, there has been work on language models which learn to compress the past context~\cite{Rae2020Compressive, Sukhbaatar2021NotAM}. The memory in these models is learned via an  autoregressive LM objective and lacks interpretation. Future work can look into adding interpretability to the memory by making it entity-centric, something which has been explored in prior work for recurrent neural network (RNN) based LMs~\cite{ji-etal-2017-dynamic, clark-etal-2018-neural}. 

\emph{Long-context Pretraining Objectives}:
There's a plethora of recent work on making the transformer architecture efficient, and thus avoiding keeping around an explicit external memory as the entire text sequence or a large portion of it can fit into a single transformer window~\citep[\emph{inter alia}]{roy-etal-2021-efficient, jaegle2022perceiver, Rae2020Compressive, Hutchins2022Block}.  
However, prior work analyzing these language models suggests that these LMs don't use the full context~\cite{sun-etal-2021-long}. This suggests that current pretraining objectives don't encourage long-context understanding such as entity tracking. Future work can use the long document entity tracking models developed in this thesis to create pretraining tasks which promote entity tracking over long contexts.

\newpage

\bibliographystyle{unsrtnat}
\bibliography{main}

\begin{thebibliography}{215}
\providecommand{\natexlab}[1]{#1}
\providecommand{\url}[1]{\texttt{#1}}
\expandafter\ifx\csname urlstyle\endcsname\relax
  \providecommand{\doi}[1]{doi: #1}\else
  \providecommand{\doi}{doi: \begingroup \urlstyle{rm}\Url}\fi

\bibitem[Brown et~al.(2020)Brown, Mann, Ryder, Subbiah, Kaplan, Dhariwal,
  Neelakantan, Shyam, Sastry, Askell, Agarwal, Herbert-Voss, Krueger, Henighan,
  Child, Ramesh, Ziegler, Wu, Winter, Hesse, Chen, Sigler, Litwin, Gray, Chess,
  Clark, Berner, McCandlish, Radford, Sutskever, and Amodei]{brown2020language}
Tom~B. Brown, Benjamin Mann, Nick Ryder, Melanie Subbiah, Jared Kaplan,
  Prafulla Dhariwal, Arvind Neelakantan, Pranav Shyam, Girish Sastry, Amanda
  Askell, Sandhini Agarwal, Ariel Herbert-Voss, Gretchen Krueger, Tom Henighan,
  Rewon Child, Aditya Ramesh, Daniel~M. Ziegler, Jeffrey Wu, Clemens Winter,
  Christopher Hesse, Mark Chen, Eric Sigler, Mateusz Litwin, Scott Gray,
  Benjamin Chess, Jack Clark, Christopher Berner, Sam McCandlish, Alec Radford,
  Ilya Sutskever, and Dario Amodei.
\newblock {Language Models are Few-Shot Learners}, 2020.

\bibitem[Graves et~al.(2014)Graves, Wayne, and Danihelka]{graves2014neural}
Alex Graves, Greg Wayne, and Ivo Danihelka.
\newblock Neural {T}uring machines.
\newblock \emph{arXiv preprint arXiv:1410.5401}, 2014.

\bibitem[Weischedel et~al.(2013)Weischedel, Palmer, Marcus, Hovy, Pradhan,
  Ramshaw, Xue, Taylor, Kaufman, Franchini, et~al.]{weischedel2013ontonotes}
Ralph Weischedel, Martha Palmer, Mitchell Marcus, Eduard Hovy, Sameer Pradhan,
  Lance Ramshaw, Nianwen Xue, Ann Taylor, Jeff Kaufman, Michelle Franchini,
  et~al.
\newblock {OntoNotes} release 5.0.
\newblock \emph{Linguistic Data Consortium, Philadelphia, PA}, 2013.
\newblock \doi{10.35111/xmhb-2b84}.

\bibitem[Li et~al.(2021)Li, Nye, and Andreas]{li-etal-2021-implicit}
Belinda~Z. Li, Maxwell Nye, and Jacob Andreas.
\newblock Implicit representations of meaning in neural language models.
\newblock In \emph{Proceedings of the 59th Annual Meeting of the Association
  for Computational Linguistics and the 11th International Joint Conference on
  Natural Language Processing (Volume 1: Long Papers)}, pages 1813--1827,
  Online, August 2021. Association for Computational Linguistics.
\newblock \doi{10.18653/v1/2021.acl-long.143}.
\newblock URL \url{https://aclanthology.org/2021.acl-long.143}.

\bibitem[Radford et~al.(2019)Radford, Wu, Child, Luan, Amodei, and
  Sutskever]{radford2019language}
Alec Radford, Jeff Wu, Rewon Child, David Luan, Dario Amodei, and Ilya
  Sutskever.
\newblock {Language Models are Unsupervised Multitask Learners}.
\newblock \emph{CoRR}, 2019.

\bibitem[Hoang et~al.(2018)Hoang, Wiseman, and Rush]{hoang-etal-2018-entity}
Luong Hoang, Sam Wiseman, and Alexander Rush.
\newblock Entity {T}racking {I}mproves {C}loze-style {R}eading {C}omprehension.
\newblock In \emph{{EMNLP}}, 2018.

\bibitem[Clark et~al.(2018)Clark, Ji, and Smith]{clark-etal-2018-neural}
Elizabeth Clark, Yangfeng Ji, and Noah~A. Smith.
\newblock {Neural Text Generation in Stories Using Entity Representations as
  Context}.
\newblock In \emph{NAACL-HLT}, 2018.

\bibitem[Hobbs(1978)]{Hobbs1978ResolvingPR}
Jerry~R. Hobbs.
\newblock Resolving pronoun references.
\newblock \emph{Lingua}, 44, 1978.

\bibitem[Brennan et~al.(1987)Brennan, Walker, and Pollard]{brennan87centering}
Susan Brennan, Marilyn Walker, and Carl Pollard.
\newblock {A Centering Approach to Pronouns}.
\newblock In \emph{ACL}, 1987.

\bibitem[Carter(1987)]{Carter1987InterpretingAI}
David Carter.
\newblock \emph{Interpreting anaphors in natural language texts}.
\newblock Ellis Horwood, 1987.

\bibitem[Webster et~al.(2018)Webster, Recasens, Axelrod, and
  Baldridge]{webster2018gap}
Kellie Webster, Marta Recasens, Vera Axelrod, and Jason Baldridge.
\newblock Mind the {GAP}: {A} {B}alanced {C}orpus of {G}endered {A}mbiguous
  {P}ronouns.
\newblock In \emph{{TACL}}, volume~6, 2018.

\bibitem[Bamman et~al.(2020)Bamman, Lewke, and Mansoor]{bamman2019annotated}
David Bamman, Olivia Lewke, and Anya Mansoor.
\newblock An {A}nnotated {D}ataset of {C}oreference in {E}nglish {L}iterature.
\newblock In \emph{LREC}, 2020.

\bibitem[Wiseman et~al.(2015)Wiseman, Rush, Shieber, and
  Weston]{wiseman-etal-2015-learning}
Sam Wiseman, Alexander~M. Rush, Stuart Shieber, and Jason Weston.
\newblock Learning {A}naphoricity and {A}ntecedent {R}anking {F}eatures for
  {C}oreference {R}esolution.
\newblock In \emph{{ACL-IJCNLP}}, 2015.

\bibitem[Clark and Manning(2016{\natexlab{a}})]{clark-manning-2016-deep}
Kevin Clark and Christopher~D. Manning.
\newblock Deep {R}einforcement {L}earning for {M}ention-{R}anking {C}oreference
  {M}odels.
\newblock In \emph{{EMNLP}}, 2016{\natexlab{a}}.

\bibitem[Lee et~al.(2017)Lee, He, Lewis, and Zettlemoyer]{lee-etal-2017-end}
Kenton Lee, Luheng He, Mike Lewis, and Luke Zettlemoyer.
\newblock End-to-end {N}eural {C}oreference {R}esolution.
\newblock In \emph{{EMNLP}}, 2017.

\bibitem[Lee et~al.(2018)Lee, He, and Zettlemoyer]{lee-etal-2018-higher}
Kenton Lee, Luheng He, and Luke Zettlemoyer.
\newblock Higher-{O}rder {C}oreference {R}esolution with {C}oarse-to-{F}ine
  {I}nference.
\newblock In \emph{{NAACL-HLT}}, 2018.

\bibitem[Joshi et~al.(2019)Joshi, Levy, Zettlemoyer, and
  Weld]{joshi-etal-2019-bert}
Mandar Joshi, Omer Levy, Luke Zettlemoyer, and Daniel Weld.
\newblock {BERT} for {C}oreference {R}esolution: {B}aselines and {A}nalysis.
\newblock In \emph{{EMNLP}}, 2019.

\bibitem[Joshi et~al.(2020)Joshi, Chen, Liu, Weld, Zettlemoyer, and
  Levy]{joshi-etal-2020-spanbert}
Mandar Joshi, Danqi Chen, Yinhan Liu, Daniel~S. Weld, Luke Zettlemoyer, and
  Omer Levy.
\newblock {SpanBERT: Improving Pre-training by Representing and Predicting
  Spans}.
\newblock \emph{{TACL}}, 8, 2020.

\bibitem[Wu et~al.(2020{\natexlab{a}})Wu, Wang, Yuan, Wu, and
  Li]{wu2019coreference}
Wei Wu, Fei Wang, Arianna Yuan, Fei Wu, and Jiwei Li.
\newblock Coreference {R}esolution as {Q}uery-based {S}pan {P}rediction.
\newblock In \emph{{ACL}}, 2020{\natexlab{a}}.

\bibitem[Durrett and Klein(2013)]{durrett2013easy}
Greg Durrett and Dan Klein.
\newblock Easy victories and uphill battles in coreference resolution.
\newblock In \emph{Proceedings of the 2013 Conference on Empirical Methods in
  Natural Language Processing}, pages 1971--1982, 2013.

\bibitem[Xu and Choi(2020)]{xu-choi-2020-revealing}
Liyan Xu and Jinho~D. Choi.
\newblock {Revealing the Myth of Higher-Order Inference in Coreference
  Resolution}.
\newblock In \emph{EMNLP}, 2020.

\bibitem[Kantor and Globerson(2019)]{kantor-globerson-2019-coreference}
Ben Kantor and Amir Globerson.
\newblock {Coreference Resolution with Entity Equalization}.
\newblock In \emph{ACL}, 2019.

\bibitem[Xia et~al.(2020)Xia, Sedoc, and Van~Durme]{xia-etal-2020-incremental}
Patrick Xia, Jo{\~a}o Sedoc, and Benjamin Van~Durme.
\newblock Incremental neural coreference resolution in constant memory.
\newblock In \emph{Proceedings of the 2020 Conference on Empirical Methods in
  Natural Language Processing (EMNLP)}, pages 8617--8624, Online, November
  2020. Association for Computational Linguistics.
\newblock \doi{10.18653/v1/2020.emnlp-main.695}.
\newblock URL \url{https://aclanthology.org/2020.emnlp-main.695}.

\bibitem[Heim(1983)]{heim1983file}
Irene Heim.
\newblock File change semantics and the familiarity theory of definiteness.
\newblock In \emph{Formal Semantics}, 1983.

\bibitem[Dalvi et~al.(2018)Dalvi, Huang, Tandon, Yih, and
  Clark]{dalvi-etal-2018-tracking}
Bhavana Dalvi, Lifu Huang, Niket Tandon, Wen-tau Yih, and Peter Clark.
\newblock Tracking state changes in procedural text: a challenge dataset and
  models for process paragraph comprehension.
\newblock In \emph{Proceedings of the 2018 Conference of the North {A}merican
  Chapter of the Association for Computational Linguistics: Human Language
  Technologies, Volume 1 (Long Papers)}, pages 1595--1604, New Orleans,
  Louisiana, June 2018. Association for Computational Linguistics.
\newblock \doi{10.18653/v1/N18-1144}.
\newblock URL \url{https://aclanthology.org/N18-1144}.

\bibitem[Weston et~al.(2015{\natexlab{a}})Weston, Bordes, Chopra, Rush, van
  Merriënboer, Joulin, and Mikolov]{weston2015aicomplete}
Jason Weston, Antoine Bordes, Sumit Chopra, Alexander~M. Rush, Bart van
  Merriënboer, Armand Joulin, and Tomas Mikolov.
\newblock {Towards AI-Complete Question Answering: A Set of Prerequisite Toy
  Tasks}, 2015{\natexlab{a}}.

\bibitem[Tandon et~al.(2020)Tandon, Sakaguchi, Dalvi, Rajagopal, Clark,
  Guerquin, Richardson, and Hovy]{tandon-etal-2020-dataset}
Niket Tandon, Keisuke Sakaguchi, Bhavana Dalvi, Dheeraj Rajagopal, Peter Clark,
  Michal Guerquin, Kyle Richardson, and Eduard Hovy.
\newblock A dataset for tracking entities in open domain procedural text.
\newblock In \emph{Proceedings of the 2020 Conference on Empirical Methods in
  Natural Language Processing (EMNLP)}, pages 6408--6417, Online, November
  2020. Association for Computational Linguistics.
\newblock \doi{10.18653/v1/2020.emnlp-main.520}.
\newblock URL \url{https://aclanthology.org/2020.emnlp-main.520}.

\bibitem[Henaff et~al.(2017)Henaff, Weston, Szlam, Bordes, and
  LeCun]{henaff2016tracking}
Mikael Henaff, Jason Weston, Arthur Szlam, Antoine Bordes, and Yann LeCun.
\newblock Tracking the world state with recurrent entity networks.
\newblock In \emph{ICLR}, 2017.

\bibitem[Bosselut et~al.(2018)Bosselut, Levy, Holtzman, Ennis, Fox, and
  Choi]{bosselut-18}
Antoine Bosselut, Omer Levy, Ari Holtzman, Corin Ennis, Dieter Fox, and Yejin
  Choi.
\newblock Simulating {A}ction {D}ynamics with {N}eural {P}rocess {N}etworks.
\newblock In \emph{{ICLR}}, 2018.

\bibitem[Gupta and Durrett(2019{\natexlab{a}})]{gupta-durrett-2019-effective}
Aditya Gupta and Greg Durrett.
\newblock Effective use of transformer networks for entity tracking.
\newblock In \emph{Proceedings of the 2019 Conference on Empirical Methods in
  Natural Language Processing and the 9th International Joint Conference on
  Natural Language Processing (EMNLP-IJCNLP)}, pages 759--769, Hong Kong,
  China, November 2019{\natexlab{a}}. Association for Computational
  Linguistics.
\newblock \doi{10.18653/v1/D19-1070}.
\newblock URL \url{https://aclanthology.org/D19-1070}.

\bibitem[Dhingra et~al.(2018)Dhingra, Jin, Yang, Cohen, and
  Salakhutdinov]{dhingra-etal-2018-neural}
Bhuwan Dhingra, Qiao Jin, Zhilin Yang, William Cohen, and Ruslan Salakhutdinov.
\newblock Neural {M}odels for {R}easoning over {M}ultiple {M}entions {U}sing
  {C}oreference.
\newblock In \emph{{NAACL}}, 2018.

\bibitem[Cheng and Erk(2020)]{cheng2020entity}
Pengxiang Cheng and Katrin Erk.
\newblock {Attending to Entities for Better Text Understanding}.
\newblock In \emph{AAAI}, 2020.

\bibitem[Dasigi et~al.(2019)Dasigi, Liu, Marasovi{\'c}, Smith, and
  Gardner]{dasigi-etal-2019-quoref}
Pradeep Dasigi, Nelson~F. Liu, Ana Marasovi{\'c}, Noah~A. Smith, and Matt
  Gardner.
\newblock {{Q}uoref: A Reading Comprehension Dataset with Questions Requiring
  Coreferential Reasoning}.
\newblock In \emph{EMNLP-IJCNLP}, 2019.

\bibitem[Gao et~al.(2019)Gao, Li, King, and Lyu]{gao-etal-2019-interconnected}
Yifan Gao, Piji Li, Irwin King, and Michael~R. Lyu.
\newblock {Interconnected Question Generation with Coreference Alignment and
  Conversation Flow Modeling}.
\newblock In \emph{ACL}, 2019.

\bibitem[Sharma et~al.(2019)Sharma, Huang, Hu, and
  Wang]{sharma-etal-2019-entity}
Eva Sharma, Luyang Huang, Zhe Hu, and Lu~Wang.
\newblock {An Entity-Driven Framework for Abstractive Summarization}.
\newblock In \emph{EMNLP-IJCNLP}, 2019.

\bibitem[Narayan et~al.(2021)Narayan, Zhao, Maynez, Sim{\~o}es, Nikolaev, and
  McDonald]{narayan-etal-2021-planning}
Shashi Narayan, Yao Zhao, Joshua Maynez, Gon{\c{c}}alo Sim{\~o}es, Vitaly
  Nikolaev, and Ryan McDonald.
\newblock {Planning with Learned Entity Prompts for Abstractive Summarization}.
\newblock \emph{Transactions of the Association for Computational Linguistics},
  9:\penalty0 1475--1492, 2021.
\newblock \doi{10.1162/tacl_a_00438}.
\newblock URL \url{https://aclanthology.org/2021.tacl-1.88}.

\bibitem[Devlin et~al.(2019{\natexlab{a}})Devlin, Chang, Lee, and
  Toutanova]{devlin-etal-2019-bert}
Jacob Devlin, Ming-Wei Chang, Kenton Lee, and Kristina Toutanova.
\newblock {{BERT}: Pre-training of Deep Bidirectional Transformers for Language
  Understanding}.
\newblock In \emph{NAACL}, 2019{\natexlab{a}}.

\bibitem[Tenney et~al.(2019{\natexlab{a}})Tenney, Xia, Chen, Wang, Poliak,
  McCoy, Kim, Durme, Bowman, Das, and Pavlick]{tenney2019probing}
Ian Tenney, Patrick Xia, Berlin Chen, Alex Wang, Adam Poliak, R.~Thomas McCoy,
  Najoung Kim, Benjamin~Van Durme, Samuel~R. Bowman, Dipanjan Das, and Ellie
  Pavlick.
\newblock {What do you learn from context? Probing for sentence structure in
  contextualized word representations}.
\newblock In \emph{ICLR}, 2019{\natexlab{a}}.

\bibitem[Liu et~al.(2019{\natexlab{a}})Liu, Gardner, Belinkov, Peters, and
  Smith]{liu2019linguistic}
Nelson~F. Liu, Matt Gardner, Yonatan Belinkov, Matthew~E. Peters, and Noah~A.
  Smith.
\newblock Linguistic {K}nowledge and {T}ransferability of {C}ontextual
  {R}epresentations.
\newblock In \emph{NAACL-HLT}, 2019{\natexlab{a}}.

\bibitem[Sorodoc et~al.(2020)Sorodoc, Gulordava, and
  Boleda]{sorodoc-etal-2020-probing}
Ionut-Teodor Sorodoc, Kristina Gulordava, and Gemma Boleda.
\newblock {Probing for Referential Information in Language Models}.
\newblock In \emph{ACL}, 2020.

\bibitem[Schuster and Linzen(2022)]{schuster-linzen-2022-sentence}
Sebastian Schuster and Tal Linzen.
\newblock When a sentence does not introduce a discourse entity,
  {T}ransformer-based models still sometimes refer to it.
\newblock In \emph{Proceedings of the 2022 Conference of the North American
  Chapter of the Association for Computational Linguistics: Human Language
  Technologies}, pages 969--982, Seattle, United States, July 2022. Association
  for Computational Linguistics.
\newblock URL \url{https://aclanthology.org/2022.naacl-main.71}.

\bibitem[Ribeiro et~al.(2020)Ribeiro, Wu, Guestrin, and
  Singh]{tulio2020checklist}
Marco~Tulio Ribeiro, Tongshuang Wu, Carlos Guestrin, and Sameer Singh.
\newblock {Beyond Accuracy: Behavioral Testing of NLP models with CheckList}.
\newblock In \emph{ACL}, 2020.

\bibitem[Kryscinski et~al.(2019)Kryscinski, Keskar, McCann, Xiong, and
  Socher]{kryscinski-etal-2019-neural}
Wojciech Kryscinski, Nitish~Shirish Keskar, Bryan McCann, Caiming Xiong, and
  Richard Socher.
\newblock {Neural Text Summarization: A Critical Evaluation}.
\newblock In \emph{EMNLP-IJCNLP}, 2019.

\bibitem[Wu et~al.(2021{\natexlab{a}})Wu, Ouyang, Ziegler, Stiennon, Lowe,
  Leike, and Christiano]{wu2021recursive}
Jeff Wu, Long Ouyang, Daniel~M. Ziegler, Nisan Stiennon, Ryan Lowe, Jan Leike,
  and Paul~F. Christiano.
\newblock Recursively summarizing books with human feedback.
\newblock \emph{CoRR}, abs/2109.10862, 2021{\natexlab{a}}.
\newblock URL \url{https://arxiv.org/abs/2109.10862}.

\bibitem[Shaham et~al.(2022)Shaham, Segal, Ivgi, Efrat, Yoran, Haviv, Gupta,
  Xiong, Geva, Berant, and Levy]{shaham-etal-2022-scrolls}
Uri Shaham, Elad Segal, Maor Ivgi, Avia Efrat, Ori Yoran, Adi Haviv, Ankit
  Gupta, Wenhan Xiong, Mor Geva, Jonathan Berant, and Omer Levy.
\newblock {SCROLLS:} standardized comparison over long language sequences.
\newblock \emph{CoRR}, abs/2201.03533, 2022.
\newblock URL \url{https://arxiv.org/abs/2201.03533}.

\bibitem[Beltagy et~al.(2020)Beltagy, Peters, and Cohan]{beltagy2020longformer}
Iz~Beltagy, Matthew~E. Peters, and Arman Cohan.
\newblock {Longformer: The Long-Document Transformer}.
\newblock \emph{arXiv:2004.05150}, 2020.

\bibitem[Roy et~al.(2021)Roy, Saffar, Vaswani, and
  Grangier]{roy-etal-2021-efficient}
Aurko Roy, Mohammad Saffar, Ashish Vaswani, and David Grangier.
\newblock Efficient content-based sparse attention with routing transformers.
\newblock \emph{Transactions of the Association for Computational Linguistics},
  9:\penalty0 53--68, 2021.
\newblock \doi{10.1162/tacl_a_00353}.
\newblock URL \url{https://aclanthology.org/2021.tacl-1.4}.

\bibitem[Kočiský et~al.(2018)Kočiský, Schwarz, Blunsom, Dyer, Hermann,
  Melis, and Grefenstette]{kocisky2018narrative}
Tomáš Kočiský, Jonathan Schwarz, Phil Blunsom, Chris Dyer, Karl~Moritz
  Hermann, Gábor Melis, and Edward Grefenstette.
\newblock {The NarrativeQA Reading Comprehension Challenge}.
\newblock \emph{TACL}, 6, 2018.

\bibitem[Chen et~al.(2022)Chen, Chu, Wiseman, and
  Gimpel]{chen-etal-2022-summscreen}
Mingda Chen, Zewei Chu, Sam Wiseman, and Kevin Gimpel.
\newblock {S}umm{S}creen: A dataset for abstractive screenplay summarization.
\newblock In \emph{Proceedings of the 60th Annual Meeting of the Association
  for Computational Linguistics (Volume 1: Long Papers)}, pages 8602--8615,
  Dublin, Ireland, May 2022. Association for Computational Linguistics.
\newblock \doi{10.18653/v1/2022.acl-long.589}.
\newblock URL \url{https://aclanthology.org/2022.acl-long.589}.

\bibitem[Pang et~al.(2022)Pang, Parrish, Joshi, Nangia, Phang, Chen,
  Padmakumar, Ma, Thompson, He, and Bowman]{pang-etal-2022-quality}
Richard~Yuanzhe Pang, Alicia Parrish, Nitish Joshi, Nikita Nangia, Jason Phang,
  Angelica Chen, Vishakh Padmakumar, Johnny Ma, Jana Thompson, He~He, and
  Samuel Bowman.
\newblock {Q}u{ALITY}: Question answering with long input texts, yes!
\newblock In \emph{Proceedings of the 2022 Conference of the North American
  Chapter of the Association for Computational Linguistics: Human Language
  Technologies}, pages 5336--5358, Seattle, United States, July 2022.
  Association for Computational Linguistics.
\newblock URL \url{https://aclanthology.org/2022.naacl-main.391}.

\bibitem[Xu et~al.(2022)Xu, Szlam, and Weston]{xu-etal-2022-beyond}
Jing Xu, Arthur Szlam, and Jason Weston.
\newblock Beyond goldfish memory: Long-term open-domain conversation.
\newblock In \emph{Proceedings of the 60th Annual Meeting of the Association
  for Computational Linguistics (Volume 1: Long Papers)}, pages 5180--5197,
  Dublin, Ireland, May 2022. Association for Computational Linguistics.
\newblock \doi{10.18653/v1/2022.acl-long.356}.
\newblock URL \url{https://aclanthology.org/2022.acl-long.356}.

\bibitem[Sun et~al.(2021)Sun, Krishna, Mattarella-Micke, and
  Iyyer]{sun-etal-2021-long}
Simeng Sun, Kalpesh Krishna, Andrew Mattarella-Micke, and Mohit Iyyer.
\newblock Do long-range language models actually use long-range context?
\newblock In \emph{Proceedings of the 2021 Conference on Empirical Methods in
  Natural Language Processing}, pages 807--822, Online and Punta Cana,
  Dominican Republic, November 2021. Association for Computational Linguistics.
\newblock \doi{10.18653/v1/2021.emnlp-main.62}.
\newblock URL \url{https://aclanthology.org/2021.emnlp-main.62}.

\bibitem[Shuster et~al.(2022)Shuster, Urbanek, Szlam, and
  Weston]{shuster-etal-2022-state}
Kurt Shuster, Jack Urbanek, Arthur Szlam, and Jason Weston.
\newblock Am {I} me or you? state-of-the-art dialogue models cannot maintain an
  identity.
\newblock In \emph{Findings of the Association for Computational Linguistics:
  NAACL 2022}, pages 2367--2387, Seattle, United States, July 2022. Association
  for Computational Linguistics.
\newblock URL \url{https://aclanthology.org/2022.findings-naacl.182}.

\bibitem[Lewis et~al.(2020{\natexlab{a}})Lewis, Perez, Piktus, Petroni,
  Karpukhin, Goyal, K\"{u}ttler, Lewis, Yih, Rockt\"{a}schel, Riedel, and
  Kiela]{lewis-etal-2020-retrieval}
Patrick Lewis, Ethan Perez, Aleksandra Piktus, Fabio Petroni, Vladimir
  Karpukhin, Naman Goyal, Heinrich K\"{u}ttler, Mike Lewis, Wen-tau Yih, Tim
  Rockt\"{a}schel, Sebastian Riedel, and Douwe Kiela.
\newblock Retrieval-augmented generation for knowledge-intensive nlp tasks.
\newblock In H.~Larochelle, M.~Ranzato, R.~Hadsell, M.F. Balcan, and H.~Lin,
  editors, \emph{Advances in Neural Information Processing Systems}, volume~33,
  pages 9459--9474. Curran Associates, Inc., 2020{\natexlab{a}}.
\newblock URL
  \url{https://proceedings.neurips.cc/paper/2020/file/6b493230205f780e1bc26945df7481e5-Paper.pdf}.

\bibitem[Karpukhin et~al.(2020)Karpukhin, Oguz, Min, Lewis, Wu, Edunov, Chen,
  and Yih]{karpukhin-etal-2020-dense}
Vladimir Karpukhin, Barlas Oguz, Sewon Min, Patrick Lewis, Ledell Wu, Sergey
  Edunov, Danqi Chen, and Wen-tau Yih.
\newblock {Dense Passage Retrieval for Open-Domain Question Answering}.
\newblock In \emph{EMNLP}, 2020.

\bibitem[Mou et~al.(2021)Mou, Yang, Yu, Yao, Guo, Potdar, and
  Su]{mou-etal-2021-narrative}
Xiangyang Mou, Chenghao Yang, Mo~Yu, Bingsheng Yao, Xiaoxiao Guo, Saloni
  Potdar, and Hui Su.
\newblock Narrative question answering with cutting-edge open-domain {QA}
  techniques: A comprehensive study.
\newblock \emph{Transactions of the Association for Computational Linguistics},
  9:\penalty0 1032--1046, 2021.
\newblock \doi{10.1162/tacl_a_00411}.
\newblock URL \url{https://aclanthology.org/2021.tacl-1.61}.

\bibitem[Sukhbaatar et~al.(2021)Sukhbaatar, Ju, Poff, Roller, Szlam, Weston,
  and Fan]{Sukhbaatar2021NotAM}
Sainbayar Sukhbaatar, Da~Ju, Spencer Poff, Stephen Roller, Arthur~D. Szlam,
  Jason Weston, and Angela Fan.
\newblock Not all memories are created equal: Learning to forget by expiring.
\newblock In \emph{ICML}, 2021.

\bibitem[Ji et~al.(2017)Ji, Tan, Martschat, Choi, and
  Smith]{ji-etal-2017-dynamic}
Yangfeng Ji, Chenhao Tan, Sebastian Martschat, Yejin Choi, and Noah~A. Smith.
\newblock {Dynamic Entity Representations in Neural Language Models}.
\newblock In \emph{EMNLP}, 2017.

\bibitem[Ye et~al.(2020)Ye, Lin, Du, Liu, Li, Sun, and
  Liu]{ye-etal-2020-coreferential}
Deming Ye, Yankai Lin, Jiaju Du, Zhenghao Liu, Peng Li, Maosong Sun, and
  Zhiyuan Liu.
\newblock Coreferential reasoning learning for language representation.
\newblock In Bonnie Webber, Trevor Cohn, Yulan He, and Yang Liu, editors,
  \emph{EMNLP}, 2020.

\bibitem[Wu et~al.(2021{\natexlab{b}})Wu, Peng, and Smith]{wu2021infusing}
Zhaofeng Wu, Hao Peng, and Noah~A. Smith.
\newblock {Infusing Finetuning with Semantic Dependencies}.
\newblock \emph{Transactions of the Association for Computational Linguistics},
  9:\penalty0 226--242, 03 2021{\natexlab{b}}.
\newblock ISSN 2307-387X.
\newblock \doi{10.1162/tacl_a_00363}.
\newblock URL \url{https://doi.org/10.1162/tacl\_a\_00363}.

\bibitem[Chen et~al.(2018)Chen, Fan, Lu, Yuille, and
  Rong]{chen-etal-2018-preco}
Hong Chen, Zhenhua Fan, Hao Lu, Alan Yuille, and Shu Rong.
\newblock "{P}re{C}o: A large-scale dataset in preschool vocabulary for
  coreference resolution".
\newblock In \emph{EMNLP}, 2018.

\bibitem[Ghaddar and Langlais(2016)]{ghaddar-langlais-2016-wikicoref}
Abbas Ghaddar and Phillippe Langlais.
\newblock {{W}iki{C}oref: An {E}nglish Coreference-annotated Corpus of
  {W}ikipedia Articles}.
\newblock In \emph{LREC}, 2016.

\bibitem[Srivastava et~al.(2022)Srivastava, Rastogi, Rao, Shoeb, Abid, Fisch,
  Brown, Santoro, Gupta, Garriga-Alonso, Kluska, Lewkowycz, Agarwal, Power,
  Ray, Warstadt, Kocurek, Safaya, Tazarv, Xiang, Parrish, Nie, Hussain, Askell,
  Dsouza, Slone, Rahane, Iyer, Andreassen, Madotto, Santilli, Stuhlmuller, Dai,
  La, Lampinen, Zou, Jiang, Chen, Vuong, Gupta, Gottardi, Norelli, Venkatesh,
  Gholamidavoodi, Tabassum, Menezes, Kirubarajan, Mullokandov, Sabharwal,
  Herrick, Efrat, Erdem, Karakas, Roberts, Loe, Zoph, Bojanowski, Ozyurt,
  Hedayatnia, Neyshabur, Inden, Stein, Ekmekci, Lin, Howald, Diao, Dour,
  Stinson, Argueta, Ramirez, Singh, Rathkopf, Meng, Baral, Wu, Callison-Burch,
  Waites, Voigt, Manning, Potts, Ramirez, Rivera, Siro, Raffel, Ashcraft,
  Garbacea, Sileo, Garrette, Hendrycks, Kilman, Roth, Freeman, Khashabi, Levy,
  Gonzalez, Perszyk, Hernandez, Chen, Ippolito, Gilboa, Dohan, Drakard,
  Jurgens, Datta, Ganguli, Emelin, Kleyko, Yuret, Chen, Tam, Hupkes, Misra,
  Buzan, Mollo, Yang, Lee, Shutova, Cubuk, Segal, Hagerman, Barnes, Donoway,
  Pavlick, Rodola, Lam, Chu, Tang, Erdem, Chang, Chi, Dyer, Jerzak, Kim,
  Manyasi, Zheltonozhskii, Xia, Siar, Martinez-Plumed, Happe, Chollet, Rong,
  Mishra, Winata, de~Melo, Kruszewski, Parascandolo, Mariani, Wang,
  Jaimovitch-Lopez, Betz, Gur-Ari, Galijasevic, Kim, Rashkin, Hajishirzi,
  Mehta, Bogar, Shevlin, Schutze, Yakura, Zhang, Wong, Ng, Noble, Jumelet,
  Geissinger, Kernion, Hilton, Lee, Fisac, Simon, Koppel, Zheng, Zou, Kocon,
  Thompson, Kaplan, Radom, Sohl-Dickstein, Phang, Wei, Yosinski, Novikova,
  Bosscher, Marsh, Kim, Taal, Engel, Alabi, Xu, Song, Tang, Waweru, Burden,
  Miller, Balis, Berant, Frohberg, Rozen, Hernandez-Orallo, Boudeman, Jones,
  Tenenbaum, Rule, Chua, Kanclerz, Livescu, Krauth, Gopalakrishnan, Ignatyeva,
  Markert, Dhole, Gimpel, Omondi, Mathewson, Chiafullo, Shkaruta, Shridhar,
  McDonell, Richardson, Reynolds, Gao, Zhang, Dugan, Qin, Contreras-Ochando,
  Morency, Moschella, Lam, Noble, Schmidt, He, Colon, Metz, Senel, Bosma, Sap,
  ter Hoeve, Farooqi, Faruqui, Mazeika, Baturan, Marelli, Maru, Quintana,
  Tolkiehn, Giulianelli, Lewis, Potthast, Leavitt, Hagen, Schubert,
  Baitemirova, Arnaud, McElrath, Yee, Cohen, Gu, Ivanitskiy, Starritt, Strube,
  Swedrowski, Bevilacqua, Yasunaga, Kale, Cain, Xu, Suzgun, Tiwari, Bansal,
  Aminnaseri, Geva, Gheini, T, Peng, Chi, Lee, Krakover, Cameron, Roberts,
  Doiron, Nangia, Deckers, Muennighoff, Keskar, Iyer, Constant, Fiedel, Wen,
  Zhang, Agha, Elbaghdadi, Levy, Evans, Casares, Doshi, Fung, Liang, Vicol,
  Alipoormolabashi, Liao, Liang, Chang, Eckersley, Htut, Hwang, Milkowski,
  Patil, Pezeshkpour, Oli, Mei, Lyu, Chen, Banjade, Rudolph, Gabriel, Habacker,
  Delgado, Milliere, Garg, Barnes, Saurous, Arakawa, Raymaekers, Frank, Sikand,
  Novak, Sitelew, LeBras, Liu, Jacobs, Zhang, Salakhutdinov, Chi, Lee, Stovall,
  Teehan, Yang, Singh, Mohammad, Anand, Dillavou, Shleifer, Wiseman, Gruetter,
  Bowman, Schoenholz, Han, Kwatra, Rous, Ghazarian, Ghosh, Casey, Bischoff,
  Gehrmann, Schuster, Sadeghi, Hamdan, Zhou, Srivastava, Shi, Singh, Asaadi,
  Gu, Pachchigar, Toshniwal, Upadhyay, Debnath, Shakeri, Thormeyer, Melzi,
  Reddy, Makini, Lee, Torene, Hatwar, Dehaene, Divic, Ermon, Biderman, Lin,
  Prasad, Piantadosi, Shieber, Misherghi, Kiritchenko, Mishra, Linzen,
  Schuster, Li, Yu, Ali, Hashimoto, Wu, Desbordes, Rothschild, Phan, Wang,
  Nkinyili, Schick, Kornev, Telleen-Lawton, Tunduny, Gerstenberg, Chang,
  Neeraj, Khot, Shultz, Shaham, Misra, Demberg, Nyamai, Raunak, Ramasesh,
  Prabhu, Padmakumar, Srikumar, Fedus, Saunders, Zhang, Vossen, Ren, Tong,
  Zhao, Wu, Shen, Yaghoobzadeh, Lakretz, Song, Bahri, Choi, Yang, Hao, Chen,
  Belinkov, Hou, Hou, Bai, Seid, Zhao, Wang, Wang, Wang, and Wu]{bigbench2022}
Aarohi Srivastava, Abhinav Rastogi, Abhishek Rao, Abu Awal~Md Shoeb, Abubakar
  Abid, Adam Fisch, Adam~R. Brown, Adam Santoro, Aditya Gupta, Adria
  Garriga-Alonso, Agnieszka Kluska, Aitor Lewkowycz, Akshat Agarwal, Alethea
  Power, Alex Ray, Alex Warstadt, Alexander~W. Kocurek, Ali Safaya, Ali Tazarv,
  Alice Xiang, Alicia Parrish, Allen Nie, Aman Hussain, Amanda Askell, Amanda
  Dsouza, Ambrose Slone, Ameet Rahane, Anantharaman~S. Iyer, Anders Andreassen,
  Andrea Madotto, Andrea Santilli, Andreas Stuhlmuller, Andrew Dai, Andrew La,
  Andrew Lampinen, Andy Zou, Angela Jiang, Angelica Chen, Anh Vuong, Animesh
  Gupta, Anna Gottardi, Antonio Norelli, Anu Venkatesh, Arash Gholamidavoodi,
  Arfa Tabassum, Arul Menezes, Arun Kirubarajan, Asher Mullokandov, Ashish
  Sabharwal, Austin Herrick, Avia Efrat, Aykut Erdem, Ayla Karakas, B.~Ryan
  Roberts, Bao~Sheng Loe, Barret Zoph, Bartlomiej Bojanowski, Batuhan Ozyurt,
  Behnam Hedayatnia, Behnam Neyshabur, Benjamin Inden, Benno Stein, Berk
  Ekmekci, Bill~Yuchen Lin, Blake Howald, Cameron Diao, Cameron Dour, Catherine
  Stinson, Cedrick Argueta, Cesar~Ferri Ramirez, Chandan Singh, Charles
  Rathkopf, Chenlin Meng, Chitta Baral, Chiyu Wu, Chris Callison-Burch, Chris
  Waites, Christian Voigt, Christopher~D. Manning, Christopher Potts, Cindy
  Ramirez, Clara~E. Rivera, Clemencia Siro, Colin Raffel, Courtney Ashcraft,
  Cristina Garbacea, Damien Sileo, Dan Garrette, Dan Hendrycks, Dan Kilman, Dan
  Roth, Daniel Freeman, Daniel Khashabi, Daniel Levy, Daniel~Mosegui Gonzalez,
  Danielle Perszyk, Danny Hernandez, Danqi Chen, Daphne Ippolito, Dar Gilboa,
  David Dohan, David Drakard, David Jurgens, Debajyoti Datta, Deep Ganguli,
  Denis Emelin, Denis Kleyko, Deniz Yuret, Derek Chen, Derek Tam, Dieuwke
  Hupkes, Diganta Misra, Dilyar Buzan, Dimitri~Coelho Mollo, Diyi Yang, Dong-Ho
  Lee, Ekaterina Shutova, Ekin~Dogus Cubuk, Elad Segal, Eleanor Hagerman,
  Elizabeth Barnes, Elizabeth Donoway, Ellie Pavlick, Emanuele Rodola, Emma
  Lam, Eric Chu, Eric Tang, Erkut Erdem, Ernie Chang, Ethan~A. Chi, Ethan Dyer,
  Ethan Jerzak, Ethan Kim, Eunice~Engefu Manyasi, Evgenii Zheltonozhskii,
  Fanyue Xia, Fatemeh Siar, Fernando Martinez-Plumed, Francesca Happe, Francois
  Chollet, Frieda Rong, Gaurav Mishra, Genta~Indra Winata, Gerard de~Melo,
  German Kruszewski, Giambattista Parascandolo, Giorgio Mariani, Gloria Wang,
  Gonzalo Jaimovitch-Lopez, Gregor Betz, Guy Gur-Ari, Hana Galijasevic, Hannah
  Kim, Hannah Rashkin, Hannaneh Hajishirzi, Harsh Mehta, Hayden Bogar, Henry
  Shevlin, Hinrich Schutze, Hiromu Yakura, Hongming Zhang, Hugh~Mee Wong, Ian
  Ng, Isaac Noble, Jaap Jumelet, Jack Geissinger, Jackson Kernion, Jacob
  Hilton, Jaehoon Lee, Jaime~Fernandez Fisac, James~B. Simon, James Koppel,
  James Zheng, James Zou, Jan Kocon, Jana Thompson, Jared Kaplan, Jarema Radom,
  Jascha Sohl-Dickstein, Jason Phang, Jason Wei, Jason Yosinski, Jekaterina
  Novikova, Jelle Bosscher, Jennifer Marsh, Jeremy Kim, Jeroen Taal, Jesse
  Engel, Jesujoba Alabi, Jiacheng Xu, Jiaming Song, Jillian Tang, Joan Waweru,
  John Burden, John Miller, John~U. Balis, Jonathan Berant, Jorg Frohberg, Jos
  Rozen, Jose Hernandez-Orallo, Joseph Boudeman, Joseph Jones, Joshua~B.
  Tenenbaum, Joshua~S. Rule, Joyce Chua, Kamil Kanclerz, Karen Livescu, Karl
  Krauth, Karthik Gopalakrishnan, Katerina Ignatyeva, Katja Markert,
  Kaustubh~D. Dhole, Kevin Gimpel, Kevin Omondi, Kory Mathewson, Kristen
  Chiafullo, Ksenia Shkaruta, Kumar Shridhar, Kyle McDonell, Kyle Richardson,
  Laria Reynolds, Leo Gao, Li~Zhang, Liam Dugan, Lianhui Qin, Lidia
  Contreras-Ochando, Louis-Philippe Morency, Luca Moschella, Lucas Lam, Lucy
  Noble, Ludwig Schmidt, Luheng He, Luis~Oliveros Colon, Luke Metz, Lutfi~Kerem
  Senel, Maarten Bosma, Maarten Sap, Maartje ter Hoeve, Maheen Farooqi, Manaal
  Faruqui, Mantas Mazeika, Marco Baturan, Marco Marelli, Marco Maru, Maria
  Jose~Ramirez Quintana, Marie Tolkiehn, Mario Giulianelli, Martha Lewis,
  Martin Potthast, Matthew~L. Leavitt, Matthias Hagen, Matyas Schubert,
  Medina~Orduna Baitemirova, Melody Arnaud, Melvin McElrath, Michael~A. Yee,
  Michael Cohen, Michael Gu, Michael Ivanitskiy, Michael Starritt, Michael
  Strube, Michal Swedrowski, Michele Bevilacqua, Michihiro Yasunaga, Mihir
  Kale, Mike Cain, Mimee Xu, Mirac Suzgun, Mo~Tiwari, Mohit Bansal, Moin
  Aminnaseri, Mor Geva, Mozhdeh Gheini, Mukund~Varma T, Nanyun Peng, Nathan
  Chi, Nayeon Lee, Neta Gur-Ari Krakover, Nicholas Cameron, Nicholas Roberts,
  Nick Doiron, Nikita Nangia, Niklas Deckers, Niklas Muennighoff,
  Nitish~Shirish Keskar, Niveditha~S. Iyer, Noah Constant, Noah Fiedel, Nuan
  Wen, Oliver Zhang, Omar Agha, Omar Elbaghdadi, Omer Levy, Owain Evans, Pablo
  Antonio~Moreno Casares, Parth Doshi, Pascale Fung, Paul~Pu Liang, Paul Vicol,
  Pegah Alipoormolabashi, Peiyuan Liao, Percy Liang, Peter Chang, Peter
  Eckersley, Phu~Mon Htut, Pinyu Hwang, Piotr Milkowski, Piyush Patil, Pouya
  Pezeshkpour, Priti Oli, Qiaozhu Mei, Qing Lyu, Qinlang Chen, Rabin Banjade,
  Rachel~Etta Rudolph, Raefer Gabriel, Rahel Habacker, Ramon~Risco Delgado,
  Raphael Milliere, Rhythm Garg, Richard Barnes, Rif~A. Saurous, Riku Arakawa,
  Robbe Raymaekers, Robert Frank, Rohan Sikand, Roman Novak, Roman Sitelew,
  Ronan LeBras, Rosanne Liu, Rowan Jacobs, Rui Zhang, Ruslan Salakhutdinov,
  Ryan Chi, Ryan Lee, Ryan Stovall, Ryan Teehan, Rylan Yang, Sahib Singh,
  Saif~M. Mohammad, Sajant Anand, Sam Dillavou, Sam Shleifer, Sam Wiseman,
  Samuel Gruetter, Samuel~R. Bowman, Samuel~S. Schoenholz, Sanghyun Han,
  Sanjeev Kwatra, Sarah~A. Rous, Sarik Ghazarian, Sayan Ghosh, Sean Casey,
  Sebastian Bischoff, Sebastian Gehrmann, Sebastian Schuster, Sepideh Sadeghi,
  Shadi Hamdan, Sharon Zhou, Shashank Srivastava, Sherry Shi, Shikhar Singh,
  Shima Asaadi, Shixiang~Shane Gu, Shubh Pachchigar, Shubham Toshniwal, Shyam
  Upadhyay, Shyamolima~(Shammie) Debnath, Siamak Shakeri, Simon Thormeyer,
  Simone Melzi, Siva Reddy, Sneha~Priscilla Makini, Soo-Hwan Lee, Spencer
  Torene, Sriharsha Hatwar, Stanislas Dehaene, Stefan Divic, Stefano Ermon,
  Stella Biderman, Stephanie Lin, Stephen Prasad, Steven~T. Piantadosi,
  Stuart~M. Shieber, Summer Misherghi, Svetlana Kiritchenko, Swaroop Mishra,
  Tal Linzen, Tal Schuster, Tao Li, Tao Yu, Tariq Ali, Tatsu Hashimoto, Te-Lin
  Wu, Theo Desbordes, Theodore Rothschild, Thomas Phan, Tianle Wang, Tiberius
  Nkinyili, Timo Schick, Timofei Kornev, Timothy Telleen-Lawton, Titus Tunduny,
  Tobias Gerstenberg, Trenton Chang, Trishala Neeraj, Tushar Khot, Tyler
  Shultz, Uri Shaham, Vedant Misra, Vera Demberg, Victoria Nyamai, Vikas
  Raunak, Vinay Ramasesh, Vinay~Uday Prabhu, Vishakh Padmakumar, Vivek
  Srikumar, William Fedus, William Saunders, William Zhang, Wout Vossen, Xiang
  Ren, Xiaoyu Tong, Xinran Zhao, Xinyi Wu, Xudong Shen, Yadollah Yaghoobzadeh,
  Yair Lakretz, Yangqiu Song, Yasaman Bahri, Yejin Choi, Yichi Yang, Yiding
  Hao, Yifu Chen, Yonatan Belinkov, Yu~Hou, Yufang Hou, Yuntao Bai, Zachary
  Seid, Zhuoye Zhao, Zijian Wang, Zijie~J. Wang, Zirui Wang, and Ziyi Wu.
\newblock {Beyond the Imitation Game: Quantifying and extrapolating the
  capabilities of language models}, 2022.

\bibitem[Winograd(1972)]{Winograd1972Understanding}
Terry Winograd.
\newblock Understanding natural language.
\newblock \emph{Cognitive Psychology}, 3, 1972.

\bibitem[Charniak(1972)]{charniak72thesis}
Eugene Charniak.
\newblock \emph{Toward a model of children's story comprehension}.
\newblock PhD thesis, MIT, 1972.

\bibitem[McCarthy and Lehnert(1995)]{mccarthy95decision}
Joseph McCarthy and Wendy Lehnert.
\newblock {Using Decision Trees for Coreference Resolution}.
\newblock In \emph{IJCAI}, 1995.

\bibitem[Ge et~al.(1998)Ge, Hale, and Charniak]{ge-etal-1998-statistical}
Niyu Ge, John Hale, and Eugene Charniak.
\newblock {A Statistical Approach to Anaphora Resolution}.
\newblock In \emph{Sixth Workshop on Very Large Corpora}, 1998.

\bibitem[Soon et~al.(2001)Soon, Ng, and Lim]{soon-etal-2001-machine}
Wee~Meng Soon, Hwee~Tou Ng, and Daniel Chung~Yong Lim.
\newblock {A Machine Learning Approach to Coreference Resolution of Noun
  Phrases}.
\newblock \emph{Computational Linguistics}, 27\penalty0 (4), 2001.

\bibitem[Bengtson and Roth(2008)]{bengtson-roth-2008-understanding}
Eric Bengtson and Dan Roth.
\newblock {Understanding the Value of Features for Coreference Resolution}.
\newblock In \emph{EMNLP}, 2008.

\bibitem[Rahman and Ng(2011)]{rahman2011narrowing}
Altaf Rahman and Vincent Ng.
\newblock {Narrowing the Modeling Gap: {A} Cluster-Ranking Approach to
  Coreference Resolution}.
\newblock \emph{JAIR}, 40, 2011.

\bibitem[Ng and Cardie(2002)]{ng-cardie-2002-improving}
Vincent Ng and Claire Cardie.
\newblock {Improving Machine Learning Approaches to Coreference Resolution}.
\newblock In \emph{ACL}, 2002.

\bibitem[Luo et~al.(2004)Luo, Ittycheriah, Jing, Kambhatla, and
  Roukos]{luo-etal-2004-mention}
Xiaoqiang Luo, Abe Ittycheriah, Hongyan Jing, Nanda Kambhatla, and Salim
  Roukos.
\newblock {{A Mention-Synchronous Coreference Resolution Algorithm Based On the
  Bell Tree}}.
\newblock In \emph{{ACL}}, 2004.

\bibitem[Yang et~al.(2008)Yang, Su, Lang, Tan, Liu, and
  Li]{yang-etal-2008-entity}
Xiaofeng Yang, Jian Su, Jun Lang, Chew~Lim Tan, Ting Liu, and Sheng Li.
\newblock {An Entity-Mention Model for Coreference Resolution with Inductive
  Logic Programming}.
\newblock In \emph{ACL-HLT}, 2008.

\bibitem[Stoyanov and Eisner(2012)]{stoyanov-eisner-2012-easy}
Veselin Stoyanov and Jason Eisner.
\newblock {{Easy-first Coreference Resolution}}.
\newblock In \emph{{COLING}}, 2012.

\bibitem[Webster and Curran(2014)]{websterC14}
Kellie Webster and James~R. Curran.
\newblock Limited memory incremental coreference resolution.
\newblock In Jan Hajic and Junichi Tsujii, editors, \emph{{COLING}}, 2014.

\bibitem[Clark and Manning(2016{\natexlab{b}})]{clark-manning-2016-improving}
Kevin Clark and Christopher~D Manning.
\newblock Improving {C}oreference {R}esolution by {L}earning {E}ntity-{L}evel
  {D}istributed {R}epresentations.
\newblock In \emph{ACL}, 2016{\natexlab{b}}.

\bibitem[Clark and Manning(2015)]{clark-manning-2015-entity}
Kevin Clark and Christopher~D. Manning.
\newblock {{Entity-Centric Coreference Resolution with Model Stacking}}.
\newblock In \emph{ACL}, 2015.

\bibitem[Wiseman et~al.(2016{\natexlab{a}})Wiseman, Rush, and
  Shieber]{wiseman-etal-2016-learning}
Sam Wiseman, Alexander~M. Rush, and Stuart~M. Shieber.
\newblock Learning {G}lobal {F}eatures for {C}oreference {R}esolution.
\newblock In \emph{{NAACL}}, 2016{\natexlab{a}}.

\bibitem[Paolini et~al.(2021)Paolini, Athiwaratkun, Krone, Ma, Achille,
  ANUBHAI, dos Santos, Xiang, and Soatto]{paolini2021structured}
Giovanni Paolini, Ben Athiwaratkun, Jason Krone, Jie Ma, Alessandro Achille,
  RISHITA ANUBHAI, Cicero~Nogueira dos Santos, Bing Xiang, and Stefano Soatto.
\newblock {Structured Prediction as Translation between Augmented Natural
  Languages}.
\newblock In \emph{International Conference on Learning Representations}, 2021.
\newblock URL \url{https://openreview.net/forum?id=US-TP-xnXI}.

\bibitem[Kirstain et~al.(2021)Kirstain, Ram, and
  Levy]{kirstain-etal-2021-coreference}
Yuval Kirstain, Ori Ram, and Omer Levy.
\newblock Coreference resolution without span representations.
\newblock In \emph{ACL}, Online, 2021.

\bibitem[Dobrovolskii(2021)]{dobrovolskii-2021-word}
Vladimir Dobrovolskii.
\newblock {Word-Level Coreference Resolution}.
\newblock In \emph{Proceedings of the 2021 Conference on Empirical Methods in
  Natural Language Processing}, pages 7670--7675, Online and Punta Cana,
  Dominican Republic, November 2021. Association for Computational Linguistics.
\newblock \doi{10.18653/v1/2021.emnlp-main.605}.
\newblock URL \url{https://aclanthology.org/2021.emnlp-main.605}.

\bibitem[Vilain et~al.(1995)Vilain, Burger, Aberdeen, Connolly, and
  Hirschman]{vilain-etal-1995-model}
Marc Vilain, John Burger, John Aberdeen, Dennis Connolly, and Lynette
  Hirschman.
\newblock {A Model-Theoretic Coreference Scoring Scheme}.
\newblock In \emph{Sixth Message Understanding Conference ({MUC}-6)}, 1995.

\bibitem[Bagga and Baldwin(1998)]{Bagga98algorithmsfor}
Amit Bagga and Breck Baldwin.
\newblock {Algorithms for Scoring Coreference Chains}.
\newblock In \emph{LREC}, 1998.

\bibitem[Luo(2005)]{luo-2005-coreference}
Xiaoqiang Luo.
\newblock {On Coreference Resolution Performance Metrics}.
\newblock In \emph{EMNLP-HLT}, 2005.

\bibitem[Moosavi and Strube(2016)]{moosavi-strube-2016-coreference}
Nafise~Sadat Moosavi and Michael Strube.
\newblock {Which Coreference Evaluation Metric Do You Trust? A Proposal for a
  Link-based Entity Aware Metric}.
\newblock In \emph{ACL}, 2016.

\bibitem[Holen(2013)]{holen-2013-critical}
Gordana~Ili{\'c} Holen.
\newblock {Critical Reflections on Evaluation Practices in Coreference
  Resolution}.
\newblock In \emph{NAACL-HLT Student Research Workshop}, 2013.

\bibitem[Denis and Baldridge(2009)]{denis09global}
Pascal Denis and Jason Baldridge.
\newblock Global joint models for coreference resolution and named entity
  classification.
\newblock \emph{Procesamiento del lenguaje natural}, 42, 2009.

\bibitem[Kiddon et~al.(2016)Kiddon, Zettlemoyer, and
  Choi]{kiddon-etal-2016-globally}
Chlo{\'e} Kiddon, Luke Zettlemoyer, and Yejin Choi.
\newblock Globally coherent text generation with neural checklist models.
\newblock In \emph{Proceedings of the 2016 Conference on Empirical Methods in
  Natural Language Processing}, pages 329--339, Austin, Texas, November 2016.
  Association for Computational Linguistics.
\newblock \doi{10.18653/v1/D16-1032}.
\newblock URL \url{https://aclanthology.org/D16-1032}.

\bibitem[Gupta and Durrett(2019{\natexlab{b}})]{gupta-durrett-2019-tracking}
Aditya Gupta and Greg Durrett.
\newblock Tracking discrete and continuous entity state for process
  understanding.
\newblock In \emph{Proceedings of the Third Workshop on Structured Prediction
  for {NLP}}, pages 7--12, Minneapolis, Minnesota, June 2019{\natexlab{b}}.
  Association for Computational Linguistics.
\newblock \doi{10.18653/v1/W19-1502}.
\newblock URL \url{https://aclanthology.org/W19-1502}.

\bibitem[Paperno et~al.(2016)Paperno, Kruszewski, Lazaridou, Pham, Bernardi,
  Pezzelle, Baroni, Boleda, and Fern{\'a}ndez]{paperno-etal-2016-lambada}
Denis Paperno, Germ{\'a}n Kruszewski, Angeliki Lazaridou, Ngoc~Quan Pham,
  Raffaella Bernardi, Sandro Pezzelle, Marco Baroni, Gemma Boleda, and Raquel
  Fern{\'a}ndez.
\newblock {The {LAMBADA} dataset: Word prediction requiring a broad discourse
  context}.
\newblock In \emph{ACL}, 2016.

\bibitem[Chu et~al.(2017)Chu, Wang, Gimpel, and
  McAllester]{chu-etal-2017-broad}
Zewei Chu, Hai Wang, Kevin Gimpel, and David McAllester.
\newblock Broad context language modeling as reading comprehension.
\newblock In \emph{EACL}, 2017.

\bibitem[Liu et~al.(2019{\natexlab{b}})Liu, Zettlemoyer, and
  Eisenstein]{liu2019referential}
Fei Liu, Luke Zettlemoyer, and Jacob Eisenstein.
\newblock The {R}eferential {R}eader: {A} {R}ecurrent {E}ntity {N}etwork for
  {A}naphora {R}esolution.
\newblock In \emph{ACL}, 2019{\natexlab{b}}.

\bibitem[Bender and Koller(2020)]{bender-koller-2020-climbing}
Emily~M. Bender and Alexander Koller.
\newblock {Climbing towards {NLU}: {On} Meaning, Form, and Understanding in the
  Age of Data}.
\newblock In \emph{ACL}, 2020.

\bibitem[Bender et~al.(2021)Bender, Gebru, McMillan-Major, and
  Shmitchell]{Bender2021OnTD}
Emily~M. Bender, Timnit Gebru, Angelina McMillan-Major, and Shmargaret
  Shmitchell.
\newblock {On the Dangers of Stochastic Parrots: Can Language Models Be Too
  Big?}
\newblock \emph{Proceedings of the 2021 ACM Conference on Fairness,
  Accountability, and Transparency}, 2021.

\bibitem[Ettinger et~al.(2016)Ettinger, Elgohary, and
  Resnik]{ettinger-etal-2016-probing}
Allyson Ettinger, Ahmed Elgohary, and Philip Resnik.
\newblock Probing for semantic evidence of composition by means of simple
  classification tasks.
\newblock In \emph{{1st Workshop on Evaluating Vector-Space Representations for
  NLP}}, 2016.

\bibitem[Adi et~al.(2017)Adi, Kermany, Belinkov, Lavi, and
  Goldberg]{adi17probing}
Yossi Adi, Einat Kermany, Yonatan Belinkov, Ofer Lavi, and Yoav Goldberg.
\newblock {Fine-grained Analysis of Sentence Embeddings Using Auxiliary
  Prediction Tasks}.
\newblock In \emph{ICLR}, 2017.

\bibitem[Belinkov et~al.(2017)Belinkov, Durrani, Dalvi, Sajjad, and
  Glass]{belinkov-etal-2017-neural}
Yonatan Belinkov, Nadir Durrani, Fahim Dalvi, Hassan Sajjad, and James Glass.
\newblock What do neural machine translation models learn about morphology?
\newblock In \emph{Proceedings of the 55th Annual Meeting of the Association
  for Computational Linguistics (Volume 1: Long Papers)}, pages 861--872,
  Vancouver, Canada, July 2017. Association for Computational Linguistics.
\newblock \doi{10.18653/v1/P17-1080}.
\newblock URL \url{https://aclanthology.org/P17-1080}.

\bibitem[Liu et~al.(2019{\natexlab{c}})Liu, Gardner, Belinkov, Peters, and
  Smith]{liu-etal-2019-linguistic}
Nelson~F. Liu, Matt Gardner, Yonatan Belinkov, Matthew~E. Peters, and Noah~A.
  Smith.
\newblock Linguistic knowledge and transferability of contextual
  representations.
\newblock In \emph{Proceedings of the 2019 Conference of the North {A}merican
  Chapter of the Association for Computational Linguistics: Human Language
  Technologies, Volume 1 (Long and Short Papers)}, pages 1073--1094,
  Minneapolis, Minnesota, June 2019{\natexlab{c}}. Association for
  Computational Linguistics.
\newblock \doi{10.18653/v1/N19-1112}.
\newblock URL \url{https://aclanthology.org/N19-1112}.

\bibitem[Tenney et~al.(2019{\natexlab{b}})Tenney, Das, and
  Pavlick]{tenney-etal-2019-bert}
Ian Tenney, Dipanjan Das, and Ellie Pavlick.
\newblock {BERT} rediscovers the classical {NLP} pipeline.
\newblock In \emph{Proceedings of the 57th Annual Meeting of the Association
  for Computational Linguistics}, pages 4593--4601, Florence, Italy, July
  2019{\natexlab{b}}. Association for Computational Linguistics.
\newblock \doi{10.18653/v1/P19-1452}.
\newblock URL \url{https://aclanthology.org/P19-1452}.

\bibitem[Hewitt and Liang(2019)]{hewitt-liang-2019-designing}
John Hewitt and Percy Liang.
\newblock {Designing and Interpreting Probes with Control Tasks}.
\newblock In \emph{EMNLP-IJCNLP}, 2019.

\bibitem[Pimentel et~al.(2020)Pimentel, Valvoda, Hall~Maudslay, Zmigrod,
  Williams, and Cotterell]{pimentel-etal-2020-information}
Tiago Pimentel, Josef Valvoda, Rowan Hall~Maudslay, Ran Zmigrod, Adina
  Williams, and Ryan Cotterell.
\newblock {Information-Theoretic Probing for Linguistic Structure}.
\newblock In \emph{{ACL}}, 2020.

\bibitem[Belinkov(2022)]{belinkov2022probing}
Yonatan Belinkov.
\newblock {Probing Classifiers: Promises, Shortcomings, and Advances}.
\newblock \emph{Computational Linguistics}, 48\penalty0 (1):\penalty0 207--219,
  04 2022.
\newblock ISSN 0891-2017.
\newblock \doi{10.1162/coli_a_00422}.
\newblock URL \url{https://doi.org/10.1162/coli\_a\_00422}.

\bibitem[Goldberg(2019)]{goldberg2019assessing}
Yoav Goldberg.
\newblock Assessing bert's syntactic abilities.
\newblock \emph{CoRR}, abs/1901.05287, 2019.
\newblock URL \url{http://arxiv.org/abs/1901.05287}.

\bibitem[Ettinger(2020{\natexlab{a}})]{ettinger-2020-bert}
Allyson Ettinger.
\newblock What {BERT} is not: Lessons from a new suite of psycholinguistic
  diagnostics for language models.
\newblock \emph{Transactions of the Association for Computational Linguistics},
  8:\penalty0 34--48, 2020{\natexlab{a}}.
\newblock \doi{10.1162/tacl_a_00298}.
\newblock URL \url{https://aclanthology.org/2020.tacl-1.3}.

\bibitem[Gulordava et~al.(2018)Gulordava, Bojanowski, Grave, Linzen, and
  Baroni]{gulordava-etal-2018-colorless}
Kristina Gulordava, Piotr Bojanowski, Edouard Grave, Tal Linzen, and Marco
  Baroni.
\newblock Colorless green recurrent networks dream hierarchically.
\newblock In \emph{Proceedings of the 2018 Conference of the North {A}merican
  Chapter of the Association for Computational Linguistics: Human Language
  Technologies, Volume 1 (Long Papers)}, pages 1195--1205, New Orleans,
  Louisiana, June 2018. Association for Computational Linguistics.
\newblock \doi{10.18653/v1/N18-1108}.
\newblock URL \url{https://aclanthology.org/N18-1108}.

\bibitem[Linzen et~al.(2016)Linzen, Dupoux, and
  Goldberg]{linzen-etal-2016-assessing}
Tal Linzen, Emmanuel Dupoux, and Yoav Goldberg.
\newblock Assessing the ability of {LSTM}s to learn syntax-sensitive
  dependencies.
\newblock \emph{Transactions of the Association for Computational Linguistics},
  4:\penalty0 521--535, 2016.
\newblock \doi{10.1162/tacl_a_00115}.
\newblock URL \url{https://aclanthology.org/Q16-1037}.

\bibitem[Marvin and Linzen(2018)]{marvin-linzen-2018-targeted}
Rebecca Marvin and Tal Linzen.
\newblock Targeted syntactic evaluation of language models.
\newblock In \emph{Proceedings of the 2018 Conference on Empirical Methods in
  Natural Language Processing}, pages 1192--1202, Brussels, Belgium,
  October-November 2018. Association for Computational Linguistics.
\newblock \doi{10.18653/v1/D18-1151}.
\newblock URL \url{https://aclanthology.org/D18-1151}.

\bibitem[Fischler et~al.(1983)Fischler, Bloom, Childers, Roukos, and
  Perry]{Fischler1983BrainPR}
Ira Fischler, Paul~Alexander Bloom, Donald~G. Childers, Salim Roukos, and
  Nathan~W. Perry.
\newblock Brain potentials related to stages of sentence verification.
\newblock \emph{Psychophysiology}, 20 4:\penalty0 400--9, 1983.

\bibitem[Chow et~al.(2015)Chow, Smith, Lau, and Phillips]{chow2015bag}
Wing~Yee Chow, Cybelle Smith, Ellen Lau, and Colin Phillips.
\newblock A “bag-of-arguments” mechanism for initial verb predictions.
\newblock \emph{Language, Cognition and Neuroscience}, 31:\penalty0 1--20, 09
  2015.
\newblock \doi{10.1080/23273798.2015.1066832}.

\bibitem[Lialin et~al.(2022)Lialin, Zhao, Shivagunde, and
  Rumshisky]{lialin-etal-2022-life}
Vladislav Lialin, Kevin Zhao, Namrata Shivagunde, and Anna Rumshisky.
\newblock Life after {BERT}: What do other muppets understand about language?
\newblock In \emph{Proceedings of the 60th Annual Meeting of the Association
  for Computational Linguistics (Volume 1: Long Papers)}, pages 3180--3193,
  Dublin, Ireland, May 2022. Association for Computational Linguistics.
\newblock \doi{10.18653/v1/2022.acl-long.227}.
\newblock URL \url{https://aclanthology.org/2022.acl-long.227}.

\bibitem[Liu et~al.(2021)Liu, Yuan, Fu, Jiang, Hayashi, and
  Neubig]{Liu2021PretrainPA}
Pengfei Liu, Weizhe Yuan, Jinlan Fu, Zhengbao Jiang, Hiroaki Hayashi, and
  Graham Neubig.
\newblock Pre-train, prompt, and predict: A systematic survey of prompting
  methods in natural language processing.
\newblock \emph{ArXiv}, abs/2107.13586, 2021.

\bibitem[Toshniwal et~al.(2022)Toshniwal, Wiseman, Livescu, and
  Gimpel]{toshniwal-etal-2022-chess}
Shubham Toshniwal, Sam Wiseman, Karen Livescu, and Kevin Gimpel.
\newblock {Chess as a Testbed for Language Model State Tracking}.
\newblock In \emph{Proceedings of the Thirty-Sixth AAAI Conference on
  Artificial Intelligence (AAAI 2022)}, 2022.

\bibitem[Li et~al.(2022)Li, Cotterell, and Sachan]{li-etal-2022-probing-via}
Jiaoda Li, Ryan Cotterell, and Mrinmaya Sachan.
\newblock Probing via prompting.
\newblock In \emph{Proceedings of the 2022 Conference of the North American
  Chapter of the Association for Computational Linguistics: Human Language
  Technologies}, pages 1144--1157, Seattle, United States, July 2022.
  Association for Computational Linguistics.
\newblock \doi{10.18653/v1/2022.naacl-main.84}.
\newblock URL \url{https://aclanthology.org/2022.naacl-main.84}.

\bibitem[Graves et~al.(2016)Graves, Wayne, Reynolds, Harley, Danihelka,
  Grabska-Barwi{\'n}ska, Colmenarejo, Grefenstette, Ramalho, Agapiou, Badia,
  Hermann, Zwols, Ostrovski, Cain, King, Summerfield, Blunsom, Kavukcuoglu, and
  Hassabis]{graves2016hybrid}
Alex Graves, Greg Wayne, Malcolm Reynolds, Tim Harley, Ivo Danihelka, Agnieszka
  Grabska-Barwi{\'n}ska, Sergio~G{\'o}mez Colmenarejo, Edward Grefenstette,
  Tiago Ramalho, John Agapiou, Adri{\`a}Puigdom{\`e}nech Badia, Karl~Moritz
  Hermann, Yori Zwols, Georg Ostrovski, Adam Cain, Helen King, Christopher
  Summerfield, Phil Blunsom, Koray Kavukcuoglu, and Demis Hassabis.
\newblock Hybrid computing using a neural network with dynamic external memory.
\newblock \emph{Nature}, 538\penalty0 (7626):\penalty0 471--476, 2016.

\bibitem[Turing(1950)]{turing1950mind}
A.~M. Turing.
\newblock Computing machinery and intelligence.
\newblock \emph{Mind}, 59\penalty0 (236):\penalty0 433--460, 1950.
\newblock ISSN 00264423, 14602113.
\newblock URL \url{http://www.jstor.org/stable/2251299}.

\bibitem[Baddeley(1986)]{baddeley1986}
Alan Baddeley.
\newblock \emph{Working {M}emory}.
\newblock Oxford University Press, 1986.

\bibitem[Nematzadeh et~al.(2020)Nematzadeh, Ruder, and
  Yogatama]{Nematzadeh2020ONMI}
Aida Nematzadeh, Sebastian Ruder, and Dani Yogatama.
\newblock On memory in human and artificial language processing systems.
\newblock In \emph{Workshop on Bridging AI and Cognitive Science}, 2020.

\bibitem[Hochreiter and Schmidhuber(1997)]{hochreiter1997long}
Sepp Hochreiter and J{\"u}rgen Schmidhuber.
\newblock Long short-term memory.
\newblock \emph{Neural computation}, 9, 1997.

\bibitem[Weston et~al.(2015{\natexlab{b}})Weston, Chopra, and
  Bordes]{weston2014memory}
Jason Weston, Sumit Chopra, and Antoine Bordes.
\newblock {Memory Networks}.
\newblock In \emph{ICLR}, 2015{\natexlab{b}}.

\bibitem[Sukhbaatar et~al.(2015)Sukhbaatar, Szlam, Weston, and
  Fergus]{sukhbaatar-15}
Sainbayar Sukhbaatar, Arthur Szlam, Jason Weston, and Rob Fergus.
\newblock End-to-end memory networks.
\newblock In \emph{NeurIPS}, 2015.

\bibitem[Miller et~al.(2016)Miller, Fisch, Dodge, Karimi, Bordes, and
  Weston]{miller-etal-2016-key}
Alexander Miller, Adam Fisch, Jesse Dodge, Amir-Hossein Karimi, Antoine Bordes,
  and Jason Weston.
\newblock Key-{V}alue {M}emory {N}etworks for {D}irectly {R}eading {D}ocuments.
\newblock In \emph{{EMNLP}}, 2016.

\bibitem[Rae et~al.(2016)Rae, Hunt, Danihelka, Harley, Senior, Wayne, Graves,
  and Lillicrap]{rae2016scaling}
Jack~W. Rae, Jonathan~J. Hunt, Ivo Danihelka, Timothy Harley, Andrew~W. Senior,
  Gregory Wayne, Alex Graves, and Tim Lillicrap.
\newblock {Scaling Memory-Augmented Neural Networks with Sparse Reads and
  Writes}.
\newblock In \emph{NeurIPS}, 2016.

\bibitem[Kumar et~al.(2016)Kumar, Irsoy, Ondruska, Iyyer, Bradbury, Gulrajani,
  Zhong, Paulus, and Socher]{kumar2016ask}
Ankit Kumar, Ozan Irsoy, Peter Ondruska, Mohit Iyyer, James Bradbury, Ishaan
  Gulrajani, Victor Zhong, Romain Paulus, and Richard Socher.
\newblock Ask me anything: {D}ynamic memory networks for natural language
  processing.
\newblock In \emph{ICML}, 2016.

\bibitem[Liu et~al.(2018{\natexlab{a}})Liu, Cohn, and
  Baldwin]{liu-etal-2018-recurrent}
Fei Liu, Trevor Cohn, and Timothy Baldwin.
\newblock Recurrent {E}ntity {N}etworks with {D}elayed {M}emory {U}pdate for
  {T}argeted {A}spect-{B}ased {S}entiment {A}nalysis.
\newblock In \emph{{NAACL-HLT}}, 2018{\natexlab{a}}.

\bibitem[Maruf and Haffari(2018)]{maruf-haffari-2018-document}
Sameen Maruf and Gholamreza Haffari.
\newblock Document {C}ontext {N}eural {M}achine {T}ranslation with {M}emory
  {N}etworks.
\newblock In \emph{{ACL}}, 2018.

\bibitem[Liu et~al.(2018{\natexlab{b}})Liu, Cohn, and
  Baldwin]{liu-etal-2018-narrative}
Fei Liu, Trevor Cohn, and Timothy Baldwin.
\newblock Narrative {M}odeling with {M}emory {C}hains and {S}emantic
  {S}upervision.
\newblock In \emph{{ACL}}, 2018{\natexlab{b}}.

\bibitem[Perez and Liu(2017)]{perez-liu-2017-dialog}
Julien Perez and Fei Liu.
\newblock Dialog state tracking, a machine reading approach using {M}emory
  {N}etwork.
\newblock In \emph{{EACL}}, 2017.

\bibitem[Rashkin et~al.(2020)Rashkin, Celikyilmaz, Choi, and
  Gao]{rashkin-etal-2020-plotmachines}
Hannah Rashkin, Asli Celikyilmaz, Yejin Choi, and Jianfeng Gao.
\newblock {P}lot{M}achines: Outline-conditioned generation with dynamic plot
  state tracking.
\newblock In \emph{Proceedings of the 2020 Conference on Empirical Methods in
  Natural Language Processing (EMNLP)}, pages 4274--4295, Online, November
  2020. Association for Computational Linguistics.
\newblock \doi{10.18653/v1/2020.emnlp-main.349}.
\newblock URL \url{https://aclanthology.org/2020.emnlp-main.349}.

\bibitem[Raffel et~al.(2020)Raffel, Shazeer, Roberts, Lee, Narang, Matena,
  Zhou, Li, and Liu]{raffel-etal-2020-exploring}
Colin Raffel, Noam Shazeer, Adam Roberts, Katherine Lee, Sharan Narang, Michael
  Matena, Yanqi Zhou, Wei Li, and Peter~J. Liu.
\newblock Exploring the limits of transfer learning with a unified text-to-text
  transformer.
\newblock \emph{Journal of Machine Learning Research}, 21\penalty0
  (140):\penalty0 1--67, 2020.
\newblock URL \url{http://jmlr.org/papers/v21/20-074.html}.

\bibitem[Devlin et~al.(2019{\natexlab{b}})Devlin, Chang, Lee, and
  Toutanova]{devlin2019bert}
Jacob Devlin, Ming{-}Wei Chang, Kenton Lee, and Kristina Toutanova.
\newblock {BERT:} {P}re-training of {D}eep {B}idirectional {T}ransformers for
  {L}anguage {U}nderstanding.
\newblock In \emph{NAACL-HLT}, 2019{\natexlab{b}}.

\bibitem[Lewis et~al.(2020{\natexlab{b}})Lewis, Liu, Goyal, Ghazvininejad,
  Mohamed, Levy, Stoyanov, and Zettlemoyer]{lewis-etal-2020-bart}
Mike Lewis, Yinhan Liu, Naman Goyal, Marjan Ghazvininejad, Abdelrahman Mohamed,
  Omer Levy, Veselin Stoyanov, and Luke Zettlemoyer.
\newblock {{BART}: Denoising Sequence-to-Sequence Pre-training for Natural
  Language Generation, Translation, and Comprehension}.
\newblock In \emph{ACL}, 2020{\natexlab{b}}.

\bibitem[Qiu et~al.(2020)Qiu, Sun, Xu, Shao, Dai, and Huang]{qiu2020ptmsurvey}
Xipeng Qiu, Tianxiang Sun, Yige Xu, Yunfan Shao, Ning Dai, and Xuanjing Huang.
\newblock Pre-trained models for natural language processing: {A} survey.
\newblock \emph{CoRR}, abs/2003.08271, 2020.
\newblock URL \url{https://arxiv.org/abs/2003.08271}.

\bibitem[Vaswani et~al.(2017)Vaswani, Shazeer, Parmar, Uszkoreit, Jones, Gomez,
  Kaiser, and Polosukhin]{vaswani2017attention}
Ashish Vaswani, Noam Shazeer, Niki Parmar, Jakob Uszkoreit, Llion Jones,
  Aidan~N Gomez, \L~ukasz Kaiser, and Illia Polosukhin.
\newblock {Attention is All you Need}.
\newblock In \emph{NeurIPS}, 2017.

\bibitem[Peters et~al.(2018)Peters, Neumann, Iyyer, Gardner, Clark, Lee, and
  Zettlemoyer]{peters-etal-2018-deep}
Matthew~E. Peters, Mark Neumann, Mohit Iyyer, Matt Gardner, Christopher Clark,
  Kenton Lee, and Luke Zettlemoyer.
\newblock Deep contextualized word representations.
\newblock In \emph{Proceedings of the 2018 Conference of the North {A}merican
  Chapter of the Association for Computational Linguistics: Human Language
  Technologies, Volume 1 (Long Papers)}, pages 2227--2237, New Orleans,
  Louisiana, June 2018. Association for Computational Linguistics.
\newblock \doi{10.18653/v1/N18-1202}.
\newblock URL \url{https://aclanthology.org/N18-1202}.

\bibitem[McCann et~al.(2017)McCann, Bradbury, Xiong, and
  Socher]{mccann2017cove}
Bryan McCann, James Bradbury, Caiming Xiong, and Richard Socher.
\newblock Learned in translation: Contextualized word vectors.
\newblock In \emph{Proceedings of the 31st International Conference on Neural
  Information Processing Systems}, NIPS'17, page 6297–6308, Red Hook, NY,
  USA, 2017. Curran Associates Inc.
\newblock ISBN 9781510860964.

\bibitem[Liu et~al.(2019{\natexlab{d}})Liu, Ott, Goyal, Du, Joshi, Chen, Levy,
  Lewis, Zettlemoyer, and Stoyanov]{liu-etal-2019-roberta}
Yinhan Liu, Myle Ott, Naman Goyal, Jingfei Du, Mandar Joshi, Danqi Chen, Omer
  Levy, Mike Lewis, Luke Zettlemoyer, and Veselin Stoyanov.
\newblock Roberta: {A} robustly optimized {BERT} pretraining approach.
\newblock \emph{CoRR}, abs/1907.11692, 2019{\natexlab{d}}.
\newblock URL \url{http://arxiv.org/abs/1907.11692}.

\bibitem[Sun et~al.(2019)Sun, Qiu, Xu, and Huang]{sun2019fine}
Chi Sun, Xipeng Qiu, Yige Xu, and Xuanjing Huang.
\newblock How to fine-tune {BERT} for text classification?
\newblock In \emph{China National Conference on Chinese Computational
  Linguistics}, pages 194--206. Springer, 2019.

\bibitem[Radford and Sutskever(2018)]{radford2018improving}
Alec Radford and Ilya Sutskever.
\newblock {Improving Language Understanding by Generative Pre-Training}.
\newblock In \emph{arXiv}, 2018.

\bibitem[Lester et~al.(2021)Lester, Al-Rfou, and
  Constant]{lester-etal-2021-power}
Brian Lester, Rami Al-Rfou, and Noah Constant.
\newblock The power of scale for parameter-efficient prompt tuning.
\newblock In \emph{Proceedings of the 2021 Conference on Empirical Methods in
  Natural Language Processing}, pages 3045--3059, Online and Punta Cana,
  Dominican Republic, November 2021. Association for Computational Linguistics.
\newblock \doi{10.18653/v1/2021.emnlp-main.243}.
\newblock URL \url{https://aclanthology.org/2021.emnlp-main.243}.

\bibitem[Li and Liang(2021)]{li-liang-2021-prefix}
Xiang~Lisa Li and Percy Liang.
\newblock Prefix-tuning: Optimizing continuous prompts for generation.
\newblock In \emph{Proceedings of the 59th Annual Meeting of the Association
  for Computational Linguistics and the 11th International Joint Conference on
  Natural Language Processing (Volume 1: Long Papers)}, pages 4582--4597,
  Online, August 2021. Association for Computational Linguistics.
\newblock \doi{10.18653/v1/2021.acl-long.353}.
\newblock URL \url{https://aclanthology.org/2021.acl-long.353}.

\bibitem[Toshniwal et~al.(2020{\natexlab{a}})Toshniwal, Ettinger, Gimpel, and
  Livescu]{toshniwal2020petra}
Shubham Toshniwal, Allyson Ettinger, Kevin Gimpel, and Karen Livescu.
\newblock {{PeTra}: A Sparsely Supervised Memory Model for People Tracking}.
\newblock In \emph{ACL}, 2020{\natexlab{a}}.

\bibitem[Cho et~al.(2014)Cho, van Merri{\"e}nboer, Gulcehre, Bahdanau,
  Bougares, Schwenk, and Bengio]{cho2014learning}
Kyunghyun Cho, Bart van Merri{\"e}nboer, Caglar Gulcehre, Dzmitry Bahdanau,
  Fethi Bougares, Holger Schwenk, and Yoshua Bengio.
\newblock Learning {P}hrase {R}epresentations using {RNN}
  {E}ncoder{--}{D}ecoder for {S}tatistical {M}achine {T}ranslation.
\newblock In \emph{{EMNLP}}, 2014.

\bibitem[Jang et~al.(2017)Jang, Gu, and Poole]{jang2017categorical}
Eric Jang, Shixiang Gu, and Ben Poole.
\newblock Categorical {R}eparameterization with {G}umbel-{S}oftmax.
\newblock In \emph{{ICLR}}, 2017.

\bibitem[Kingma and Ba(2015)]{Kingma2015AdamAM}
Diederik~P. Kingma and Jimmy Ba.
\newblock Adam: {A} {M}ethod for {S}tochastic {O}ptimization.
\newblock In \emph{ICLR}, 2015.

\bibitem[Toshniwal et~al.(2020{\natexlab{b}})Toshniwal, Wiseman, Ettinger,
  Livescu, and Gimpel]{toshniwal-etal-2020-learning}
Shubham Toshniwal, Sam Wiseman, Allyson Ettinger, Karen Livescu, and Kevin
  Gimpel.
\newblock {Learning to Ignore: Long Document Coreference with Bounded Memory
  Neural Networks}.
\newblock In \emph{EMNLP}, 2020{\natexlab{b}}.

\bibitem[Toshniwal et~al.(2021)Toshniwal, Xia, Wiseman, Livescu, and
  Gimpel]{toshniwal-etal-2021-generalization}
Shubham Toshniwal, Patrick Xia, Sam Wiseman, Karen Livescu, and Kevin Gimpel.
\newblock On generalization in coreference resolution.
\newblock In \emph{Proceedings of the Fourth Workshop on Computational Models
  of Reference, Anaphora and Coreference}, pages 111--120, Punta Cana,
  Dominican Republic, November 2021. Association for Computational Linguistics.
\newblock \doi{10.18653/v1/2021.crac-1.12}.
\newblock URL \url{https://aclanthology.org/2021.crac-1.12}.

\bibitem[Tanenhaus et~al.(1995)Tanenhaus, Spivey-Knowlton, Eberhard, and
  Sedivy]{Tanenhaus1632}
MK~Tanenhaus, MJ~Spivey-Knowlton, KM~Eberhard, and JC~Sedivy.
\newblock Integration of visual and linguistic information in spoken language
  comprehension.
\newblock \emph{Science}, 268\penalty0 (5217), 1995.

\bibitem[Keller(2010)]{keller2010cognitively}
Frank Keller.
\newblock Cognitively {P}lausible {M}odels of {H}uman {L}anguage {P}rocessing.
\newblock In \emph{{ACL}}, 2010.

\bibitem[Dai et~al.(2019)Dai, Yang, Yang, Carbonell, Le, and
  Salakhutdinov]{dai-etal-2019-transformer}
Zihang Dai, Zhilin Yang, Yiming Yang, Jaime Carbonell, Quoc Le, and Ruslan
  Salakhutdinov.
\newblock {Transformer-{XL}: Attentive Language Models beyond a Fixed-Length
  Context}.
\newblock In \emph{Proceedings of the 57th Annual Meeting of the Association
  for Computational Linguistics}, pages 2978--2988, Florence, Italy, July 2019.
  Association for Computational Linguistics.
\newblock \doi{10.18653/v1/P19-1285}.
\newblock URL \url{https://aclanthology.org/P19-1285}.

\bibitem[Toshniwal et~al.(2020{\natexlab{c}})Toshniwal, Shi, Shi, Gao, Livescu,
  and Gimpel]{toshniwal2020cross}
Shubham Toshniwal, Haoyue Shi, Bowen Shi, Lingyu Gao, Karen Livescu, and Kevin
  Gimpel.
\newblock {A Cross-Task Analysis of Text Span Representations}.
\newblock In \emph{RepL4NLP}, 2020{\natexlab{c}}.

\bibitem[Wu and Gardner(2021)]{wu2020understanding}
Zhaofeng Wu and Matt Gardner.
\newblock {Understanding Mention Detector-Linker Interaction for Neural
  Coreference Resolution}.
\newblock In \emph{CRAC (EMNLP)}, 2021.

\bibitem[Santoro et~al.(2016)Santoro, Bartunov, Botvinick, Wierstra, and
  Lillicrap]{santoro2016one}
Adam Santoro, Sergey Bartunov, Matthew Botvinick, Daan Wierstra, and Timothy~P.
  Lillicrap.
\newblock {One-shot Learning with Memory-Augmented Neural Networks}.
\newblock In \emph{ICML}, 2016.

\bibitem[Wu et~al.(2020{\natexlab{b}})Wu, Wang, Yuan, Wu, and
  Li]{wu-etal-2020-corefqa}
Wei Wu, Fei Wang, Arianna Yuan, Fei Wu, and Jiwei Li.
\newblock {{C}oref{QA}: Coreference Resolution as Query-based Span Prediction}.
\newblock In \emph{ACL}, 2020{\natexlab{b}}.

\bibitem[Xia and Van~Durme(2021)]{xia-van-durme-2021-moving}
Patrick Xia and Benjamin Van~Durme.
\newblock {Moving on from {O}nto{N}otes: Coreference Resolution Model
  Transfer}.
\newblock In \emph{EMNLP}, 2021.

\bibitem[Thirukovalluru et~al.(2021)Thirukovalluru, Monath, Shridhar, Zaheer,
  Sachan, and McCallum]{thirukovalluru-etal-2021-scaling}
Raghuveer Thirukovalluru, Nicholas Monath, Kumar Shridhar, Manzil Zaheer,
  Mrinmaya Sachan, and Andrew McCallum.
\newblock {Scaling Within Document Coreference to Long Texts}.
\newblock In \emph{Findings of the ACL-IJCNLP}, pages 3921--3931, Online,
  August 2021. Association for Computational Linguistics.
\newblock \doi{10.18653/v1/2021.findings-acl.343}.
\newblock URL \url{https://aclanthology.org/2021.findings-acl.343}.

\bibitem[Kummerfeld and Klein(2013)]{kummerfeld2013error}
Jonathan~K. Kummerfeld and Dan Klein.
\newblock {Error-Driven Analysis of Challenges in Coreference Resolution}.
\newblock In \emph{EMNLP}, 2013.

\bibitem[Li et~al.(2020)Li, Zareian, Lin, Pan, Whitehead, Chen, Wu, Ji, Chang,
  Voss, Napierski, and Freedman]{li-etal-2020-gaia}
Manling Li, Alireza Zareian, Ying Lin, Xiaoman Pan, Spencer Whitehead, Brian
  Chen, Bo~Wu, Heng Ji, Shih-Fu Chang, Clare Voss, Daniel Napierski, and
  Marjorie Freedman.
\newblock {{GAIA}: A Fine-grained Multimedia Knowledge Extraction System}.
\newblock In \emph{ACL: System Demonstrations}, 2020.

\bibitem[Yang et~al.(2012)Yang, Mao, Xiang, Tsang, Chai, and
  Chieu]{yang-etal-2012-domain}
Jian~Bo Yang, Qi~Mao, Qiao~Liang Xiang, Ivor Wai-Hung Tsang, Kian Ming~Adam
  Chai, and Hai~Leong Chieu.
\newblock {Domain Adaptation for Coreference Resolution: An Adaptive Ensemble
  Approach}.
\newblock In \emph{EMNLP}, 2012.

\bibitem[Zhao and Ng(2014)]{zhao-ng-2014-domain}
Shanheng Zhao and Hwee~Tou Ng.
\newblock {Domain Adaptation with Active Learning for Coreference Resolution}.
\newblock In \emph{Proceedings of the 5th International Workshop on Health Text
  Mining and Information Analysis (Louhi)}, 2014.

\bibitem[Poot and van Cranenburgh(2020)]{poot-van-cranenburgh-2020-benchmark}
Corb{\`e}n Poot and Andreas van Cranenburgh.
\newblock {A Benchmark of Rule-Based and Neural Coreference Resolution in
  {D}utch Novels and News}.
\newblock In \emph{CRAC}, 2020.

\bibitem[Akta{\c{s}} et~al.(2020)Akta{\c{s}}, Solopova, Kohnert, and
  Stede]{aktas-etal-2020-adapting}
Berfin Akta{\c{s}}, Veronika Solopova, Annalena Kohnert, and Manfred Stede.
\newblock {Adapting Coreference Resolution to {T}witter Conversations}.
\newblock In \emph{Findings of EMNLP}, 2020.

\bibitem[Moosavi(2020)]{moosavi-thesis}
Nafise~Sadat Moosavi.
\newblock \emph{Robustness in Coreference Resolution}.
\newblock PhD thesis, Heidelberg University, 2020.

\bibitem[Wiseman et~al.(2016{\natexlab{b}})Wiseman, Rush, and
  Shieber]{wiseman-etal-2016-antecedent}
Sam Wiseman, Alexander~M. Rush, and Stuart Shieber.
\newblock {Antecedent Prediction Without a Pipeline}.
\newblock In \emph{Proceedings of the Workshop on Coreference Resolution Beyond
  {O}nto{N}otes ({CORBON} 2016)}, 2016{\natexlab{b}}.

\bibitem[Poesio et~al.(2018)Poesio, Grishina, Kolhatkar, Moosavi, Roesiger,
  Roussel, Simonjetz, Uma, Uryupina, Yu, and
  Zinsmeister]{poesio-etal-2018-anaphora}
Massimo Poesio, Yulia Grishina, Varada Kolhatkar, Nafise Moosavi, Ina Roesiger,
  Adam Roussel, Fabian Simonjetz, Alexandra Uma, Olga Uryupina, Juntao Yu, and
  Heike Zinsmeister.
\newblock {Anaphora Resolution with the {ARRAU} Corpus}.
\newblock In \emph{Proceedings of the First Workshop on Computational Models of
  Reference, Anaphora and Coreference}, 2018.

\bibitem[Zhu et~al.(2021)Zhu, Pradhan, and Zeldes]{zhu-etal-2021-ontogum}
Yilun Zhu, Sameer Pradhan, and Amir Zeldes.
\newblock {{O}nto{GUM}: Evaluating Contextualized {SOTA} Coreference Resolution
  on 12 More Genres}.
\newblock In \emph{ACL}, 2021.

\bibitem[Zhou and Choi(2018)]{zhou-choi-2018-exist}
Ethan Zhou and Jinho~D. Choi.
\newblock {They Exist! Introducing Plural Mentions to Coreference Resolution
  and Entity Linking}.
\newblock In \emph{Proceedings of the 27th International Conference on
  Computational Linguistics}, 2018.

\bibitem[Guha et~al.(2015)Guha, Iyyer, Bouman, and
  Boyd-Graber]{guha-etal-2015-removing}
Anupam Guha, Mohit Iyyer, Danny Bouman, and Jordan Boyd-Graber.
\newblock {Removing the Training Wheels: A Coreference Dataset that Entertains
  Humans and Challenges Computers}.
\newblock In \emph{NAACL-HLT}, 2015.

\bibitem[Levesque et~al.(2012)Levesque, Davis, and
  Morgenstern]{levasque2012wino}
Hector~J. Levesque, Ernest Davis, and Leora Morgenstern.
\newblock {The Winograd Schema Challenge}.
\newblock In \emph{Proceedings of the International Conference on Knowledge
  Representation and Reasoning}, 2012.

\bibitem[K{\"u}bler and Zhekova(2011)]{kubler-zhekova-2011-singletons}
Sandra K{\"u}bler and Desislava Zhekova.
\newblock {Singletons and Coreference Resolution Evaluation}.
\newblock In \emph{Proceedings of the International Conference Recent Advances
  in Natural Language Processing}, 2011.

\bibitem[Stymne et~al.(2018)Stymne, de~Lhoneux, Smith, and
  Nivre]{stymne-etal-2018-parser}
Sara Stymne, Miryam de~Lhoneux, Aaron Smith, and Joakim Nivre.
\newblock {Parser Training with Heterogeneous Treebanks}.
\newblock In \emph{ACL}, 2018.

\bibitem[Kobus et~al.(2017)Kobus, Crego, and Senellart]{kobus-etal-2017-domain}
Catherine Kobus, Josep Crego, and Jean Senellart.
\newblock {Domain Control for Neural Machine Translation}.
\newblock In \emph{RANLP}, 2017.

\bibitem[Tan et~al.(2019)Tan, Chen, He, Xia, Qin, and
  Liu]{tan-etal-2019-multilingual}
Xu~Tan, Jiale Chen, Di~He, Yingce Xia, Tao Qin, and Tie-Yan Liu.
\newblock {Multilingual Neural Machine Translation with Language Clustering}.
\newblock In \emph{EMNLP-IJCNLP}, 2019.

\bibitem[Aralikatte et~al.(2019)Aralikatte, Lent, Gonzalez, Herschcovich, Qiu,
  Sandholm, Ringaard, and S{\o}gaard]{aralikatte-etal-2019-rewarding}
Rahul Aralikatte, Heather Lent, Ana~Valeria Gonzalez, Daniel Herschcovich, Chen
  Qiu, Anders Sandholm, Michael Ringaard, and Anders S{\o}gaard.
\newblock {Rewarding Coreference Resolvers for Being Consistent with World
  Knowledge}.
\newblock In \emph{EMNLP-IJCNLP}, 2019.

\bibitem[Subramanian and Roth(2019)]{subramanian-roth-2019-improving}
Sanjay Subramanian and Dan Roth.
\newblock Improving generalization in coreference resolution via adversarial
  training.
\newblock In \emph{Proceedings of the Eighth Joint Conference on Lexical and
  Computational Semantics (*{SEM} 2019)}, pages 192--197, Minneapolis,
  Minnesota, June 2019. Association for Computational Linguistics.
\newblock \doi{10.18653/v1/S19-1021}.
\newblock URL \url{https://aclanthology.org/S19-1021}.

\bibitem[Moosavi and Strube(2018)]{moosavi-strube-2018-using}
Nafise~Sadat Moosavi and Michael Strube.
\newblock {Using Linguistic Features to Improve the Generalization Capability
  of Neural Coreference Resolvers}.
\newblock In \emph{{EMNLP}}, 2018.

\bibitem[Recasens et~al.(2013)Recasens, de~Marneffe, and
  Potts]{recasens-etal-2013-life}
Marta Recasens, Marie-Catherine de~Marneffe, and Christopher Potts.
\newblock The life and death of discourse entities: Identifying singleton
  mentions.
\newblock In \emph{Proceedings of the 2013 Conference of the North American
  Chapter of the Association for Computational Linguistics: Human Language
  Technologies}, pages 627--633, 2013.

\bibitem[Yu et~al.(2020)Yu, Uma, and Poesio]{yu-etal-2020-cluster}
Juntao Yu, Alexandra Uma, and Massimo Poesio.
\newblock A cluster ranking model for full anaphora resolution.
\newblock In \emph{Proceedings of the 12th Language Resources and Evaluation
  Conference}, pages 11--20, Marseille, France, May 2020. European Language
  Resources Association.
\newblock ISBN 979-10-95546-34-4.
\newblock URL \url{https://aclanthology.org/2020.lrec-1.2}.

\bibitem[Zhang et~al.(2018)Zhang, Lapata, Wei, and
  Zhou]{zhang-etal-2018-neural}
Xingxing Zhang, Mirella Lapata, Furu Wei, and Ming Zhou.
\newblock Neural latent extractive document summarization.
\newblock In \emph{Proceedings of the 2018 Conference on Empirical Methods in
  Natural Language Processing}, pages 779--784, Brussels, Belgium,
  October-November 2018. Association for Computational Linguistics.
\newblock \doi{10.18653/v1/D18-1088}.
\newblock URL \url{https://aclanthology.org/D18-1088}.

\bibitem[Swayamdipta et~al.(2018)Swayamdipta, Thomson, Lee, Zettlemoyer, Dyer,
  and Smith]{swayamdipta-etal-2018-syntactic}
Swabha Swayamdipta, Sam Thomson, Kenton Lee, Luke Zettlemoyer, Chris Dyer, and
  Noah~A. Smith.
\newblock Syntactic scaffolds for semantic structures.
\newblock In \emph{Proceedings of the 2018 Conference on Empirical Methods in
  Natural Language Processing}, pages 3772--3782, Brussels, Belgium,
  October-November 2018. Association for Computational Linguistics.
\newblock \doi{10.18653/v1/D18-1412}.
\newblock URL \url{https://aclanthology.org/D18-1412}.

\bibitem[Manning et~al.(2008)Manning, Raghavan, and Sch\"{u}tze]{ir-book}
Christopher~D. Manning, Prabhakar Raghavan, and Hinrich Sch\"{u}tze.
\newblock \emph{Introduction to Information Retrieval}.
\newblock Cambridge University Press, 2008.

\bibitem[Fiekas(2012)]{python-chess}
Niklas Fiekas.
\newblock {python-chess: a pure Python chess library}, 2012.
\newblock URL \url{https://python-chess.readthedocs.io/en/latest/index.html}.

\bibitem[Kitaev et~al.(2020)Kitaev, Kaiser, and Levskaya]{kitaev2020reformer}
Nikita Kitaev, Lukasz Kaiser, and Anselm Levskaya.
\newblock {Reformer: The Efficient Transformer}.
\newblock In \emph{ICLR}, 2020.

\bibitem[Choromanski et~al.(2021)Choromanski, Likhosherstov, Dohan, Song, Gane,
  Sarlos, Hawkins, Davis, Mohiuddin, Kaiser, Belanger, Colwell, and
  Weller]{choromanski2021rethinking}
Krzysztof~Marcin Choromanski, Valerii Likhosherstov, David Dohan, Xingyou Song,
  Andreea Gane, Tamas Sarlos, Peter Hawkins, Jared~Quincy Davis, Afroz
  Mohiuddin, Lukasz Kaiser, David~Benjamin Belanger, Lucy~J Colwell, and Adrian
  Weller.
\newblock {Rethinking Attention with Performers}.
\newblock In \emph{ICLR}, 2021.

\bibitem[Kingma and Ba(2014)]{kingma2014adam}
Diederik~P. Kingma and Jimmy Ba.
\newblock Adam: A method for stochastic optimization.
\newblock In \emph{ICLR}, 2014.

\bibitem[Micikevicius et~al.(2018)Micikevicius, Narang, Alben, Diamos, Elsen,
  Garcia, Ginsburg, Houston, Kuchaiev, Venkatesh, and
  Wu]{micikevicius2018mixed}
Paulius Micikevicius, Sharan Narang, Jonah Alben, Gregory Diamos, Erich Elsen,
  David Garcia, Boris Ginsburg, Michael Houston, Oleksii Kuchaiev, Ganesh
  Venkatesh, and Hao Wu.
\newblock {Mixed Precision Training}.
\newblock In \emph{{ICLR}}, 2018.

\bibitem[Falcon(2019)]{falcon2019pytorch}
WA~Falcon.
\newblock {PyTorch Lightning}, 2019.
\newblock URL \url{https://github.com/PyTorchLightning/pytorch-lightning}.

\bibitem[Paszke et~al.(2019)Paszke, Gross, Massa, Lerer, Bradbury, Chanan,
  Killeen, Lin, Gimelshein, Antiga, Desmaison, Kopf, Yang, DeVito, Raison,
  Tejani, Chilamkurthy, Steiner, Fang, Bai, and Chintala]{pytorch}
Adam Paszke, Sam Gross, Francisco Massa, Adam Lerer, James Bradbury, Gregory
  Chanan, Trevor Killeen, Zeming Lin, Natalia Gimelshein, Luca Antiga, Alban
  Desmaison, Andreas Kopf, Edward Yang, Zachary DeVito, Martin Raison, Alykhan
  Tejani, Sasank Chilamkurthy, Benoit Steiner, Lu~Fang, Junjie Bai, and Soumith
  Chintala.
\newblock {PyTorch: An Imperative Style, High-Performance Deep Learning
  Library}.
\newblock In \emph{NeurIPS}, 2019.

\bibitem[Wolf et~al.(2019)Wolf, Debut, Sanh, Chaumond, Delangue, Moi, Cistac,
  Rault, Louf, Funtowicz, Davison, Shleifer, von Platen, Ma, Jernite, Plu, Xu,
  Scao, Gugger, Drame, Lhoest, and Rush]{Wolf2019HuggingFacesTS}
Thomas Wolf, Lysandre Debut, Victor Sanh, Julien Chaumond, Clement Delangue,
  Anthony Moi, Pierric Cistac, Tim Rault, Rémi Louf, Morgan Funtowicz, Joe
  Davison, Sam Shleifer, Patrick von Platen, Clara Ma, Yacine Jernite, Julien
  Plu, Canwen Xu, Teven~Le Scao, Sylvain Gugger, Mariama Drame, Quentin Lhoest,
  and Alexander~M. Rush.
\newblock Huggingface's transformers: State-of-the-art natural language
  processing.
\newblock \emph{ArXiv}, abs/1910.03771, 2019.

\bibitem[Katharopoulos et~al.(2020)Katharopoulos, Vyas, Pappas, and
  Fleuret]{katharopoulos20}
Angelos Katharopoulos, Apoorv Vyas, Nikolaos Pappas, and Francois Fleuret.
\newblock {Transformers are RNNs: Fast Autoregressive Transformers with Linear
  Attention}.
\newblock In \emph{ICML}, 2020.

\bibitem[C\^ot\'e et~al.(2018)C\^ot\'e, K\'ad\'ar, Yuan, Kybartas, Barnes,
  Fine, Moore, Tao, Hausknecht, Asri, Adada, Tay, and
  Trischler]{cote18textworld}
Marc-Alexandre C\^ot\'e, \'Akos K\'ad\'ar, Xingdi Yuan, Ben Kybartas, Tavian
  Barnes, Emery Fine, James Moore, Ruo~Yu Tao, Matthew Hausknecht, Layla~El
  Asri, Mahmoud Adada, Wendy Tay, and Adam Trischler.
\newblock {TextWorld: A Learning Environment for Text-based Games}.
\newblock \emph{CoRR}, abs/1806.11532, 2018.

\bibitem[Hermann et~al.(2017)Hermann, Hill, Green, Wang, Faulkner, Soyer,
  Szepesvari, Czarnecki, Jaderberg, Teplyashin, Wainwright, Apps, Hassabis, and
  Blunsom]{hermann17grounded}
Karl~Moritz Hermann, Felix Hill, Simon Green, Fumin Wang, Ryan Faulkner, Hubert
  Soyer, David Szepesvari, Wojciech~Marian Czarnecki, Max Jaderberg, Denis
  Teplyashin, Marcus Wainwright, Chris Apps, Demis Hassabis, and Phil Blunsom.
\newblock {Grounded Language Learning in a Simulated 3D World}.
\newblock \emph{CoRR}, abs/1706.06551, 2017.

\bibitem[Hill et~al.(2017)Hill, Hermann, Blunsom, and
  Clark]{hill17understanding}
Felix Hill, Karl~Moritz Hermann, Phil Blunsom, and Stephen Clark.
\newblock {Understanding Grounded Language Learning Agents}.
\newblock \emph{CoRR}, abs/1710.09867, 2017.
\newblock URL \url{http://arxiv.org/abs/1710.09867}.

\bibitem[Hermann et~al.(2015)Hermann, Ko\v{c}isk\'{y}, Grefenstette, Espeholt,
  Kay, Suleyman, and Blunsom]{hermann2015cnn}
Karl~Moritz Hermann, Tom\'{a}\v{s} Ko\v{c}isk\'{y}, Edward Grefenstette, Lasse
  Espeholt, Will Kay, Mustafa Suleyman, and Phil Blunsom.
\newblock {Teaching Machines to Read and Comprehend}.
\newblock In \emph{NeurIPS}, 2015.

\bibitem[Hill et~al.(2016)Hill, Bordes, Chopra, and Weston]{hill2016cbt}
Felix Hill, Antoine Bordes, Sumit Chopra, and Jason Weston.
\newblock {The Goldilocks Principle: Reading Children's Books with Explicit
  Memory Representations.}
\newblock In \emph{ICLR}, 2016.

\bibitem[Mostafazadeh et~al.(2016)Mostafazadeh, Chambers, He, Parikh, Batra,
  Vanderwende, Kohli, and Allen]{mostafazadeh-etal-2016-corpus}
Nasrin Mostafazadeh, Nathanael Chambers, Xiaodong He, Devi Parikh, Dhruv Batra,
  Lucy Vanderwende, Pushmeet Kohli, and James Allen.
\newblock {A Corpus and Cloze Evaluation for Deeper Understanding of
  Commonsense Stories}.
\newblock In \emph{NAACL}, 2016.

\bibitem[Ettinger(2020{\natexlab{b}})]{ettinger2020bert}
Allyson Ettinger.
\newblock {What BERT is Not: Lessons from a New Suite of Psycholinguistic
  Diagnostics for Language Models}.
\newblock \emph{TACL}, 8\penalty0 (0), 2020{\natexlab{b}}.

\bibitem[Mostafazadeh et~al.(2017)Mostafazadeh, Roth, Louis, Chambers, and
  Allen]{mostafazadeh-etal-2017-lsdsem}
Nasrin Mostafazadeh, Michael Roth, Annie Louis, Nathanael Chambers, and James
  Allen.
\newblock {{LSDSem 2017 Shared Task: The Story Cloze Test}}.
\newblock In \emph{2nd Workshop on Linking Models of Lexical, Sentential and
  Discourse-level Semantics}, 2017.

\bibitem[Petroni et~al.(2019)Petroni, Rockt{\"a}schel, Riedel, Lewis, Bakhtin,
  Wu, and Miller]{petroni-etal-2019-language}
Fabio Petroni, Tim Rockt{\"a}schel, Sebastian Riedel, Patrick Lewis, Anton
  Bakhtin, Yuxiang Wu, and Alexander Miller.
\newblock {Language Models as Knowledge Bases?}
\newblock In \emph{EMNLP-IJCNLP}, 2019.

\bibitem[David et~al.(2016)David, Netanyahu, and Wolf]{david16deepchess}
Eli David, Nathan~S. Netanyahu, and Lior Wolf.
\newblock {DeepChess: End-to-End Deep Neural Network for Automatic Learning in
  Chess}.
\newblock In \emph{International Conference on Artificial Neural Networks
  (ICANN)}, 2016.

\bibitem[Oshri and Khandwala(2015)]{Oshri2015PredictingMI}
Barak Oshri and Nishith Khandwala.
\newblock {Predicting Moves in Chess using Convolutional Neural Networks},
  2015.

\bibitem[Silver et~al.(2018)Silver, Hubert, Schrittwieser, Antonoglou, Lai,
  Guez, Lanctot, Sifre, Kumaran, Graepel, Lillicrap, Simonyan, and
  Hassabis]{silver18general}
David Silver, Thomas Hubert, Julian Schrittwieser, Ioannis Antonoglou, Matthew
  Lai, Arthur Guez, Marc Lanctot, Laurent Sifre, Dharshan Kumaran, Thore
  Graepel, Timothy Lillicrap, Karen Simonyan, and Demis Hassabis.
\newblock {A general reinforcement learning algorithm that masters chess,
  shogi, and Go through self-play}.
\newblock \emph{Science}, 362\penalty0 (6419), 2018.

\bibitem[Presser and Branwen(2020)]{presser2020chess}
Shawn Presser and Gwern Branwen.
\newblock {A Very Unlikely Chess Game}, 2020.
\newblock URL
  \url{https://slatestarcodex.com/2020/01/06/a-very-unlikely-chess-game/}.

\bibitem[Cheng(2020)]{cheng2020chess}
Ricson Cheng.
\newblock {Transformers Play Chess}, 2020.
\newblock URL \url{https://github.com/ricsonc/transformers-play-chess}.

\bibitem[Noever et~al.(2020)Noever, Ciolino, and Kalin]{noever2020chess}
David Noever, Matt Ciolino, and Josh Kalin.
\newblock The chess transformer: Mastering play using generative language
  models, 2020.

\bibitem[Kaplan et~al.(2020)Kaplan, McCandlish, Henighan, Brown, Chess, Child,
  Gray, Radford, Wu, and Amodei]{kaplan2020scaling}
Jared Kaplan, Sam McCandlish, Tom Henighan, Tom~B. Brown, Benjamin Chess, Rewon
  Child, Scott Gray, Alec Radford, Jeffrey Wu, and Dario Amodei.
\newblock {Scaling Laws for Neural Language Models}, 2020.

\bibitem[Bisk et~al.(2020)Bisk, Holtzman, Thomason, Andreas, Bengio, Chai,
  Lapata, Lazaridou, May, Nisnevich, Pinto, and
  Turian]{bisk-etal-2020-experience}
Yonatan Bisk, Ari Holtzman, Jesse Thomason, Jacob Andreas, Yoshua Bengio, Joyce
  Chai, Mirella Lapata, Angeliki Lazaridou, Jonathan May, Aleksandr Nisnevich,
  Nicolas Pinto, and Joseph Turian.
\newblock Experience grounds language.
\newblock In \emph{Proceedings of the 2020 Conference on Empirical Methods in
  Natural Language Processing (EMNLP)}, pages 8718--8735, Online, November
  2020. Association for Computational Linguistics.
\newblock \doi{10.18653/v1/2020.emnlp-main.703}.
\newblock URL \url{https://aclanthology.org/2020.emnlp-main.703}.

\bibitem[Nye et~al.(2021)Nye, Andreassen, Gur{-}Ari, Michalewski, Austin,
  Bieber, Dohan, Lewkowycz, Bosma, Luan, Sutton, and Odena]{nye2021show}
Maxwell~I. Nye, Anders~Johan Andreassen, Guy Gur{-}Ari, Henryk Michalewski,
  Jacob Austin, David Bieber, David Dohan, Aitor Lewkowycz, Maarten Bosma,
  David Luan, Charles Sutton, and Augustus Odena.
\newblock {Show Your Work: Scratchpads for Intermediate Computation with
  Language Models}.
\newblock \emph{CoRR}, abs/2112.00114, 2021.
\newblock URL \url{https://arxiv.org/abs/2112.00114}.

\bibitem[Lampinen et~al.(2022)Lampinen, Dasgupta, Chan, Matthewson, Tessler,
  Creswell, McClelland, Wang, and Hill]{lampinen2022explanation}
Andrew~K. Lampinen, Ishita Dasgupta, Stephanie C.~Y. Chan, Kory Matthewson,
  Michael~Henry Tessler, Antonia Creswell, James~L. McClelland, Jane~X. Wang,
  and Felix Hill.
\newblock Can language models learn from explanations in context?, 2022.
\newblock URL \url{https://arxiv.org/abs/2204.02329}.

\bibitem[Long et~al.(2016)Long, Pasupat, and Liang]{long-etal-2016-simpler}
Reginald Long, Panupong Pasupat, and Percy Liang.
\newblock Simpler context-dependent logical forms via model projections.
\newblock In \emph{Proceedings of the 54th Annual Meeting of the Association
  for Computational Linguistics (Volume 1: Long Papers)}, pages 1456--1465,
  Berlin, Germany, August 2016. Association for Computational Linguistics.
\newblock \doi{10.18653/v1/P16-1138}.
\newblock URL \url{https://aclanthology.org/P16-1138}.

\bibitem[Keskar et~al.(2019)Keskar, McCann, Varshney, Xiong, and
  Socher]{keskar2019ctrl}
Nitish~Shirish Keskar, Bryan McCann, Lav~R. Varshney, Caiming Xiong, and
  Richard Socher.
\newblock {CTRL:} {A} conditional transformer language model for controllable
  generation.
\newblock \emph{CoRR}, abs/1909.05858, 2019.
\newblock URL \url{http://arxiv.org/abs/1909.05858}.

\bibitem[See et~al.(2019)See, Roller, Kiela, and Weston]{see-etal-2019-makes}
Abigail See, Stephen Roller, Douwe Kiela, and Jason Weston.
\newblock What makes a good conversation? how controllable attributes affect
  human judgments.
\newblock In \emph{Proceedings of the 2019 Conference of the North {A}merican
  Chapter of the Association for Computational Linguistics: Human Language
  Technologies, Volume 1 (Long and Short Papers)}, pages 1702--1723,
  Minneapolis, Minnesota, June 2019. Association for Computational Linguistics.
\newblock \doi{10.18653/v1/N19-1170}.
\newblock URL \url{https://aclanthology.org/N19-1170}.

\bibitem[Zhu et~al.(2015)Zhu, Kiros, Zemel, Salakhutdinov, Urtasun, Torralba,
  and Fidler]{Zhu2015AligningBA}
Yukun Zhu, Ryan Kiros, Richard~S. Zemel, Ruslan Salakhutdinov, Raquel Urtasun,
  Antonio Torralba, and Sanja Fidler.
\newblock Aligning books and movies: Towards story-like visual explanations by
  watching movies and reading books.
\newblock \emph{2015 IEEE International Conference on Computer Vision (ICCV)},
  pages 19--27, 2015.

\bibitem[Rae et~al.(2020)Rae, Potapenko, Jayakumar, Hillier, and
  Lillicrap]{Rae2020Compressive}
Jack~W. Rae, Anna Potapenko, Siddhant~M. Jayakumar, Chloe Hillier, and
  Timothy~P. Lillicrap.
\newblock {Compressive Transformers for Long-Range Sequence Modelling}.
\newblock In \emph{ICLR}, 2020.

\bibitem[Jaegle et~al.(2022)Jaegle, Borgeaud, Alayrac, Doersch, Ionescu, Ding,
  Koppula, Zoran, Brock, Shelhamer, Henaff, Botvinick, Zisserman, Vinyals, and
  Carreira]{jaegle2022perceiver}
Andrew Jaegle, Sebastian Borgeaud, Jean-Baptiste Alayrac, Carl Doersch, Catalin
  Ionescu, David Ding, Skanda Koppula, Daniel Zoran, Andrew Brock, Evan
  Shelhamer, Olivier~J Henaff, Matthew Botvinick, Andrew Zisserman, Oriol
  Vinyals, and Joao Carreira.
\newblock Perceiver {IO}: A general architecture for structured inputs \&
  outputs.
\newblock In \emph{International Conference on Learning Representations}, 2022.
\newblock URL \url{https://openreview.net/forum?id=fILj7WpI-g}.

\bibitem[Hutchins et~al.(2022)Hutchins, Schlag, Wu, Dyer, and
  Neyshabur]{Hutchins2022Block}
DeLesley Hutchins, Imanol Schlag, Yuhuai Wu, Ethan Dyer, and Behnam Neyshabur.
\newblock Block-recurrent transformers, 2022.
\newblock URL \url{https://arxiv.org/abs/2203.07852}.

\end{thebibliography}

\end{document}